\documentclass{article}

\usepackage[margin=1in]{geometry}
\usepackage[T1]{fontenc}
\usepackage{amsmath, amssymb, amsthm, mathtools}
\usepackage[cal=euler]{mathalpha}
\usepackage{cochineal}
\usepackage{microtype}
\usepackage{graphicx}
\usepackage[numbers,sort&compress]{natbib}
\usepackage{hyperref}
\usepackage{booktabs}
\usepackage{array}
\usepackage{longtable}
\usepackage{xcolor}

\newcommand{\akshay}[1]{}
\newcommand{\vac}[1]{}

\newcommand{\KL}{\mathrm{KL}}
\newcommand{\kl}{\mathrm{kl}}
\newcommand{\TV}{\mathrm{TV}}
\newcommand{\Alt}{\mathrm{Alt}}
\newcommand{\Kinf}{\mathcal{K}_{\inf}}

\newtheorem{theorem}{Theorem}[section]
\newtheorem{proposition}[theorem]{Proposition}
\newtheorem{lemma}[theorem]{Lemma}
\newtheorem{corollary}[theorem]{Corollary}

\title{Instance-dependent sequential limits as an exact information ledger}
\author{Akshay Balsubramani \\ {\small \texttt{akshay@vac.bio}}}
\date{}

\makeatletter
\def\bvapx@title{}
\AtBeginDocument{\providecommand{\phantomsection}{}}
\let\bvapx@maketitle\maketitle
\def\maketitle{\global\let\bvapx@title\@title\bvapx@maketitle}
\newcommand{\appendixtitle}{%
  \clearpage
  \suppressfloats[t]%
  \phantomsection
  \begin{center}
    {\LARGE\bfseries Appendices\ifx\bvapx@title\@empty\else\space to ``\bvapx@title''\fi\par}%
  \end{center}
  \vspace{1.5\baselineskip}%
}
\makeatother

\providecommand{\akshay}[1]{}\renewcommand{\akshay}[1]{}%
\providecommand{\vac}[1]{}\renewcommand{\vac}[1]{}%
\begin{document}
\maketitle

\begin{abstract}
Instance-dependent limits for bandit regret, sequential tests, best-arm identification and Markov decision processes are proved by one chain.
It runs from a divergence decomposition through data processing to a statistic, a testing step that turns a class guarantee into a requirement on the history, and optimization over alternatives and allocations.
Only the first step is an identity.
This work writes the chain as an exact ledger.
By the chain rule, data processing loses exactly a conditional divergence, which vanishes for arms in natural exponential families at the per-arm counts and reward sums.
A one-parameter family of martingale identities extends the decomposition to every tilt between two instances.
The classical limits follow by relaxations with named defects.
All defects but the asymptotic step's can be computed from an algorithm's own runs, the uniform quantification's only when the class guarantee is a number.
On seven published sources, from movie ratings to engine replacement, the ledger locates the difference between standard algorithms and the classical limits.
For identification among trained models and for Wald's test the largest term is usually the uniform quantification's.
For regret the asymptotic limit overstates what practical horizons require.
At the midpoint tilt the finite-horizon limit improves the exponent of the Bretagnolle--Huber bound by a factor close to two for fixed allocations at small separations, and by less under adaptive play.
\end{abstract}

\section{Introduction}
\label{sec:intro}

An instance-dependent limit says how much sampling a sequential problem requires on a given instance.
A bandit algorithm must pull each suboptimal arm about $\log T$ times, divided by that arm's divergence to the best mean \cite{lai1985advances,burnetas1996optimal}.
A sequential test must draw about $\log(1/\alpha)$ observations, divided by the divergence between its hypotheses \cite{wald1945sequential}.
A best-arm identification procedure must sample for about $\log(1/\delta)$ times a characteristic time of the instance \cite{garivier2016optimal}.
An agent exploring a Markov decision process must visit each suboptimal state-action pair a matching number of times \cite{burnetas1997optimal,ok2018exploration}.

These limits share one proof.
For bandit regret and best-arm identification, that proof appears in \cite{kaufmann2016journal,garivier2019mathematics}.
It starts from the divergence decomposition, which writes the relative entropy between the histories under two instances as a sum over arms of expected pull counts times per-pull divergences.
It processes the history down to a statistic, replaces the algorithm's behavior at the two instances by a guarantee that the algorithm's class makes over its domain (the uniform quantification), and optimizes over alternatives and allocations.
The first step is an identity.
The later steps are inequalities.
The classical proofs leave the loss at each of them unmeasured.

This paper writes the chain as an exact ledger.
By the chain rule for relative entropy, data processing to a statistic loses exactly a conditional divergence \cite{polyanskiywu2025information}.
For arms in natural exponential families it vanishes when the statistic is the vector of per-arm pull counts and reward sums (Theorem~\ref{thm:sufficiency}).
A one-parameter family of martingale identities, the coincidence ledger, extends the divergence decomposition to every tilt between the two instances (Theorem~\ref{thm:ledger}).
Its endpoint is the divergence decomposition.
Its midpoint, the Bhattacharyya point, gives a finite-horizon limit (Theorem~\ref{thm:limit}).
Each remaining step is a relaxation with a named defect.
Section~\ref{sec:classical} lists $33$ classical results and one result of this paper, $29$ of them limits, each exact or obtained by a short sequence of these relaxations.
The appendix proves each of them in that form.

Every defect except that of the asymptotic step can be computed from an algorithm's own runs.
For the uniform quantification over the algorithm class this needs a guarantee that is a number, such as $\delta$ or a nominal level, and a change of measure to the alternative.
Under an asymptotic guarantee, such as consistency, the uniform quantification and the asymptotic step can be measured only together, as what the other defects leave of the difference from the limit at a practical horizon.
Section~\ref{sec:eval} computes these defects on published sources: movie ratings, the accuracies of trained models, an oat variety trial, an e-mail marketing experiment, the items of a national science assessment, a textbook maintenance problem and an engine-replacement model fitted to fleet data.
For identification among trained models and for Wald's test the largest term is usually the uniform quantification's.
In best-arm identification among trained models, with a $\delta$-correct stopping threshold it is the largest term of the adaptive identification algorithms at every risk $\delta$ tested.
At an algorithm's own error rates the chain gives a bound that lies much closer than the limit to the measured cost of the sequential tests and of Chernoff's rule for sequential design on the assessment items.
For identification no estimate of the error probability at the most confusing alternative passed an exact control, so that bound is not reported there (Section~\ref{sec:eval}).
In best-arm identification the ranking of the uniform quantification's term rests on such an estimate.
The estimate reads low in the control, so that term is drawn too large (Section~\ref{sec:eval-bai}).
For regret the asymptotically optimal algorithms stay below the limit at every horizon tested.
For these algorithms the joint term of the uniform quantification and the asymptotic step subtracts more than their whole shortfall below the limit: the asymptotic requirement overstates what practical horizons require.
With its uniform quantification computed at the algorithms' own regrets, the finite-horizon regret bound of the chain certifies far fewer pulls than they make on exact one-armed reductions (Section~\ref{sec:eval-regret}).
The midpoint's gain in the exponent of the Bretagnolle--Huber bound is close to a factor of two for fixed allocations at small separations.
On movie ratings, under the adaptive allocations of standard algorithms it falls with the horizon (Figure~\ref{fig:factor-two}), to between $1.03$ and $1.40$ at $T = 3{,}000$.

\section{Setting}
\label{sec:setting}

A bandit instance $\nu = (\nu_a)_{a \in [K]}$ assigns to each arm a law on a common space.
An algorithm interacts with $\nu$ for $T$ rounds, choosing the arm $A_t$ measurably with respect to the observed history and any independent randomization, and observing $X_t \sim \nu_{A_t}$.
Write $N_a(t)$ for the number of pulls of arm $a$ by time $t$, $S_a(t)$ for the sum of the observations collected from it, $\mathcal{F}_t$ for the observed filtration, and $P_\nu$ for the induced law on histories.
Throughout, $\lambda$ denotes a second instance with $\nu_a \ll \lambda_a$ for every arm, and $L_t := \log (dP_\nu / dP_\lambda)|_{\mathcal{F}_t}$.

Because the action probabilities are $\mathcal{F}_{t-1}$-measurable and do not depend on the instance, they cancel in $L_t$.
Every ledger statement of Sections~\ref{sec:allocation}--\ref{sec:mdp} therefore holds under adaptive allocation,
with no independence, fixed-design, or consistency assumption.

\subsection{Notation}
\label{sec:notation}

Each symbol below names one object throughout the paper.
Table~\ref{tab:notation}, at the head of the appendices, lists each symbol with the site of its definition.
Symbols defined inside a result, such as the error budget $\varepsilon$ of Theorem~\ref{thm:limit}, keep their meaning wherever they reappear.
Bound variables, such as the arguments of $\kl$, and letters introduced inside one proof hold only there.

The mean of arm $a$ under an instance $\lambda$ is $\mu_a(\lambda)$.
Its gap is $\Delta_a(\lambda) := \max_{a'} \mu_{a'}(\lambda) - \mu_a(\lambda)$.
Under $\nu$ they are $\mu_a := \mu_a(\nu)$ and $\Delta_a := \Delta_a(\nu)$.
The optimal mean is $\mu^\star := \max_a \mu_a$, attained by an arm $a^\star$.
The letter $b$ marks a suboptimal arm singled out for perturbation; sums over $b$ run over the suboptimal arms.
When an alternative $\lambda$ moves the law of arm $b$, the separation of its two means is $g := \mu_b(\lambda) - \mu_b$.
The regret on an instance $\lambda$ is $R_T(\lambda) := \sum_a \Delta_a(\lambda)\, \mathbb{E}_\lambda[N_a(T)]$, and $R_T := R_T(\nu)$.
An algorithm is consistent on a class of instances if $R_T(\lambda) = o(T^p)$ for every $p > 0$ and every instance $\lambda$ of the class.
A stopping time for $(\mathcal{F}_t)$ is written $\tau$.
An event in $\mathcal{F}_T$ or $\mathcal{F}_\tau$ is written $E$.
The law of a statistic $Y$ under $P_\nu$ is $P_\nu^Y$.
A statistic written $Z$ takes values in $[0,1]$.
The vector of pull counts is $N(t) := (N_a(t))_{a \in [K]}$.
The log-likelihood ratio of one observation from arm $a$ is $\ell_a := \log(d\nu_a / d\lambda_a)$.
The path log-likelihood ratio $L_t$ is the sum of $\ell_{A_{t'}}(X_{t'})$ over rounds $t' \le t$.

Relative entropy between two laws is written $\KL(\cdot \,\|\, \cdot)$.
Its value on the history is $D_T := \KL(P_\nu|_{\mathcal{F}_T} \,\|\, P_\lambda|_{\mathcal{F}_T})$; $D_\tau$ is the same quantity on $\mathcal{F}_\tau$.
The total variation distance is $\TV$.
For $u, v \in [0,1]$, $\kl(u, v)$ is the relative entropy between the Bernoulli laws with means $u$ and $v$.
In this paper a one-parameter exponential family is a natural exponential family: its laws have densities $e^{\theta y - \psi(\theta)}$ in the observation $y$ with respect to one base measure, so the reward is its natural statistic.
When each arm's family is one, $\theta$ is its natural parameter, $\psi$ its log-partition function and $\psi^\star$ the convex conjugate of $\psi$.
The natural parameters of $\nu_a$ and $\lambda_a$ are then $\theta_a$ and $\eta_a$.
In mean coordinates, $d(\mu, \mu')$ is the relative entropy from the law with mean $\mu$ to the law with mean $\mu'$.
The coincidence divergence of arm $a$ at tilt $s$ is $c_a(s)$; its path counterpart is $C_T(s)$ (Section~\ref{sec:coincidence}).
The Bhattacharyya coefficient of $P_\nu$ and $P_\lambda$ on $\mathcal{F}_T$ is $\rho_T := e^{-C_T(1/2)}$.

A star marks an optimal or extremal object, such as $a^\star$, $\mu^\star$ and $T^\star(\nu)$; $\psi^\star$ is the convex conjugate of $\psi$.
A bar marks an average.
The notation of regret classes, of identification and of Markov decision processes is added in Sections~\ref{sec:certificate}, \ref{sec:alternative} and \ref{sec:mdp}, where each is first used.
That of the sequential tests is added in Section~\ref{sec:eval-ab}.

\section{The divergence decomposition}
\label{sec:allocation}

The relative entropy between the two histories is determined by the expected allocation, at every stopping time.

\begin{theorem}[Divergence decomposition at a stopping time]
\label{thm:allocation}
Suppose $\KL(\nu_a \,\|\, \lambda_a) < \infty$ for every arm.
For any stopping time $\tau$ with $\mathbb{E}_\nu[\tau] < \infty$,
\begin{equation}
\label{eq:allocation}
\KL\!\left(P_\nu|_{\mathcal{F}_\tau} \,\big\|\, P_\lambda|_{\mathcal{F}_\tau}\right)
 \;=\; \sum_{a \in [K]} \mathbb{E}_\nu[N_a(\tau)] \, \KL(\nu_a \,\|\, \lambda_a)
\end{equation}
\end{theorem}

The proof is in Appendix~\ref{app:proofs-the-allocation-ledger}.

Equation~\eqref{eq:allocation} is the divergence decomposition \cite{lattimore2020bandit}.
It is the one step of the classical chain already in identity form.
Writing $n_a := \mathbb{E}_\nu[N_a(\tau)]$ and $k_a(\lambda) := \KL(\nu_a \,\|\, \lambda_a)$, it reads $\langle n, k(\lambda)\rangle = D_\tau$ for every admissible $\lambda$, so $\lambda \mapsto D_\tau$ is an exact linear functional of the expected allocation.
Evaluating it at $K$ alternatives whose vectors $k(\lambda^{(j)})$ are linearly independent determines $n$ outright.
The regret $\sum_a \Delta_a n_a$ is then a linear read-off, for every algorithm and every horizon.
Perturbing a single arm $b$ is the rank-one case,
$\mathbb{E}_\nu[N_b(\tau)] = D_\tau / \KL(\nu_b \,\|\, \lambda_b)$.

\section{Sufficiency and the data-processing step}
\label{sec:sufficiency}

Processing the history down to a statistic loses exactly one conditional divergence, which vanishes at the per-arm counts and reward sums for arms in exponential families.

\begin{theorem}[Sufficiency]
\label{thm:sufficiency}
Let $\tau$ be a stopping time.
For any $\mathcal{F}_\tau$-measurable statistic $Y$,
\begin{equation}
\label{eq:chain}
\KL\!\left(P_\nu|_{\mathcal{F}_\tau} \,\big\|\, P_\lambda|_{\mathcal{F}_\tau}\right)
 \;=\; \KL\!\left(P_\nu^{Y} \,\big\|\, P_\lambda^{Y}\right)
 \;+\; \mathbb{E}_\nu\!\left[\KL\!\left(P_\nu(\cdot \mid Y) \,\big\|\, P_\lambda(\cdot \mid Y)\right)\right]
\end{equation}
Suppose further that each arm's family is a one-parameter exponential family in its natural parameter and that $\tau$ is finite almost surely under both $P_\nu$ and $P_\lambda$.
Then $Y^\star := (N_a(\tau), S_a(\tau))_{a \in [K]}$ is sufficient for $\{P_\nu, P_\lambda\}$ on $\mathcal{F}_\tau$.
The second term vanishes if and only if $L_\tau$ coincides $P_\nu$-almost surely with a measurable function of $Y$.
\end{theorem}

The proof is in Appendix~\ref{app:proofs-sufficiency-closes-the-data-processing-s}.

The equivalence is the sufficiency theorem of Kullback and Leibler \cite[Theorem 4.1]{kullbackleibler1951information}, written for the conditional term.

The classical arguments at the horizon use the event $\{N_b(T) > T/2\}$ or the normalized count $N_b(T)/T$.
Both are strict coarsenings of $Y^\star$.
Theorem~\ref{thm:sufficiency} therefore identifies the term discarded by the data-processing step as one conditional divergence, which vanishes for the sufficient statistic.
The classical chains pay it because their class guarantee constrains only the coarser statistic.
Sections~\ref{sec:eval-regret} and~\ref{sec:eval-mdp-regret} measure this discarded term on movie ratings and on two Markov decision processes.

\section{The coincidence ledger}
\label{sec:coincidence}

For $s \in \mathbb{R}$ set $c_a(s) := -\log \int \nu_a^{\,s} \lambda_a^{\,1-s}$, the integral taken against any measure dominating both laws.
At $s = 1/2$ it is the Bhattacharyya divergence.
Its maximum over $s$ is the Chernoff information \cite{chernoff1952measure}.
The coincidence divergence of several laws is defined in \cite{balsubramani2026information}.
For $s \in [0,1]$, $c_a(s)$ is its case of two laws with weights $(s, 1 - s)$.
The path counterpart is $C_T(s) := -\log \int (dP_\nu)^{s} (dP_\lambda)^{1-s}$ on $\mathcal{F}_T$.
Where $c_a(s)$ is finite, the probability measure $e^{c_a(s)}\,\nu_a^{\,s} \lambda_a^{\,1-s}$ is the exponential tilt of $\lambda_a$ by $s\,\ell_a$.
The exponent $s$ is the tilt.

\begin{theorem}[Coincidence ledger]
\label{thm:ledger}
Fix $s$ such that $c_a(s)$ is finite for every $a$. Then
\begin{equation}
\label{eq:ledger}
M_t(s) \;:=\; \exp\!\Big( s\,L_t \;+\; \sum_{a \in [K]} N_a(t)\, c_a(s) \Big)
\end{equation}
is a $P_\lambda$-martingale with $\mathbb{E}_\lambda[M_t(s)] = 1$ for every $t$.
\end{theorem}

The proof is in Appendix~\ref{app:proofs-the-coincidence-ledger}.

For exponential-family arms, $M_t(s)$ is the product over arms of the test martingales of Kaufmann and Koolen \cite[eq.~(7)]{kaufmann2021mixture}, with exponent $s(\theta_a - \eta_a)$ on arm $a$.
For a single stream it descends from Wald's fundamental identity for one random walk \cite{wald1944cumulative}.
Its proof uses no parametric form.
Equivalently, $M_t(s)$ is the likelihood ratio on $\mathcal{F}_t$ of the instance whose arms are the tilted laws $e^{c_a(s)}\,\nu_a^{\,s} \lambda_a^{\,1-s}$, taken against $\lambda$.

\begin{corollary}[The coincidence ledger at a stopping time]
\label{cor:ledger-stopped}
Fix $s$ such that $c_a(s)$ is finite for every $a$, and let $\tau$ be a stopping time.
Then $\mathbb{E}_\lambda[M_{\tau \wedge t}(s)] = 1$ for every $t$, and $\mathbb{E}_\lambda[M_\tau(s)] \le 1$, with equality whenever $\tau$ is bounded.
\end{corollary}

The proof is in Appendix~\ref{app:proofs-the-coincidence-ledger}.

\begin{corollary}[The divergence decomposition is an endpoint]
\label{cor:endpoint}
Suppose $c_a(s)$ is finite for every $a$ and every $s$ in an open interval containing $[0,1]$, and suppose $\lambda_a \ll \nu_a$ for every $a$.
Then $c_a(1) = c_a(0) = 0$, $c_a'(0) = \KL(\lambda_a \,\|\, \nu_a)$ and $c_a'(1) = -\KL(\nu_a \,\|\, \lambda_a)$, and differentiating \eqref{eq:ledger} at $s = 1$ returns \eqref{eq:allocation} at every fixed horizon and at every bounded stopping time.
\end{corollary}

The proof is in Appendix~\ref{app:proofs-the-coincidence-ledger}.

\begin{corollary}[Anytime validity]
\label{cor:ville}
Fix $s$ such that $c_a(s)$ is finite for every $a$. For every $\alpha \in (0,1)$,
$P_\lambda\big( \exists t \ge 0 : s L_t + \sum_a N_a(t) c_a(s) \ge \log(1/\alpha) \big) \le \alpha$.
\end{corollary}

The proof is in Appendix~\ref{app:proofs-the-coincidence-ledger}.

\section{Per-pull divergences at the Bhattacharyya point}
\label{sec:tightening}

The testing step converts a constraint on play into a requirement that the two instances be hard to tell apart at the horizon.
Theorem~\ref{thm:limit} states that requirement in $2\,C_T(1/2)$, twice the path coincidence divergence at the Bhattacharyya point, where the Bretagnolle--Huber bound states it in the relative entropy $D_T$.
Proposition~\ref{prop:half} shows the former is never larger, so the bound of Theorem~\ref{thm:limit} on the error sum is never weaker than the Bretagnolle--Huber bound.
As a requirement on $D_T$, the quantity determined by the expected allocation, Theorem~\ref{thm:limit} gains at most $\log 2$ nats over the Bretagnolle--Huber bound, and the best-split bound of Proposition~\ref{prop:cl-split} is larger than both at every error sum below $1$.
The per-pull divergences satisfy the per-arm form of Proposition~\ref{prop:half}, $2\,c_b(1/2) \le \KL(\nu_b \,\|\, \lambda_b)$ for each arm.
Their ratio $\KL(\nu_b \,\|\, \lambda_b)/(2\,c_b(1/2))$ is the per-pull factor, computed at small separations in Proposition~\ref{prop:ratio}.
How much of the per-pull factor reaches the whole path depends on the allocation (Section~\ref{sec:limit}).

\begin{proposition}[The per-pull factor at small separations]
\label{prop:ratio}
Let arm $b$ have Bernoulli laws with mean $\mu_b$ under $\nu$ and mean $\mu_b + g$ under $\lambda$, and write $v := \mu_b(1-\mu_b)$. Then
\begin{equation}
\label{eq:ratio}
\frac{\kl(\mu_b,\, \mu_b+g)}{2\,c_b(1/2)} \;=\; 2 \;+\; \frac{2\mu_b-1}{3v}\,g \;+\; O(g^2)
\end{equation}
If instead both laws of arm $b$ are Gaussian with common variance $\sigma^2$ and means $g$ apart, then $c_b(s) = s(1-s)g^2/2\sigma^2$ for every $s$, so the per-pull factor equals $2$ at every $g$.
\end{proposition}

The proof is in Appendix~\ref{app:proofs-per-pull-divergences-at-the-bhattacharyy}.

The factor of two is a small-separation statement and not a bound in either direction.
Since $\KL(\nu_b \,\|\, \lambda_b)$ and $2\,c_b(1/2)$ are the R\'enyi divergences of orders one and one half of $\nu_b$ from $\lambda_b$, the value two is their local ratio \cite[(49)--(50)]{erven2014b}.
The first-order term of \eqref{eq:ratio} has the sign of $(2\mu_b - 1)g$.
At small separations the per-pull factor therefore exceeds $2$ when the alternative moves the mean away from $1/2$.
It falls below $2$ when the alternative moves the mean toward $1/2$.
The per-pull factor grows without bound as $\lambda_b$ assigns vanishing mass to an outcome that $\nu_b$ does not.

\section{The finite-horizon limit}
\label{sec:limit}

At the Bhattacharyya point the path coincidence divergence is $C_T(1/2) = -\log \int \sqrt{dP_\nu \, dP_\lambda}$.
Its relative-entropy endpoint is $D_T$.

\begin{theorem}[Finite-horizon limit]
\label{thm:limit}
For every $E \in \mathcal{F}_T$,
\begin{equation}
\label{eq:limit}
P_\nu(E^c) + P_\lambda(E) \;\ge\; 1 - \sqrt{1 - e^{-2 C_T(1/2)}}
\end{equation}
Consequently any algorithm and event with $P_\nu(E^c) + P_\lambda(E) \le \varepsilon \le 1$ satisfy $2 C_T(1/2) \ge \log\!\big(1/(\varepsilon(2-\varepsilon))\big)$.
The constant in \eqref{eq:limit} cannot be improved: equality holds for the pair of two-point laws $(q, 1-q)$ and $(1-q, q)$.
\end{theorem}

The proof is in Appendix~\ref{app:proofs-the-finite-horizon-limit}.

Inequality~\eqref{eq:limit} is Le Cam's affinity bound \cite[(14.11)]{lattimore2020bandit}.
The ledger reads its exponent as one point of the family of Theorem~\ref{thm:ledger}.
The Cauchy--Schwarz step in the proof of Theorem~\ref{thm:limit}, $\TV(P_\nu, P_\lambda) \le \sqrt{1 - \rho_T^2}$, is the upper edge of the joint range of total variation and Hellinger distance \cite{polyanskiywu2025information}.
Inequality~\eqref{eq:limit} is written as an identity in Lemma~\ref{lem:affinity}: the minimum error sum exceeds its right side by a non-negative defect, which vanishes exactly when $|L_T|$ is almost surely constant under $P_\lambda$.

Of the family's H\"older bounds on the error sum, the one at the Bhattacharyya point is the largest.

\begin{proposition}[The Bhattacharyya point maximizes the H\"older bound]
\label{prop:s-half}
For every $s \in (0,1)$, with $\underline{s} := s \wedge (1-s)$,
\begin{equation}
\label{eq:family}
\inf_{E \in \mathcal{F}_T} \big( P_\nu(E^c) + P_\lambda(E) \big)
 \;\ge\; 2^{\,1 - 1/\underline{s}}\, e^{-C_T(s)/\underline{s}}
\end{equation}
and the right side is maximized at $s = 1/2$, where it equals $\tfrac12 e^{-2 C_T(1/2)}$.
\end{proposition}

The proof is in Appendix~\ref{app:proofs-the-finite-horizon-limit}.

The H\"older route of its proof recovers the weaker constant $\tfrac12 e^{-2 C_T(1/2)}$ at that point, so Theorem~\ref{thm:limit} is the sharper statement there.
Proposition~\ref{prop:s-half} settles that no bound of the form \eqref{eq:family} at another tilt improves on it.
It also shows that the Chernoff point, the tilt that maximizes the exponent $C_T(s)$, weakens the H\"older bound.

Theorem~\ref{thm:limit} requires no consistency class, no asymptotics, and no restriction on the allocation rule.
It differs from the classical route in the functional of the pair $(P_\nu, P_\lambda)$ in which the requirement is stated.
Proposition~\ref{prop:half} measures that difference exactly.

\begin{proposition}[Relative entropy minus twice the Bhattacharyya divergence]
\label{prop:half}
Let $P^{(1/2)}$ denote the probability measure with $dP^{(1/2)} = e^{\,C_T(1/2)}\sqrt{dP_\nu\, dP_\lambda}$. Then
\begin{equation}
\label{eq:half}
D_T - 2\,C_T(1/2) \;=\; 2\,\KL\!\left(P_\nu \,\big\|\, P^{(1/2)}\right) \;\ge\; 0
\end{equation}
The same identity holds per arm with $\nu_a^{(1/2)} := e^{c_a(1/2)}\,\nu_a^{1/2}\lambda_a^{1/2}$ in place of $P^{(1/2)}$, giving $2\,c_a(1/2) \le \KL(\nu_a \,\|\, \lambda_a)$.
\end{proposition}

The proof is in Appendix~\ref{app:proofs-the-finite-horizon-limit}.

Equation~\eqref{eq:half} is the case $s = 1/2$, $P' = P_\nu$ of the identity $s \KL(P' \,\|\, P_\nu) + (1-s)\KL(P' \,\|\, P_\lambda) = \KL(P' \,\|\, P^{(s)}) + C_T(s)$, with $P^{(s)} \propto (dP_\nu)^s (dP_\lambda)^{1-s}$ \cite[(53)]{erven2014b}, which van Erven and Harremo\"es credit to Csisz\'ar \cite{csiszarmatus2003}.

The Bretagnolle--Huber bound on the same error sum is $\tfrac12 e^{-D_T}$ (Proposition~\ref{prop:cl-twopoint}), so Theorem~\ref{thm:limit} improves on it in the functional and in the constant.
Against the original form of the Bretagnolle--Huber inequality, $\TV(P_\nu, P_\lambda) \le \sqrt{1 - e^{-D_T}}$ (Appendix~\ref{app:cl-twopoint}), Theorem~\ref{thm:limit} improves in the functional alone.
Equation~\eqref{eq:half} measures the gain in the functional: the path factor $D_T/(2 C_T(1/2))$ equals $1 + \KL(P_\nu \,\|\, P^{(1/2)})/C_T(1/2)$.
The term $\KL(P_\nu \,\|\, P^{(1/2)}) = \tfrac12\,\mathbb{E}_\nu[L_T] + \log \mathbb{E}_\nu[e^{-L_T/2}]$ is a Jensen defect (often called the Jensen gap), zero exactly when $L_T$ is constant under $P_\nu$.
Under a fixed allocation of $n_b$ pulls to the perturbed arm, $L_T$ is a sum of $n_b$ independent terms, and the path factor equals the per-pull factor for every $n_b$, close to $2$ at small separations.
An asymptotically optimal algorithm allocates adaptively instead.
It stops sampling the perturbed arm once the evidence against it reaches about $\log t$, so $L_T$ concentrates and the path factor falls toward $1$.
On the MovieLens arm closest to the best (Section~\ref{sec:eval-factor-two}), the path factor of KL-UCB is $2.02$ at $T = 25$ and $1.09$ at $T = 3{,}000$ under the smaller of two perturbations (Figure~\ref{fig:factor-two}).
The factor of two therefore belongs to fixed designs and to short horizons on arms near the best.
Lattimore and Szepesv\'ari describe the bound $1 - \sqrt{1 - e^{-D_T}}$ as a modest improvement on their Theorem~14.2, the bound $\tfrac12 e^{-D_T}$ \cite[Chapter~14, Note~4]{lattimore2020bandit}.
The gain in the functional measured in Figure~\ref{fig:factor-two} is additional to that improvement.

The per-arm form of Theorem~\ref{thm:limit} does not decompose the way \eqref{eq:allocation} does.
The measure in the next proposition is a probability measure by Theorem~\ref{thm:ledger}.

\begin{proposition}[Per-arm form at the Bhattacharyya point]
\label{prop:tilt}
Let $Q_s$ be the measure with $dQ_s / dP_\lambda = M_T(s)$, a probability measure, and let $P^{(1/2)}$ be as in Proposition~\ref{prop:half}. Then
\begin{equation}
\label{eq:tilt}
\sum_{a \in [K]} \mathbb{E}_{Q_{1/2}}\!\left[ N_a(T) \right] c_a(1/2) \;-\; C_T(1/2)
 \;=\; \KL\!\left(Q_{1/2} \,\big\|\, P^{(1/2)}\right) \;\ge\; 0
\end{equation}
\end{proposition}

The proof is in Appendix~\ref{app:proofs-the-finite-horizon-limit}.

Equation~\eqref{eq:tilt} differs from \eqref{eq:allocation} in where the allocation is measured: under the tilted law $Q_{1/2}$, whose per-arm law is the geometric mixture $\nu_a^{(1/2)}$ of the two endpoints.
The two allocations do not agree.
On the MovieLens reductions of Section~\ref{sec:eval-factor-two} the perturbed arm is pulled several times as often under the tilted law as under $P_\nu$ (Figure~\ref{fig:factor-two}b).
A comparison between \eqref{eq:tilt} and the classical per-arm bound is therefore not like for like.

\section{Per-arm transported fractions}
\label{sec:certificate}

For regret, the law of arm $a$ ranges over a class $\mathcal{M}_a$.
The instances range over the product class $\mathcal{M}_1 \times \dots \times \mathcal{M}_K$ of the $\lambda$ with $\lambda_a \in \mathcal{M}_a$ for every arm.
With one class $\mathcal{M}$ for every arm, the product class is $\mathcal{M}^K$.
For a set $\mathcal{H}$ of laws, $\Kinf(\cdot, \mathcal{H})$ is the infimum of the relative entropy from its first argument to a law in $\mathcal{H}$.
For a suboptimal arm $b$, $\Kinf(\nu_b, \mu^\star)$ takes $\mathcal{H}$ to be the laws in $\mathcal{M}_b$ with mean above $\mu^\star$.
The Lai--Robbins constant is $c^\star(\nu) := \sum_b \Delta_b / \Kinf(\nu_b, \mu^\star)$; when every $\mathcal{M}_a$ is the same one-parameter exponential family and some law of the family has mean above $\mu^\star$, $\Kinf(\nu_b, \mu^\star) = d(\mu_b, \mu^\star)$.

Fix an instance whose arms belong to a one-parameter exponential family with a law of mean above $\mu^\star$, and let $\lambda$ raise a suboptimal arm $b$ to $\mu^\star$.
Writing $d_b := d(\mu_b, \mu^\star)$, the corresponding term of \eqref{eq:allocation} is
\begin{equation}
\label{eq:cert}
I_b(T) \;:=\; \mathbb{E}_\nu[N_b(T)] \cdot d_b
\end{equation}
the information transported through arm $b$ by time $T$.
It is exact given the counts.
The transported fraction $I_b(T)/\log T$ of arm $b$ measures how much of the asymptotic requirement an algorithm has met.
For KL-UCB on a Lipschitz bandit with Bernoulli arms, $\mathbb{E}_\nu[N_b(T)]/\log T$ is plotted for each suboptimal arm in \cite{magureanu2014lipschitz}.
It is reported to approach $1/d_b$, a transported fraction of one.

\begin{proposition}[Relation to the regret ratio]
\label{prop:cert}
Here the Lai--Robbins constant is $c^\star(\nu) = \sum_{b} \Delta_b / d_b$. Then
\begin{equation}
\label{eq:cert-avg}
\frac{R_T}{c^\star(\nu) \log T} \;=\; \frac{\sum_b (\Delta_b/d_b)\, \big(I_b(T)/\log T\big)}
{\sum_b \Delta_b/d_b}
\end{equation}
so the regret ratio is the $(\Delta_b/d_b)$-weighted average of the per-arm transported fractions.
When every suboptimal arm has the same gap the weights are equal and the regret ratio is their unweighted mean; it equals each transported fraction individually only if the algorithm also allocates equally across those arms.
\end{proposition}

The proof is in Appendix~\ref{app:proofs-a-per-arm-efficiency-certificate}.

Proposition~\ref{prop:cert} places the transported fraction against the quantity it generalizes.
On a homogeneous instance that the algorithm treats symmetrically it is no more informative than the regret ratio.
Off one it separates arms that are averaged together in the regret ratio, and it distinguishes two situations that are conflated in the regret ratio: an algorithm that transports more information than it needs, and a horizon too short for the asymptotic requirement to have accrued.
Both occur on the MovieLens arms of Section~\ref{sec:eval-regret} at $T = 20{,}000$.
On the liked-or-not instance UCB1's transported fractions run from $2.9$ to $4.7$ across arms, behind a regret ratio of $4.0$.
The asymptotically optimal algorithms' transported fractions stay below one; for IMED they run from $0.42$ to $0.77$, behind a regret ratio of $0.56$.

\section{The alternative set and the allocation supremum}
\label{sec:alternative}

The remaining two steps of the classical chain optimize over the alternatives and over allocations.

For best-arm identification, $\nu$ has a unique optimal arm, and $\Alt(\nu)$ is the set of instances in which some arm other than $a^\star$ is at least as good as $a^\star$.
An allocation $w$ is a point of the simplex over arms.
The characteristic time $T^\star(\nu)$ satisfies $T^\star(\nu)^{-1} = \sup_w \inf_{\lambda \in \Alt(\nu)} \sum_a w_a \KL(\nu_a \,\|\, \lambda_a)$.
A best-arm identification algorithm, with stopping time $\tau$ and recommendation $\hat a$, is $\delta$-correct if on every instance $\lambda$ with a unique optimal arm $a^\star(\lambda)$ it stops almost surely and $P_\lambda(\hat a \ne a^\star(\lambda)) \le \delta$; $\delta$ is its risk.
Other alternative sets and their characteristic times are marked by a subscript, such as $\Alt_m(\nu)$ and $T^\star_m(\nu)$ for top-$m$ identification.
For a structured class $\Phi$ of bandit instances the alternative set is $\Alt_\Phi(\nu)$ and the regret constant $c^\star_\Phi(\nu)$ (Corollary~\ref{cor:cl-structured}); without a subscript $c^\star(\nu)$ is that of the product class $\mathcal{M}_1 \times \dots \times \mathcal{M}_K$.

\begin{proposition}[The alternative-set infimum is a Bregman information]
\label{prop:mbary}
For a one-parameter exponential family with conjugate log-partition $\psi^\star$, and weights $w$ with $w_i \ge 0$ and $\sum_i w_i = 1$,
\begin{equation}
\label{eq:mbary}
\min_{\mu'} \sum_i w_i \, d(\mu_i, \mu')
 \;=\; \sum_i w_i \psi^\star(\mu_i) - \psi^\star\!\Big(\sum_i w_i \mu_i\Big)
\end{equation}
attained at the barycentre $\sum_i w_i \mu_i$ of the means.
\end{proposition}

The proof is in Appendix~\ref{app:proofs-the-alternative-set-and-the-allocation-s}.

The minimization in \eqref{eq:mbary} is over the second argument of the divergence.
Its value is the Jensen defect of $\psi^\star$ in the mean parameters, the Bregman information of the means \cite{banerjee2005journal}.
For two means the Bregman information is the divergence through which the infimum over the alternative set is computed in \cite[Lemma~3]{garivier2016optimal}.
The coincidence divergence $c_a(s)$ of Section~\ref{sec:coincidence} is the Jensen defect of $\psi$ in the natural parameters, attained at the geometric mixture \cite[Theorem~3]{nielsen2011burbea}.
The two are Jensen defects of the Legendre-dual pair $\psi$, $\psi^\star$.
Lemma~\ref{lem:jensen-pair} compares the two at matching weights.
For Gaussian arms with common variance $\sigma^2$ and means $g$ apart they are equal, both $s(1-s)g^2/2\sigma^2$.
For exponentially distributed arms the coincidence divergence at tilt $s$ equals the Bregman information at the reflected weights $(1-s, s)$, so the two agree at $s = 1/2$.
In general they agree to second order in the separation $g$, and at $s = 1/2$ to third order.
On Bernoulli arms their disagreement is unbounded.
There the Bregman information is a Jensen defect of the negative entropy and never exceeds $\log 2$.
For means $q \le 1/2$ and $1-q$ the coincidence divergence lies within $\log 2$ of $(s \wedge (1-s))\log(1/q)$.
The relative discrepancy therefore diverges as the two laws approach opposite vertices of the simplex.
They enter the chain at different steps, the coincidence divergence at the testing step and the Bregman information at the alternative-set step.
Where the two differ, as on Bernoulli arms, conflating them misprices whichever step receives the wrong one.
In fixed-budget identification the divergences of the lower bound run from the alternative to the instance.
There the alternative-set step returns a coincidence divergence between the laws of two arms (Proposition~\ref{prop:cl-fb-rate}).
The largest coincidence divergence over the tilt and the largest Bregman information over the weights are the information quantities of the fixed-budget and the fixed-confidence lower bounds for two arms, in that order \cite[Appendix~C]{kaufmann2016journal}.
They are equal for Gaussian and for exponentially distributed arms, but on Bernoulli arms the first is the larger \cite[Section~5.1]{kaufmann2016journal}.

For best-arm identification the characteristic time $T^\star(\nu)^{-1} = \sup_w \inf_{\lambda \in \Alt(\nu)} \sum_a w_a \KL(\nu_a \,\|\,
\lambda_a)$ inherits \eqref{eq:mbary} pairwise.
The supremum then collapses.

\begin{proposition}[Scalar form of the characteristic time]
\label{prop:scalar}
For Gaussian arms with common variance $\sigma^2$, where $d_b = \Delta_b^2/2\sigma^2$,
\begin{equation}
\label{eq:scalar}
T^\star(\nu) \;=\; \min_{\gamma \in (0, \min_b d_b)} \Big[ \frac{1}{\gamma} + \sum_{b \ne a^\star}
\frac{1}{d_b - \gamma} \Big]
\end{equation}
with $\gamma^\star$ the minimizer, $w_{a^\star} = 1/(\gamma^\star T^\star(\nu))$, $w_b = 1/\big(T^\star(\nu) (d_b - \gamma^\star)\big)$, and stationarity $\gamma^{-2} = \sum_b (d_b - \gamma)^{-2}$, equivalently $w_{a^\star}^2 = \sum_b w_b^2$.
\end{proposition}

The proof is in Appendix~\ref{app:proofs-the-alternative-set-and-the-allocation-s}.

Garivier and Kaufmann \cite[Lemmas 3 and 4, Theorem 5 and p.~6]{garivier2016optimal} proved that at the optimal allocation the $K-1$ pairwise infima are equal, reduced the optimal weights to a one-dimensional root-finding problem, and stated the Gaussian condition $w_{a^\star}^2 = \sum_b w_b^2$.
For Gaussian arms with a common variance, Glynn and Juneja \cite[Theorem~1 and Example~1]{glynn2004large} had derived the same equalization and condition for the fixed-budget problem over static allocations.
Equation~\eqref{eq:scalar} writes the reduction of Garivier and Kaufmann for Gaussian arms as a convex minimization in one variable.
For unit-variance Gaussian arms, Barrier, Garivier and Koc\'ak \cite[Lemma~7 and Proposition~8]{barrier2022nonasymptotic} obtain the weights and the characteristic time in closed form from the unique root of a convex, strictly decreasing function of one variable.
In the variable $1/(2\gamma)$, that root equation is the stationarity condition of \eqref{eq:scalar} with $\sigma = 1$.
Proposition~\ref{prop:scalar-general} writes the reduction for any one-parameter family.
The remarks after Proposition~\ref{prop:cl-weights} map it to the notation of Garivier and Kaufmann.

\begin{proposition}[Scalar form outside the Gaussian case]
\label{prop:scalar-general}
For one-parameter exponential-family arms, let
\[
G_b(r) \;:=\; \min_{\mu'} \big[ d(\mu_{a^\star}, \mu') + r\, d(\mu_b, \mu') \big]
\]
denote the pairwise alternative-set infimum at weights $(1, r)$, which is continuous and strictly increasing on $[0, \infty)$ with supremum $d(\mu_{a^\star}, \mu_b)$. Then
\begin{equation}
\label{eq:scalar-general}
T^\star(\nu) \;=\; \min_{\gamma \in (0,\, \min_b d(\mu_{a^\star}, \mu_b))}
 \frac{1 + \sum_{b \ne a^\star} G_b^{-1}(\gamma)}{\gamma}
\end{equation}
with $w_b/w_{a^\star} = G_b^{-1}(\gamma^\star)$ at the minimizer.
\end{proposition}

The proof is in Appendix~\ref{app:proofs-the-alternative-set-and-the-allocation-s}.

The object that replaces $1/(d_b - \gamma)$ is $G_b^{-1}(\gamma)/\gamma$. The program stays one-dimensional with a monotone inner solve; convexity in $\gamma$, however, is a feature of the Gaussian case and is not claimed in general.
The divergences $d(\mu_{a^\star}, \mu_b)$ of Proposition~\ref{prop:scalar-general} run from the optimal arm, in the reverse order of $d_b = d(\mu_b, \mu^\star)$; outside the Gaussian case the two need not be equal.

\begin{proposition}[Union and intersection alternative sets]
\label{prop:altsets}
Let $w$ be an allocation, $k(\lambda)$ the vector of per-pull divergences of Section~\ref{sec:allocation}, and $\mathcal{C}$ a set of alternatives.
If $\mathcal{C} = \bigcup_j \mathcal{B}_j$ then $\inf_{\lambda \in \mathcal{C}} \langle w, k(\lambda)\rangle = \min_j \inf_{\lambda \in \mathcal{B}_j} \langle w, k(\lambda)\rangle$.
If $\mathcal{C} = \bigcap_j \mathcal{B}_j$ then $\inf_{\lambda \in \mathcal{C}} \langle w, k(\lambda)\rangle \ge \max_j \inf_{\lambda \in \mathcal{B}_j} \langle w, k(\lambda)\rangle \ge \min_j \inf_{\lambda \in \mathcal{B}_j} \langle w, k(\lambda)\rangle$.
\end{proposition}

The proof is in Appendix~\ref{app:proofs-the-alternative-set-and-the-allocation-s}.

The alternative set of a top-$m$ identification problem is a union of single-swap half-spaces, each constraining two coordinates \cite[Lemma~1]{reda2021dealing}, so Proposition~\ref{prop:altsets} makes its infimum a minimum of pairwise infima of the form \eqref{eq:mbary} (Proposition~\ref{prop:cl-topm-exact}).
A conjunctive alternative set, where several arms must move together, is an intersection;
the maximum of the pairwise infima is then the better of the two surrogates and both underestimate, so the infimum over the intersection itself is needed there.

\section{Markov decision processes}
\label{sec:mdp}

Nothing in Section~\ref{sec:coincidence} used the bandit structure beyond the condition that the probabilities of the sampled component be $\mathcal{F}_{t-1}$-measurable and the same under both instances.
In a Markov decision process the state is observed before the action is drawn, so the state and the action of each round meet this condition and the argument transfers with visit counts in place of pull counts.
The condition also holds inside an episode, a finite trajectory logged by a known policy that may depend on the whole history of the episode.
The policy's probabilities are the same under both instances, so they cancel from the episode's likelihood ratio.
Within an episode the divergence decomposition then holds with each history at which the environment draws an observation in place of an arm, weighted by the probability that the episode reaches it.
A one-sided sequential test on independent logged episodes is therefore covered by Proposition~\ref{prop:cl-power-one} for any null set of environments, with each environment identified with the law of one logged episode.
For one trajectory of an ergodic finite-state Markov chain, a non-asymptotic lower bound on the expected stopping time of one-sided power-one tests is proved for any composite null in \cite[Theorem~3.3]{sethi2026asymptotically}.
There the per-state relative entropies are averaged under the stationary distribution of the chain.
Under a fixed stationary policy the states and actions of a Markov decision process form a finite-state Markov chain, which is assumed ergodic in \cite[Section~5.2]{sethi2026asymptotically}.

An instance $\nu$ of a Markov decision process with finite state and action sets $\mathcal{X}$ and $\mathcal{A}$ assigns to each state-action pair $(x,a)$ a law $\nu_{x,a}$ of the next state and the reward.
Its probability of a move from $x$ to $y$ under $a$ is $\nu(y \mid x,a)$.
Its mean reward at $(x,a)$ is $r(x,a)$.
In each round an algorithm observes the state and draws an action given the observed history and any independent randomization.
It then observes the reward and the next state, drawn from the law of the instance at that state-action pair.
The observed filtration $\mathcal{F}_t$ is generated by the initial state and by the actions, rewards and next states of the first $t$ rounds.
The law of the history under $\nu$ is $P_\nu$.
Write $\nu \ll \lambda$ when $\nu_{x,a} \ll \lambda_{x,a}$ at every pair.
For two instances with the same initial-state law and $\nu \ll \lambda$, $L_t := \log(dP_\nu / dP_\lambda)|_{\mathcal{F}_t}$.
The visits to $(x,a)$ by round $t$ are counted by $N_{x,a}(t)$, and the visits to $x$ by $N_x(t)$.
The instance is ergodic if every stationary deterministic policy makes the sequence of states an irreducible Markov chain \cite{ok2018exploration}.
An ergodic instance has an optimal gain $g^\star$ and a bias $h^\star$.
The Bellman gap of $(x,a)$ is $\Delta_{x,a} := g^\star + h^\star(x) - r(x,a) - \sum_y \nu(y \mid x,a)\, h^\star(y)$.
The actions with $\Delta_{x,a} = 0$ form the set $\mathcal{A}^\star_x$ of optimal actions at $x$, and the optimal stationary policies form $\Pi^\star$; under $\lambda$ these objects are $\Delta_{x,a}(\lambda)$, $\mathcal{A}^\star_x(\lambda)$ and $\Pi^\star(\lambda)$.
The instance belongs to a known class $\Phi$ of instances.
Its alternative set $\Alt_\Phi(\nu)$ holds the $\lambda \in \Phi$ with $\nu \ll \lambda$ that agree with $\nu$ at the optimal pairs and share no optimal stationary policy with it.
The regret constant of the class is $c^\star_\Phi(\nu) := \inf_\zeta \sum_{x,a} \zeta(x,a)\,\Delta_{x,a}$, where the infimum is over the rates $\zeta \ge 0$ with $\sum_{x,a} \zeta(x,a)\,\KL(\nu_{x,a} \,\|\, \lambda_{x,a}) \ge 1$ for every $\lambda \in \Alt_\Phi(\nu)$ (Proposition~\ref{prop:cl-mdp}).
For a suboptimal pair, $\Kinf(x,a)$ is the infimum of $\KL(\nu_{x,a} \,\|\, \lambda_{x,a})$ over the instances $\lambda \in \Phi$ that differ from $\nu$ only at $(x,a)$ and under which the mean reward at $(x,a)$ plus the mean of $h^\star$ at the next state exceeds $g^\star + h^\star(x)$; such an instance makes the pair optimal \cite{ok2018exploration}.
The information transported through the pair by round $t$ is $I_{x,a}(t) := \mathbb{E}_\nu[N_{x,a}(t)]\,\Kinf(x,a)$.
Its transported fraction is $I_{x,a}(t)/\log t$.
From an initial state, the regret $R_T(\lambda)$ of an algorithm on an instance $\lambda$ is $T$ times the optimal gain of $\lambda$ minus the expected total reward over $T$ rounds, and $R_T := R_T(\nu)$; for a bandit it is the regret of Section~\ref{sec:notation}.
An algorithm is uniformly good on $\Phi$ if its regret is $o(T^p)$ for every $p > 0$, every ergodic instance in $\Phi$ and every initial state.

\begin{theorem}[Coincidence ledger on a Markov decision process]
\label{thm:mdp}
Let $\nu$ and $\lambda$ share their initial-state law, let $N_{x,a}(t)$ count visits to the state-action pair $(x,a)$, and let $c_{x,a}(s)$ be the coincidence divergence of $\nu_{x,a}$ and $\lambda_{x,a}$.
Fix $s$ such that $c_{x,a}(s)$ is finite for every pair.
Then $\exp\big( s L_t + \sum_{x,a} N_{x,a}(t)\, c_{x,a}(s) \big)$ is a mean-one $P_\lambda$-martingale.
\end{theorem}

The proof is that of Theorem~\ref{thm:ledger} with the index set enlarged.
Its $s \to 1$ endpoint is the divergence decomposition over state-action pairs (Proposition~\ref{prop:cl-mdp-decomposition}), on which the regret limit of Proposition~\ref{prop:cl-mdp} is built.

The alternative-set step transfers with the same caveat as in Section~\ref{sec:alternative}: its object is not the coincidence divergence.
Over categorical transition kernels the infimum over the alternative set $\Alt_\Phi(\nu)$ in Proposition~\ref{prop:cl-mdp} is again a minimization over the second argument, but it produces no barycentre.
The alternative set holds the laws of the optimal pairs fixed and lets each suboptimal pair's kernel move on its own.
An alternative that changes only the suboptimal pair $(x,a)$ moves its law by at least $\Kinf(x,a)$ in relative entropy, which is infinite when no law makes the pair optimal.
In a class where each pair's law ranges over its own set, such as all categorical transition kernels, the regret constant is set by these single-pair alternatives: $c^\star_\Phi(\nu) = \sum_x \sum_{a \notin \mathcal{A}^\star_x} \Delta_{x,a}/\Kinf(x,a)$, with $1/0 := \infty$ and $1/\infty := 0$ (Proposition~\ref{prop:cl-mdp-decoupled}).

\section{Classical limits as relaxations of the ledger}
\label{sec:classical}

Table~\ref{tab:classical} lists 34 results: 33 classical results from the bandit, sequential-testing and reinforcement-learning literature and one result of this paper (Proposition~\ref{prop:cl-own}).
Of the 34, 29 are instance-dependent limits, three are ingredients of their proofs and two are exact results on Markov decision processes.
The ingredients are the divergence decomposition, the fundamental inequality and the characteristic time.
Each result starts from an exact quantity: the divergence decomposition of Theorem~\ref{thm:allocation} or its Markov decision process form (Proposition~\ref{prop:cl-mdp-decomposition}), the minimum error sum $1 - \TV(P_\nu, P_\lambda)$ of a two-point test, or the allocation supremum of Section~\ref{sec:alternative}.
A short sequence of relaxations then gives the result.
For four of the rows the sequence is empty.
The small capitals \textsc{process}, \textsc{test}, \textsc{binary}, \textsc{markov}, \textsc{uniform}, \textsc{asymptotic}, \textsc{boundary} and \textsc{allocation} name the relaxations, and the loss of each is its defect.
Every relaxation but \textsc{asymptotic} is an inequality whose defect is an exact quantity, \textsc{uniform} only when the class guarantee is a number:
\begin{itemize}
\item \textsc{process}: data processing to a statistic $Y$, $D_\tau \ge \KL(P_\nu^Y \,\|\, P_\lambda^Y)$, or in total variation, $\TV(P_\nu, P_\lambda) \ge \TV(P_\nu^Y, P_\lambda^Y)$.
The defect is the conditional divergence in the chain rule \eqref{eq:chain}, zero for a sufficient statistic, or the difference of the two total variations.
\item \textsc{test}: a two-point testing inequality in place of the exact value $1 - \TV(P_\nu, P_\lambda)$.
The defects are the Jensen defect of Lemma~\ref{lem:affinity}, the term $2\,\KL(P_\nu \,\|\, P^{(1/2)})$ of Proposition~\ref{prop:half}, and the loss of any weaker form of the affinity bound, which the proofs name.
The first is the loss of the Cauchy--Schwarz step in the proof of Theorem~\ref{thm:limit}.
\item \textsc{binary}: an elementary lower bound on the Bernoulli relative entropy $\kl$.
\item \textsc{markov}: Markov's inequality, which turns a bound on an expected count, on the regret or on a risk into a bound on the probability of an event.
\item \textsc{uniform}: the uniform quantification.
The algorithm's behavior at the instance and at the alternative is replaced by the guarantee of its class over the class's whole domain: consistency, $\delta$-correctness, a regret bound or a risk bound.
\item \textsc{asymptotic}: the horizon taken to infinity, or an error probability or the sampling cost taken to zero.
\item \textsc{boundary}: alternatives inside the alternative set that approach the infimum of the divergence over the set.
For a one-parameter exponential family the infimum lies on the set's boundary.
A proof lets the alternatives approach the infimum unless the bound degrades as they do.
In that case it keeps one alternative at a fixed margin beyond the boundary and names the difference between its divergence and the infimum.
\item \textsc{allocation}: the algorithm's own allocation replaced by the best allocation.
\end{itemize}

{\small
\begin{longtable}{>{\raggedright\arraybackslash}p{0.27\linewidth}>{\raggedright\arraybackslash}p{0.24\linewidth}>{\raggedright\arraybackslash}p{0.27\linewidth}>{\raggedright\arraybackslash}p{0.11\linewidth}}
\caption{\textbf{Each of 34 results is an exact quantity or follows from one by named relaxations.} Each row names a result and its source, its setting, the relaxations its proof applies (defined in Section~\ref{sec:classical} and listed in the order given there) and where the paper states it. Appendix~\ref{app:proofs-classical} proves the appendix statements.}
\label{tab:classical}\\
\toprule
Result & Setting & Relaxations & Statement \\
\midrule
\endfirsthead
\toprule
Result & Setting & Relaxations & Statement \\
\midrule
\endhead
\midrule
\multicolumn{4}{r}{\textit{Continued on the next page}} \\
\endfoot
\bottomrule
\endlastfoot
Divergence decomposition \cite{lattimore2020bandit} & any algorithm, stopping time & none (identity) & Thm.~\ref{thm:allocation} \\
Divergence decomposition on a Markov decision process \cite{ok2018exploration} & any algorithm, stopping time & none (identity) & Prop.~\ref{prop:cl-mdp-decomposition} \\
Fundamental inequality \cite{garivier2019mathematics}; transportation lemma \cite{kaufmann2016journal} & statistic in $[0,1]$, or an event & \textsc{process} & Lem.~\ref{lem:relax}(b) \\
Le Cam, Bretagnolle--Huber \cite{bretagnolle1979estimation} and Pinsker bounds \cite[Chapter~14]{lattimore2020bandit} & two instances, fixed horizon & \textsc{process}, \textsc{test}; Pinsker: \textsc{process}, \textsc{binary} & Prop.~\ref{prop:cl-twopoint}, Lem.~\ref{lem:affinity} \\
Best-split bound \cite{fedotov2003refinements,berend2014minimum} & an event, through its error sum & \textsc{process}, \textsc{binary} & Prop.~\ref{prop:cl-split} \\
Chernoff's exponent and its converse \cite{chernoff1952measure,polyanskiywu2025information} & two instances, fixed horizon, deterministic counts, mutually absolutely continuous laws & \textsc{test} & Lem.~\ref{lem:chernoff-point} \\
Lai--Robbins \cite{lai1985advances} & one-parameter family, consistent algorithms & \textsc{process}, \textsc{binary}, \textsc{markov}, \textsc{uniform}, \textsc{asymptotic}, \textsc{boundary} & Prop.~\ref{prop:cl-lai-robbins} \\
Burnetas--Katehakis \cite{burnetas1996optimal} & any product class, consistent algorithms & as for Lai--Robbins & Prop.~\ref{prop:cl-lai-robbins} \\
Finite-horizon regret bound \cite{lattimore2020bandit} & fixed horizon, regret bound on a neighborhood & \textsc{process}, \textsc{test}, \textsc{markov}, \textsc{uniform}, \textsc{boundary} & Prop.~\ref{prop:cl-finite}, Cor.~\ref{cor:cl-gauss} \\
Initial-phase count bound \cite{garivier2019mathematics} & any class, optimal arms pulled at least $T/K$ times & \textsc{process}, \textsc{binary}, \textsc{uniform}, \textsc{boundary} & Prop.~\ref{prop:cl-gms-initial} \\
Finite-horizon Lai--Robbins bound \cite{garivier2019mathematics} & well-behaved class, counts at most $\Upsilon\log T/\Delta_a^2$ & \textsc{process}, \textsc{binary}, \textsc{uniform}, \textsc{boundary} & Prop.~\ref{prop:cl-gms-finite} \\
Explore-then-commit \cite{garivier2016neurips} & two Gaussian arms, consistent algorithms & \textsc{process}, \textsc{binary}, \textsc{uniform}, \textsc{asymptotic}; \textsc{boundary} for an unknown gap & Prop.~\ref{prop:cl-etc} \\
Fixed-design explore-then-commit \cite{garivier2016neurips} & two Gaussian arms, known gap, consistent algorithms & \textsc{process}, \textsc{uniform}, \textsc{asymptotic} & Prop.~\ref{prop:cl-etc-fixed} \\
Wald's bound \cite{wald1945sequential} and the SPRT's defect & one stream, two simple hypotheses & \textsc{process}; \textsc{uniform} for nominal levels & Prop.~\ref{prop:cl-wald} \\
Wald--Wolfowitz optimality of the SPRT \cite{waldwolfowitz1948optimum}, zero-defect case & one stream, two simple hypotheses & \textsc{process}, \textsc{uniform} & Prop.~\ref{prop:cl-ww} \\
Tests of several or composite hypotheses \cite{hoeffding1960lower,lai1988nearly} & one stream, finitely many disjoint hypotheses, truth in one of them & \textsc{process}, \textsc{uniform}, \textsc{boundary}; \textsc{binary} and \textsc{asymptotic} for the logarithmic form & Prop.~\ref{prop:cl-multitest} \\
Power-one tests with a composite null \cite{agrawal2025stopping} & one stream, any composite null & \textsc{process}, \textsc{uniform}, \textsc{boundary} & Prop.~\ref{prop:cl-power-one} \\
Sequential design \cite{chernoff1959sequential} & finitely many states and experiments & \textsc{process}, \textsc{binary}, \textsc{markov}, \textsc{uniform}, \textsc{asymptotic}, \textsc{allocation} & Prop.~\ref{prop:cl-chernoff} \\
Active hypothesis testing, first-order term \cite{naghshvar2013active} & finitely many states and experiments, a prior on the states & \textsc{process}, \textsc{binary}, \textsc{markov}, \textsc{uniform}, \textsc{asymptotic}, \textsc{allocation} & Cor.~\ref{cor:cl-active} \\
Sequential design at fixed confidence \cite{kaufmann2021mixture} & finitely many experiments, any set of states & \textsc{process}, \textsc{uniform}, \textsc{boundary}, \textsc{allocation}; \textsc{binary} for the $\log(1/(3\delta))$ form & Prop.~\ref{prop:cl-design} \\
Best-arm identification \cite{garivier2016optimal} & fixed confidence & \textsc{process}, \textsc{uniform}, \textsc{boundary}, \textsc{allocation} & Prop.~\ref{prop:cl-bai} \\
Top-$m$ identification \cite{kaufmann2016journal} & fixed confidence & \textsc{process}, \textsc{uniform}, \textsc{boundary}; \textsc{binary} for the $\log(1/(2.4\delta))$ form & Prop.~\ref{prop:cl-topm} \\
Top-$m$ characteristic time \cite{you2022informationdirected,reda2021dealing} & fixed confidence, unique top-$m$ set & \textsc{process}, \textsc{uniform}, \textsc{boundary}, \textsc{allocation} & Prop.~\ref{prop:cl-topm-exact} \\
Identification at the algorithm's own error rates & one-parameter family, any stopping rule; best-arm or top-$m$ & none for the identity; \textsc{process}, \textsc{allocation} for the bound & Prop.~\ref{prop:cl-own} \\
Characteristic time and optimal weights \cite{garivier2016optimal} & one-parameter family & none (exact) & Prop.~\ref{prop:cl-weights} \\
Identification within a tolerance \cite[Theorem~5]{mannor2004sample}; second term of \cite[Remark~5]{kaufmann2016journal} & fixed confidence, tolerance $\Delta_{\mathrm{tol}}$ & \textsc{process}, \textsc{binary}, \textsc{uniform}; for the $1/\Delta_a^2$ form, an upper bound on each per-pull divergence, which no tag names & Prop.~\ref{prop:cl-tolerance} \\
Two-point fixed-budget bound \cite[Lemma~15]{kaufmann2016journal} & fixed budget, two instances & \textsc{process}, \textsc{test}; \textsc{allocation} for the form in $\max_a c_a(1/2)$ & Prop.~\ref{prop:cl-fb-twopoint} \\
Gaussian fixed-budget bounds, corrected \cite[Theorems~16 and~17]{kaufmann2016journal} & fixed budget, Gaussian arms, common variance & \textsc{process}, \textsc{test}, \textsc{boundary}, \textsc{allocation} & Prop.~\ref{prop:cl-fb-gauss} \\
Fixed-budget exponent \cite[Theorem~12]{kaufmann2016journal} and pairwise rate \cite[Section~2]{glynn2004large} & fixed budget, error probability tending to $0$ & \textsc{process}, \textsc{binary}, \textsc{uniform}, \textsc{asymptotic}, \textsc{boundary}, \textsc{allocation}; the rate: none (identity) & Prop.~\ref{prop:cl-fb-rate} \\
Regret in ergodic MDPs \cite{burnetas1997optimal,ok2018exploration} & ergodic MDP, uniformly good algorithms & \textsc{process}, \textsc{binary}, \textsc{markov}, \textsc{uniform}, \textsc{asymptotic}, \textsc{allocation} & Prop.~\ref{prop:cl-mdp} \\
Decoupled regret constant \cite{burnetas1997optimal,ok2018exploration} & ergodic MDP, product class & none (exact) & Prop.~\ref{prop:cl-mdp-decoupled} \\
Structured bandits \cite{combes2017minimal}; Graves--Lai with independent observations \cite{graves1997asymptotically} & known class of instances, consistent algorithms & as for regret in ergodic MDPs & Cor.~\ref{cor:cl-structured} \\
Linear bandits \cite{tlattimore2017end} & unit-variance Gaussian arms with linear means, consistent algorithms & as for structured bandits & Cor.~\ref{cor:cl-linear} \\
Lipschitz bandits \cite{magureanu2014lipschitz} & Bernoulli arms with Lipschitz means, consistent algorithms & as for structured bandits & Cor.~\ref{cor:cl-lipschitz} \\
\end{longtable}
}

Appendix~\ref{app:classical} states each result in this paper's notation.
Appendix~\ref{app:proofs-classical} proves the statements of Appendix~\ref{app:classical} and names the defect at every step.
Every step except \textsc{asymptotic} can be evaluated for a given algorithm on a given instance, \textsc{uniform} when the class guarantee is a number, such as $\delta$ or a nominal level.
At a finite horizon the difference between an algorithm's cost and a classical limit is a signed sum of the defects of its chain.
Under an asymptotic guarantee, such as consistency, \textsc{uniform} and \textsc{asymptotic} enter this sum only together, as what the other defects leave of the difference.
Most regret rows turn counts into event probabilities by \textsc{markov}.
The count bounds of Garivier, M\'enard and Stoltz and the explore-then-commit bounds work with a normalized count or the exact error sum instead.
The lower bounds for best-arm identification, sequential design and Markov decision processes replace the algorithm's allocation by the best one (\textsc{allocation}).

The chain does not recover every lower bound in the literature.
The iterated-logarithm term in the expected stopping time of power-one tests \cite[Theorem~3.2]{agrawal2025stopping} is one.
Its proof divides one null's error budget among many alternatives at once, while the chain proves a bound for each pair $(\nu, \lambda)$ and then optimizes over the alternative.
Minimax bounds by the methods of Fano and Assouad \cite{chen2024assouad} are another.
They bound the worst case over a prior or a packing of many instances.
Fano's method processes the history to a decision among many hypotheses, where the chain processes it to an event that separates two instances.
The testing step at the Bhattacharyya point does not reach the converse of Chernoff's exponent \cite{chernoff1952measure,polyanskiywu2025information}.
For a fixed design the bound of Theorem~\ref{thm:limit} is of order $e^{-2C_T(1/2)}$, while the minimum error sum is $e^{-\max_s C_T(s) + O(\sqrt{T})}$, and $2C_T(1/2)$ lies between $\max_s C_T(s)$ and twice it.
The coincidence ledger at the maximizing tilt reaches the converse when the pull counts are deterministic (Lemma~\ref{lem:chernoff-point}).
The $1/\Delta_a^2$ forms of the bounds of Mannor and Tsitsiklis need an upper bound on a per-pull divergence, which none of the relaxations supplies (Proposition~\ref{prop:cl-tolerance}).
The coincidence family does not sharpen the fixed-budget exponent of a single instance (Proposition~\ref{prop:cl-fb-rate}).
Four further results are recovered only in part.
The bound of Graves and Lai for controlled Markov chains \cite{graves1997asymptotically} is recovered only for independent observations (Appendix~\ref{app:cl-structured}).
Of Lorden's optimality of a matrix of one-sided sequential probability ratio tests \cite{lorden1977nearlyoptimal} only the first-order lower bound is recovered (Appendix~\ref{app:cl-composite}).
Of the bound in Remark~5 of Kaufmann, Capp\'e and Garivier \cite{kaufmann2016journal} only the second term is reached (Appendix~\ref{app:cl-tolerance}).
The optimality of Wald and Wolfowitz \cite{waldwolfowitz1948optimum} follows from Proposition~\ref{prop:cl-ww} only when its defect is zero.

\section{Evaluation on realistic instances}
\label{sec:eval}

The instances below come from published sources: movie ratings, the accuracies of trained models, an oat variety trial, an e-mail marketing experiment, the items of a national science assessment, a textbook maintenance problem and an engine-replacement model fitted to fleet data.
On each, standard algorithms run against the classical limit of their problem.
For regret on movie ratings and on the two Markov decision processes, for identification, for the sequential tests and for sequential design, the ledger splits the difference into terms named by the relaxations of Section~\ref{sec:classical}, each computed from the algorithm's own runs.
Where the class guarantee is a number, as for $\delta$-correct identification and tests at nominal levels, the \textsc{uniform} term is computed by a change of measure or by an exact program.
For identification no change-of-measure estimate passed an exact control (Appendix~\ref{app:eval-own-error}).
The \textsc{uniform} and \textsc{process} terms are split only for best-arm identification among trained models, with the measured bias stated, and are reported together on the other identification instances.
For regret the guarantee is asymptotic, so \textsc{uniform} and \textsc{asymptotic} enter the split together, as what the other terms leave of the difference.
Appendix~\ref{app:eval} gives the protocols.

\subsection{Regret on movie ratings}
\label{sec:eval-regret}

\begin{figure}[t]
\centering
\includegraphics[width=\linewidth]{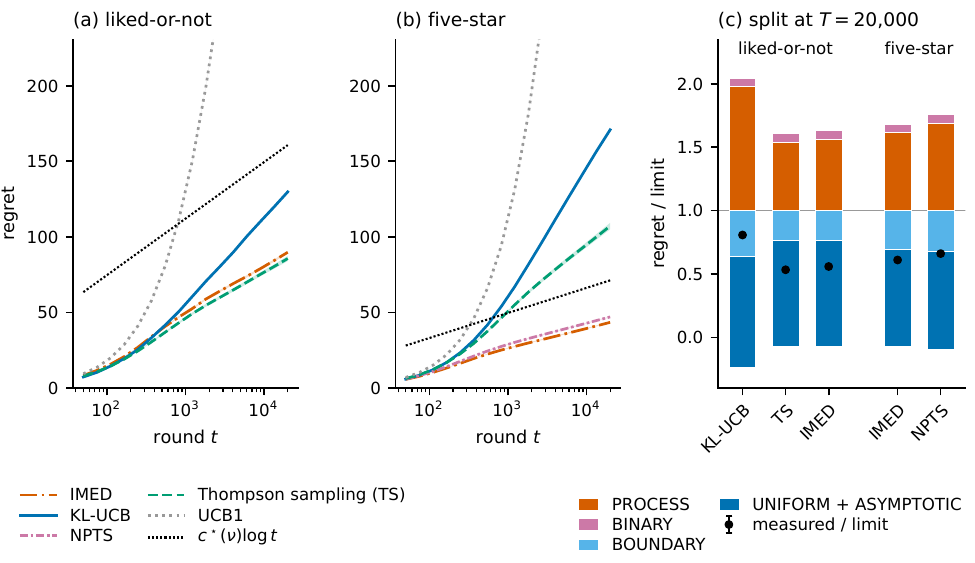}
\caption{\textbf{On MovieLens arms, the asymptotic regret limit overstates what asymptotically optimal algorithms need at $T = 20{,}000$.} (a, b) Regret against $c^\star(\nu)\log t$, for the liked-or-not and the five-star instances. The $95\%$ intervals have half-widths of at most $2\%$ of the regret. UCB1 leaves both panels; it reaches $643$ and $746$ at $T = 20{,}000$. On the five-star instance KL-UCB and Thompson sampling treat each arm as Bernoulli, so they are not asymptotically optimal there. They incur $1.5$ to $2.4$ times the limit at $T = 20{,}000$. (c) The regret of each asymptotically optimal algorithm relative to the limit, split along the chain of Proposition~\ref{prop:cl-lai-robbins} with the normalized count and the closest alternatives. Terms above the line at $1$ add to the regret ratio and terms below it subtract. Markers show the measured regret ratio, which the line at $1$ plus the four signed terms reproduce exactly; its $95\%$ interval is narrower than the marker.}
\label{fig:regret}
\end{figure}

The arms are the $10$ most-rated movies of the MovieLens 100K ratings \cite{harper2015movielens}.
On the liked-or-not instance the reward of a pull is $1$ when the rating is at least $4$.
On the five-star instance it is the rating rescaled to $[0,1]$, drawn from the movie's empirical rating distribution.
The algorithms are UCB1, KL-UCB, Thompson sampling and IMED, with NPTS on the five-star instance (Appendix~\ref{app:eval-regret}).
Figure~\ref{fig:regret} compares them with the limit $c^\star(\nu)\log t$, where $c^\star(\nu) = \sum_b \Delta_b / \Kinf(\nu_b, \mu^\star)$.
At $T = 20{,}000$ the asymptotically optimal algorithms incur $53\%$ to $81\%$ of $c^\star(\nu)\log T$ on the liked-or-not instance, and IMED and NPTS $61\%$ to $66\%$ on the five-star instance.
The ledger splits each of these ratios exactly along the chain of Proposition~\ref{prop:cl-lai-robbins}, with the normalized count $N_b(T)/T$ as the statistic (Figure~\ref{fig:regret}(c)).
With the closest alternatives of Appendix~\ref{app:eval-regret}, which raise arm $b$ to $\mu^\star + 0.02$, the \textsc{process} term adds $0.54$ to $0.98$ of the limit.
The \textsc{binary} term adds $0.06$ to $0.07$, just under the $\log 2 / \log T = 0.07$ of Lemma~\ref{lem:relax}(c).
The \textsc{boundary} term subtracts $0.24$ to $0.36$.
Consistency is an asymptotic guarantee, so \textsc{uniform} and \textsc{asymptotic} enter only together.
Their joint term subtracts $0.77$ to $0.87$, more than the regret ratio falls short of one: the asymptotic requirement overstates the finite-horizon one.
Regret below the asymptotic limit at practical horizons is reported for KL-UCB \cite{cappeGarivierMaillardMunosStoltz2013klucb} and attributed to second-order terms \cite[Section~1 and Theorem~5]{garivier2019mathematics}.
With farther alternatives the \textsc{process} and \textsc{boundary} terms both grow, by nearly equal amounts, and the joint term subtracts $0.55$ to $0.75$.
On the five-star instance, KL-UCB and Thompson sampling treat each arm as Bernoulli and incur $1.5$ to $2.4$ times the limit.
Their asymptotic constant $\sum_b \Delta_b/\kl(\mu_b, \mu^\star)$ uses the Bernoulli divergence where the limit uses $\Kinf(\nu_b, \mu^\star)$.
On this instance it is $3.1$ times $c^\star(\nu)$.
The data-processing step also reads as a pull count.
For each suboptimal arm, the fundamental inequality with the normalized count certifies a pull count from the algorithm's own behavior at the alternatives.
For the asymptotically optimal algorithms it certifies $16\%$ to $37\%$ of the pulls actually made.
On the exact one-armed reductions of Section~\ref{sec:eval-factor-two}, the same computation certifies $37\%$ to $67\%$, so the shortfall belongs to the statistic.
By Lemma~\ref{lem:relax}(b), it is the conditional divergence of the history given a coin whose mean is the normalized count.

On the exact reductions the law of $N_b(T)$ recovers most of that conditional divergence.
At $T = 3{,}000$, with the alternative that raises arm $b$ to $\mu^\star + 0.01$, it keeps at least $82\%$ of $D_T$ (Figure~\ref{fig:regret-statistics}, Appendix~\ref{app:eval-regret-statistics}).
The finite-horizon regret bound of Proposition~\ref{prop:cl-finite} certifies less.
Its regret form, which uses only $R_T(\nu)$ and $R_T(\lambda)$, is negative there for every arm and both algorithms.
That form is positive only when $R_T(\nu) + R_T(\lambda) < T\underline{\Delta}/4$.
The algorithms' own regrets sum to more on every arm.
Under KL-UCB the \textsc{markov} step replaces the probability of $\{N_b(T) > T/2\}$ by $2\mathbb{E}_\nu[N_b(T)]/T$.
With the replacement of both gaps by $\underline{\Delta}$, it removes more than the error-sum form keeps.
Under Thompson sampling the error-sum form is already small, because under $\lambda$ the algorithm still pulls arm $b$ at most $T/2$ times with probability $0.354$.
Over all margins the regret form certifies at most $7\%$ of KL-UCB's pulls of arm $b$ and $27\%$ of Thompson sampling's.
On Gaussian reductions with the same standardized gaps, Corollary~\ref{cor:cl-gauss} certifies at most $13.1\%$ of the measured regret.

\subsection{The factor of two under adaptive play}
\label{sec:eval-factor-two}

\begin{figure}[t]
\centering
\includegraphics[width=\linewidth]{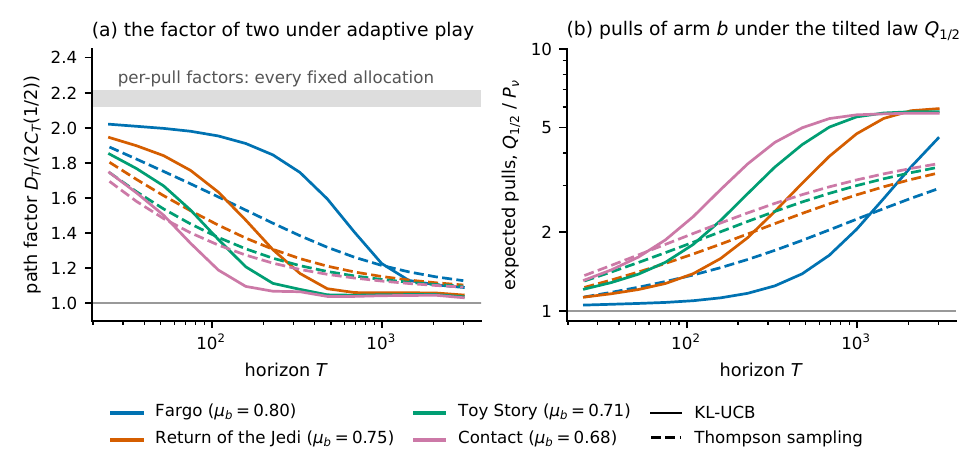}
\caption{\textbf{Under adaptive play the path factor falls from the per-pull factor toward one as the horizon grows.} (a) The path factor $D_T/(2C_T(1/2))$ of KL-UCB and Thompson sampling on four one-armed reductions of MovieLens arms, $\xi = 0.01$. The gray band marks the per-pull factors, which every fixed allocation attains. (b) At the same $\xi$, the expected pulls of the perturbed arm under the tilted law $Q_{1/2}$, relative to $P_\nu$. At $T = 3{,}000$ this ratio lies between $2.9$ and $5.9$. All values are exact.}
\label{fig:factor-two}
\end{figure}

Proposition~\ref{prop:half} writes the gain of Theorem~\ref{thm:limit} over the Bretagnolle--Huber bound as a Jensen defect of the log-likelihood ratio.
Figure~\ref{fig:factor-two} computes it exactly on one-armed reductions of MovieLens ratings \cite{harper2015movielens}.
The best arm $a^\star$ has like-rate $0.859$ and is known.
The perturbed arm $b$ is one of the next four movies.
The alternative $\lambda$ raises arm $b$ to $0.859 + \xi$, with margin $\xi \in \{0.01, 0.03\}$.
Every fixed allocation attains the per-pull factor, between $2.12$ and $2.26$ on these arms.
Under KL-UCB and Thompson sampling the path factor falls toward $1$ as the horizon grows.
At $T = 3{,}000$ it lies between $1.03$ and $1.40$ over four arms, two margins and both algorithms.
The same computation gives the allocation under the tilted law $Q_{1/2}$ of Proposition~\ref{prop:tilt}.
At $T = 3{,}000$ and $\xi = 0.01$ the tilted law pulls the perturbed arm $2.9$ to $5.9$ times as often as $P_\nu$ does, and up to $11$ times at $\xi = 0.03$.
The per-pull and path factors on Gaussian reductions of an oat variety trial are given in Section~\ref{sec:eval-bai}.

\subsection{Identification on model accuracies and an oat variety trial}
\label{sec:eval-bai}

\begin{figure}[t]
\centering
\includegraphics[width=\linewidth]{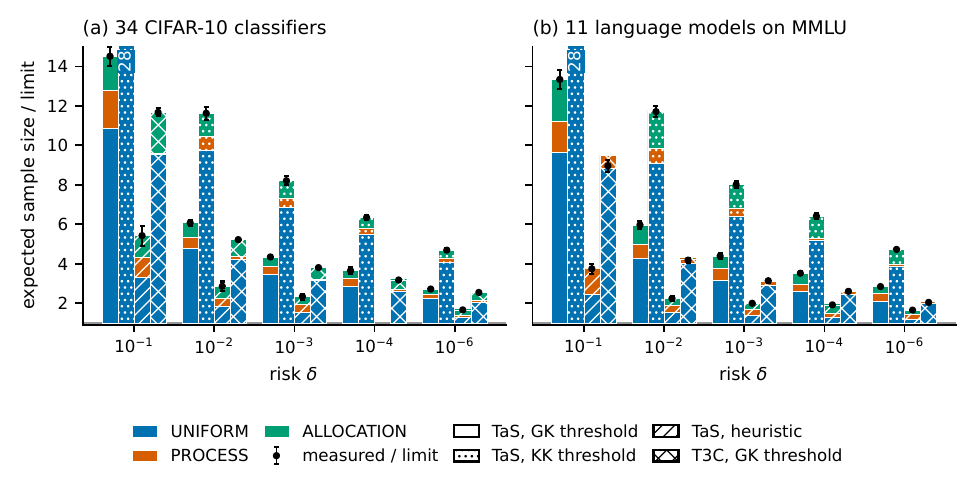}
\caption{\textbf{Among trained models, the largest term of the excess over the best-arm identification limit is the uniform quantification over the class, for every adaptive algorithm tested under a $\delta$-correct stopping threshold.} Each bar splits the expected sample size relative to $T^\star(\nu)\kl(\delta, 1-\delta)$ into the \textsc{uniform} term, the \textsc{process} term of data processing to the recommendation and the \textsc{allocation} term; markers show the measured ratio to the limit with a $95\%$ interval. The boundary between the \textsc{uniform} and \textsc{process} terms in each bar is set by an estimate whose logarithm is biased low (Section~\ref{sec:eval-bai}). Hatching marks the algorithm and its stopping threshold: TaS is Track-and-Stop, GK the threshold of Garivier and Kaufmann and KK that of Kaufmann and Koolen. Bars that leave the axis show their value in white. Cross-fitting, in which the most confusing alternative is chosen on each half of the runs and evaluated on the other half, makes the \textsc{allocation} estimate run low where two alternatives are about equally confusing. The \textsc{uniform} and \textsc{process} terms together then run high by the same amount. A negative estimate is drawn as zero. Round-robin sampling reaches $11$ to $88$ times the limit and is omitted. Track-and-Stop with the heuristic threshold at $\delta = 10^{-4}$ on the CIFAR-10 instance is omitted too: one of its $400$ runs had not stopped after $237\,T^\star(\nu)$ pulls.}
\label{fig:bai}
\end{figure}

A pull scores one trained model on one test item drawn uniformly.
The first instance takes $34$ image classifiers evaluated on the reproduced CIFAR-10 test set \cite{recht2019imagenet}.
The second takes $11$ language models on MMLU \cite{touvron2023llama}.
The algorithms are Track-and-Stop, T3C and round-robin sampling (Appendix~\ref{app:eval-bai}).
Each algorithm stops by the generalized likelihood ratio rule.
Figure~\ref{fig:bai} splits its expected sample size into the limit $T^\star(\nu) \kl(\delta, 1-\delta)$ \cite{garivier2016optimal} and three terms.
The \textsc{uniform} term compares the algorithm's own error probability at the most confusing alternative with the guarantee $\delta$ of its class.
For Track-and-Stop and T3C with a $\delta$-correct threshold it is the largest term at every $\delta$.
It is $13$ to $17$ nats under the threshold of Garivier and Kaufmann and $36$ to $42$ nats under that of Kaufmann and Koolen.
At $\delta = 0.1$ these two algorithms draw $9$ to $15$ times the limit under the first threshold.
A common heuristic threshold cuts the term to $2.4$ to $4.1$ nats.
It is not proved $\delta$-correct.
On these instances its error rates stay below $\delta$, as in earlier experiments \cite{garivier2016optimal,kaufmann2021mixture}.
The \textsc{process} term, of data processing to the recommendation, is at most $5.7$ nats.
The split between the \textsc{uniform} and \textsc{process} terms is set by the logarithm of a change-of-measure estimate of the error probability at the most confusing alternative.
That logarithm is biased low.
In an exact two-arm control with an under-sampled challenger, at the same thresholds and risks, the bias is $0.07$ to $1.98$ nats (Appendix~\ref{app:eval-own-error}).
With a bias of that size, the drawn \textsc{uniform} term is too large, and the \textsc{process} term too small, by the same number of nats.
The \textsc{allocation} term is at most $10.6$ nats for Track-and-Stop and T3C.
For round-robin sampling it is $75$ to $215$ nats.

Before the \textsc{uniform} step, the chain gives the bound $\mathbb{E}_\nu[\tau] \ge T^\star(\nu)\,\kl(P_\nu(E), P_{\lambda^\star}(E))$ at the algorithm's own error rates (Proposition~\ref{prop:cl-own}).
Under $\lambda^\star$ a run almost never stops, so $P_{\lambda^\star}(E)$ is estimated by a change of measure from the runs under $\nu$.
That estimate was tested against an exact control before any reading (Appendix~\ref{app:eval-own-error}).
No estimate fixed before the runs met its tolerance in every combination of reduction, threshold and confidence level.
The bound at the algorithms' own error rates is therefore not reported for identification.

\begin{figure}[t]
\centering
\includegraphics[width=\linewidth]{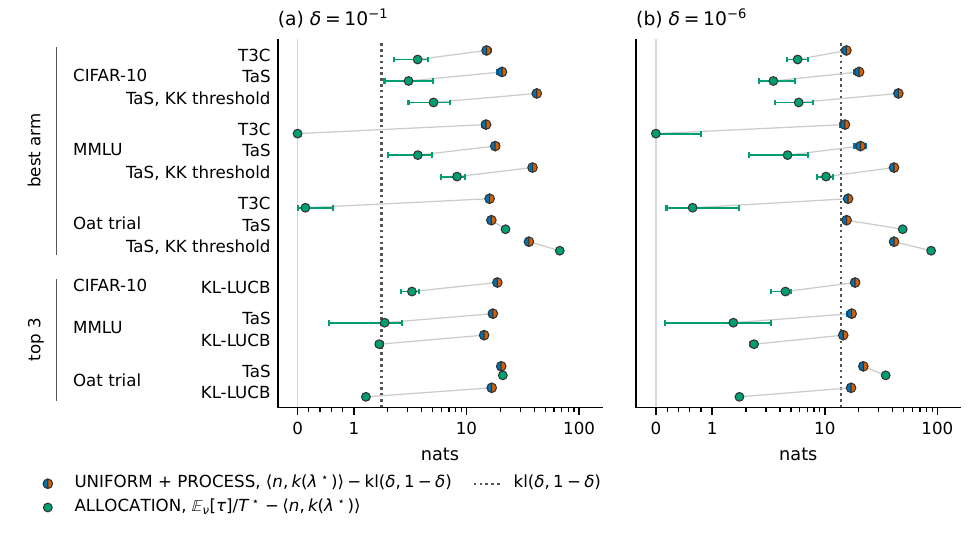}
\caption{\textbf{From $\delta = 10^{-1}$ to $10^{-6}$ the information that the identification algorithms collect beyond $\kl(\delta, 1-\delta)$ changes little, while $\kl(\delta, 1-\delta)$ itself grows nearly $8$-fold.} In each row the excess of the expected sample size over the limit, $\mathbb{E}_\nu[\tau]/T^\star(\nu) - \kl(\delta, 1-\delta)$, is split into two terms in nats. The \textsc{uniform} and \textsc{process} terms together are $\langle n, k(\lambda^\star)\rangle - \kl(\delta, 1-\delta)$, the information collected against the most confusing alternative $\lambda^\star$ beyond $\kl(\delta, 1-\delta)$ (two-colored markers). The \textsc{allocation} term is $\mathbb{E}_\nu[\tau]/T^\star(\nu) - \langle n, k(\lambda^\star)\rangle$ (green markers). The value of $\kl(\delta, 1-\delta)$ is marked by the dotted lines. The best-arm rows are the CIFAR-10 and MMLU instances of Figure~\ref{fig:bai} and the oat variety trial of Appendix~\ref{app:eval-field-trial}. The top-$3$ rows use $T^\star_m(\nu)$ with $m = 3$ (Appendices~\ref{app:eval-topm} and~\ref{app:eval-field-trial}). TaS is Track-and-Stop. KK marks the threshold of Kaufmann and Koolen. Every other row uses the first $\delta$-correct threshold: that of Garivier and Kaufmann on CIFAR-10 and MMLU, with a union over the swap pairs for the top~$3$, and a Gaussian calibration of Kaufmann and Koolen on the oat trial. Under these thresholds the \textsc{uniform}-plus-\textsc{process} term is $14$ to $22$ nats at both risks, and under KK it is $36$ to $45$ nats. Between the two risks it changes by at most $5.4$ nats, while $\kl(\delta, 1-\delta)$ grows from $1.8$ to $13.8$ nats. The \textsc{allocation} term is at most $10.3$ nats, below every \textsc{uniform}-plus-\textsc{process} term, except for Track-and-Stop on the oat trial, where it is $21$ to $88$ nats. The alternative $\lambda^\star$ is chosen on one half of the runs and evaluated on the other. The \textsc{allocation} term is nonnegative. Its cross-fitted estimate runs low where two alternatives are about equally confusing. The \textsc{uniform}-plus-\textsc{process} estimate then runs high by the same amount. For T3C on MMLU the \textsc{allocation} estimate is $-0.9$ and $-0.8$ nats at the two risks, against $0.03$ and $0.17$ nats when $\lambda^\star$ is chosen and evaluated on all runs. A negative estimate is drawn as zero, as in Figure~\ref{fig:bai}, and every interval is cut at zero. The axis is linear between $0$ and $1$ nat and logarithmic above. Error bars are $95\%$ percentile bootstrap intervals, with the runs resampled within each half and $\lambda^\star$ chosen again in every resample; where none shows, the interval is narrower than the marker. Round-robin sampling, whose \textsc{allocation} term is $58$ to $249$ nats, is omitted. So is Track-and-Stop on the CIFAR-10 top-$3$ problem, where $7$ of its $200$ runs at each risk reached the step cap.}
\label{fig:ident-tightness}
\end{figure}

The same models identify a top-$m$ set (Appendix~\ref{app:eval-topm}).
The limit is $T^\star_m(\nu)\kl(\delta, 1-\delta)$ of Proposition~\ref{prop:cl-topm-exact}.
Each characteristic time is certified by a duality bracket, an upper bound from an allocation and a lower bound from the dual program.
The explicit bound of Proposition~\ref{prop:cl-topm}, which uses single-arm alternatives, lies $1.65$ to $1.72$ times below the limit.
On the MMLU instance Track-and-Stop, KL-LUCB and LUCB draw $2.2$ to $2.4$ times the limit at $\delta = 10^{-6}$ under the union threshold of Garivier and Kaufmann (Figure~\ref{fig:topm}).
Most of their excess is the information collected at the most confusing alternative beyond $\kl(\delta, 1-\delta)$, which is $13$ to $17$ nats at every risk (Appendix~\ref{app:eval-topm}).
That information is drawn in Figure~\ref{fig:ident-tightness} for Track-and-Stop and KL-LUCB with $m = 3$, at the largest and the smallest risk.
The excess of round-robin sampling is mostly its \textsc{allocation} term.
On the CIFAR-10 instance some runs of Track-and-Stop at each risk reach the step cap, so its cost there is a lower bound, while KL-LUCB stops every run.
The cause is a pair of models tied just below the top-$3$ boundary, on which almost all the weight at the empirical means is placed.
Neither breaking the tie nor recomputing the weights more often removes the deficit in the allocation (Appendix~\ref{app:eval-topm}).

The 2024 oat variety trial at Minot gives a Gaussian instance with $22$ arms, one variety per arm and one plot per pull \cite{bortolon20242024}.
Its characteristic time is $T^\star(\nu) = 48.2$ plots, from the one-variable program of Proposition~\ref{prop:scalar}.
Under the first $\delta$-correct threshold, T3C and KL-LUCB collect $15$ to $17$ nats at $\lambda^\star$ beyond $\kl(\delta, 1-\delta)$ (Figure~\ref{fig:ident-tightness}).
That information is most of their excess over the limit.
Their average allocations have characteristic times within $4\%$ and $17\%$ of the optimum.
For Track-and-Stop in best-arm identification the largest term is \textsc{allocation}, at $22$ to $49$ nats.
It exceeds the sum of the other two terms, so its rank needs no estimate of $P_{\lambda^\star}(E)$.
The forced exploration of Track-and-Stop holds $19$ or $20$ of the $22$ varieties at $\sqrt t - K/2$ pulls.
On this instance that floor takes $20\%$ to $35\%$ of the pulls, against $0.8\%$ to $3.6\%$ on the two instances above, whose characteristic times are $72$ and $116$ times larger.
The per-pull factor of Proposition~\ref{prop:ratio} is $2$.
At $T = 3{,}000$ the path factor of KL-UCB and Thompson sampling is $1.03$ to $1.18$ on $15$ of $16$ one-armed reductions.
On the remaining one, KL-UCB on the variety closest to the best at a margin of $0.3$ standard deviations, it is $1.44$ (Appendix~\ref{app:eval-field-trial}).

\subsection{Sequential A/B tests}
\label{sec:eval-ab}

\begin{figure}[tp]
\centering
\includegraphics[width=\linewidth]{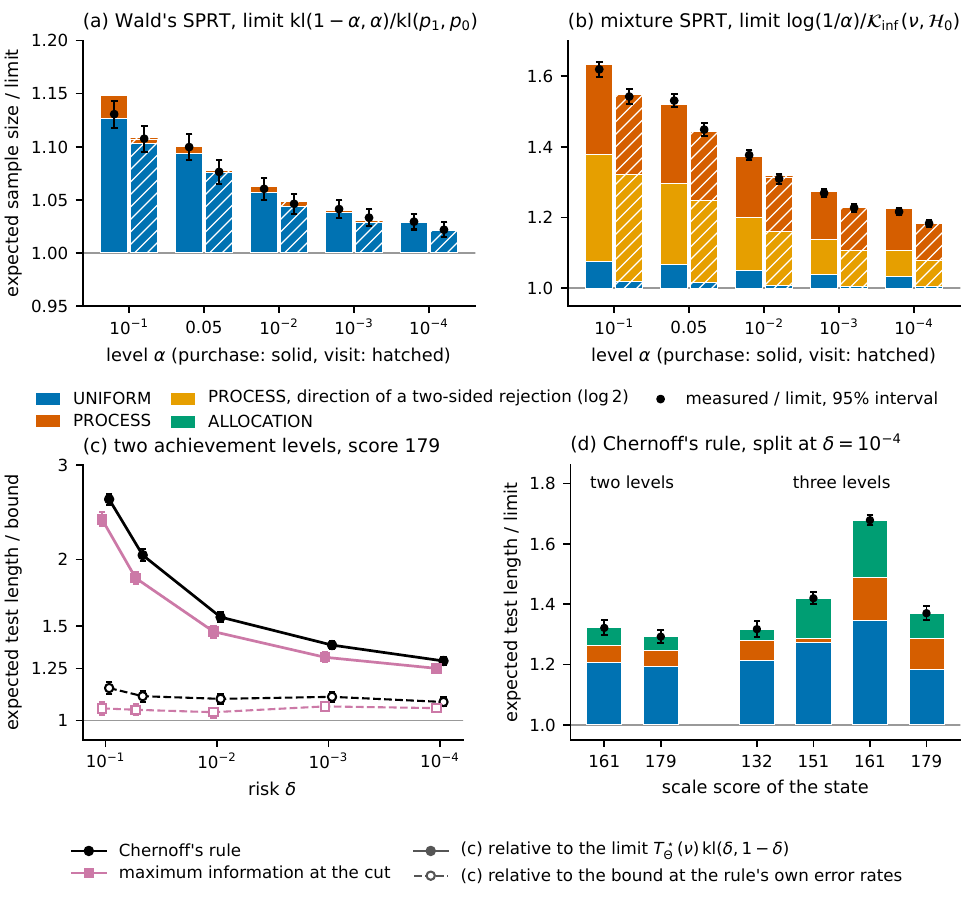}
\caption{\textbf{Both the sequential tests and Chernoff's rule run above their classical limits, by $2\%$ to $62\%$ for the tests and by $1.3$ to $2.6$ times for Chernoff's rule next to the Proficient cut.} (a, b) The sequential A/B tests on both metrics of the Hillstrom experiment. In each bar the expected sample size relative to the limit is split into the \textsc{uniform} term and the \textsc{process} term, computed from $\mathbb{E}_\nu[L_\tau]$. For the mixture test the $\log 2$ of its two-sided decision is drawn apart from the rest of the \textsc{process} term, and its \textsc{uniform} term uses the error rate at the barycentre null of Appendix~\ref{app:eval-ab}. For Wald's SPRT the \textsc{uniform} term is the largest part of the excess at every $\alpha$ and on both metrics. For the mixture SPRT most of the excess is the \textsc{process} term. Markers show the measured ratio to the limit, computed from $\mathbb{E}_\nu[\tau]$, with a $95\%$ interval from the normal approximation. The two agree in expectation by Theorem~\ref{thm:allocation}, and in the data their largest difference is $1.9$ times its standard error. Solid bars show the purchase rate and hatched bars the site-visit rate. Panels (a) and (b) have different vertical scales. (c, d) Sequential design on NAEP science items (Section~\ref{sec:eval-design}), with every rule at the $\delta$-correct threshold. (c) Two achievement levels, the state just above the Proficient cut (scale score $179$): the expected test length relative to the limit $T^\star_\Theta(\nu)\,\kl(\delta, 1-\delta)$ (solid) and relative to the bound at the rule's own error rates (dashed), for Chernoff's rule and maximum Fisher information at the cut. Chernoff's rule sits at most $15\%$ above that bound at every risk. (d) At $\delta = 10^{-4}$, Chernoff's expected test length relative to the limit at the six states next to a cut, split into the \textsc{uniform}, \textsc{process} and \textsc{allocation} terms. The \textsc{uniform} term is the largest of the three at every state. At the three-level state of score $179$ the split rests on change-of-measure weights with effective sample sizes of $31$ and $14$ in the two halves of the runs. Error bars in (c, d) are $95\%$ bootstrap intervals over runs; where none shows, the interval is narrower than the marker.}
\label{fig:ab}
\end{figure}

In the Hillstrom e-mail experiment, customers who received an e-mail featuring men's merchandise bought at a rate of $1.25\%$ against $0.57\%$ in the control group, and visited the site at $18.28\%$ against $10.62\%$ \cite{radcliffe2008hillstroms}.
Each metric has a treatment rate $p_1$ and a control rate $p_0$.
The nominal error probabilities of a sequential test are written $\alpha$ (type I) and $\beta$ (type II), and its realized ones $\alpha'$ and $\beta'$ (Appendix~\ref{app:cl-wald}).
Two sequential tests detect each effect (Figure~\ref{fig:ab}(a, b)).
Wald's SPRT observes the treatment stream and tests its rate against the known control rate.
Its expected sample size is $2\%$ to $14\%$ above Wald's bound $\kl(1-\alpha, \alpha)/\kl(p_1, p_0)$ (Proposition~\ref{prop:cl-wald}) at levels from $10^{-1}$ to $10^{-4}$.
Most of the excess is the \textsc{uniform} term, the difference between the relative entropy at the realized error rates and at the nominal levels.
Wald's thresholds put the type-I error at $0.77$ to $0.82$ of its nominal level and the type-II error at $0.96$ to $1.02$ of its nominal level.
The \textsc{process} term, of data processing to the decision, is at most $0.04$ nats.
The mixture SPRT \cite{johari2017peeking} observes pairs and tests whether the two rates are equal.
Its limit is $\log(1/\alpha)/\Kinf(\nu, \mathcal{H}_0)$ \cite[Theorem~3.1]{agrawal2025stopping}, where $\mathcal{H}_0$ is the null of equal rates and $\Kinf(\nu, \mathcal{H}_0)$ is the per-pair relative entropy from the truth to the null, attained at the barycentre null of Proposition~\ref{prop:mbary}.
Proposition~\ref{prop:cl-power-one} writes the chain of this limit with the defect of each step.
The test draws $18\%$ to $62\%$ more pairs than this limit.
Three parts account for the excess.
A two-sided test spends half its error budget on rejections in the direction the truth never takes, which costs $\log 2$ nats.
In the terms of Section~\ref{sec:classical} this is a \textsc{process} defect: the decision to reject discards the direction of the rejection.
Proposition~\ref{prop:cl-power-one}(ii) gives this defect as the relative entropy between the laws of the direction of rejection under the truth and under the null given a rejection.
By that proposition it equals $\log 2$ when the null rejects in the two directions equally often and the truth never rejects in the other direction.
Here $3$ of the $100{,}000$ runs under the truth reject in the other direction.
The estimated defect lies within $0.002$ nats of $\log 2$ on both metrics and at every level.
The test's error rate at the barycentre null is $0.74$ to $0.96$ times $\alpha$.
Processing the history to a rejection in the truth's direction loses a further $0.52$ to $1.07$ nats.

At the tests' own error rates the bound comes within $10\%$ of the measured cost.
Wald's SPRT is computed exactly, by a program over the success count.
At its realized error rates $\alpha'$ and $\beta'$, its expected sample size exceeds $\kl(1-\beta', \alpha')/\kl(p_1, p_0)$ by $0.1\%$ to $0.8\%$ on both metrics and at every level.
For the mixture SPRT, $\alpha'_{+,\mathrm{cap}}$ is simulated under the barycentre null with $10^4$ to $2 \times 10^6$ runs, each capped at $10$ times the limit plus $100$ pairs (Appendix~\ref{app:eval-own-error}).
At that rate the test draws $4.7\%$ to $9.1\%$ more pairs than $\kl(P_\nu(E_+), \alpha'_{+,\mathrm{cap}})/\Kinf(\nu, \mathcal{H}_0)$ at every level and metric with a direct simulation, against $18\%$ to $62\%$ more than its limit.
The upper ends of the $95\%$ intervals are at most $1.13$ times these bounds.

\subsection{Sequential design on test items}
\label{sec:eval-design}

An adaptive test places a grade-8 student in an achievement level of the 2019 science assessment of the National Assessment of Educational Progress (NAEP), using the published parameters of its $268$ dichotomous items \cite{statistics2019irt}.
The items are the experiments of Chernoff's setting.
Abilities on a grid are its states (Appendix~\ref{app:cl-chernoff}).
One design separates Proficient students from the rest, and a second places students below Basic, at Basic, or at Proficient and above.
Chernoff's rule, maximum Fisher information at the estimated ability and at the nearest cut, and random selection run with a $\delta$-correct likelihood-ratio stopping rule (Appendix~\ref{app:eval-design}).
Figure~\ref{fig:ab}(c, d) compares their expected test lengths with the limit $T^\star_\Theta(\nu)\,\kl(\delta, 1-\delta)$ of Proposition~\ref{prop:cl-design}.
At $\delta = 10^{-4}$ Chernoff's rule needs $1.29$ to $1.68$ times the limit at the six states next to a cut.
At $\delta = 0.1$ it needs $2.59$ to $3.98$ times the limit.
In both designs the largest part of the difference is the \textsc{uniform} term.
At the three-level state of score $179$ its split rests on change-of-measure weights with small effective sample sizes (Figure~\ref{fig:ab}(d)).
The stopping threshold holds the error probability at the most confusing alternative to at most a fifth of $\delta$.
The bound at the rule's own error rates is therefore much closer.
In the two-level design Chernoff's rule sits $8\%$ to $15\%$ above it at every risk.
Maximum information at the cut sits $1.5\%$ to $6.2\%$ above its bound.
An exact computation of that rule's law agrees.
In the three-level design Chernoff's rule sits $9\%$ to $31\%$ above the bound, with the largest ratios at the Basic state where the maximin allocation mixes two items.
At the two Basic states maximum information at the cut sits $33\%$ to $74\%$ above its bound, because its draws follow the cut nearest the estimate.
A check of the change of measure against direct simulation failed at one of $72$ configurations.
A repeat with $10$ times as many runs agrees at all $72$ (Appendix~\ref{app:eval-design}).
The \textsc{allocation} term measures how far a rule's draws stray from the maximin allocation.
In the two-level design the most informative item at the cut is the maximin item at both states next to the cut.
The cut rule therefore loses nothing there, while Chernoff's rule loses information to its early wrong estimates.
At the Basic state where the maximin allocation mixes two items, Chernoff's rule needs $5.6$ to $25.7$ fewer items than the cut rule.

\subsection{Regret on two Markov decision processes}
\label{sec:eval-mdp-regret}

\begin{figure}[tp]
\centering
\includegraphics[width=\linewidth]{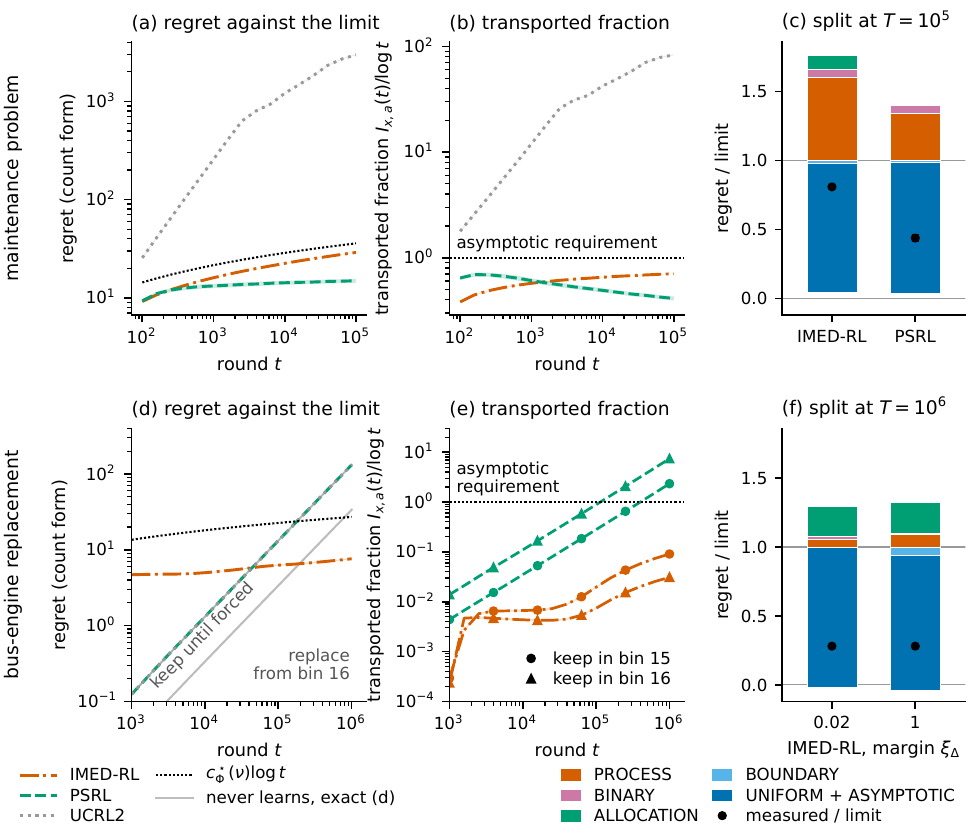}
\caption{\textbf{On both models the asymptotic regret limit overstates what IMED-RL needs at the horizons reached.} (a--c) The machine-maintenance problem of Section~\ref{sec:eval-mdp-regret}. (d--f) Rust's engine-replacement model, fitted to fleet data (Appendix~\ref{app:eval-bus-engine}). (a, d) Regret in count form, $\sum_{x,a} \mathbb{E}_\nu[N_{x,a}(t)]\,\Delta_{x,a}$, with $95\%$ intervals, against $c^\star_\Phi(\nu) \log t$. In (a) UCRL2 exceeds the limit by a factor of $83$ at $T = 10^5$. The gray lines in (d) are the exact regret of two policies that never learn. The one that replaces from bin $16$ crosses above the limit at $t = 7.9 \times 10^5$, and the one that keeps until forced, which PSRL and UCRL2 follow, at $t = 1.8 \times 10^5$. (b, e) The transported fraction $I_{x,a}(t)/\log t$ of each suboptimal pair with finite $\Kinf(x,a)$, with $95\%$ intervals: doing nothing at major deterioration in (b), and keeping the engine in bin $15$ or $16$ in (e), where UCRL2 coincides with PSRL. (c, f) The regret relative to the limit at the last horizon, split along the chain of Proposition~\ref{prop:cl-mdp} with the closest alternative. In (c) its margin is $\xi_\Delta = 0.02$ of the Bellman gap, and the statistic is the fraction of visits to major deterioration at which nothing is done. UCRL2 is omitted from (c): its ratio of $84$ there leaves the shared scale. The split in (c) comes from a further $1{,}000$ runs per algorithm, on which the regret ratio of PSRL is $0.44$, against $0.42$ in (a). In (f) the split is that of IMED-RL, at the margins $\xi_\Delta = 0.02$ and $1$. The statistic of each pair is the fraction of visits to its bin at which the engine is kept. Terms above the line at $1$ add to the ratio and terms below it subtract. Markers show the measured regret ratio, which the line at $1$ plus the five signed terms reproduce exactly, with its $95\%$ interval; where no bar shows, the interval is narrower than the marker. Panels (c) and (f) share one scale.}
\label{fig:mdp}
\end{figure}

The machine-maintenance example of Hillier and Lieberman \cite{hillier2001introduction} is an ergodic Markov decision process with four states.
Its limit is $c^\star_\Phi(\nu)\log t$, with $c^\star_\Phi(\nu)$ the constant of Proposition~\ref{prop:cl-mdp} for the class $\Phi$ of all transition laws \cite{burnetas1997optimal,ok2018exploration}.
In this class it equals the sum of $\Delta_{x,a}/\Kinf(x,a)$ over the suboptimal pairs, with $1/\infty := 0$ (Proposition~\ref{prop:cl-mdp-decoupled}).
Only one suboptimal state-action pair has finite $\Kinf(x,a)$: doing nothing when the machine shows major deterioration.
Replacing the machine in either deteriorated state already leads to the state of largest bias, so no transition law makes that action optimal.
Figure~\ref{fig:mdp}(a--c) follows three algorithms over $1{,}000$ runs of $10^5$ steps.
UCRL2 \cite{jaksch2010} spends about $83$ times the limit.
IMED-RL \cite{pesquerel2022imedrl} and PSRL \cite{osband2013more} stay below $c^\star_\Phi(\nu) \log t$ over the whole range.
Uniform goodness is an asymptotic guarantee, so \textsc{uniform} and \textsc{asymptotic} enter this difference only together.
At $T = 10^5$ the transported fraction of IMED-RL through that pair is $0.70$ and still rising.
That of PSRL rises until $t \approx 160$ and then falls, to $0.41$ at $T = 10^5$.
On a further $1{,}000$ runs per algorithm, the ledger splits each regret ratio exactly along the chain of Proposition~\ref{prop:cl-mdp} (Figure~\ref{fig:mdp}(c)).
The statistic is the fraction of visits to major deterioration at which nothing is done.
The closest alternative changes only the law of doing nothing there, so that doing nothing becomes optimal by a margin of $2\%$ of its Bellman gap $\Delta_{x,a}$.
The two pairs with infinite $\Kinf(x,a)$ enter as the \textsc{allocation} term.
At $T = 10^5$ the \textsc{process} term adds $0.60$ of the limit for IMED-RL and $0.34$ for PSRL.
The \textsc{allocation} term adds $0.10$ and $0.003$.
The joint \textsc{uniform}-and-\textsc{asymptotic} term subtracts $0.93$ and $0.95$, more than the regret ratio falls short of one.
Under that alternative the mean fraction of visits to major deterioration at which IMED-RL and PSRL overhaul is still $23\%$ and $30\%$.
UCRL2's excess sits in the \textsc{process} term, which adds $86$ times the limit.

The second instance is Rust's model of engine replacement for a fleet of city buses \cite{rust1987optimal}.
Its mileage law and costs are his estimates for one group of buses, as restated by Ferrall \cite{ferrall2023was}.
Mileage is grouped in $18$ bins of $25{,}000$ miles, numbered from $0$ (Appendix~\ref{app:eval-bus-engine}).
Replacement is forced in the top bin.
A small wear shock in the lowest bin makes every stationary policy irreducible without changing the law of any suboptimal pair.
The optimal policy replaces the engine from $375{,}000$ miles.
Among the suboptimal pairs, only keeping the engine in the two bins above that threshold has finite $\Kinf(x,a)$.
For the class of all transition laws, $c^\star_\Phi(\nu) = 1.965$.
Until $t = 7.9 \times 10^5$ a policy that never learns and replaces one bin late lies below $c^\star_\Phi(\nu)\log t$, so the comparison is made at $T = 10^6$.
There IMED-RL replaces from the optimal threshold in all $500$ runs and incurs $0.281$ of the limit (Figure~\ref{fig:mdp}(d--f)).
With the closest alternatives, the joint \textsc{uniform}-and-\textsc{asymptotic} term subtracts $1.015$ of the limit, more than the regret ratio falls short of one.
The \textsc{allocation} term adds $0.220$, most of it the single replacement IMED-RL tries in each low bin.
PSRL with the uniform prior and UCRL2 never replace before the forced bin and incur $4.87$ times the limit.
UCRL2's confidence radius makes keeping look better than replacing until about $10^9$ rounds.
PSRL never tries an early replacement, so its sampled law of replacing stays at the uniform prior over the $18$ bins.
With the laws of keeping of $\nu$, a replacement sampled from that prior beats keeping with probability $7.6 \times 10^{-28}$ at each planning step.
Under a lighter Dirichlet prior most runs of PSRL replace the engine before the forced bin at least once.
The regret barely falls, since few runs leave the line of keeping until forced (Appendix~\ref{app:eval-bus-engine}).

\section{Prior work}
\label{sec:prior}

The instance-dependent limit for bandits originates with Lai and Robbins \cite{lai1985advances},
whose asymptotic $\log T / \KL$ rate the chain of Section~\ref{sec:classical} reproduces at its relative-entropy endpoint.
The divergence decomposition in the form used here, and the two-point argument built on it, are standard \cite{lattimore2020bandit}.
In the classical proofs each later step is an inequality whose loss is not named.
A non-negative remainder is discarded by the data processing of \cite{garivier2019mathematics} and by the conditional Jensen step of \cite{kaufmann2016journal}.
The algorithm's behavior at the two instances is then replaced by its class guarantee.
At a finite horizon a second-order term of order $\log\log T$ is subtracted in the bound of \cite{garivier2019mathematics}.
It bounds the shortfall below the asymptotic rate.
No individual step's loss is resolved in that term.
Lumbreras, Haapasalo and Tomamichel \cite[Lemma~7]{lumbreras2022multiarmed} use the middle bound of Proposition~\ref{prop:cl-twopoint}, written with the R\'enyi divergence of order $1/2$, which equals $2C_T(1/2)$, in regret lower bounds for quantum bandits.
They note that R\'enyi divergences admit no decomposition of the form \eqref{eq:allocation} \cite[Section~4]{lumbreras2022multiarmed}.
For the Bhattacharyya coefficient, $1 - \rho_T$ is bounded by $7 \sum_a \mathbb{E}_\nu[N_a(T)]\,(1 - e^{-c_a(1/2)})$ in a sub-additivity inequality \cite{chen2024assouad}.
Proposition~\ref{prop:tilt} bounds $C_T(1/2) = -\log \rho_T$ by $\sum_a \mathbb{E}_{Q_{1/2}}[N_a(T)]\, c_a(1/2)$, with constant $1$ and a named defect, but with the counts taken under the tilted law.
The characteristic time of Section~\ref{sec:alternative} and its equalizing characterization for fixed-confidence identification are due to Garivier and Kaufmann \cite{garivier2016optimal}, whose lower bound holds at every risk $\delta \le 1/2$, with no asymptotics in $\delta$;
on that side only achievability is asymptotic.
For Gaussian arms with a common variance, the static allocation that maximizes the decay rate of the probability of false selection under a fixed budget, with the empirical best arm recommended, solves the max-min program defining $T^\star(\nu)^{-1}$ \cite{glynn2004large}.
For other one-parameter families the fixed-budget program has a coincidence divergence between the laws of two arms in place of the Bregman information of Proposition~\ref{prop:mbary} (Proposition~\ref{prop:cl-fb-rate}).
The coincidence divergence $c_a(s)$ is minus the logarithm of the Chernoff coefficient \cite{chernoff1952measure}.

Finite-horizon regret bounds follow from the fundamental inequality, the \textsc{process} step applied to a count normalized to $[0,1]$ \cite{garivier2019mathematics}.
Garivier, M\'enard and Stoltz use it to bound the pull counts in the initial phase and at large horizons.
Garivier, Lattimore and Kaufmann use it to bound the regret of explore-then-commit algorithms \cite{garivier2016neurips}.
The Lai--Robbins limit extends to general families \cite{burnetas1996optimal}, to controlled Markov chains \cite{graves1997asymptotically} and to ergodic Markov decision processes \cite{burnetas1997optimal,ok2018exploration}.
For structured bandits its constant is the value of a semi-infinite linear program \cite{combes2017minimal}, which reduces to simpler programs for linear \cite{tlattimore2017end} and Lipschitz \cite{magureanu2014lipschitz} structure.
For structured bandit and reinforcement-learning problems, Wagenmaker and Foster \cite{ajwagenmaker2023instanceoptimality} give lower bounds on the number of rounds needed to learn an approximately optimal Graves--Lai allocation, in terms of the allocation-estimation coefficient they introduce.
These bounds are minimax over a subclass of instances, so none of them is a single-instance limit.
Track-and-Stop asymptotically attains the characteristic time of best-arm identification \cite{garivier2016optimal}.
For Gaussian rewards, Top-Two Thompson Sampling and its variant T3C asymptotically attain its analogue in which the best arm receives a fixed share of the samples \cite{shang2020fixed}.
For rewards in a known bounded interval, with distinct means in its interior, T3C with a Dirichlet sampler does the same under the stopping rule of \cite[Theorem~1]{jourdan2022top}.
For top-$m$ identification the exact characteristic time, a max-min over single-swap pairs, is stated for unit-variance Gaussian arms in possibly misspecified linear models in \cite{reda2021dealing} and for one-parameter exponential families in \cite{you2022informationdirected}.
The best split of Proposition~\ref{prop:cl-split} is Vajda's tight lower bound on relative entropy at a given total variation \cite{fedotov2003refinements,berend2014minimum}.
For sequential tests, Agrawal, Ram and Ramdas \cite{agrawal2025stopping} prove the composite-null bound for power-one tests by data processing to the rejection event.
Naghshvar and Javidi \cite{naghshvar2013active} bound the Bayes risk of active hypothesis testing with many hypotheses.
Kaufmann, Capp\'e and Garivier \cite{kaufmann2016journal} bound the error exponent of two-armed identification under a fixed budget and the error probability of Gaussian identification with a common variance.
Mannor and Tsitsiklis \cite{mannor2004sample} bound the expected sample size of identification within a tolerance.
The single-instance lower bounds among these results appear as rows of Table~\ref{tab:classical}, that of Graves and Lai in its case of independent observations.

\appendix
\appendixtitle
\section*{Notation}

Table~\ref{tab:notation} lists each symbol of the paper with the site of its definition.
Letters that appear only in proofs are omitted, and so are some bound variables and some letters used in only one result, paragraph or evaluation protocol.

{\small
\begin{longtable}{>{\raggedright\arraybackslash}p{0.22\linewidth}>{\raggedright\arraybackslash}p{0.57\linewidth}>{\raggedright\arraybackslash}p{0.13\linewidth}}
\caption{\textbf{Each symbol of the paper names one object, defined at the site listed.} The groups follow the parts of the paper. Writing $\lambda$ as a subscript or an argument gives the same object under the alternative.}
\label{tab:notation}\\
\toprule
Symbol & Meaning & Defined in \\
\midrule
\endfirsthead
\toprule
Symbol & Meaning & Defined in \\
\midrule
\endhead
\midrule
\multicolumn{3}{r}{\textit{Continued on the next page}} \\
\endfoot
\bottomrule
\endlastfoot
\multicolumn{3}{@{}l}{\textit{Instances and histories}} \\*
$\nu$, $\lambda$ & the instance $(\nu_a)_{a \in [K]}$ and an alternative instance & Sec.~\ref{sec:setting} \\
$[K]$, $a$, $b$ & the arms; $b$ a suboptimal arm singled out & Sec.~\ref{sec:setting}, \ref{sec:notation} \\
$\mu_a(\lambda)$, $\mu_a$ & mean of arm $a$ under $\lambda$; under $\nu$ & Sec.~\ref{sec:notation} \\
$\mu^\star$, $a^\star$ & optimal mean; an optimal arm & Sec.~\ref{sec:notation} \\
$\Delta_a(\lambda)$, $\Delta_a$ & gap of arm $a$ under $\lambda$; under $\nu$ & Sec.~\ref{sec:notation} \\
$g$ & separation $\mu_b(\lambda) - \mu_b$ of the perturbed arm's two means & Sec.~\ref{sec:notation} \\
$T$, $t$, $\tau$ & horizon; round; stopping time & Sec.~\ref{sec:setting}, \ref{sec:notation} \\
$A_t$, $X_t$ & arm pulled and observation at round $t$ & Sec.~\ref{sec:setting} \\
$N_a(t)$, $N(t)$ & pulls of arm $a$ by round $t$; the vector of pull counts & Sec.~\ref{sec:setting}, \ref{sec:notation} \\
$S_a(t)$ & sum of the observations from arm $a$ by round $t$ & Sec.~\ref{sec:setting} \\
$\mathcal{F}_t$, $P_\nu$, $\mathbb{E}_\nu$ & observed filtration; law of the history under $\nu$; its expectation & Sec.~\ref{sec:setting} \\
$E$ & an event in $\mathcal{F}_T$ or $\mathcal{F}_\tau$ & Sec.~\ref{sec:notation} \\
$Y$, $P_\nu^Y$, $Z$ & a statistic; its law under $P_\nu$; a statistic with values in $[0,1]$ & Sec.~\ref{sec:notation} \\
$Y^\star$ & the sufficient statistic $(N_a(\tau), S_a(\tau))_{a \in [K]}$ & Thm.~\ref{thm:sufficiency} \\
$R_T(\lambda)$, $R_T$ & regret on $\lambda$; on $\nu$ & Sec.~\ref{sec:notation} \\
$p$ & growth exponent; a consistent algorithm has $R_T(\lambda) = o(T^p)$ for every $p > 0$ & Sec.~\ref{sec:notation} \\
\multicolumn{3}{@{}l}{\textit{Divergences and the divergence decomposition}} \\*
$\KL$, $D_T$, $D_\tau$ & relative entropy; its value on $\mathcal{F}_T$; on $\mathcal{F}_\tau$ & Sec.~\ref{sec:notation} \\
$\TV$, $\kl(u,v)$ & total variation; relative entropy between the Bernoulli laws with means $u$ and $v$ & Sec.~\ref{sec:notation} \\
$\theta_a$, $\eta_a$, $\psi$, $\psi^\star$ & natural parameters of $\nu_a$ and $\lambda_a$; log-partition function; its convex conjugate & Sec.~\ref{sec:notation} \\
$d(\mu, \mu')$, $d_b$ & relative entropy in mean coordinates; $d(\mu_b, \mu^\star)$ & Sec.~\ref{sec:notation}, \ref{sec:certificate} \\
$\ell_a$, $L_t$ & log-likelihood ratio of one observation from arm $a$; of the history on $\mathcal{F}_t$ & Sec.~\ref{sec:setting}, \ref{sec:notation} \\
$n$, $k(\lambda)$ & expected allocation $(\mathbb{E}_\nu[N_a(\tau)])_a$; per-pull divergences $(\KL(\nu_a \,\|\, \lambda_a))_a$ & Sec.~\ref{sec:allocation} \\
\multicolumn{3}{@{}l}{\textit{The coincidence ledger and the testing step}} \\*
$s$, $c_a(s)$, $C_T(s)$ & tilt; coincidence divergence of arm $a$; its path counterpart & Sec.~\ref{sec:coincidence} \\
$c_{P,Q}(s)$ & coincidence divergence $-\log\int P^s Q^{1-s}$ of two laws, so $c_a(s) = c_{\nu_a,\lambda_a}(s)$; its maximum over $s$ is their Chernoff information & App.~\ref{app:cl-fixed-budget} \\
$M_t(s)$ & the martingale of the coincidence ledger & Thm.~\ref{thm:ledger} \\
$Q_s$ & the tilted law, $dQ_s/dP_\lambda = M_T(s)$ & Prop.~\ref{prop:tilt} \\
$P^{(1/2)}$, $\nu_a^{(1/2)}$ & normalized geometric mixtures at $s = 1/2$, of the paths and of arm $a$ & Prop.~\ref{prop:half} \\
$\nu_a^{(s)}$, $s^\star$ & tilt $e^{c_a(s)}\nu_a^{s}\lambda_a^{1-s}$ of arm $a$, equal to $\nu_a^{(1/2)}$ at $s = 1/2$; a maximizer of $C_T$ over $[0,1]$, the Chernoff point & Lem.~\ref{lem:chernoff-point} \\
$\rho_T$ & Bhattacharyya coefficient $e^{-C_T(1/2)}$ & Sec.~\ref{sec:notation} \\
$\underline{s}$ & $s \wedge (1 - s)$ & Prop.~\ref{prop:s-half} \\
$\varepsilon$, $\varepsilon_T$ & a bound on the error sum $P_\nu(E^c) + P_\lambda(E)$; the error sum of $\{N_b(T) \le T/2\}$ & Thm.~\ref{thm:limit}, Prop.~\ref{prop:cl-finite} \\
$\underline{D}(\varepsilon)$ & best-split bound $\min_{0 \le u \le \varepsilon} \kl(1-u, \varepsilon-u)$ on $D_\tau$ at error sum $\varepsilon$ & Prop.~\ref{prop:cl-split} \\
per-pull factor & $\KL(\nu_b \,\|\, \lambda_b)/(2c_b(1/2))$ & Sec.~\ref{sec:tightening} \\
$\sigma^2$ & common variance of Gaussian arms & Prop.~\ref{prop:ratio} \\
path factor & $D_T/(2C_T(1/2))$ & Sec.~\ref{sec:limit} \\
\multicolumn{3}{@{}l}{\textit{Regret}} \\*
$\mathcal{M}_a$, $\mathcal{M}$ & class of the laws of arm $a$; one class for every arm & Sec.~\ref{sec:certificate} \\
$\mathcal{M}_1 \times \dots \times \mathcal{M}_K$, $\mathcal{M}^K$ & product class of the instances with $\lambda_a \in \mathcal{M}_a$ for every arm; its case of one class & Sec.~\ref{sec:certificate} \\
$\Kinf(\cdot, \mathcal{H})$ & infimum of the relative entropy from a law to a set $\mathcal{H}$ of laws & Sec.~\ref{sec:certificate} \\
$\Kinf(\nu_b, \mu^\star)$ & the same, to the laws in $\mathcal{M}_b$ with mean above $\mu^\star$ & Sec.~\ref{sec:certificate} \\
$c^\star(\nu)$ & Lai--Robbins constant $\sum_b \Delta_b / \Kinf(\nu_b, \mu^\star)$ & Sec.~\ref{sec:certificate} \\
$I_b(T)$ & information transported through arm $b$, $\mathbb{E}_\nu[N_b(T)]\, d_b$ & Sec.~\ref{sec:certificate} \\
transported fraction & $I_b(T)/\log T$; for a state-action pair, $I_{x,a}(t)/\log t$ & Sec.~\ref{sec:certificate}, \ref{sec:mdp} \\
regret ratio & $R_T/(c^\star(\nu)\log T)$, or $R_T/(c^\star_\Phi(\nu)\log T)$ & Prop.~\ref{prop:cert} \\
$\xi$ & margin of an alternative beyond the boundary of the alternative set, in the mean, or slack above the infimum; \textsc{boundary} sends it to $0$ or holds it fixed & Sec.~\ref{sec:eval-factor-two}, App.~\ref{app:classical} \\
$\xi_\Delta$ & margin of an alternative beyond the boundary of the alternative set, as a fraction of the gap $\Delta_b$ of the perturbed arm or of the Bellman gap $\Delta_{x,a}$ of the perturbed pair; \textsc{boundary} holds it fixed & Cor.~\ref{cor:cl-gauss}, App.~\ref{app:eval-mdp-regret} \\
$\underline{\Delta}$ & $\Delta_b \wedge (g - \Delta_b)$ & Prop.~\ref{prop:cl-finite} \\
$\mathcal{U}(\nu)$, $\Gamma$ & the Gaussian neighborhood of the finite-horizon bound; the constant of the regret bound $\Gamma T^p$ assumed on it & Cor.~\ref{cor:cl-gauss} \\
$\Kinf(\nu_a, \mu)$ & infimum of the relative entropy from $\nu_a$ to the laws in $\mathcal{M}$ with mean above $\mu$ & App.~\ref{app:cl-gms} \\
$\mathcal{I}$ & interior of the set of means of the laws in $\mathcal{M}$ & App.~\ref{app:cl-gms} \\
$\Upsilon$ & constant of the count guarantee $\mathbb{E}_\lambda[N_a(T)] \le \Upsilon\log T/\Delta_a(\lambda)^2$ & App.~\ref{app:cl-gms} \\
$\xi_{\mathcal{M}}$, $\mathrm{Lip}_{\mathcal{M}}$ & bound on the margin and Lipschitz constant in the definition of a well-behaved class & App.~\ref{app:cl-gms} \\
$\mathcal{M}^2_\Delta$ & two-arm unit-variance Gaussian instances whose means differ by $\Delta$ & App.~\ref{app:cl-etc} \\
$\mathrm{erfc}$ & complementary error function & App.~\ref{app:cl-etc} \\
\multicolumn{3}{@{}l}{\textit{Identification}} \\*
$\Alt(\nu)$ & instances in which some arm other than $a^\star$ is at least as good as $a^\star$ & Sec.~\ref{sec:alternative} \\
$w$, $\bar w$ & an allocation; an algorithm's average allocation & Sec.~\ref{sec:alternative}, App.~\ref{app:cl-bai} \\
$T^\star(\nu)$ & characteristic time & Sec.~\ref{sec:alternative} \\
$G_b(r)$ & pairwise alternative-set infimum at weights $(1, r)$ & Prop.~\ref{prop:scalar-general} \\
$\gamma$, $\gamma^\star$ & variable of the scalar program for $T^\star(\nu)$; its minimizer & Prop.~\ref{prop:scalar} \\
$\mathcal{C}$, $\mathcal{B}_j$ & a set of alternatives; the parts of $\mathcal{C}$ written as a union or an intersection & Prop.~\ref{prop:altsets} \\
$\mathcal{J}_a(w)$ & Bregman information of $\mu_a$ and $\mu_a(\lambda)$ at weights $(w, 1-w)$ & App.~\ref{app:cl-jensen} \\
$\delta$, $\hat a$ & risk of a $\delta$-correct algorithm; its recommendation, or the arm committed to by an explore-then-commit algorithm & Sec.~\ref{sec:alternative}, App.~\ref{app:cl-etc} \\
$\lambda^\star$ & alternative attaining the infimum at $\bar w$ & Prop.~\ref{prop:cl-own}, App.~\ref{app:eval-bai} \\
$m$ & number of arms to identify in top-$m$ identification & App.~\ref{app:cl-bai} \\
$\mathcal{S}^\star$, $\mu_{[k]}$ & the $m$ best arms; the $k$-th largest mean & App.~\ref{app:cl-bai} \\
$T^\star_m(\nu)$, $\Alt_m(\nu)$ & characteristic time and alternative set of top-$m$ identification & App.~\ref{app:cl-bai} \\
$\Delta_{\mathrm{tol}}$, $\mathcal{S}_{\mathrm{tol}}$ & tolerance on the gap of the recommended arm; the arms whose gap is below it & App.~\ref{app:cl-tolerance} \\
$\hat a_T$, $\hat{\mathcal{S}}_T$ & recommended arm and recommended top-$m$ set after a budget of $T$ pulls & App.~\ref{app:cl-fixed-budget} \\
$d^{(m)}_c$ & $d(\mu_c, \mu_{[m+1]})$ for $c \in \mathcal{S}^\star$ and $d(\mu_c, \mu_{[m]})$ otherwise & Prop.~\ref{prop:cl-fb-gauss} \\
\multicolumn{3}{@{}l}{\textit{Sequential tests}} \\*
$\alpha$, $\beta$ & nominal type-I and type-II error probabilities & Sec.~\ref{sec:eval-ab} \\
$\alpha'$, $\beta'$ & realized error probabilities & App.~\ref{app:cl-wald} \\
$\tau^\star$, $E^\star$ & stopping time and decision event of a sequential probability ratio test & Prop.~\ref{prop:cl-ww} \\
$p_0$, $p_1$ & control and treatment rates of the e-mail experiment & Sec.~\ref{sec:eval-ab} \\
$\mathcal{H}_0$, $\Kinf(\nu, \mathcal{H}_0)$ & the null of a sequential test, in Section~\ref{sec:eval-ab} the null of equal rates; the relative entropy from the truth to it, per pair in Section~\ref{sec:eval-ab} & App.~\ref{app:cl-composite}, Sec.~\ref{sec:eval-ab} \\
$q^\star$, $\alpha'_+$, $\alpha'_{+,\mathrm{cap}}$ & rate $(p_0 + p_1)/2$ of the barycentre null; realized probability of rejecting with the treatment ahead; the same before a cap & App.~\ref{app:eval-ab}, \ref{app:eval-own-error} \\
$E_+$, $E_-$ & rejections of a one-sided sequential test in its two directions & Prop.~\ref{prop:cl-power-one} \\
$\mathcal{H}_j$, $\alpha_j$ & hypotheses of a test of several hypotheses; the test's nominal level on $\mathcal{H}_j$, so $\alpha_0 = \alpha$ and $\alpha_1 = \beta$ for a test of $\mathcal{H}_0$ against $\mathcal{H}_1$ & App.~\ref{app:cl-composite} \\
\multicolumn{3}{@{}l}{\textit{Sequential design}} \\*
$\Theta$, $\Theta_i$ & a set of states, each an instance (finite in Chernoff's setting); the hypotheses that partition it (two in Chernoff's setting) & App.~\ref{app:cl-chernoff} \\
$e$, $\mathcal{E}$, $\nu_e$, $N_e(t)$ & an experiment; the experiments; law of $e$ under the state $\nu$; uses of $e$ by round $t$ & App.~\ref{app:cl-chernoff} \\
$\Alt_\Theta(\nu)$, $T^\star_\Theta(\nu)$ & the states outside the hypothesis of $\nu$; its characteristic time, in Chernoff's setting the inverse of his information number & App.~\ref{app:cl-chernoff} \\
$\kappa$, $\mathcal{R}_\kappa(\nu)$, $r_{\min}$ & cost per observation; risk at $\nu$, the expected sampling cost plus the expected terminal loss; a lower bound on the loss of a wrong decision & App.~\ref{app:cl-chernoff} \\
$\pi$ & a prior on the states, positive at every state & Cor.~\ref{cor:cl-active} \\
\multicolumn{3}{@{}l}{\textit{Markov decision processes}} \\*
$\mathcal{X}$, $\mathcal{A}$, $x$ & state and action sets; a state & Sec.~\ref{sec:mdp} \\
$\nu_{x,a}$, $\nu(y \mid x,a)$, $r(x,a)$ & law of the next state and the reward at $(x,a)$; transition probability; mean reward & Sec.~\ref{sec:mdp} \\
$N_{x,a}(t)$, $N_x(t)$ & visits to $(x,a)$; visits to $x$ & Sec.~\ref{sec:mdp} \\
$c_{x,a}(s)$ & coincidence divergence of $\nu_{x,a}$ and $\lambda_{x,a}$ & Thm.~\ref{thm:mdp} \\
$g^\star$, $h^\star$ & optimal gain; bias & Sec.~\ref{sec:mdp} \\
$\Delta_{x,a}$, $\mathcal{A}^\star_x$, $\Pi^\star$ & Bellman gap of $(x,a)$; optimal actions at $x$; optimal stationary policies & Sec.~\ref{sec:mdp} \\
$\Kinf(x,a)$, $I_{x,a}(t)$ & infimum of the relative entropy from $\nu_{x,a}$ to a law that makes the pair optimal; information transported through the pair & Sec.~\ref{sec:mdp} \\
$\Phi$, $\Alt_\Phi(\nu)$ & a class of instances; its alternative set & Sec.~\ref{sec:mdp}, Cor.~\ref{cor:cl-structured} \\
$\mathcal{M}_{x,a}$ & laws of the pair $(x,a)$ in a product class & Prop.~\ref{prop:cl-mdp-decoupled} \\
$c^\star_\Phi(\nu)$, $\zeta$ & regret constant of $\Phi$; the exploration rates of its program & Prop.~\ref{prop:cl-mdp}, Cor.~\ref{cor:cl-structured} \\
$\chi$ & rate of the visit bound & App.~\ref{app:cl-mdp} \\
\multicolumn{3}{@{}l}{\textit{Structured bandits}} \\*
$\varphi_a$, $\omega$, $V(\zeta)$ & feature vector of arm $a$; parameter with $\mu_a = \langle \varphi_a, \omega \rangle$; the matrix $\sum_a \zeta_a \varphi_a \varphi_a^\top$ & Cor.~\ref{cor:cl-linear} \\
$\mathrm{dist}$ & metric on the arms of a Lipschitz class & Cor.~\ref{cor:cl-lipschitz} \\
\multicolumn{3}{@{}l}{\textit{Relaxations and conventions}} \\*
\textsc{process}, \dots, \textsc{allocation} & relaxation tags; a tag's defect is the loss of its step, and its term the measured value of that loss in a split & Sec.~\ref{sec:classical} \\
$z_\nu$, $z_\lambda$, $\underline{\kl}$ & means of the statistic under $\nu$ and $\lambda$; the binary bound $(1 - z_\nu)\log\frac{1}{1 - z_\lambda} - \log 2$ on $\kl(z_\nu, z_\lambda)$ & App.~\ref{app:eval-regret}, \ref{app:eval-mdp-regret} \\
star, bar, underline & an optimal or extremal object, except the conjugate $\psi^\star$; an average; a lower bound & Sec.~\ref{sec:notation}, this table \\
\end{longtable}
}

\section{Classical results recovered from the ledger}
\label{app:classical}

In Appendix~\ref{app:proofs-classical}, each proof of a classical result starts from one of the exact quantities of Section~\ref{sec:classical}, then applies a sequence of the relaxations of Section~\ref{sec:classical}.
Each relaxation is marked with its tag.
Dropping a non-negative term, such as a sampling cost or an expected terminal loss, is named but not tagged.
Where the discarded defect differs from the one listed in Section~\ref{sec:classical}, the proof names it.
Throughout, $\KL(\nu_a \,\|\, \lambda_a) < \infty$ for every arm, and every stopping time satisfies $\mathbb{E}_\nu[\tau] < \infty$, the hypotheses of Theorem~\ref{thm:allocation}.

\subsection{Event, mean and binary forms of the ledger}
\label{app:cl-tools}

\begin{lemma}[Event, mean and binary forms]
\label{lem:relax}
Let $\tau$ be a stopping time with $\mathbb{E}_\nu[\tau] < \infty$.
\begin{enumerate}
\item[(a)] For every $E \in \mathcal{F}_\tau$,
\[
\sum_{a \in [K]} \mathbb{E}_\nu[N_a(\tau)]\,\KL(\nu_a \,\|\, \lambda_a)
 \;=\; \kl\big(P_\nu(E), P_\lambda(E)\big)
 + \mathbb{E}_\nu\!\left[\KL\big(P_\nu(\cdot \mid \mathbf{1}_E) \,\big\|\, P_\lambda(\cdot \mid \mathbf{1}_E)\big)\right]
\]
\item[(b)] For every $\mathcal{F}_\tau$-measurable $Z$ with values in $[0,1]$,
\[
\sum_{a \in [K]} \mathbb{E}_\nu[N_a(\tau)]\,\KL(\nu_a \,\|\, \lambda_a) \;\ge\; \kl\big(\mathbb{E}_\nu[Z], \mathbb{E}_\lambda[Z]\big)
\]
\item[(c)] For $u, v \in [0,1]$, $\kl(u, v) \ge (1-u)\log\frac{1}{1-v} - \log 2$, $\kl(u, v) \ge u \log\frac1v - \log 2$ and $\kl(u,v) \ge 2(u-v)^2$.
On $\{u \ge v\}$, $\kl(u,v)$ is non-decreasing in $u$ and non-increasing in $v$.
\item[(d)] For $0 \le u \le v < 1$ with $v > 0$,
\[
\kl(u, v) \;=\; \int_u^v \frac{y - u}{y(1-y)}\,dy \;\ge\; \frac{(v - u)^2}{2v}
\]
On $\{u \le v\}$, $\kl(u,v)$ is non-increasing in $u$ and non-decreasing in $v$.
\end{enumerate}
\end{lemma}
The proof is in Appendix~\ref{app:proofs-classical}.
\subsection{Divergence decomposition and the fundamental inequality}
\label{app:cl-fundamental}

The divergence decomposition \cite[Lemma 15.1]{lattimore2020bandit} is Theorem~\ref{thm:allocation} at a fixed horizon.
The fundamental inequality of Garivier, M\'enard and Stoltz \cite[display (6), Remark 2]{garivier2019mathematics} states that $\sum_a \mathbb{E}_\nu[N_a(\tau)]\,\KL(\nu_a \,\|\, \lambda_a) \ge \kl(\mathbb{E}_\nu[Z], \mathbb{E}_\lambda[Z])$ for every $\mathcal{F}_\tau$-measurable $Z$ with values in $[0,1]$.
This is Lemma~\ref{lem:relax}(b), a single \textsc{process} step.
With $Z = \mathbf{1}_E$ it is the transportation lemma of Kaufmann, Capp\'e and Garivier \cite[Lemma 1]{kaufmann2016journal}.
Lemma~\ref{lem:relax}(a) turns that inequality into an equality by adding back the conditional divergence.

\subsection{Two-point testing}
\label{app:cl-twopoint}

\begin{proposition}[Le Cam, Bretagnolle--Huber and Pinsker bounds]
\label{prop:cl-twopoint}
For every $E \in \mathcal{F}_T$,
\[
P_\nu(E^c) + P_\lambda(E) \;\ge\; 1 - \sqrt{1 - \rho_T^2} \;\ge\; \tfrac12\,\rho_T^2 \;\ge\; \tfrac12\, e^{-D_T},
\qquad
\TV(P_\nu, P_\lambda) \;\le\; \sqrt{D_T/2}
\]
\end{proposition}

The first bound is Le Cam's affinity bound \cite[(14.11)]{lattimore2020bandit}, \cite{polyanskiywu2025information}, the second Le Cam's inequality \cite[(14.9)]{lattimore2020bandit}, the third the Bretagnolle--Huber inequality \cite[Theorem 14.2]{lattimore2020bandit}, \cite{bretagnolle1979estimation}, and the last Pinsker's inequality.

The proof is in Appendix~\ref{app:proofs-classical}.
The affinity route alone yields $\TV(P_\nu, P_\lambda) \le \sqrt{1 - e^{-D_T}}$, which is weaker than Pinsker's bound when $D_T < 1.59$.
This is the inequality of Bretagnolle and Huber in its original form \cite{bretagnolle1979estimation}, as restated in \cite{sason2018}.
Pinsker's inequality needs the \textsc{process} and \textsc{binary} steps instead.

\begin{lemma}[The affinity bound as an identity]
\label{lem:affinity}
Let $\nu_a \ll \lambda_a$ for every arm.
On $\mathcal{F}_T$, $\rho_T = \mathbb{E}_\nu[e^{-L_T/2}]$, $\TV(P_\nu, P_\lambda) = \mathbb{E}_\nu[(1 - e^{-L_T})^+]$, and
\[
\inf_{E \in \mathcal{F}_T} \big( P_\nu(E^c) + P_\lambda(E) \big)
 \;=\; 1 - \sqrt{1 - \rho_T^2} \;+\; \Big( \sqrt{1 - \rho_T^2} - \mathbb{E}_\nu\big[(1 - e^{-L_T})^+\big] \Big)
\]
The term in parentheses is non-negative.
It vanishes if and only if $|L_T|$ is almost surely constant under $P_\lambda$.
\end{lemma}

The proof is in Appendix~\ref{app:proofs-classical}.
The term in parentheses is the defect of the \textsc{test} step in the proof of Theorem~\ref{thm:limit}.
The proof writes it as a Jensen defect of the concave map $y \mapsto \sqrt{1 - y^2}$.
Its two terms are computed from expectations under $P_\nu$, so it can be estimated from runs of the algorithm on $\nu$.
By the lemma, the pairs with $P_\nu \ll P_\lambda$ on the upper edge of the joint range of total variation and Hellinger distance \cite{polyanskiywu2025information} are those with $|L_T|$ almost surely constant under $P_\lambda$.
For $q \in (0,1)$, the two-point laws $(q, 1-q)$ and $(1-q, q)$ of Theorem~\ref{thm:limit} form such a pair with one observation.
For two or more independent observations of that pair with $q \ne 1/2$, the absolute log-likelihood ratio takes at least two values, so the affinity bound is strict for every event.

For $\varepsilon \in (0,1]$ write
\[
\underline{D}(\varepsilon) \;:=\; \min_{0 \le u \le \varepsilon} \kl(1-u,\, \varepsilon-u)
\]
for the least value of $\kl$ at a split of an error sum $\varepsilon$ into $P_\nu(E^c) = u$ and $P_\lambda(E) = \varepsilon - u$.

\begin{proposition}[Best-split bound]
\label{prop:cl-split}
\begin{enumerate}
\item[(i)] The function $\underline{D}$ is non-increasing on $(0,1]$, $\underline{D}(1) = 0$, and for every $\varepsilon \in (0,1]$
\[
\log\frac{1}{2\varepsilon} \;\le\; \log\frac{1}{\varepsilon(2-\varepsilon)} \;\le\; \underline{D}(\varepsilon) \;\le\; \log\frac{1}{\varepsilon}
\]
with the second inequality strict for $\varepsilon < 1$.
\item[(ii)] For every $\varepsilon \in (0,1]$, $\underline{D}(\varepsilon) = \inf\{\KL(P \,\|\, Q) : \TV(P,Q) = 1 - \varepsilon\}$, the infimum taken over pairs of probability measures on a common measurable space.
\item[(iii)] Let $\tau$ be a stopping time with $\mathbb{E}_\nu[\tau] < \infty$, and let $E \in \mathcal{F}_\tau$ have error sum $\varepsilon := P_\nu(E^c) + P_\lambda(E) \le 1$.
Then $\varepsilon > 0$ and
\[
D_\tau \;=\; \underline{D}(\varepsilon) + \Big(\kl\big(P_\nu(E), P_\lambda(E)\big) - \underline{D}(\varepsilon)\Big) + \mathbb{E}_\nu\!\left[\KL\big(P_\nu(\cdot \mid \mathbf{1}_E) \,\big\|\, P_\lambda(\cdot \mid \mathbf{1}_E)\big)\right]
\]
with the last two terms non-negative.
In particular $D_\tau \ge \underline{D}(\varepsilon')$ for every $\varepsilon' \in [\varepsilon, 1]$.
\end{enumerate}
\end{proposition}

The proof is in Appendix~\ref{app:proofs-classical}.
By (ii), $\underline{D}(\varepsilon)$ is the least relative entropy between two laws at total variation $1 - \varepsilon$, so no lower bound on $D_\tau$ that depends on the event only through its error sum is larger.
One pull of a Bernoulli arm with means $1-u$ and $\varepsilon-u$ under $\nu$ and $\lambda$, at a minimizing $u$, attains $\underline{D}(\varepsilon)$ with the event that the observation is $1$.
The least relative entropy at a given total variation is known as Vajda's tight lower bound \cite{berend2014minimum}.
Fedotov, Harremo\"es and Tops{\o}e give its curve in parametric form \cite{fedotov2003refinements}, as restated in \cite{berend2014minimum}, where it is also written as the minimum over binary laws used here.

The three lower bounds of (i) are the requirements on $D_T$ of three testing steps at a fixed horizon.
The Bretagnolle--Huber bound $\tfrac12 e^{-D_T}$ of Proposition~\ref{prop:cl-twopoint} gives $\log(1/(2\varepsilon))$.
Theorem~\ref{thm:limit}, with $D_T \ge 2C_T(1/2)$ from Proposition~\ref{prop:half}, gives $\log(1/(\varepsilon(2-\varepsilon)))$.
So does the original form of the Bretagnolle--Huber inequality given after Proposition~\ref{prop:cl-twopoint}.
The best split gives $\underline{D}(\varepsilon)$.
By (i), the three requirements lie within $\log 2$ nats of one another, and the best split is the largest of them at every error sum below $1$.
Since $\kl(u,v) \ge 2(u-v)^2$ (Lemma~\ref{lem:relax}(c)), $\underline{D}(\varepsilon) \ge 2(1-\varepsilon)^2$, while $\log(1/(\varepsilon(2-\varepsilon))) \le (1-\varepsilon)^2/(\varepsilon(2-\varepsilon))$.
The best split is therefore at least $2\varepsilon(2-\varepsilon)$ times the requirement of Theorem~\ref{thm:limit}, a factor that tends to $2$ as $\varepsilon \to 1$.
On the scale of the error sum the comparison differs.
Theorem~\ref{thm:limit} bounds the error sum through $2C_T(1/2)$, which can lie far below $D_T$ (Section~\ref{sec:limit}), and there its bound has the smaller exponent.

\begin{lemma}[The minimum error sum at every tilt]
\label{lem:chernoff-point}
Fix $s \in (0,1)$, and let $Q_s$ be the tilted law of Proposition~\ref{prop:tilt}.
Then on $\mathcal{F}_T$
\[
1 - \TV(P_\nu, P_\lambda) \;=\; \mathbb{E}_{Q_s}\!\left[\exp\Big(-\sum_{a \in [K]} N_a(T)\, c_a(s) - \big((s L_T) \vee (-(1-s) L_T)\big)\Big)\right]
\]
If moreover $\nu_a$ and $\lambda_a$ are mutually absolutely continuous for every $a$, the pull counts $N_a(T) = n_a$ are deterministic, and $s^\star$ maximizes $C_T$ over $[0,1]$, then
\[
e^{-C_T(s^\star)} \;\ge\; 1 - \TV(P_\nu, P_\lambda) \;\ge\; \exp\Big(-C_T(s^\star) - \big(s^\star \vee (1-s^\star)\big)\Big(\sum_{a \in [K]} n_a \operatorname{Var}_{\nu_a^{(s^\star)}}(\ell_a)\Big)^{1/2}\Big)
\]
where $\nu_a^{(s)} := e^{c_a(s)}\,\nu_a^{s}\lambda_a^{1-s}$ is the tilt of Section~\ref{sec:coincidence}.
\end{lemma}

The proof is in Appendix~\ref{app:proofs-classical}.
For $T$ pulls of one arm $b$, $C_T(s^\star) = T \max_s c_b(s)$, which is $T$ times the Chernoff information of $\nu_b$ and $\lambda_b$.
The minimum error sum is then $e^{-T \max_s c_b(s) + O(\sqrt{T})}$, the rate of Chernoff's theorem \cite{chernoff1952measure,polyanskiywu2025information}, and the lower half is its converse.
The bound of Theorem~\ref{thm:limit} decays like $e^{-2C_T(1/2)}$ instead.
Since $C_T$ is concave and non-negative on $[0,1]$, $C_T(1/2) \le C_T(s^\star) \le 2C_T(1/2)$.
The left end is attained by a symmetric pair, such as the two-point laws of Theorem~\ref{thm:limit}, whose Chernoff point is the Bhattacharyya point; for $T$ observations of such a pair the bound of Theorem~\ref{thm:limit} decays at twice the true rate.

\subsection{Lai--Robbins and Burnetas--Katehakis}
\label{app:cl-lai-robbins}

For each arm $a$ let $\mathcal{M}_a$ be the class of laws that arm may have, and let $\mathcal{M}_1 \times \dots \times \mathcal{M}_K$ be the class of instances $\lambda$ with $\lambda_a \in \mathcal{M}_a$ for every arm.
An algorithm is consistent on $\mathcal{M}_1 \times \dots \times \mathcal{M}_K$ if $R_T(\lambda) = o(T^p)$ for every $p > 0$ and every instance $\lambda$ of that class \cite[Definition 16.1]{lattimore2020bandit}.
For a law $\nu_b \in \mathcal{M}_b$ with mean below $\mu^\star$, $\Kinf(\nu_b, \mu^\star) = \inf\{\KL(\nu_b \,\|\, \lambda_b) : \lambda_b \in \mathcal{M}_b,\ \text{mean of } \lambda_b > \mu^\star\}$, as in Section~\ref{sec:certificate}.

\begin{proposition}[Lai--Robbins and Burnetas--Katehakis bounds]
\label{prop:cl-lai-robbins}
Let the algorithm be consistent on $\mathcal{M}_1 \times \dots \times \mathcal{M}_K$, and let $\nu$ be an instance of that class.
For every suboptimal arm $b$,
\[
\liminf_{T \to \infty} \frac{\mathbb{E}_\nu[N_b(T)]}{\log T} \;\ge\; \frac{1}{\Kinf(\nu_b, \mu^\star)},
\qquad
\liminf_{T \to \infty} \frac{R_T}{\log T} \;\ge\; \sum_b \frac{\Delta_b}{\Kinf(\nu_b, \mu^\star)}
\]
with $1/0 := \infty$ and $1/\infty := 0$.
\end{proposition}

When every $\mathcal{M}_a$ is the same one-parameter exponential family and some law of the family has mean above $\mu^\star$, then $\Kinf(\nu_b, \mu^\star) = d(\mu_b, \mu^\star)$.
The statement is then the lower bound of Lai and Robbins \cite{lai1985advances}, in the form restated by \cite[p.~3]{kaufmann2016journal}.
The general form, with a class $\mathcal{M}_a$ for each arm, is restated as \cite[Theorem 16.2]{lattimore2020bandit}, where it is credited to Burnetas and Katehakis \cite{burnetas1996optimal}.
Garivier, M\'enard and Stoltz describe the bound of Burnetas and Katehakis as one for a multi-dimensional parameter, under mild conditions on the model \cite[Section~1.2]{garivier2019mathematics}.

The proof is in Appendix~\ref{app:proofs-classical}.
In its proof, only step 5 refers to other algorithms and other instances.
Steps 1 to 4 hold for the algorithm at hand.
Each of their defects can be computed for it exactly.
With the normalized count $Z := N_b(T)/T$ in place of $\mathbf{1}_E$, step 2 becomes $D_T \ge \kl(\mathbb{E}_\nu[Z], \mathbb{E}_\lambda[Z])$ by Lemma~\ref{lem:relax}(b).
Step 3 becomes $\kl(\mathbb{E}_\nu[Z], \mathbb{E}_\lambda[Z]) \ge (1 - \mathbb{E}_\nu[Z])\log\big(T/\mathbb{E}_\lambda[T - N_b(T)]\big) - \log 2$ by Lemma~\ref{lem:relax}(c).
Consistency bounds both expected counts directly, so step 4 is not needed.
Section~\ref{sec:eval-regret} evaluates this form.
This route is that of the proof of Theorem~1 of Garivier, M\'enard and Stoltz \cite{garivier2019mathematics}.
That theorem is stated for every class $\mathcal{M}$ shared by the arms.
With a numeric class guarantee in place of consistency, the finite-horizon bounds of Propositions~\ref{prop:cl-gms-initial} and~\ref{prop:cl-gms-finite} follow from the fundamental inequality with a normalized count.

\subsection{A finite-horizon regret bound}
\label{app:cl-finite}

\begin{proposition}[Finite-horizon regret bound]
\label{prop:cl-finite}
Let $\lambda$ differ from $\nu$ only at arm $b$, with $b$ suboptimal under $\nu$ and uniquely optimal under $\lambda$.
Write $g := \mu_b(\lambda) - \mu_b$ for the separation of arm $b$, $\underline{\Delta} := \Delta_b \wedge (g - \Delta_b)$ and $\varepsilon_T := P_\nu(N_b(T) > T/2) + P_\lambda(N_b(T) \le T/2)$.
For every algorithm and every horizon $T$,
\[
\mathbb{E}_\nu[N_b(T)]\,\KL(\nu_b \,\|\, \lambda_b)
 \;\ge\; \log\frac{1}{\varepsilon_T(2-\varepsilon_T)}
 \;\ge\; \log\frac{T \underline{\Delta}}{4\,\big(R_T(\nu) + R_T(\lambda)\big)}
\]
\end{proposition}

The outer inequality is Lemma~16.3 of \cite{lattimore2020bandit}.

The proof is in Appendix~\ref{app:proofs-classical}.
The Bretagnolle--Huber route of \cite{lattimore2020bandit} reaches $\log(1/(2\varepsilon_T))$ directly.
The affinity route adds $\log(2/(2-\varepsilon_T))$.
This gain is at most $\log 2$ when $\varepsilon_T \le 1$.
It vanishes as $\varepsilon_T \to 0$.

\begin{corollary}[Gaussian finite-horizon bound]
\label{cor:cl-gauss}
Let $\nu$ have unit-variance Gaussian arms, let $\mathcal{U}(\nu)$ be the set of unit-variance Gaussian instances whose means satisfy $\mu_a \le \mu_a(\lambda') \le \mu_a + 2\Delta_a$ for every $a$, and suppose $R_T(\lambda') \le \Gamma T^p$ for every $\lambda' \in \mathcal{U}(\nu)$, with $\Gamma > 0$ and $p \in (0,1)$.
Then for every $\xi_\Delta \in (0, 1]$,
\[
R_T(\nu) \;\ge\; \frac{2}{(1+\xi_\Delta)^2} \sum_{b} \frac{\big((1-p)\log T + \log(\xi_\Delta \Delta_b / 8\Gamma)\big)^+}{\Delta_b}
\]
\end{corollary}

This is Theorem~16.4 of \cite{lattimore2020bandit}.

The proof is in Appendix~\ref{app:proofs-classical}.

\subsection{The normalized count at a finite horizon}
\label{app:cl-gms}

Let every arm's laws range over one class $\mathcal{M}$, as in Appendix~\ref{app:cl-lai-robbins} with $\mathcal{M}_a = \mathcal{M}$ for every arm.
For a law $\nu_a \in \mathcal{M}$ and a number $\mu$, $\Kinf(\nu_a, \mu)$ takes $\mathcal{H}$ to be the laws in $\mathcal{M}$ with mean above $\mu$.
On $\mathcal{M}^K$ an algorithm favors optimal arms if $\mathbb{E}_\lambda[N_a(T)] \ge T/K$ for every $\lambda \in \mathcal{M}^K$, every arm $a$ with $\Delta_a(\lambda) = 0$ and every $T \ge 1$.
Random play, which pulls each arm with probability $1/K$ at every round, meets this condition with equality.

\begin{proposition}[Initial-phase count bound]
\label{prop:cl-gms-initial}
Let the algorithm favor optimal arms on $\mathcal{M}^K$, and let $\nu \in \mathcal{M}^K$.
For every arm $a$ with $\Kinf(\nu_a, \mu^\star) < \infty$ and every $T \ge 1$,
\[
\mathbb{E}_\nu[N_a(T)]
 \;\ge\; \frac{T}{K}\Big(1 + T\Kinf(\nu_a,\mu^\star) - \sqrt{T^2\Kinf(\nu_a,\mu^\star)^2 + 2T\Kinf(\nu_a,\mu^\star)}\Big)
 \;\ge\; \frac{T}{K}\Big(1 - \sqrt{2T\Kinf(\nu_a,\mu^\star)}\Big)
\]
In particular $\mathbb{E}_\nu[N_a(T)] \ge T/(2K)$ whenever $T \le 1/(8\Kinf(\nu_a,\mu^\star))$, with $1/0 := \infty$.
\end{proposition}

The second bound and the particular case are Theorem~2 of Garivier, M\'enard and Stoltz \cite{garivier2019mathematics}, whose Definition~2 is the hypothesis on the algorithm.
The first bound solves exactly the quadratic requirement on which their proof rests.
Unlike the second, it stays positive at every horizon.

The proof is in Appendix~\ref{app:proofs-classical}.
Its chain uses the normalized count $N_a(T)/T$ of Appendix~\ref{app:cl-lai-robbins}, the class guarantee at the alternative as \textsc{uniform}, and Lemma~\ref{lem:relax}(d) as \textsc{binary}.
No step is asymptotic.
Since the class guarantee is a number, every defect can be computed for a given algorithm.

Write $\mathcal{I}$ for the interior of the set of means of the laws in $\mathcal{M}$.
The class $\mathcal{M}$ is well behaved if there are positive functions $\xi_{\mathcal{M}}$ on $\mathcal{I}$ and $\mathrm{Lip}_{\mathcal{M}}$ on $\mathcal{M} \times \mathcal{I}$ with $\Kinf(\nu_a, \mu + \xi) \le \Kinf(\nu_a, \mu) + \xi\,\mathrm{Lip}_{\mathcal{M}}(\nu_a, \mu)$ whenever $\nu_a \in \mathcal{M}$, $\mu \in \mathcal{I}$ lies above the mean of $\nu_a$ and $0 < \xi < \xi_{\mathcal{M}}(\mu)$.
An algorithm meets the count guarantee with constant $\Upsilon$ on $\mathcal{M}^K$ if $\mathbb{E}_\lambda[N_a(T)] \le \Upsilon \log T/\Delta_a(\lambda)^2$ for every $\lambda \in \mathcal{M}^K$, every arm $a$ with $\Delta_a(\lambda) > 0$ and every $T \ge 2$.

\begin{proposition}[Finite-horizon Lai--Robbins bound]
\label{prop:cl-gms-finite}
Let $\mathcal{M}$ be well behaved, let the algorithm meet the count guarantee with constant $\Upsilon$ on $\mathcal{M}^K$, and let $\nu \in \mathcal{M}^K$ have $\mu^\star \in \mathcal{I}$.
Let $b$ be a suboptimal arm.
For every $T \ge 2$ with $(\log T)^{-4} < \xi_{\mathcal{M}}(\mu^\star)$ at which each of the three terms subtracted from $1$ below lies in $[0,1)$,
\[
\mathbb{E}_\nu[N_b(T)]
 \;\ge\; \frac{\log T}{\Kinf(\nu_b,\mu^\star)}\bigg(1
 - \frac{\mathrm{Lip}_{\mathcal{M}}(\nu_b,\mu^\star)}{\Kinf(\nu_b,\mu^\star)\,(\log T)^4}
 - \frac{\Upsilon\log T}{T}\sum_{a : \Delta_a > 0} \frac{1}{\Delta_a^2}
 - \frac{\log\big(K\Upsilon(\log T)^9\big)}{\log T}\bigg)
 - \frac{\log 2}{\Kinf(\nu_b,\mu^\star)}
\]
with $1/0 := \infty$ and $1/\infty := 0$.
\end{proposition}

This is Theorem~5 of Garivier, M\'enard and Stoltz \cite{garivier2019mathematics}, in whose display (18) the three terms are written $a_T$, $b_T$ and $c_T$.
The count guarantee is their Definition~6.
A well-behaved class is that of their Definition~5.
Their statement asks the three terms to be smaller than $1$.
The last step of their proof, $(1 - a_T)(1 - b_T)(1 - c_T) \ge 1 - (a_T + b_T + c_T)$, needs them to be non-negative.
Only $c_T$ can be negative, so the statement here asks $c_T \ge 0$ as well.

The proof is in Appendix~\ref{app:proofs-classical}.
Its chain is that of Proposition~\ref{prop:cl-lai-robbins} kept at a finite horizon.
The statistic is the normalized count of the optimal arms, the binary bound is the second of Lemma~\ref{lem:relax}(c), the count guarantee at $\nu$ and at the alternative is \textsc{uniform}, and the alternatives are kept at the margin $(\log T)^{-4}$ above $\mu^\star$ (\textsc{boundary}).
No step is asymptotic and none uses Markov's inequality, so every defect can be computed for an algorithm whose constant $\Upsilon$ is known.
The bound is positive only when $T > 2K\Upsilon(\log T)^9$.
For $K\Upsilon = 1$ it is positive only when $T > 6.7 \times 10^{13}$.

\subsection{Explore-then-commit}
\label{app:cl-etc}

Let $K = 2$ and let $\mathcal{M}$ be the unit-variance Gaussian laws.
For $\Delta > 0$, $\mathcal{M}^2_\Delta$ is the set of instances in $\mathcal{M}^2$ whose two means differ by $\Delta$.
An algorithm explores then commits if, for an even stopping time $\tau$, it pulls arm $1$ at the odd rounds and arm $2$ at the even rounds up to $\tau$, and an $\mathcal{F}_\tau$-measurable arm $\hat a$ at every later round.
Its design is fixed if $\tau = 2n$ for a deterministic integer $n$.
The stopping time, the arm $\hat a$ and the integer $n$ may depend on $T$, and on $\Delta$ when the class is $\mathcal{M}^2_\Delta$.
The complementary error function is $\mathrm{erfc}(z) := 2\pi^{-1/2}\int_z^\infty e^{-y^2}\,dy$.

\begin{proposition}[Explore-then-commit bounds]
\label{prop:cl-etc}
Let $\nu \in \mathcal{M}^2$ have means $\Delta > 0$ apart.
\begin{enumerate}
\item[(i)] If the algorithm is consistent on $\mathcal{M}^2_\Delta$, then $\liminf_{T \to \infty} R_T/\log T \ge 1/(2\Delta)$.
\item[(ii)] If the algorithm explores then commits and is consistent on $\mathcal{M}^2_\Delta$, then $\liminf_{T \to \infty} R_T/\log T \ge 1/\Delta$.
\item[(iii)] If the algorithm explores then commits and is consistent on $\mathcal{M}^2$, then $\liminf_{T \to \infty} R_T/\log T \ge 4/\Delta$.
\end{enumerate}
\end{proposition}

These are Theorems~6, 3 and~4 of Garivier, Lattimore and Kaufmann \cite{garivier2016neurips}, who prove them with the fundamental inequality at a stopping time \cite{garivier2019mathematics}.

The proof is in Appendix~\ref{app:proofs-classical}.
Each part uses one alternative.
In (i) the alternative raises the suboptimal arm to $\Delta$ above $\mu^\star$, the only instance of $\mathcal{M}^2_\Delta$ that keeps the law of the optimal arm and makes the other arm optimal.
In (ii) it exchanges the two laws, which minimizes the mean per-pull divergence of an alternating allocation over the instances of $\mathcal{M}^2_\Delta$ with the other optimal arm.
In (iii) it approaches the instance with both means at their midpoint, where both per-pull divergences equal $\Delta^2/8$, the Bregman information of Proposition~\ref{prop:mbary} at equal weights.
That instance has no unique optimal arm, so the proof keeps a margin and then lets it vanish (\textsc{boundary}).
An algorithm that explores then commits has its allocation fixed up to $\tau$, so no \textsc{allocation} step enters.
The alternation is part of the hypothesis.
If any sampling rule were allowed before $\tau$, an index policy run with $\tau = \infty$ would explore then commit, so no bound above that of general algorithms could hold \cite[Section~6]{garivier2016neurips}.

\begin{proposition}[Fixed-design explore-then-commit bound]
\label{prop:cl-etc-fixed}
Let $\nu \in \mathcal{M}^2$ have optimal arm $a^\star$ and means $\Delta > 0$ apart, let $\lambda$ exchange its two laws, and let the algorithm explore then commit with a fixed design $\tau = 2n$.
\begin{enumerate}
\item[(i)] If $2n \le T$, then $R_T(\nu) + R_T(\lambda) = 2\Delta n + \Delta(T - 2n)\big(P_\nu(\hat a \ne a^\star) + P_\lambda(\hat a = a^\star)\big)$ and
\[
P_\nu(\hat a \ne a^\star) + P_\lambda(\hat a = a^\star)
 \;\ge\; 1 - \TV\big(P_\nu|_{\mathcal{F}_{2n}}, P_\lambda|_{\mathcal{F}_{2n}}\big)
 \;=\; \mathrm{erfc}\Big(\sqrt{C_{2n}(1/2)}\Big),
\qquad
C_{2n}(1/2) \;=\; \frac{n\Delta^2}{4}
\]
with equality when $\hat a$ is the arm with the larger sample mean.
For that arm,
\[
R_T(\nu) \;=\; \Delta n + \tfrac{\Delta}{2}(T - 2n)\,\mathrm{erfc}\big(\Delta\sqrt{n}/2\big)
\]
\item[(ii)] If the algorithm is consistent on $\mathcal{M}^2_\Delta$, then $\liminf_{T \to \infty} R_T/\log T \ge 4/\Delta$.
\end{enumerate}
\end{proposition}

For the arm with the larger sample mean, part (i) is the regret formula in the proof of Theorem~1 of Garivier, Lattimore and Kaufmann \cite{garivier2016neurips}, and part (ii) gives the lower half of that theorem's asymptotic statement.
Their summary of results states the constant $4$ for every fixed design consistent on $\mathcal{M}^2_\Delta$.

The proof is in Appendix~\ref{app:proofs-classical}.
It starts from the exact minimum error sum $1 - \TV$ and applies no \textsc{test} step.
With the affinity bound of Theorem~\ref{thm:limit} in place of the exact value (\textsc{test}), the same chain gives $2/\Delta$.
With the Bretagnolle--Huber bound $\tfrac12 e^{-D_{2n}}$, where $D_{2n} = n\Delta^2 = 4C_{2n}(1/2)$, it gives $1/\Delta$.
As $n$ grows, minus the logarithms of the exact value and of these two bounds grow like $C_{2n}(1/2)$, $2C_{2n}(1/2)$ and $4C_{2n}(1/2)$.
The relative-entropy chain of Proposition~\ref{prop:cl-etc}(ii) applies to fixed designs as well and requires only $D_{2n} \ge (1 - o(1))\log T$, which gives $1/\Delta$.
The factor of four between part (ii) and Proposition~\ref{prop:cl-etc}(ii) is the ratio $D_{2n}/C_{2n}(1/2)$ on this pair.
\subsection{Wald's bound for sequential tests}
\label{app:cl-wald}

Take one stream ($K = 1$) and identify each instance with the law of one observation.
Suppose $\nu$ and $\lambda$ are mutually absolutely continuous.
A sequential test is a stopping time $\tau$, almost surely finite under both laws, together with a decision; let $E \in \mathcal{F}_\tau$ be the event that it decides for $\nu$.
Its error probabilities are $\alpha' := P_\lambda(E)$ and $\beta' := P_\nu(E^c)$.

\begin{proposition}[Wald's bound]
\label{prop:cl-wald}
If $\mathbb{E}_\nu[\tau] < \infty$ then
\[
\mathbb{E}_\nu[\tau]\,\KL(\nu \,\|\, \lambda) \;=\; \kl(1-\beta', \alpha') + \sum_{E' \in \{E, E^c\}} P_\nu(E') \Big( \mathbb{E}_\nu[L_\tau \mid E'] + \log \mathbb{E}_\nu[e^{-L_\tau} \mid E'] \Big)
 \;\ge\; \kl(1-\beta', \alpha')
\]
If $\mathbb{E}_\lambda[\tau] < \infty$ and $\KL(\lambda \,\|\, \nu) < \infty$, the symmetric statement holds under $\lambda$ with $\kl(1-\alpha', \beta')$ and $\KL(\lambda \,\|\, \nu)$.
\end{proposition}

The inequality is Wald's bound \cite[(4.81)]{wald1945sequential}, written with the Bernoulli relative entropy.

The proof is in Appendix~\ref{app:proofs-classical}.
The defect is zero exactly when $L_\tau$ is constant on each decision set.
For the sequential probability ratio test, $L_\tau$ has just crossed one of two boundaries, so the defect measures the spread of the overshoot.
The inaccuracy of Wald's approximate thresholds for the realized error rates is also caused by the overshoot \cite{fischer2024improving}.
Wald and Wolfowitz \cite{waldwolfowitz1948optimum} proved that this test minimizes both expected sample sizes among tests with no larger error probabilities \cite[Section~2]{cpabbaraju2026simple}.
The ledger gives the defect of that test exactly.
A test held to nominal levels $\alpha \ge \alpha'$ and $\beta \ge \beta'$ with $\alpha + \beta \le 1$ satisfies $\kl(1-\beta', \alpha') \ge \kl(1-\beta, \alpha)$ by Lemma~\ref{lem:relax}(c) (\textsc{uniform}).

With the error probabilities of a sequential probability ratio test as these levels, that test is compared with every test held to them.

\begin{proposition}[Wald--Wolfowitz optimality to within the defect]
\label{prop:cl-ww}
Let $\nu \ne \lambda$, let $\tau^\star$ be the first $t \ge 1$ at which $L_t$ leaves a bounded open interval containing $0$, and let $E^\star := \{L_{\tau^\star} > 0\}$, $\alpha := P_\lambda(E^\star)$ and $\beta := P_\nu((E^\star)^c)$.
Then $\mathbb{E}_\nu[\tau^\star] < \infty$, and every sequential test with $\alpha' \le \alpha$ and $\beta' \le \beta$ satisfies
\[
\mathbb{E}_\nu[\tau^\star] - \mathbb{E}_\nu[\tau]
 \;\le\; \frac{\mathbb{E}_\nu\big[\KL\big(P_\nu(\cdot \mid \mathbf{1}_{E^\star}) \,\big\|\, P_\lambda(\cdot \mid \mathbf{1}_{E^\star})\big)\big]}{\KL(\nu \,\|\, \lambda)}
\]
If $L_{\tau^\star}$ is $P_\nu$-almost surely constant on $E^\star$ and on its complement, the right side is zero and $\mathbb{E}_\nu[\tau^\star]\, \KL(\nu \,\|\, \lambda) = \kl(1-\beta, \alpha)$.
The symmetric statements hold under $\lambda$ when $\KL(\lambda \,\|\, \nu) < \infty$.
\end{proposition}

The proof is in Appendix~\ref{app:proofs-classical}.
The defect is zero when every observation moves $L_t$ up or down by one fixed amount, as for two Bernoulli laws with means placed symmetrically about $1/2$.
On each side, $L_t$ then leaves the interval at a single point.
In that case the proposition gives the theorem of Wald and Wolfowitz.
For these Bernoulli laws and symmetric boundaries, Pabbaraju, Valiant and Verma \cite[Theorem~1]{cpabbaraju2026simple} give a geometric proof of a related optimality, stated for the sums of the two error probabilities and of the two expected sample sizes.
When the defect is not zero, the theorem of Wald and Wolfowitz rests on their Bayes argument.
In that argument the test is shown to minimize, for suitable weights, a weighted sum of the two error probabilities and the two expected sample sizes \cite[Sections~1.1 and~2]{cpabbaraju2026simple}.

\subsection{Tests of several or composite hypotheses}
\label{app:cl-composite}

Take one stream and identify each instance with the law of one observation.
Let $\mathcal{H}_j$, for $j$ in a finite index set with at least two elements, be disjoint sets of laws.
A test of these hypotheses is a stopping time $\tau$, almost surely finite under every law in their union, together with a decision among the hypotheses made at $\tau$.
Its levels are numbers $\alpha_j \in (0,1)$ such that, under every $\lambda \in \mathcal{H}_j$, the test decides a hypothesis other than $\mathcal{H}_j$ with probability at most $\alpha_j$.

\begin{proposition}[Tests of several hypotheses]
\label{prop:cl-multitest}
Let a test of the hypotheses $\mathcal{H}_j$ have levels $\alpha_j$, and let $\nu \in \mathcal{H}_i$ satisfy $\mathbb{E}_\nu[\tau] < \infty$.
\begin{enumerate}
\item[(i)] Let $j \ne i$ satisfy $\alpha_i + \alpha_j \le 1$.
Every $\lambda \in \mathcal{H}_j$ with $\KL(\nu \,\|\, \lambda) < \infty$ satisfies
\[
\mathbb{E}_\nu[\tau]\,\KL(\nu \,\|\, \lambda) \;\ge\; \kl(1 - \alpha_i, \alpha_j)
\]
so $\mathbb{E}_\nu[\tau]\,\Kinf(\nu, \mathcal{H}_j) \ge \kl(1 - \alpha_i, \alpha_j)$ when $\Kinf(\nu, \mathcal{H}_j)$ is finite.
\item[(ii)] Let $j \ne i$ satisfy $0 < \Kinf(\nu, \mathcal{H}_j) < \infty$.
Along a sequence of such tests with $\alpha_i \to 0$ and $\alpha_j \to 0$,
\[
\liminf \frac{\mathbb{E}_\nu[\tau]\,\Kinf(\nu, \mathcal{H}_j)}{\log(1/\alpha_j)} \;\ge\; 1
\]
\end{enumerate}
\end{proposition}

The proof is in Appendix~\ref{app:proofs-classical}.
With two simple hypotheses, part (i) is the bound of Proposition~\ref{prop:cl-wald} at nominal levels.
With finitely many simple hypotheses, part (ii) bounds each expected sample size below, to first order, by the largest of $\log(1/\alpha_j)/\Kinf(\nu, \mathcal{H}_j)$ over $j \ne i$.
Lorden \cite[Theorem~4 and its corollary]{lorden1977nearlyoptimal} shows that a matrix of one-sided SPRTs minimizes each expected sample size to within $o(1)$ among tests with no larger error probabilities, as those probabilities go to zero.
Part (ii) gives the first-order lower bound in that statement.
The $o(1)$ term is not recovered by the chain: the overshoot enters the chain only as a defect of the test at hand.
Lorden's proof accounts for the overshoot through constants that he relates to it \cite[Theorem~5]{lorden1977nearlyoptimal}.

For two composite hypotheses of a one-parameter exponential family, let $\mathcal{H}_0$ hold the laws of the family with mean at most that of a law $\lambda^{(0)}$, and $\mathcal{H}_1$ those with mean at least that of a law $\lambda^{(1)}$ of larger mean.
The laws between the two form an indifference zone.
At every law $\nu$ of the family, Hoeffding's lower bound \cite{hoeffding1960lower}, as restated in \cite[(94)]{agtartakovsky2026change}, has leading term $\log(1/(\alpha' + \beta'))$ divided by the larger of $\KL(\nu \,\|\, \lambda^{(0)})$ and $\KL(\nu \,\|\, \lambda^{(1)})$, with $\alpha'$ and $\beta'$ the error probabilities at $\lambda^{(0)}$ and $\lambda^{(1)}$.
Lai \cite{lai1988nearly}, as restated in \cite[Theorem~5]{agtartakovsky2026change}, showed that, when $\log(1/\alpha_0)$ and $\log(1/\alpha_1)$ grow at the same rate, a generalized likelihood ratio test attains, to first order, the least expected sample size among tests with no larger error probabilities.
That least size is $\log(1/\alpha_0)$ divided by the same larger divergence.
For $\nu \in \mathcal{H}_1$ the larger divergence is $\KL(\nu \,\|\, \lambda^{(0)}) = \Kinf(\nu, \mathcal{H}_0)$, attained at $\lambda^{(0)}$, so part (ii) is the lower half of Lai's statement and \textsc{boundary} has no defect.
Inside the indifference zone the chain, applied to each hypothesis in turn, gives a bound whose leading term is smaller by up to a factor of two, because the levels leave the probability of each decision free there.

A one-sided sequential test of a nonempty set $\mathcal{H}_0$ of laws is a stopping time $\tau$ at which the test rejects $\mathcal{H}_0$ \cite{agrawal2025stopping}.
On $\{\tau = \infty\}$ it never rejects.
It has level $\alpha \in (0,1)$ if $P_\lambda(\tau < \infty) \le \alpha$ for every $\lambda \in \mathcal{H}_0$, and power one at $\nu$ if $P_\nu(\tau < \infty) = 1$.

\begin{proposition}[Power-one tests with a composite null]
\label{prop:cl-power-one}
Take one stream and identify each instance with the law of one observation.
Let $\tau$ be a one-sided sequential test of $\mathcal{H}_0$ with level $\alpha$, and let $\nu$ be a law with $\mathbb{E}_\nu[\tau] < \infty$.
\begin{enumerate}
\item[(i)] For every $\lambda \in \mathcal{H}_0$ with $\KL(\nu \,\|\, \lambda) < \infty$,
\[
\mathbb{E}_\nu[\tau]\,\KL(\nu \,\|\, \lambda)
 \;=\; \log\frac{1}{P_\lambda(\tau < \infty)} + \KL\big(P_\nu \,\big\|\, P_\lambda(\cdot \mid \tau < \infty)\big)
 \;\ge\; \log\frac{1}{\alpha}
\]
with the last relative entropy taken on $\mathcal{F}_\tau$, so $\mathbb{E}_\nu[\tau]\,\Kinf(\nu, \mathcal{H}_0) \ge \log(1/\alpha)$.
\item[(ii)] Let $E_+, E_- \in \mathcal{F}_\tau$ partition $\{\tau < \infty\}$, and let $Y := \mathbf{1}_{E_+} - \mathbf{1}_{E_-}$.
The last relative entropy of (i) equals
\[
\kl\big(P_\nu(E_+),\, P_\lambda(E_+ \mid \tau < \infty)\big) + \mathbb{E}_\nu\big[\KL\big(P_\nu(\cdot \mid Y) \,\big\|\, P_\lambda(\cdot \mid Y)\big)\big]
\]
\end{enumerate}
\end{proposition}

The proof is in Appendix~\ref{app:proofs-classical}.
Agrawal, Ram and Ramdas \cite[Theorem~3.1]{agrawal2025stopping} prove the bound of (i) for any composite null by the same data processing to the event $\{\tau < \infty\}$.
That event has probability one under $\nu$, so the step needs no \textsc{binary} relaxation, since $\kl(1, v) = \log(1/v)$.
For the null of equal rates of Section~\ref{sec:eval-ab}, with observations in (control, treatment) pairs, the relative entropy from $\nu$ to the null law of common rate $q$ is $\kl(p_0, q) + \kl(p_1, q)$.
Its minimum over $q$ is attained at the barycentre $q^\star$ by Proposition~\ref{prop:mbary}, so the chain at that law has no \textsc{boundary} defect.
When the test also reports a direction, as a two-sided test does, part (ii) splits the \textsc{process} defect of (i) in two.
The first term is the defect of processing the direction-tagged decision $Y$ to the rejection event.
If the null law rejects in the two directions equally often, it is $\kl(P_\nu(E_+), 1/2)$, which is $\log 2$ less the binary entropy of $P_\nu(E_+)$ and equals $\log 2$ when $\nu$ rejects only on $E_+$.
The second term is the defect of processing the history to $Y$.

\subsection{Sequential design of experiments}
\label{app:cl-chernoff}

Chernoff's setting \cite{chernoff1959sequential} has a finite set $\Theta$ of states, split into two disjoint hypotheses $\Theta_1$ and $\Theta_2$, and finitely many experiments $e \in \mathcal{E}$, which play the role of arms.
Each state is an instance: under the state $\nu$, experiment $e$ yields an observation with law $\nu_e$, and $N_e(t)$ counts the uses of experiment $e$ by round $t$.
The divergence $\KL(\nu_e \,\|\, \lambda_e)$ is assumed finite for all states $\nu$ and $\lambda$.
For $\nu \in \Theta_i$ the alternative set is $\Alt_\Theta(\nu) := \Theta_{3-i}$.
Chernoff's information number is the value of a zero-sum game \cite[(4.2)]{chernoff1959sequential}.
In the notation of Section~\ref{sec:alternative} it is the inverse of the characteristic time of this alternative set,
\[
T^\star_\Theta(\nu)^{-1} \;:=\; \max_{w} \min_{\lambda \in \Alt_\Theta(\nu)} \sum_{e} w_e\, \KL(\nu_e \,\|\, \lambda_e)
\]
where $w$ ranges over the allocations on $\mathcal{E}$.
A procedure chooses experiments adaptively, stops at $\tau$ and decides a hypothesis.
Each observation costs $\kappa > 0$, and for each $\kappa$ a procedure is given, which may depend on $\kappa$.
With $\tau$ and the decision those of the procedure for $\kappa$, its risk $\mathcal{R}_\kappa(\nu)$ at $\nu$ is $\kappa\,\mathbb{E}_\nu[\tau]$ plus the expected terminal loss, where a correct decision loses $0$ and a wrong one at least $r_{\min} > 0$.
One procedure used at every $\kappa$ cannot satisfy the hypothesis of Proposition~\ref{prop:cl-chernoff}.
Its risk tends to its expected terminal loss as $\kappa \to 0$, so its error probabilities would vanish at every state.
For $\lambda \in \Alt_\Theta(\nu)$, Theorem~\ref{thm:allocation} and Lemma~\ref{lem:relax}(a) would then bound the infinite $\kl(1, 0)$ by the finite $\mathbb{E}_\nu[\tau] \max_e \KL(\nu_e \,\|\, \lambda_e)$.

\begin{proposition}[Chernoff's bound]
\label{prop:cl-chernoff}
Let a procedure be given for each $\kappa \in (0, \kappa_0)$, for some $\kappa_0 > 0$.
Suppose $T^\star_\Theta(\nu) < \infty$ and $\mathcal{R}_\kappa(\lambda) = O(\kappa\log(1/\kappa))$ as $\kappa \to 0$ for every $\lambda \in \Theta$.
Then $\mathcal{R}_\kappa(\nu) \ge (1 - o(1))\, \kappa \log(1/\kappa)\, T^\star_\Theta(\nu)$.
\end{proposition}

This is Chernoff's Theorem~2 \cite{chernoff1959sequential}.

The proof is in Appendix~\ref{app:proofs-classical}.

With finitely many states, more than two hypotheses change only the definitions.
Let $\Theta$ be partitioned into finitely many hypotheses $\Theta_i$, at least two, and for $\nu \in \Theta_i$ let $\Alt_\Theta(\nu) := \Theta \setminus \Theta_i$, with $T^\star_\Theta(\nu)$ defined by the same formula.
Proposition~\ref{prop:cl-chernoff} and its proof then hold as written.
Deciding the hypothesis of $\nu$ is an error under every $\lambda \in \Alt_\Theta(\nu)$, deciding any other hypothesis is an error under $\nu$, and the proof uses nothing else about the hypotheses.
Averaged over a prior on the states, the risks form the Bayes risk of active hypothesis testing.

\begin{corollary}[Active hypothesis testing]
\label{cor:cl-active}
Let $\Theta$ be partitioned into finitely many hypotheses, let a procedure be given for each $\kappa \in (0, \kappa_0)$, for some $\kappa_0 > 0$, and let $\pi$ be a probability on $\Theta$ with $\pi_\nu > 0$ for every $\nu \in \Theta$.
Suppose $T^\star_\Theta(\nu) < \infty$ for every $\nu \in \Theta$ and $\sum_{\nu \in \Theta} \pi_\nu\, \mathcal{R}_\kappa(\nu) = O(\kappa \log(1/\kappa))$ as $\kappa \to 0$.
Then $\sum_{\nu \in \Theta} \pi_\nu\, \mathcal{R}_\kappa(\nu) \ge (1 - o(1))\, \kappa \log(1/\kappa) \sum_{\nu \in \Theta} \pi_\nu T^\star_\Theta(\nu)$.
\end{corollary}

The proof is in Appendix~\ref{app:proofs-classical}.
With one state in each hypothesis and a loss of one for each wrong decision, the Bayes risk divided by $\kappa$ is the objective of Naghshvar and Javidi \cite{naghshvar2013active}, the expected sample size plus $1/\kappa$ times the probability of a wrong decision.
The corollary is then the first-order term of their lower bound \cite[(6)]{naghshvar2013active}, whose denominator for the hypothesis of $\nu$ is $T^\star_\Theta(\nu)^{-1}$.
Their remainder, of order $(\log(1/\kappa))^{2/3}$, uses a bound on the log-likelihood ratios \cite[Assumption~2]{naghshvar2013active}, and the chain leaves it as $o(1)$.

A procedure can instead be held to a risk $\delta$ at every state.
The set $\Theta$ of states may then be any set of instances over the experiments in $\mathcal{E}$, with $\KL(\nu_e \,\|\, \lambda_e)$ finite for all states $\nu$ and $\lambda$.
It is partitioned into hypotheses.
For $\nu$ in the hypothesis $\Theta_i$, the alternative set is $\Alt_\Theta(\nu) := \Theta \setminus \Theta_i$, and
\[
T^\star_\Theta(\nu)^{-1} \;:=\; \sup_{w} \inf_{\lambda \in \Alt_\Theta(\nu)} \sum_{e} w_e\, \KL(\nu_e \,\|\, \lambda_e)
\]
which is the definition above when $\Theta$ is finite and has two hypotheses.
The procedure is $\delta$-correct if, for every $\lambda \in \Theta$, it stops and decides the hypothesis containing $\lambda$ with $P_\lambda$-probability at least $1 - \delta$.

\begin{proposition}[Sequential design at fixed confidence]
\label{prop:cl-design}
Let $\nu \in \Theta$ satisfy $T^\star_\Theta(\nu) < \infty$.
If $\delta \in (0, 1/2]$, the procedure is $\delta$-correct and $\mathbb{E}_\nu[\tau] < \infty$, then $\mathbb{E}_\nu[\tau] \ge T^\star_\Theta(\nu)\, \kl(\delta, 1-\delta)$.
\end{proposition}

The proof is in Appendix~\ref{app:proofs-classical}.
The states may form a continuum, so the infimum over $\Alt_\Theta(\nu)$ need not be attained, and the proof reaches it by a \textsc{boundary} step.
Kaufmann and Koolen \cite[Proposition~16]{kaufmann2021mixture} prove $\mathbb{E}_\nu[\tau] \ge T^\star_\Theta(\nu) \log(1/(3\delta))$ for every $\delta$-correct procedure.
In their setting the experiments are the arms of a one-parameter exponential family, the states are the vectors of the arms' means, and the hypotheses are finitely many open sets.
Their $\delta$-correctness is the one above.
The second bound of Lemma~\ref{lem:relax}(c) gives $\kl(\delta, 1-\delta) = \kl(1-\delta, \delta) \ge (1-\delta)\log(1/\delta) - \log 2$.
This exceeds $\log(1/(3\delta))$ by $\log(3/2) - \delta\log(1/\delta)$, which is positive because $\delta\log(1/\delta) \le 1/e < \log(3/2)$ (\textsc{binary}).
Their bound is trivial when $3\delta \ge 1$.
Their hypotheses are open, so $\nu$ lies at a positive distance from $\Alt_\Theta(\nu)$ and $T^\star_\Theta(\nu) < \infty$.
The proposition therefore contains their bound, with $\kl(\delta, 1-\delta)$ in place of $\log(1/(3\delta))$.
With the instances that have a unique optimal arm as states and their optimal arms as hypotheses, it gives Proposition~\ref{prop:cl-bai}.
There $T^\star_\Theta(\nu) = T^\star(\nu)$, by the density argument in the proof of Proposition~\ref{prop:cl-bai}.
An identification algorithm that is $\delta$-correct in the sense of Section~\ref{sec:alternative} stops almost surely, so it is $\delta$-correct as a procedure.
With a single experiment the allocation is fixed, and $T^\star_\Theta(\nu)^{-1} = \Kinf(\nu, \Alt_\Theta(\nu))$.

\subsection{Best-arm and top-\texorpdfstring{$m$}{m} identification}
\label{app:cl-bai}

Let the arms belong to a one-parameter exponential family.
An algorithm with stopping time $\tau$ and recommendation $\hat a$ is $\delta$-correct if $P_\lambda(\tau < \infty) = 1$ and $P_\lambda(\hat a \ne a^\star(\lambda)) \le \delta$ for every instance $\lambda$ with a unique optimal arm $a^\star(\lambda)$.

\begin{proposition}[Best-arm identification bound]
\label{prop:cl-bai}
Let $\nu$ have a unique optimal arm.
If $\delta \in (0, 1/2]$, the algorithm is $\delta$-correct and $\mathbb{E}_\nu[\tau] < \infty$, then $\mathbb{E}_\nu[\tau] \ge T^\star(\nu)\, \kl(\delta, 1-\delta)$.
\end{proposition}

This is Theorem~1 of Garivier and Kaufmann \cite{garivier2016optimal}.
They state it for $\delta \in (0,1)$.
It needs $\delta \le 1/2$: with two close arms and $\delta > 1/2$, pulling one arm once and then recommending an arm at random is $\delta$-correct and violates the bound.

The proof is in Appendix~\ref{app:proofs-classical}.
Proposition~\ref{prop:cl-own} writes the excess of $\mathbb{E}_\nu[\tau]$ over this limit exactly, as $T^\star(\nu)$ times the sum of the \textsc{process}, \textsc{uniform} and \textsc{allocation} defects at the alternative that attains the infimum at the algorithm's average allocation $\bar w$.

For top-$m$ identification, order the means $\mu_{[1]} \ge \dots \ge \mu_{[K]}$ with $\mu_{[m]} > \mu_{[m+1]}$, let $\mathcal{S}^\star$ be the $m$ best arms, and call an algorithm $\delta$-correct if on every such instance it stops almost surely and recommends the top-$m$ set with probability at least $1 - \delta$.

\begin{proposition}[Top-$m$ identification bound]
\label{prop:cl-topm}
If $\delta \in (0, 1/2]$, the algorithm is $\delta$-correct for top-$m$ identification and $\mathbb{E}_\nu[\tau] < \infty$, then
\[
\mathbb{E}_\nu[\tau] \;\ge\; \Big( \sum_{a \in \mathcal{S}^\star} \frac{1}{d(\mu_a, \mu_{[m+1]})} + \sum_{a \notin \mathcal{S}^\star} \frac{1}{d(\mu_a, \mu_{[m]})} \Big) \kl(\delta, 1-\delta)
\]
\end{proposition}

With $\kl(\delta, 1-\delta) \ge \log(1/(2.4\delta))$ \cite[(3)]{kaufmann2016journal} (\textsc{binary}) this is Theorem~4 of Kaufmann, Capp\'e and Garivier \cite{kaufmann2016journal}, which is stated for $\delta \le 0.15$.
That inequality holds for every $\delta \in (0,1)$.
Its right side is non-positive for $\delta \ge 1/2.4$.
For smaller $\delta$ it reads $(1-2\delta)\log\frac{1}{1-\delta} + 2\delta\log\frac{1}{\delta} \le \log 2.4$.
The left side has second derivative $-2/\delta - 2/(1-\delta) - 1/(1-\delta)^2 < 0$, so it lies below its tangent line at $\delta = 0.32$.
On $(0, 1/2.4]$ that line stays below $0.872$, and $\log 2.4 > 0.875$.

The proof is in Appendix~\ref{app:proofs-classical}.
The single-arm alternatives of its proof give an explicit bound.
The whole alternative set gives the exact bound.
Write $\Alt_m(\nu)$ for the set of instances under which some arm outside $\mathcal{S}^\star$ has a mean at least that of some arm of $\mathcal{S}^\star$, so that $\mathcal{S}^\star$ is not their unique top-$m$ set.
Its characteristic time $T^\star_m(\nu)$ satisfies $T^\star_m(\nu)^{-1} = \sup_w \inf_{\lambda \in \Alt_m(\nu)} \langle w, k(\lambda)\rangle$, with $k(\lambda)$ the vector of per-pull divergences of Section~\ref{sec:allocation}.

\begin{proposition}[Top-$m$ identification bound at the exact characteristic time]
\label{prop:cl-topm-exact}
Let the arms belong to a one-parameter exponential family, and let $\nu$ have a unique top-$m$ set $\mathcal{S}^\star$.
\begin{enumerate}
\item[(i)] For every allocation $w$,
\[
\inf_{\lambda \in \Alt_m(\nu)} \langle w, k(\lambda)\rangle \;=\; \min_{a \in \mathcal{S}^\star,\ b \notin \mathcal{S}^\star}\ \min_{\mu'} \big[w_a\, d(\mu_a, \mu') + w_b\, d(\mu_b, \mu')\big]
\]
and when $w_a + w_b > 0$ the inner minimum is attained at $\mu' = (w_a\mu_a + w_b\mu_b)/(w_a + w_b)$.
\item[(ii)] If $\delta \in (0, 1/2]$, the algorithm is $\delta$-correct for top-$m$ identification and $\mathbb{E}_\nu[\tau] < \infty$, then $\mathbb{E}_\nu[\tau] \ge T^\star_m(\nu)\, \kl(\delta, 1-\delta)$.
\item[(iii)] $T^\star_m(\nu) \ge \sum_{a \in \mathcal{S}^\star} 1/d(\mu_a, \mu_{[m+1]}) + \sum_{a \notin \mathcal{S}^\star} 1/d(\mu_a, \mu_{[m]})$.
\end{enumerate}
\end{proposition}

Part (i) is Proposition~\ref{prop:altsets} for the union of the single-swap half-spaces, with the pairwise infimum of Proposition~\ref{prop:mbary}.
For one-parameter exponential families, (i) and the bound (ii) appear in \cite[Lemma~33 and Appendix~D]{you2022informationdirected}.
For unit-variance Gaussian arms in a possibly misspecified linear model, the reduction to single-swap half-spaces and the bound with $\log(1/(2.4\delta))$ appear in \cite[Lemma~1 and Theorem~1]{reda2021dealing}.
By (iii), the bound (ii) is never weaker than that of Proposition~\ref{prop:cl-topm}.

The proof is in Appendix~\ref{app:proofs-classical}.

\begin{proposition}[Identification at the algorithm's own error rates]
\label{prop:cl-own}
Let the arms belong to a one-parameter exponential family, let $\tau$ be a stopping time with $\mathbb{E}_\nu[\tau] < \infty$, and write $\bar w := \mathbb{E}_\nu[N(\tau)]/\mathbb{E}_\nu[\tau]$.
\begin{enumerate}
\item[(i)] Let $\nu$ have a unique optimal arm, let $E := \{\tau < \infty,\ \hat a = a^\star(\nu)\}$, and let $\lambda^\star \in \Alt(\nu)$ attain $\inf_{\lambda \in \Alt(\nu)} \langle \bar w, k(\lambda)\rangle$.
Then
\[
\mathbb{E}_\nu[\tau]\, \langle \bar w, k(\lambda^\star)\rangle \;=\; \kl\big(P_\nu(E), P_{\lambda^\star}(E)\big) + \mathbb{E}_\nu\!\left[\KL\big(P_\nu(\cdot \mid \mathbf{1}_E) \,\big\|\, P_{\lambda^\star}(\cdot \mid \mathbf{1}_E)\big)\right]
\]
and $\langle \bar w, k(\lambda^\star)\rangle \le T^\star(\nu)^{-1}$, so that
\[
\begin{aligned}
\frac{\mathbb{E}_\nu[\tau]}{T^\star(\nu)} \;=\;& \kl\big(P_\nu(E), P_{\lambda^\star}(E)\big) + \mathbb{E}_\nu\!\left[\KL\big(P_\nu(\cdot \mid \mathbf{1}_E) \,\big\|\, P_{\lambda^\star}(\cdot \mid \mathbf{1}_E)\big)\right] \\
&+ \mathbb{E}_\nu[\tau]\big(T^\star(\nu)^{-1} - \langle \bar w, k(\lambda^\star)\rangle\big)
\end{aligned}
\]
with each of the three terms on the right non-negative.
In particular $\mathbb{E}_\nu[\tau] \ge T^\star(\nu)\, \kl(P_\nu(E), P_{\lambda^\star}(E))$.
\item[(ii)] Part (i) holds for top-$m$ identification of a unique top-$m$ set $\mathcal{S}^\star$, with $E := \{\tau < \infty,\ \text{the recommendation is } \mathcal{S}^\star\}$, $\Alt_m(\nu)$ in place of $\Alt(\nu)$ and $T^\star_m(\nu)$ in place of $T^\star(\nu)$.
\item[(iii)] In (i) and (ii), $P_{\lambda^\star}(E) = \mathbb{E}_\nu[\mathbf{1}_E\, e^{-L_\tau}]$, where $L_\tau$ is the log-likelihood ratio of $P_\nu$ to $P_{\lambda^\star}$ on $\mathcal{F}_\tau$.
\end{enumerate}
\end{proposition}

The proof is in Appendix~\ref{app:proofs-classical}.
Such a $\lambda^\star$ exists by Proposition~\ref{prop:cl-topm-exact}(i), applied with $m = 1$ for best-arm identification.
One such alternative moves the best arm and one challenger, or one arm of $\mathcal{S}^\star$ and one arm outside it, to a common mean.
That mean is their $\bar w$-weighted mean unless both of their weights are zero.
The first display is the operating-characteristic identity at the algorithm's own error probabilities.
It needs no correctness guarantee.
In the second display the first term is the bound at those error probabilities, the second the \textsc{process} defect and the third the \textsc{allocation} defect.
If the algorithm is $\delta$-correct with $\delta \le 1/2$, then $P_\nu(E) \ge 1-\delta$, and $P_{\lambda^\star}(E) \le \delta$ by Fatou's lemma along instances in the alternative set with a unique answer that converge to $\lambda^\star$.
The difference $\kl(P_\nu(E), P_{\lambda^\star}(E)) - \kl(1-\delta, \delta)$ is then the \textsc{uniform} defect, so the proposition splits the excess of $\mathbb{E}_\nu[\tau]$ over the limit of Proposition~\ref{prop:cl-bai} or Proposition~\ref{prop:cl-topm-exact} exactly.
At any other alternative of the set the identity still holds, and a fourth term, the \textsc{boundary} defect, enters with a minus sign.
For a sequential test one stream is sampled and the alternative set is one instance, so no allocation term arises.
The identity is then that of Proposition~\ref{prop:cl-wald}.
Its bound at the realized error probabilities is $\kl(1-\beta', \alpha')/\KL(\nu \,\|\, \lambda)$.
Part (iii) computes $P_{\lambda^\star}(E)$ from runs under $\nu$.

\subsection{Characteristic time and optimal weights}
\label{app:cl-weights}

\begin{proposition}[Characteristic time and optimal weights]
\label{prop:cl-weights}
Let the arms belong to a one-parameter exponential family and let $\nu$ have a unique optimal arm $a^\star$.
\begin{enumerate}
\item[(i)] For every allocation $w$ with $w_{a^\star} > 0$, $\inf_{\lambda \in \Alt(\nu)} \sum_a w_a\, d(\mu_a, \mu_a(\lambda)) = \min_{b \ne a^\star} w_{a^\star}\, G_b(w_b / w_{a^\star})$, with $G_b$ as in Proposition~\ref{prop:scalar-general}.
\item[(ii)] At the optimal allocation the $K-1$ terms of (i) are equal.
\item[(iii)] Equation~\eqref{eq:scalar-general} holds.
\end{enumerate}
\end{proposition}

These are Lemma~3, Lemma~4 and Theorem~5 of Garivier and Kaufmann \cite{garivier2016optimal}.
In their notation, with their arm $1$ as $a^\star$ and their arm $a$ as $b$, $g_a(x) = (1+x)\, I_{1/(1+x)}(\mu_1, \mu_a)$ equals $G_b(x)$, $x_a(y)$ equals $G_b^{-1}(\gamma)$ at $y = \gamma$, and the scalar program of (iii) is the reciprocal of the maximization of $y/(1 + \sum_a x_a(y))$ in their Appendix~A.2.
For Gaussian arms with a common variance the condition $w_{a^\star}^2 = \sum_b w_b^2$ \cite[Example~1]{glynn2004large}, \cite[p.~6]{garivier2016optimal} is the stationarity condition of Proposition~\ref{prop:scalar}.

The proof is in Appendix~\ref{app:proofs-classical}.
No relaxation enters this row: the characteristic time is exact.

\subsection{Coincidence divergence and Bregman information}
\label{app:cl-jensen}

Section~\ref{sec:alternative} compares the two Jensen defects of an exponential family at matching weights.
For arms $\nu_a$ and $\lambda_a$ in one one-parameter exponential family and $w \in [0,1]$, write
\[
\mathcal{J}_a(w) \;:=\; w\, \psi^\star(\mu_a) + (1-w)\, \psi^\star(\mu_a(\lambda)) - \psi^\star\big(w \mu_a + (1-w) \mu_a(\lambda)\big)
\]
for the Bregman information \eqref{eq:mbary} of the two means at weights $(w, 1-w)$.

\begin{lemma}[Coincidence divergence and Bregman information]
\label{lem:jensen-pair}
Let $\nu_a$ and $\lambda_a$ lie in one one-parameter exponential family, with natural parameters $\theta_a$ and $\eta_a$ interior to its natural parameter space, and fix $s \in (0,1)$.
\begin{enumerate}
\item[(i)] As $\eta_a \to \theta_a$,
\begin{gather*}
c_a(s) = \tfrac12 s(1-s)\, \psi''(\theta_a)\, (\eta_a - \theta_a)^2 + O\big((\eta_a - \theta_a)^3\big) \\
c_a(s) - \mathcal{J}_a(s) = \tfrac16 s(1-s)(1-2s)\, \psi'''(\theta_a)\, (\eta_a - \theta_a)^3 + O\big((\eta_a - \theta_a)^4\big)
\end{gather*}
\item[(ii)] If the laws are Gaussian with common variance $\sigma^2$ and means $g$ apart, then $c_a(s) = \mathcal{J}_a(s) = s(1-s)g^2/2\sigma^2$.
\item[(iii)] If the laws are exponential, then $c_a(s) = \mathcal{J}_a(1-s)$.
\item[(iv)] If the laws are Bernoulli, then $\mathcal{J}_a(s) \le -s\log s - (1-s)\log(1-s) \le \log 2$.
If moreover $\mu_a = q$ and $\mu_a(\lambda) = 1-q$ with $q \in (0,1)$, then
\[
\underline{s}\log\frac1q - \log 2 \;\le\; c_a(s) \;\le\; \underline{s}\log\frac1q + \log\frac{1}{1-q}
\]
with $\underline{s} := s \wedge (1-s)$.
\end{enumerate}
\end{lemma}

The proof is in Appendix~\ref{app:proofs-classical}.

\subsection{Identification within a tolerance}
\label{app:cl-tolerance}

Let the arms belong to a one-parameter exponential family, and let $\Delta_{\mathrm{tol}} > 0$.
An algorithm with stopping time $\tau$ and recommendation $\hat a$ is $(\Delta_{\mathrm{tol}}, \delta)$-correct if $P_\lambda(\tau < \infty) = 1$ and $P_\lambda(\Delta_{\hat a}(\lambda) < \Delta_{\mathrm{tol}}) \ge 1 - \delta$ for every instance $\lambda$.
The arms with $\Delta_a < \Delta_{\mathrm{tol}}$ form the set $\mathcal{S}_{\mathrm{tol}}$ of arms acceptable under $\nu$.
For each arm $a$ let $\lambda^{(a)}$ equal $\nu$ except that arm $a$ has mean $\mu^\star + \Delta_{\mathrm{tol}}$.
Under $\lambda^{(a)}$ arm $a$ is the only acceptable arm.

\begin{proposition}[Identification within a tolerance]
\label{prop:cl-tolerance}
Suppose that $\mu^\star + \Delta_{\mathrm{tol}}$ is the mean of a law of the family, that $\delta \in (0, 1/2]$, and that the algorithm is $(\Delta_{\mathrm{tol}}, \delta)$-correct with $\mathbb{E}_\nu[\tau] < \infty$.
\begin{enumerate}
\item[(i)] Every arm $a \notin \mathcal{S}_{\mathrm{tol}}$ has $\mathbb{E}_\nu[N_a(\tau)]\, k_a(\lambda^{(a)}) \ge \kl(\delta, 1-\delta)$.
\item[(ii)] Every arm $a \in \mathcal{S}_{\mathrm{tol}}$ has $\mathbb{E}_\nu[N_a(\tau)]\, k_a(\lambda^{(a)}) \ge \big(1 - P_\nu(\hat a = a)\big)\log(1/\delta) - \log 2$.
\item[(iii)] The expected sample size satisfies
\[
\mathbb{E}_\nu[\tau] \;\ge\; \kl(\delta, 1-\delta) \sum_{a \notin \mathcal{S}_{\mathrm{tol}}} \frac{1}{k_a(\lambda^{(a)})}
 + \bigg( \log\frac{1}{\delta} \Big( \sum_{a \in \mathcal{S}_{\mathrm{tol}}} \frac{1}{k_a(\lambda^{(a)})} - \max_{a \in \mathcal{S}_{\mathrm{tol}}} \frac{1}{k_a(\lambda^{(a)})} \Big) - \log 2 \sum_{a \in \mathcal{S}_{\mathrm{tol}}} \frac{1}{k_a(\lambda^{(a)})} \bigg)^{\!+}
\]
\end{enumerate}
\end{proposition}

Here $k_a(\lambda^{(a)}) = d(\mu_a, \mu^\star + \Delta_{\mathrm{tol}})$, the per-pull divergence of Section~\ref{sec:allocation}.
For Bernoulli arms with means in $(0, 1/2]$, Theorem~5 of Mannor and Tsitsiklis \cite{mannor2004sample} bounds $\mathbb{E}_\nu[\tau]$ below by a constant multiple of $\log(1/(8\delta))$ times a sum of two terms.
The first is the positive part of the number of acceptable arms less one, over $\Delta_{\mathrm{tol}}^2$.
The second is the sum of $\Delta_a^{-2}$ over the arms outside $\mathcal{S}_{\mathrm{tol}}$.
Both counts omit arms with small means.
Parts (i) and (ii), applied arm by arm after an upper bound on each per-pull divergence such as $\kl(u, v) \le (u - v)^2/(v(1-v))$ at $v = \mu^\star + \Delta_{\mathrm{tol}}$, take that form.
The first term also uses that the probabilities $P_\nu(\hat a = a)$ sum to at most one.
Here $\mu^\star + \Delta_{\mathrm{tol}} - \mu_a$ is below $2\Delta_{\mathrm{tol}}$ for an acceptable arm and at most $2\Delta_a$ for the others.
The restrictions of Mannor and Tsitsiklis on the means bound $v(1-v)$ below by a positive number.
Their bound then holds with a quarter of that number as the constant, for every $\delta \in (0, 1/2]$.
Their theorem takes $\delta$ below $e^{-8}/8$.
An upper bound on a per-pull divergence is not among the relaxations of Section~\ref{sec:classical}.
Its defect, the excess of the bound over the divergence, is exact.

Consider Bernoulli arms with one mean $1/2 + \Delta_{\mathrm{tol}}$ and $K - 1$ means $1/2$, with $\Delta_{\mathrm{tol}} \le 1/8$.
There part (i), summed over the $K - 1$ arms with mean $1/2$, gives $\mathbb{E}_\nu[\tau] \ge 2(K-1)\kl(\delta, 1-\delta)/\log\big(1/(1 - 16\Delta_{\mathrm{tol}}^2)\big)$, which is at least $\tfrac{3}{32}(K-1)\Delta_{\mathrm{tol}}^{-2}\kl(\delta, 1-\delta)$.
This is the form of Theorem~1 of Mannor and Tsitsiklis \cite{mannor2004sample}, with explicit constants.

Summed over the arms outside $\mathcal{S}_{\mathrm{tol}}$, part (i) with $\kl(\delta, 1-\delta) \ge \log(1/(2.4\delta))$ (\textsc{binary}) gives the second term of the bound in Remark~5 of Kaufmann, Capp\'e and Garivier \cite{kaufmann2016journal}.
The first term of that bound, with denominator $d(\mu^\star, \mu^\star - \Delta_{\mathrm{tol}})$, is not reached.
Raising an acceptable arm to $\mu^\star + \Delta_{\mathrm{tol}}$ costs $k_a(\lambda^{(a)})$ per pull.
For Bernoulli arms with $\mu^\star > 1/2$ this cost exceeds that denominator.
Lowering an acceptable arm $a \ne a^\star$ to $\mu^\star - \Delta_{\mathrm{tol}}$ costs at most that denominator per pull.
It rules out only the recommendation of $a$.
The bounds from these alternatives are weighted by the probabilities $P_\nu(\hat a = a)$, which sum to at most one.
The remark is stated without proof.

The proof is in Appendix~\ref{app:proofs-classical}.

\subsection{Fixed-budget identification}
\label{app:cl-fixed-budget}

A fixed-budget algorithm samples for $T$ rounds and then recommends a set $\hat{\mathcal{S}}_T$ of $m$ arms, drawn given $\mathcal{F}_T$ with probabilities that do not depend on the instance.
For $m = 1$ it recommends one arm $\hat a_T$.
On an instance $\lambda$ with a unique top-$m$ set $\mathcal{S}^\star(\lambda)$ its error probability is $P_\lambda(\hat{\mathcal{S}}_T \ne \mathcal{S}^\star(\lambda))$.
For two laws $P$ and $Q$ write $c_{P,Q}(s) := -\log \int P^s Q^{1-s}$, so that $c_a(s) = c_{\nu_a, \lambda_a}(s)$.
Its maximum over $s \in [0,1]$ is the Chernoff information of $P$ and $Q$ (Section~\ref{sec:coincidence}).

\begin{proposition}[Two-point fixed-budget bound]
\label{prop:cl-fb-twopoint}
Let $\nu$ and $\lambda$ have unique top-$m$ sets $\mathcal{S}^\star(\nu) \ne \mathcal{S}^\star(\lambda)$, and let $Q_{1/2}$ and $P^{(1/2)}$ be as in Propositions~\ref{prop:tilt} and~\ref{prop:half}.
\begin{enumerate}
\item[(i)] Every algorithm with budget $T$ satisfies
\[
\max\Big( P_\nu\big(\hat{\mathcal{S}}_T \ne \mathcal{S}^\star(\nu)\big),\ P_\lambda\big(\hat{\mathcal{S}}_T \ne \mathcal{S}^\star(\lambda)\big) \Big)
 \;\ge\; \tfrac12\Big(1 - \sqrt{1 - e^{-2C_T(1/2)}}\Big) \;\ge\; \tfrac14\, e^{-2C_T(1/2)} \;\ge\; \tfrac14\, e^{-D_T}
\]
\item[(ii)] Every algorithm satisfies
\[
T \max_{a \in [K]} c_a(1/2) - C_T(1/2) \;=\; \KL\!\left(Q_{1/2} \,\big\|\, P^{(1/2)}\right) + \sum_{a \in [K]} \mathbb{E}_{Q_{1/2}}[N_a(T)] \Big( \max_{a' \in [K]} c_{a'}(1/2) - c_a(1/2) \Big)
\]
\end{enumerate}
\end{proposition}

The last inequality of (i) is Lemma~15 of Kaufmann, Capp\'e and Garivier \cite{kaufmann2016journal}.
Proposition~\ref{prop:half} gives $2C_T(1/2) \le D_T$.
The midpoint form is therefore never weaker than Lemma~15.
With deterministic pull counts the first term of (ii) vanishes.
Then $C_T(1/2) = \sum_a N_a(T)\, c_a(1/2)$.
For Gaussian arms with a common variance the per-pull factor is $2$ (Proposition~\ref{prop:ratio}).
With deterministic counts the exponent of the midpoint form is then half that of Lemma~15.
A \textsc{test} step and an \textsc{allocation} step drop the two terms of (ii).
Every algorithm then has an error probability of at least $\tfrac14 e^{-2T\max_a c_a(1/2)}$ on one of the two instances.
The \textsc{allocation} step applied to Lemma~15 gives the weaker bound $\tfrac14 e^{-T\max_a \KL(\nu_a \| \lambda_a)}$, since $2c_a(1/2) \le \KL(\nu_a \,\|\, \lambda_a)$ for every arm.
Since $\min(u, v) \le u^s v^{1-s}$, the exact minimum error sum $1 - \TV(P_\nu, P_\lambda)$ is at most $e^{-C_T(s)}$ at every tilt $s \in [0,1]$.
So $2C_T(1/2) \le -2\log\big(1 - \TV(P_\nu, P_\lambda)\big)$.
The exponent of the midpoint form is therefore at most twice that of the minimum error sum.
No such factor holds for $D_T$.
For $T = 1$ and an arm whose laws under $\nu$ and $\lambda$ are Bernoulli with means $1/2$ and $q \le 1/2$, the minimum error sum is $1/2 + q$, while $D_T = \kl(1/2, q)$ grows without bound as $q \to 0$.

The proof is in Appendix~\ref{app:proofs-classical}.

\begin{proposition}[Gaussian fixed-budget bounds]
\label{prop:cl-fb-gauss}
Let the arms be Gaussian with a common variance, and let an algorithm have budget $T$.
\begin{enumerate}
\item[(i)] Let $\nu$ have a unique optimal arm $a^\star$, and for each suboptimal arm $b$ let $\lambda^{(b)}$ equal $\nu$ except that arm $b$ has mean $\mu_b + 2\Delta_b$.
Every $\lambda^{(b)}$ has the unique optimal arm $b$.
Some suboptimal arm $b$ satisfies
\[
\max\Big( P_\nu\big(\hat a_T \ne a^\star\big),\ P_{\lambda^{(b)}}\big(\hat a_T \ne b\big) \Big)
 \;\ge\; \tfrac14 \exp\Big( -\frac{4T}{\sum_{b' \ne a^\star} 1/d_{b'}} \Big)
\]
\item[(ii)] Let $\nu$ have a unique top-$m$ set $\mathcal{S}^\star$, with $1 \le m < K$, and write $d^{(m)}_c := d(\mu_c, \mu_{[m+1]})$ for $c \in \mathcal{S}^\star$ and $d^{(m)}_c := d(\mu_c, \mu_{[m]})$ for $c \notin \mathcal{S}^\star$.
For $a \in \mathcal{S}^\star$ and $b \notin \mathcal{S}^\star$ let $\lambda^{(a,b)}$ equal $\nu$ except that arm $a$ has mean $2\mu_{[m+1]} - \mu_a$ and arm $b$ has mean $2\mu_{[m]} - \mu_b$.
Every $\lambda^{(a,b)}$ has the unique top-$m$ set $\mathcal{S}^{(a,b)} := (\mathcal{S}^\star \setminus \{a\}) \cup \{b\}$.
Some such pair satisfies
\[
\begin{aligned}
&\max\Big( P_\nu\big(\hat{\mathcal{S}}_T \ne \mathcal{S}^\star\big),\ P_{\lambda^{(a,b)}}\big(\hat{\mathcal{S}}_T \ne \mathcal{S}^{(a,b)}\big) \Big) \\
&\qquad \ge\; \tfrac14 \exp\bigg( -\frac{4T}{\sum_{c \in [K]} 1/d^{(m)}_c} - \frac{4T}{\min\big(\sum_{c \in \mathcal{S}^\star} 1/d^{(m)}_c,\ \sum_{c \notin \mathcal{S}^\star} 1/d^{(m)}_c\big)} \bigg)
\end{aligned}
\]
\end{enumerate}
\end{proposition}

For Gaussian arms with variance $\sigma^2$, $1/d_b = 2\sigma^2/\Delta_b^2$, and $1/d^{(m)}_c$ is $2\sigma^2$ over the squared distance of $\mu_c$ from the boundary mean on the other side.
Part (i) is Theorem~16 of Kaufmann, Capp\'e and Garivier \cite{kaufmann2016journal}, whose $H'(\nu)$ is the sum of the $1/d_{b'}$.
Part (ii) is their Theorem~17, whose $H(\nu)$, $H^+(\nu)$ and $H^-(\nu)$ are the three sums of the $1/d^{(m)}_c$.
As printed, Theorem~16 omits the factor $\tfrac14$ of Lemma~15.
Only the proof of Theorem~17 is printed.
The factor appears there.
Without that factor Theorem~16 fails.
With two arms, pulling the optimal arm once and recommending it when the observation exceeds its mean makes both error probabilities $1/2$, while $e^{-4T/H'(\nu)}$ tends to $1$ at $T = 1$ as the gap shrinks.
Theorem~17 as printed moves arm $a$ by $2(\mu_{[m]} - \mu_b)$ and arm $b$ by $2(\mu_a - \mu_{[m+1]})$, while its proof uses the divergences of the moves in part (ii).
Each alternative is no harder than $\nu$, as both theorems state.
The sum of the $1/d_{b'}$ does not increase from $\nu$ to $\lambda^{(b)}$, because there the old optimal arm trails arm $b$ by $\Delta_b$ and every other suboptimal arm $c$ trails arm $b$ by $\Delta_c + \Delta_b$.
The sum of the $1/d^{(m)}_c$ decreases from $\nu$ to $\lambda^{(a,b)}$, because arms $a$ and $b$ end farther from the new boundary means and no other arm comes closer to the boundary.

With deterministic pull counts the pigeonhole step of the proof applies to the midpoint form of Proposition~\ref{prop:cl-fb-twopoint} as well.
There it halves both exponents.
For an adaptive algorithm it does not apply, because the tilted law $Q_{1/2}$ of the pair $(\nu, \lambda^{(b)})$ changes with $b$ and the expected pull counts under these laws need not sum to at most $T$.
Without the pigeonhole step, Proposition~\ref{prop:cl-fb-twopoint}(ii) applied to $\lambda^{(b)}$ gives every algorithm the bound $\tfrac14 e^{-2T d_b}$, since $c_b(1/2) = d_b$ there.
At the suboptimal arm $b$ closest to the best this bound is stronger than part (i) exactly when $\sum_{b' \ne a^\star} d_b/d_{b'} < 2$, as always for two arms.

The proof is in Appendix~\ref{app:proofs-classical}.

\begin{proposition}[Fixed-budget exponent]
\label{prop:cl-fb-rate}
Let the arms belong to a one-parameter exponential family, and let $\nu$ have a unique optimal arm $a^\star$.
\begin{enumerate}
\item[(i)] Suppose that the algorithm, run with each budget $T$, satisfies $P_\lambda(\hat a_T \ne a^\star(\lambda)) \to 0$ as $T \to \infty$ for every instance $\lambda$ with a unique optimal arm.
Then
\[
\limsup_{T \to \infty} -\frac1T \log P_\nu\big(\hat a_T \ne a^\star\big) \;\le\; \inf_{\lambda \in \Alt(\nu)} \max_{a \in [K]} \KL(\lambda_a \,\|\, \nu_a)
\]
If moreover its pull counts are deterministic, with $N_a(T)/T \to w_a$ for every arm, then the right side may be replaced by $\inf_{\lambda \in \Alt(\nu)} \sum_a w_a \KL(\lambda_a \,\|\, \nu_a)$.
\item[(ii)] For every arm $b$, every mean $\mu'$ of the family and every $s \in [0,1]$,
\[
s\, d(\mu', \mu_{a^\star}) + (1-s)\, d(\mu', \mu_b) \;=\; c_{\nu_{a^\star}, \nu_b}(s) + d\big(\mu', \psi'(s\theta_{a^\star} + (1-s)\theta_b)\big)
\]
\item[(iii)] For every allocation $w$, with the term of $b$ read as $0$ when $w_{a^\star} = w_b = 0$,
\[
\inf_{\lambda \in \Alt(\nu)} \sum_a w_a \KL(\lambda_a \,\|\, \nu_a) \;=\; \min_{b \ne a^\star} (w_{a^\star} + w_b)\, c_{\nu_{a^\star}, \nu_b}\Big(\frac{w_{a^\star}}{w_{a^\star} + w_b}\Big)
\]
and
\[
\inf_{\lambda \in \Alt(\nu)} \max_{a \in [K]} \KL(\lambda_a \,\|\, \nu_a) \;=\; \min_{b \ne a^\star} \max_{s \in [0,1]} c_{\nu_{a^\star}, \nu_b}(s)
\]
\end{enumerate}
\end{proposition}

For two arms, part (i) is Theorem~12 of Kaufmann, Capp\'e and Garivier \cite{kaufmann2016journal}.
The constant of their first bound is the first right side.
The constant of their bound for uniform sampling is the second at $w = (1/2, 1/2)$.
By (iii) these are the Chernoff information and the Bhattacharyya divergence $c_{\nu_1, \nu_2}(1/2)$ of the laws of the two arms, as they note in their own terms \cite[Section~5.1 and Appendix~C]{kaufmann2016journal}.
Part (ii) is the identity of van Erven and Harremo\"es \cite[(53)]{erven2014b} quoted in Section~\ref{sec:limit}, for the laws $\nu_{a^\star}$ and $\nu_b$ and the law of the family with mean $\mu'$.
Its minimum over $\mu'$ is their Theorem~30 \cite{erven2014b}.
Under a fixed budget the divergences of the bound run from the alternative to the instance.
The alternative-set step therefore returns a coincidence divergence between the laws of two arms, where under a fixed confidence it returns the Bregman information of Proposition~\ref{prop:mbary}.
For a static allocation with every $w_a > 0$ and the empirical best arm recommended, Glynn and Juneja show that $-\frac1T \log P_\nu(\hat a_T \ne a^\star)$ converges \cite[Section~2]{glynn2004large}.
This rate is the minimum over $b \ne a^\star$ of the smallest value over $\mu'$ of a weighted sum of the Cram\'er rate functions of arms $a^\star$ and $b$ at $\mu'$, with weights $w_{a^\star}$ and $w_b$.
In a one-parameter exponential family the Cram\'er rate function of arm $c$ at $\mu'$ is $d(\mu', \mu_c)$.
By (iii) this rate equals the bound of (i) at a static allocation.
The bound is therefore attained by that procedure.

The coincidence family does not sharpen part (i).
At a tilt $s \in (0,1)$, data processing of $\int (dP_\nu)^s (dP_\lambda)^{1-s}$ to the recommendation gives $e^{-C_T(s)} \le P_\nu(\hat a_T \ne a^\star)^s + P_\lambda(\hat a_T \ne a^\star(\lambda))^{1-s}$.
The hypothesis of (i) gives no rate at which the second term vanishes.
Only the endpoint $s \to 0$ remains, which is the divergence decomposition read under $\lambda$.

The proof is in Appendix~\ref{app:proofs-classical}.

\subsection{Regret in ergodic Markov decision processes}
\label{app:cl-mdp}

Take the setting of Section~\ref{sec:mdp}, in which an instance $\nu$ assigns a law $\nu_{x,a}$ of the next state and the reward to every pair $(x,a) \in \mathcal{X} \times \mathcal{A}$.
Let $\Phi$ be a class of such instances.
The relative entropy at a pair is $\KL(\nu_{x,a} \,\|\, \lambda_{x,a})$, and the optimal actions $\mathcal{A}^\star_x$, the Bellman gaps $\Delta_{x,a}$ and the optimal stationary policies $\Pi^\star$ are those of Section~\ref{sec:mdp}.
The alternative set is $\Alt_\Phi(\nu) := \{\lambda \in \Phi : \nu \ll \lambda,\ \KL(\nu_{x,a} \,\|\, \lambda_{x,a}) = 0 \text{ for } a \in \mathcal{A}^\star_x,\ \Pi^\star \cap \Pi^\star(\lambda) = \emptyset\}$.
From an initial state, the regret $R_T(\lambda)$ of an algorithm on an instance $\lambda$ is $T$ times the optimal gain of $\lambda$ minus the expected total reward over $T$ rounds, and $R_T := R_T(\nu)$; for a bandit it is the regret of Section~\ref{sec:notation}.
An algorithm is uniformly good if its regret is $o(T^p)$ for every $p > 0$, every ergodic $\lambda \in \Phi$ and every initial state.

\begin{proposition}[Divergence decomposition on a Markov decision process]
\label{prop:cl-mdp-decomposition}
Let $\nu$ and $\lambda$ have the same initial-state law, and let $\KL(\nu_{x,a} \,\|\, \lambda_{x,a}) < \infty$ for every pair $(x,a)$.
For every algorithm and every stopping time $\tau$ with $\mathbb{E}_\nu[\tau] < \infty$,
\[
\KL\!\left(P_\nu|_{\mathcal{F}_\tau} \,\big\|\, P_\lambda|_{\mathcal{F}_\tau}\right)
 \;=\; \sum_{x \in \mathcal{X}} \sum_{a \in \mathcal{A}} \mathbb{E}_\nu[N_{x,a}(\tau)]\, \KL(\nu_{x,a} \,\|\, \lambda_{x,a})
\]
\end{proposition}

At a fixed horizon, and with the reward independent of the next state, this is step~(i) of the proof of Theorem~1 of Ok, Prouti\`ere and Tranos \cite[Appendix~C, (9)]{ok2018exploration}.
The proof is in Appendix~\ref{app:proofs-classical}.
It is the proof of Theorem~\ref{thm:allocation} with state-action pairs as arms.
The next state is random and its law depends on the action, but it belongs to the observation of its round, whose law given the history and the action is that of the instance at the pair.

\begin{proposition}[Regret in ergodic MDPs]
\label{prop:cl-mdp}
Let $\nu \in \Phi$ be ergodic and the algorithm uniformly good.
Then
\[
\liminf_{T \to \infty} \frac{R_T}{\log T} \;\ge\; c^\star_\Phi(\nu) := \inf_{\zeta} \sum_{x,a} \zeta(x,a)\, \Delta_{x,a}
\]
where the infimum is over the $\zeta \ge 0$ with $\sum_{x,a} \zeta(x,a) \KL(\nu_{x,a} \,\|\, \lambda_{x,a}) \ge 1$ for every $\lambda \in \Alt_\Phi(\nu)$.
\end{proposition}

This is Theorem~1 of Ok, Prouti\`ere and Tranos \cite{ok2018exploration}, which extends the bound of Burnetas and Katehakis \cite{burnetas1997optimal} for ergodic Markov decision processes; Graves and Lai \cite{graves1997asymptotically} proved the analogous bound for controlled Markov chains.
The proof uses two facts about ergodic instances, taken from \cite[Appendix C, (7) and (8)]{ok2018exploration}, where they are derived as in \cite{burnetas1997optimal}.
For every ergodic $\lambda \in \Phi$ and every algorithm, the regret has the count form $R_T(\lambda) = \sum_x \sum_{a \notin \mathcal{A}^\star_x(\lambda)} \mathbb{E}_\lambda[N_{x,a}(T)]\, \Delta_{x,a}(\lambda) + O(1)$.
The visits obey the visit bound: for every ergodic $\lambda \in \Phi$ and every algorithm there is $\chi > 0$ with $P_\lambda(N_x(T) \le \chi T) = O(e^{-\chi T/2})$ as $T \to \infty$, for every state $x$.

The proof is in Appendix~\ref{app:proofs-classical}.
Let $(x,a)$ be the pair whose law a single-pair alternative $\lambda$ changes.
With the fraction $Z := N_{x,a}(T)/N_x(T)$, and $Z := 0$ when $N_x(T) = 0$, in place of $\mathbf{1}_E$, step 2 of the proof becomes $\mathbb{E}_\nu[L_T] \ge \kl(\mathbb{E}_\nu[Z], \mathbb{E}_\lambda[Z])$ by Lemma~\ref{lem:relax}(b).
Step 3 is not needed.
Step 4 applies the first bound of Lemma~\ref{lem:relax}(c) to $\mathbb{E}_\nu[Z]$ and $\mathbb{E}_\lambda[Z]$.
A fixed alternative at a margin beyond the boundary adds a \textsc{boundary} step.
The suboptimal pairs with infinite $\Kinf$ enter through \textsc{allocation}.
Section~\ref{sec:eval-mdp-regret} evaluates this form.

The class $\Phi$ is a product class if each pair's law ranges over its own set: there are sets $\mathcal{M}_{x,a}$ of laws such that $\lambda \in \Phi$ exactly when $\lambda_{x,a} \in \mathcal{M}_{x,a}$ for every pair.
The class of all transition laws with known rewards, used in Section~\ref{sec:eval-mdp-regret}, is one: there $\mathcal{M}_{x,a}$ is the set of laws under which the next state has any law and the reward has the known law of the pair, independently of the next state.
So is the class in which the next state and the reward at every pair have arbitrary laws.

\begin{proposition}[Decoupled regret constant]
\label{prop:cl-mdp-decoupled}
Let $\nu \in \Phi$ be ergodic, and let $\Kinf(x,a)$ be as in Section~\ref{sec:mdp}, with $1/0 := \infty$ and $1/\infty := 0$.
Then
\[
c^\star_\Phi(\nu) \;\ge\; \sum_{x \in \mathcal{X}} \sum_{a \notin \mathcal{A}^\star_x} \frac{\Delta_{x,a}}{\Kinf(x,a)}
\]
with equality when $\Phi$ is a product class.
\end{proposition}

For the class of Ok, Prouti\`ere and Tranos, in which the next-state law and the reward law at every pair are arbitrary, the constraints of the program reduce to the single-pair alternatives by their Theorem~2 \cite{ok2018exploration}, which gives the equality.
They describe their Theorem~1 as extending the bound of Burnetas and Katehakis \cite{burnetas1997optimal} for unstructured ergodic instances with known rewards \cite[Section~4]{ok2018exploration}.
The proof is in Appendix~\ref{app:proofs-classical}.
Its equality part rests on display~(18) of the proof of their Theorem~2 \cite[Appendix~D]{ok2018exploration}, with the product structure in place of their decoupling lemma.
No relaxation enters this row: for a product class the decoupled constant is exact.
For a class that ties the pairs together the inequality can be strict.
In the linear class of Corollary~\ref{cor:cl-linear} with three pairwise independent feature vectors in the plane, and for an instance $\nu$ with a unique optimal arm, no other instance of the class differs from $\nu$ at one arm only, so the right side is $0$ while $c^\star_\Phi(\nu) > 0$.

\subsection{Structured, linear and Lipschitz bandits}
\label{app:cl-structured}

In a structured bandit problem the instance is known to lie in a class $\Phi$ of bandit instances.
For $\nu \in \Phi$ the alternative set $\Alt_\Phi(\nu)$ holds the $\lambda \in \Phi$ with $\nu \ll \lambda$ and with $\lambda_a = \nu_a$ for every optimal arm $a$ of $\nu$, under which no optimal arm of $\nu$ is optimal.

\begin{corollary}[Structured bandits]
\label{cor:cl-structured}
Let $\nu \in \Phi$, and let the algorithm be consistent on $\Phi$.
Then
\[
\liminf_{T \to \infty} \frac{R_T}{\log T} \;\ge\; c^\star_\Phi(\nu) := \inf_{\zeta} \sum_{a \in [K]} \zeta_a \Delta_a
\]
where the infimum is over the $\zeta \in [0,\infty)^K$ with $\sum_{a} \zeta_a \KL(\nu_a \,\|\, \lambda_a) \ge 1$ for every $\lambda \in \Alt_\Phi(\nu)$.
\end{corollary}

This is Proposition~\ref{prop:cl-mdp} with one state.
Its relaxations are those of that proposition.
With a unique optimal arm in every instance and mutually absolutely continuous laws, $\Alt_\Phi(\nu)$ is the set of confusing parameters of Combes, Magureanu and Prouti\`ere.
The corollary is then their Theorem~1 \cite{combes2017minimal} for a payoff equal to the mean of the observation, as in their classical, linear and Lipschitz examples.
They also allow a known payoff map whose values are not observed.
They derive the bound from that of Graves and Lai \cite{graves1997asymptotically}, with the arms as control laws that ignore the state \cite[Appendix~A]{combes2017minimal}.
For $\Phi = \mathcal{M}_1 \times \dots \times \mathcal{M}_K$, Proposition~\ref{prop:cl-mdp-decoupled} with one state gives $c^\star_\Phi(\nu) = c^\star(\nu)$.
The corollary is then the regret bound of Proposition~\ref{prop:cl-lai-robbins}.
The proof is in Appendix~\ref{app:proofs-classical}.

The general bound of Graves and Lai is not recovered.
In their framework, as restated by Combes, Magureanu and Prouti\`ere \cite[Appendix~A]{combes2017minimal}, a choice is one of finitely many stationary control laws, each of which makes the chain ergodic.
Regret is measured against the best of these laws, through their long-run average rewards under their stationary laws.
In Proposition~\ref{prop:cl-mdp-decomposition} visits to state-action pairs are counted.
In Proposition~\ref{prop:cl-mdp} regret is measured against an optimal policy of the Markov decision process.
Turning visit counts into uses of the control laws, weighted by their stationary laws, takes an ergodic average that is performed by none of the relaxations of Section~\ref{sec:classical}.
When every control law ignores the state and the law of the next state does not depend on the current one, every control law is an arm with independent observations, the case of Corollary~\ref{cor:cl-structured}.

\begin{corollary}[Linear bandits]
\label{cor:cl-linear}
Fix a feature vector $\varphi_a \in \mathbb{R}^d$ for each arm $a$, the vectors spanning $\mathbb{R}^d$.
Let $\Phi$ be the class of instances whose arms have unit-variance Gaussian laws with means $\langle \varphi_a, \omega \rangle$ for some $\omega \in \mathbb{R}^d$.
Let $\nu \in \Phi$ have a unique optimal arm $a^\star$, and let the algorithm be consistent on $\Phi$.
Then $\liminf_{T \to \infty} R_T/\log T \ge c^\star_\Phi(\nu)$, and $c^\star_\Phi(\nu)$ equals the infimum of $\sum_{b \ne a^\star} \zeta_b \Delta_b$ over the $\zeta \in [0,\infty)^K$ such that $V(\zeta) := \sum_{a} \zeta_a \varphi_a \varphi_a^\top$ is invertible and $\varphi_b^\top V(\zeta)^{-1} \varphi_b \le \Delta_b^2/2$ for every $b \ne a^\star$.
\end{corollary}

This is Corollary~2 of Lattimore and Szepesv\'ari \cite{tlattimore2017end}.
Combes, Magureanu and Prouti\`ere restate its program as the linear case of their Theorem~1 \cite[Section~4]{combes2017minimal}.
The proof is in Appendix~\ref{app:proofs-classical}.
In it the infimum over $\Alt_\Phi(\nu)$ is evaluated exactly, so no relaxation is added to those of Corollary~\ref{cor:cl-structured}.
The weight $\zeta_{a^\star}$ costs nothing.
As it grows, each constraint approaches the one set by the alternatives that keep the mean of $a^\star$.

\begin{corollary}[Lipschitz bandits]
\label{cor:cl-lipschitz}
Let $\mathrm{dist}$ be a metric on $[K]$, and let $\Phi$ be the class of instances whose arms have Bernoulli laws with means satisfying $|\mu_a(\lambda) - \mu_{a'}(\lambda)| \le \mathrm{dist}(a,a')$ for all arms $a$ and $a'$.
Let $\nu \in \Phi$ have a unique optimal arm $a^\star$ and every mean in $(0,1)$, and let the algorithm be consistent on $\Phi$.
Then $\liminf_{T \to \infty} R_T/\log T \ge c^\star_\Phi(\nu)$, and $c^\star_\Phi(\nu)$ equals the infimum of $\sum_{a} \zeta_a \Delta_a$ over the $\zeta \in [0,\infty)^K$ with
\[
\sum_{a \in [K]} \zeta_a\, \kl\big(\mu_a,\, \mu_a \vee (\mu^\star - \mathrm{dist}(a,b))\big) \;\ge\; 1
\]
for every $b \ne a^\star$.
\end{corollary}

For arms at points of $[0,1]$ and a metric proportional to their distance, this is Theorem~1 of Magureanu, Combes and Prouti\`ere \cite{magureanu2014lipschitz}, on instances with a unique optimal arm and means in $(0,1)$.
Combes, Magureanu and Prouti\`ere state it for any metric \cite[Section~4]{combes2017minimal}.
The proof is in Appendix~\ref{app:proofs-classical}.
In it the infimum over each part of $\Alt_\Phi(\nu)$ is evaluated exactly, so no relaxation is added to those of Corollary~\ref{cor:cl-structured}.
The infimum is approached by alternatives that raise arm $b$ just above $\mu^\star$ and every other arm only as far as the Lipschitz condition requires.
\section{Evaluation protocols}
\label{app:eval}

\subsection{Regret on movie ratings}
\label{app:eval-regret}

\paragraph{Instance.}
The arms are the $10$ most-rated movies of the MovieLens 100K ratings \cite{harper2015movielens}, between $429$ and $583$ ratings each.
The liked-or-not instance rewards a pull with $\mathbf{1}\{\text{rating} \ge 4\}$; its means run from $0.406$ to $0.859$.
The five-star instance rewards a pull with $(\text{rating} - 1)/4$ on $\{0, 0.25, 0.5, 0.75, 1\}$, drawn from each movie's empirical rating distribution; its means run from $0.539$ to $0.840$.
Each instance is the empirical law of the ratings.
The results are statements about that law.
On the liked-or-not instance the class of the Lai--Robbins limit is the Bernoulli family, with $\Kinf(\nu_b, \mu^\star) = \kl(\mu_b, \mu^\star)$.
On the five-star instance it is every law on the five-point support.
There $\Kinf(\nu_b, \mu^\star)$ is computed from its one-dimensional dual \cite{honda2015nonasymptotic}.

\paragraph{Algorithms.}
UCB1 \cite{auer2002}, KL-UCB with exploration $\log t$ \cite{cappeGarivierMaillardMunosStoltz2013klucb}, Thompson sampling with uniform priors \cite{thompson1933likelihood,kaufmann2012thompson}, binarized on the five-star instance, and IMED \cite{honda2015nonasymptotic}.
On the five-star instance NPTS \cite{riou2020bandit} runs as well.
Every algorithm runs $2{,}000$ times for $T = 20{,}000$ rounds, and the counts are read at $16$ log-spaced checkpoints.

\paragraph{Alternatives.}
For every suboptimal arm $b$ and margin $\xi \in \{0.02, 0.05, 0.1\}$, the alternative $\lambda^{(b,\xi)}$ replaces arm $b$ alone by its closest law with mean $\mu^\star + \xi$: the Bernoulli law on the liked-or-not instance and the $\Kinf$ minimizer on the five-star instance.
KL-UCB, Thompson sampling and IMED on the liked-or-not instance, and KL-UCB, IMED and NPTS on the five-star one, run $1{,}000$ times under each alternative.
They record $\mathbb{E}_\lambda[N_b(t)]$.

\paragraph{Read-outs.}
The limit is $c^\star(\nu) \log t$ with $c^\star(\nu) = \sum_b \Delta_b / \Kinf(\nu_b, \mu^\star)$.
For each arm $b$ and alternative, the term $\mathbb{E}_\nu[N_b(t)]\,\KL(\nu_b \,\|\, \lambda_b)$ of the divergence decomposition is exact, and the fundamental inequality \cite{garivier2019mathematics} bounds it below by $\kl(\mathbb{E}_\nu[N_b(t)]/t,\ \mathbb{E}_\lambda[N_b(t)]/t)$.
Their difference is the \textsc{process} defect of data processing to the normalized count.
The algorithm's certified pull count is the least $\underline{n}_b$ with $\underline{n}_b\, \KL(\nu_b \,\|\, \lambda_b) \ge \kl(\underline{n}_b/t,\ \mathbb{E}_\lambda[N_b(t)]/t)$, maximized over the margin $\xi$.
It is the number of pulls the fundamental inequality certifies from the algorithm's behavior at the alternatives alone.
The regret ratio splits exactly.
For arm $b$ and alternative $\lambda$ write $n_b := \mathbb{E}_\nu[N_b(T)]$, $z_\nu := n_b/T$, $z_\lambda := \mathbb{E}_\lambda[N_b(T)]/T$ and $\underline{\kl} := (1-z_\nu)\log\frac{1}{1-z_\lambda} - \log 2$.
Then
\begin{align*}
n_b\,\Kinf(\nu_b, \mu^\star) - \log T
 &= \big(n_b \KL(\nu_b \,\|\, \lambda_b) - \kl(z_\nu, z_\lambda)\big) + \big(\kl(z_\nu, z_\lambda) - \underline{\kl}\big) \\
 &\quad + \big(\underline{\kl} - \log T\big) - n_b\big(\KL(\nu_b \,\|\, \lambda_b) - \Kinf(\nu_b, \mu^\star)\big)
\end{align*}
The four terms are the \textsc{process}, \textsc{binary}, joint \textsc{uniform}-and-\textsc{asymptotic} and \textsc{boundary} terms.
The weight of arm $b$ is $\Delta_b / \Kinf(\nu_b, \mu^\star)$.
With these weights, and divided by $c^\star(\nu)\log T$, the left side becomes the regret ratio minus one.
Figure~\ref{fig:regret}(c) shows the split at $\xi = 0.02$.

\paragraph{Readings.}
On the liked-or-not instance the certified pull count is $16\%$ to $23\%$ of the actual count for KL-UCB, $28\%$ to $36\%$ for Thompson sampling and $30\%$ to $33\%$ for IMED.
On the five-star instance it is $29\%$ to $37\%$ for IMED and NPTS.
At $\xi = 0.1$ the \textsc{process} and \textsc{boundary} terms reach $4.48$ and $-4.10$ of the limit for KL-UCB on the liked-or-not instance.

\subsection{The factor of two under adaptive play}
\label{app:eval-factor-two}

\paragraph{Instance.}
The arms come from the MovieLens 100K ratings \cite{harper2015movielens}.
They are the $10$ most-rated movies.
A movie's reward is $\mathbf{1}\{\text{rating} \ge 4\}$.
Star Wars has the highest like-rate, $\mu^\star = 0.859$.
Each instance is a one-armed reduction in which the best arm $a^\star$ is known to have mean $\mu^\star$.
The perturbed arm $b$ is one of the four next movies: Fargo ($0.799$), Return of the Jedi ($0.748$), Toy Story ($0.710$) or Contact ($0.676$).
The alternative $\lambda$ raises arm $b$ to $\mu^\star + \xi$, with margin $\xi \in \{0.01, 0.03\}$.

\paragraph{Algorithms.}
KL-UCB \cite{cappeGarivierMaillardMunosStoltz2013klucb} pulls arm $b$ when $N_b = 0$, when its empirical mean $\hat\mu_b := S_b/N_b$ is at least $\mu^\star$, or when $N_b\, \kl(\hat\mu_b, \mu^\star) \le \log t$.
Thompson sampling \cite{thompson1933likelihood} with a uniform prior pulls arm $b$ with the posterior probability that its mean exceeds $\mu^\star$, under the posterior $\mathrm{Beta}(S_b + 1, N_b - S_b + 1)$.

\paragraph{Computation.}
Both algorithms act through $(N_b, S_b)$, so the law of $(N_b(t), S_b(t))$ under $\nu$, under $\lambda$ and under $Q_{1/2}$ is propagated exactly by dynamic programming up to $T = 3{,}000$.
Under $Q_{1/2}$, arm $b$ draws from the normalized geometric mixture $\nu_b^{(1/2)}$ of $\nu_b$ and $\lambda_b$ (Proposition~\ref{prop:tilt}).
All quantities are read at $14$ log-spaced horizons from $25$ to $3{,}000$, with no sampling error.

\paragraph{Readings.}
The per-pull factor $\KL(\nu_b \,\|\, \lambda_b) / (2c_b(1/2))$ lies between $2.12$ and $2.26$ on these instances.
At $T = 3{,}000$ the largest adaptive path factor, $1.40$, belongs to KL-UCB on Fargo at $\xi = 0.03$, where $L_T$ still has standard deviation $4.1$ nats under $\nu$.

\subsection{Statistics and finite-horizon regret bounds on the exact reductions}
\label{app:eval-regret-statistics}

\paragraph{Instance.}
The instances are the one-armed reductions of Appendix~\ref{app:eval-factor-two}, with the same two algorithms.
The best arm $a^\star$ is known at $\mu^\star = 0.859$.
The perturbed arm $b$ is Fargo ($\Delta_b = 0.060$), Return of the Jedi ($0.112$), Toy Story ($0.149$) or Contact ($0.184$).
The alternative $\lambda$ raises arm $b$ to $\mu^\star + \xi$, with margin $\xi \in \{0.01, 0.03\}$.
The measured cost is $D_T = \mathbb{E}_\nu[N_b(T)]\,\KL(\nu_b \,\|\, \lambda_b)$.
Each lower bound below is reported as a share of $D_T$.
That share is also the share of the expected pulls of arm $b$, and of the regret, that the bound certifies.

\paragraph{Computation.}
The laws of $(N_b(t), S_b(t))$ under $\nu$ and $\lambda$ are propagated exactly up to $T = 3{,}000$ and read at $16$ rounds from $25$ to $3{,}000$, with no sampling error.
The first statistic is the count $N_b(T)$.
The second pairs it with the round of the last pull of arm $b$, at $T \in \{300, 1{,}000, 3{,}000\}$.
The others are the best tail event $\{N_b(T) \le x\}$, the best event of the count, the best event of the whole history and the event $\{N_b(T) > T/2\}$.
By Theorem~\ref{thm:sufficiency} none of them keeps more than $D_T$.
For the normalized count $N_b(T)/T$ the quantity is $\kl(\mathbb{E}_\nu[N_b(T)/T], \mathbb{E}_\lambda[N_b(T)/T])$, as in the fundamental inequality.
By Lemma~\ref{lem:relax}(b) it is at most $D_T$.
The best event of a statistic maximizes $\kl(P_\nu(E), P_\lambda(E))$ over the events it determines.
It is found among the likelihood-ratio level sets of the statistic, because $\kl$ is jointly convex and the probability pairs of those sets are the extreme points of the convex hull of the attainable pairs $(P_\nu(E), P_\lambda(E))$.
Proposition~\ref{prop:cl-finite} is evaluated at the algorithm's own error sum and regrets.
Its error-sum form is $\log(1/(\varepsilon_T(2-\varepsilon_T)))$, with $\varepsilon_T$ the error sum of $\{N_b(T) > T/2\}$.
Its regret form is $\log(T\underline{\Delta}/(4(R_T(\nu) + R_T(\lambda))))$, with $\underline{\Delta} = \Delta_b \wedge \xi$.
Its class-level form puts twice the largest regret over the neighborhood of Corollary~\ref{cor:cl-gauss} in place of $R_T(\nu) + R_T(\lambda)$.
The neighborhood holds the instances whose arm $b$ has its mean in $[\mu_b, \mu_b + 2\Delta_b]$, cut at $1$ for Toy Story and Contact.
Its largest regret is taken over seven grid points, $\nu$ among them, and over $\lambda$, each regret computed exactly.
On Fargo and Contact at $\xi = 0.01$, the pair $(N_b(50), N_b(T))$ is propagated to $T = 1{,}000$.
On Gaussian one-armed reductions with unit variance, $a^\star$ is known at $0$ and arm $b$ has mean $-\Delta_b/\sqrt{\mu_b(1-\mu_b)}$, a gap of $0.150$ to $0.392$.
Gaussian UCB pulls arm $b$ while its empirical mean plus $\sqrt{2\log t/N_b}$ is at least $0$.
Gaussian Thompson sampling, after one pull, draws arm $b$'s mean from the Gaussian law centered at its empirical mean with variance $1/N_b$.
Each algorithm runs $2{,}000$ times at each of seven instances of the neighborhood $\mathcal{U}(\nu)$.
The bound of Corollary~\ref{cor:cl-gauss} is evaluated with $\Gamma T^p$ replaced by the largest measured regret over these instances, at $\xi_\Delta \in \{0.1, 0.5, 1\}$ and on a grid of $1{,}000$ margins in $(0,1]$.

\paragraph{Validation.}
The relative entropy between the two laws of $(N_b, S_b)$ equals $D_T$, as Theorem~\ref{thm:sufficiency} requires.
The remaining controls pass: agreement with the quantities of Appendix~\ref{app:eval-factor-two} at its rounds, with simulated runs of each algorithm, and with an independent recomputation of every read-out that depends only on the two laws of $N_b(T)$.

\begin{figure}[tp]
\centering
\includegraphics[width=\linewidth]{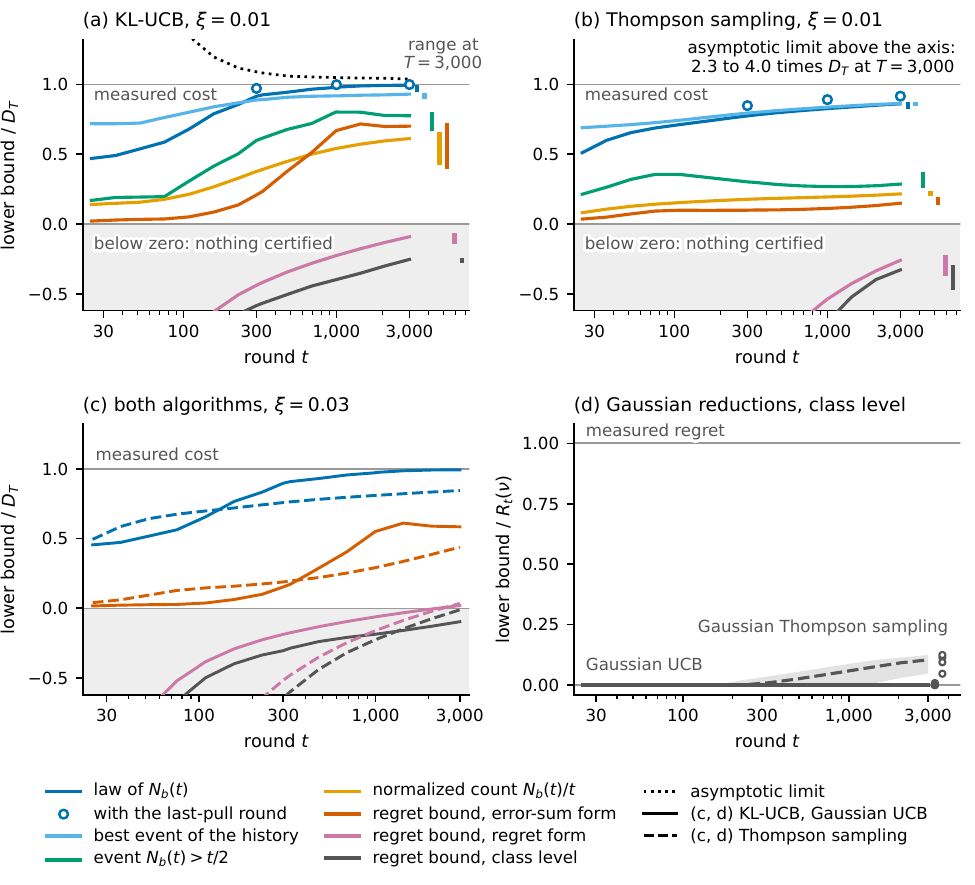}
\caption{\textbf{On exact one-armed reductions at $\xi = 0.01$, the law of the pull count keeps at least $82\%$ of $D_T$ at $T = 3{,}000$, while the regret form of the finite-horizon regret bound certifies none of it.} (a, b) Lower bounds on $D_T = \mathbb{E}_\nu[N_b(T)]\,\KL(\nu_b \,\|\, \lambda_b)$, divided by $D_T$, at $\xi = 0.01$. Each color is one bound, named in the key below the panels. The three regret-bound curves are the error-sum and regret forms of Proposition~\ref{prop:cl-finite} and its class-level form. The markers show $N_b(t)$ together with the round of the last pull of arm $b$. Lines and markers are medians over the four arms, and the bars right of the last round give the range at $T = 3{,}000$. The dotted line is the asymptotic requirement $\log t\,\KL(\nu_b \,\|\, \lambda_b)/d(\mu_b, \mu^\star)$, which is not a bound at a finite horizon. A value below zero certifies nothing. (c) Four of the same bounds at $\xi = 0.03$, as medians, with KL-UCB solid and Thompson sampling dashed. (d) The bound of Corollary~\ref{cor:cl-gauss} at the best of $\xi_\Delta \in \{0.1, 0.5, 1\}$, with $\Gamma T^p$ replaced by the largest measured regret over seven instances of the neighborhood, over the measured regret on Gaussian reductions with the same standardized gaps. Gaussian UCB is solid and Gaussian Thompson sampling dashed. Lines are medians and bands ranges over the four arms. Markers give each arm's value at $T = 3{,}000$ with its $95\%$ percentile bootstrap interval, which is narrower than the marker. Values in (a) to (c) are exact.}
\label{fig:regret-statistics}
\end{figure}

\paragraph{Readings.}
Figure~\ref{fig:regret-statistics} shows the read-outs.
At $T = 3{,}000$ and $\xi = 0.01$, the law of $N_b(T)$ keeps $0.947$ to $0.997$ of $D_T$ under KL-UCB.
Under Thompson sampling it keeps $0.825$ to $0.872$.
Jointly with the round of the last pull of arm $b$, it keeps $0.997$ to $1.000$ under KL-UCB and $0.900$ to $0.917$ under Thompson sampling.
The best event of the history keeps $0.843$ to $0.940$.
So processing the history to a single event discards $6\%$ to $16\%$ of $D_T$ for these arms and algorithms.
The classical statistics keep less.
Under KL-UCB the event $\{N_b(T) > T/2\}$ keeps $0.667$ to $0.803$ and the normalized count $0.425$ to $0.657$.
Under Thompson sampling they keep $0.256$ to $0.371$ and $0.201$ to $0.239$.
The law of $N_b(T)$ thus recovers $77\%$ to $99\%$ of the \textsc{process} defect of the normalized count.

The best tail event $\{N_b(T) \le x\}$ keeps at least $0.851$ of the count's share for seven of the eight pairs of arm and algorithm at $\xi = 0.01$.
Under KL-UCB on Fargo it keeps $0.706$.
Under KL-UCB the likelihood ratio between the two laws of the count is not monotone in the count.
Its increments change sign about $1{,}800$ times over the $2{,}820$ counts with positive mass under both laws, against at most $3$ times under Thompson sampling.
So no tail event selects the counts at which the two laws differ most.
The best event of the count, a likelihood-ratio level set, keeps $0.867$ of the count's share on Fargo.
At $\xi = 0.03$ the tail events of KL-UCB keep between $0.474$ and $0.828$ of the count's share.

At the smaller margin, the error-sum form of Proposition~\ref{prop:cl-finite} keeps $0.393$ to $0.725$ of $D_T$ under KL-UCB and $0.133$ to $0.193$ under Thompson sampling.
Its regret form is negative for every arm and both algorithms, between $-1.50$ and $-0.57$ nats.
Over the same pairs $D_T$ lies between $2.80$ and $10.65$ nats.
The class-level form lies between $-2.66$ and $-1.20$ nats.
It cannot exceed the regret form, since its largest regret is taken over a set that holds $\nu$ and $\lambda$.
The regret form is positive only when $R_T(\nu) + R_T(\lambda) < T\underline{\Delta}/4$.
At $T = 3{,}000$ and $\xi = 0.01$ that threshold is $7.5$, against measured sums of $13.3$ to $33.7$.
Under KL-UCB the \textsc{markov} step bounds $P_\nu(N_b(T) > T/2)$, at most $0.007$, by $2\mathbb{E}_\nu[N_b(T)]/T$, at least $0.048$.
With the matching step under $\lambda$, it removes $3.9$ to $5.1$ nats.
Bounding both gaps below by $\underline{\Delta}$ removes another $1.8$ to $2.8$ nats.
Together these exceed the error-sum form, which is $4.2$ to $6.5$ nats.
Under Thompson sampling the error-sum form is $0.54$ nats for every arm, because $P_\lambda(N_b(T) \le T/2) = 0.354$ at $T = 3{,}000$.
At $\xi = 0.03$ the regret form is positive for seven of the eight pairs and keeps at most $0.045$ of $D_T$.
The class-level form stays negative for all eight.
Since $R_T(\lambda) \ge 0$ at every alternative that raises arm $b$ alone, the regret form at margin $\xi$ is at most $\log(T(\Delta_b \wedge \xi)/(4R_T(\nu)))$.
Over all margins it certifies at most $7\%$ of the expected pulls of arm $b$ under KL-UCB and at most $27\%$ under Thompson sampling.
The asymptotic requirement $\log T\,\KL(\nu_b \,\|\, \lambda_b)/d(\mu_b, \mu^\star)$ lies above $D_T$ for every arm and both algorithms.
It exceeds $D_T$ by a factor of $1.03$ to $1.06$ under KL-UCB and of $2.3$ to $4.0$ under Thompson sampling.

On the Gaussian reductions at $T = 3{,}000$, the bound of Corollary~\ref{cor:cl-gauss} is $0$ for Gaussian UCB on three of the four arms and $0.008$ of the measured regret on Contact.
For Gaussian Thompson sampling it reaches $0.047$ to $0.125$ of the measured regret.
Over the grid of margins it reaches at most $0.131$.
The bound is positive only when the largest regret over the neighborhood is below $\xi_\Delta T\Delta_b/8$, with $\Delta_b$ now the gap of the Gaussian reduction.
For Gaussian UCB that largest regret exceeds $T\Delta_b/8$ on three of the arms.

Adding $N_b(50)$ to $N_b(T)$ raises the share kept at $T = 1{,}000$ by $0.002$ to $0.026$ of $D_T$ on Fargo and Contact.

\paragraph{Scope.}
The reductions know $a^\star$, so the shares describe one perturbed arm against a known mean.
None of them is transferred to the instance with $10$ arms.
The largest regret over the neighborhood is taken over eight of its instances, so each class-level form reported is an upper bound on its value over the whole neighborhood.
For the Gaussian bound it is taken over seven instances, so each Gaussian value is, up to Monte Carlo error, an upper bound on the share certified by Corollary~\ref{cor:cl-gauss}.
The read-outs stop at $T = 3{,}000$, and the pair $(N_b(50), N_b(T))$ at $T = 1{,}000$.
The Gaussian reductions are constructed with standardized gaps matched to the four arms.
Their regrets are Monte Carlo estimates.

\subsection{Best-arm identification among trained models}
\label{app:eval-bai}

\paragraph{Instances.}
An arm is a trained model.
A pull scores it on one test item drawn uniformly, so the arm is Bernoulli with the model's accuracy as its mean.
The first instance takes the accuracies of $34$ image classifiers on the reproduced CIFAR-10 test set, all evaluated by one team \cite{recht2019imagenet}.
Its two best models differ by $1.6$ points of accuracy.
Its characteristic time is $T^\star(\nu) = 3{,}481$.
The second takes the five-shot MMLU accuracies of $11$ language models collected by Touvron et al.\ \cite{touvron2023llama}, which come from several sources.
Its characteristic time is $T^\star(\nu) = 5{,}602$.

\paragraph{Algorithms.}
Track-and-Stop \cite{garivier2016optimal} uses D-tracking with forced exploration and recomputes its target weights whenever $t$ grows by $5\%$.
T3C \cite{shang2020fixed} draws a leader from the posterior, takes as challenger the arm of least transportation cost from the leader, and pulls the leader with probability $1/2$.
Round-robin sampling cycles through the arms.
Every algorithm stops by the generalized likelihood ratio rule, at a threshold that depends on $t$ and $\delta$.
Two thresholds are $\delta$-correct: $\log(2t(K-1)/\delta)$ \cite[Theorem~10]{garivier2016optimal} and the threshold of Kaufmann and Koolen \cite[(25)]{kaufmann2021mixture}.
The third, $\log((\log t + 1)/\delta)$, is a common practical choice that is not proved $\delta$-correct \cite[Section~6]{garivier2016optimal}.
Track-and-Stop runs with all three thresholds, and T3C and round-robin sampling run with the first.
The risks are $\delta \in \{10^{-1}, 10^{-2}, 10^{-3}, 10^{-4}, 10^{-6}\}$, which gives $50$ configurations over the two instances.
Each configuration runs $400$ times.

\paragraph{Split.}
Let $E$ be the event that the recommendation is the best arm of $\nu$, and let $\lambda^\star$ be the alternative that attains $\inf_{\lambda \in \Alt(\nu)} \sum_a \bar w_a\, d(\mu_a, \mu_a(\lambda))$ at the algorithm's average allocation $\bar w$ (Appendix~\ref{app:cl-bai}).
In $\lambda^\star$ the best arm and one challenger move to their allocation-weighted mean (Proposition~\ref{prop:mbary}).
Then
\[
\frac{\mathbb{E}_\nu[\tau]}{T^\star(\nu)} = \kl(\delta, 1-\delta) + \textsc{uniform} + \textsc{process} + \textsc{allocation}
\]
Here $\textsc{allocation} = \mathbb{E}_\nu[\tau]/T^\star(\nu) - \sum_a \mathbb{E}_\nu[N_a(\tau)]\, d(\mu_a, \mu_a(\lambda^\star))$, $\textsc{process}$ is that sum minus $\kl(P_\nu(E), P_{\lambda^\star}(E))$, and $\textsc{uniform}$ is the rest.
The probability $P_{\lambda^\star}(E) = \mathbb{E}_\nu[\mathbf{1}_E\, e^{-L_\tau}]$ is computed from the per-run counts and reward sums at the stopping time, since for these arms $L_\tau = \sum_a \big[(\theta_a - \eta_a) S_a(\tau) - N_a(\tau)(\psi(\theta_a) - \psi(\eta_a))\big]$.
The alternative $\lambda^\star$ is chosen on half of the runs and evaluated on the other half, and the halves then swap.
Choosing it on the runs that evaluate it would favor a challenger whose realized rewards fell short, which biases the likelihood ratio upward.
The logarithm of each half's estimate of $P_{\lambda^\star}(E)$ from its $200$ runs is biased low.
In an exact two-arm control at the same thresholds and risks, the bias is $0.07$ to $2.39$ nats (Appendix~\ref{app:eval-own-error}).
The bias makes the drawn \textsc{uniform} term too large, and the \textsc{process} term too small, by the same number of nats.
Their sum needs no estimate of $P_{\lambda^\star}(E)$.

\paragraph{Readings.}
The ledger identity $\mathbb{E}_\nu[L_\tau] = \sum_a \mathbb{E}_\nu[N_a(\tau)]\, d(\mu_a, \mu_a(\lambda^\star))$ holds within sampling error in all $49$ configurations with a measured mean, with the largest deviation $2.0$ standard errors.
Under the threshold of Garivier and Kaufmann, the algorithm's error probability at $\lambda^\star$ is $3 \times 10^{-8}$ to $2 \times 10^{-6}$ times $\delta$.
Under that of Kaufmann and Koolen it is below $10^{-15}\,\delta$.
Under the heuristic threshold it is $2\%$ to $14\%$ of $\delta$.
Of the $19{,}999$ runs that stop, one errs: with the heuristic threshold at $\delta = 0.1$ on the MMLU instance.
Round-robin sampling's \textsc{uniform} term is within $1.1$ nats of Track-and-Stop's under the same threshold.

On the CIFAR-10 instance the stopping time of Track-and-Stop has a heavy tail.
Its longest run lasts up to $7$ times the configuration's mean under the first threshold and up to $15$ times under the heuristic one.
On that instance the longest run of T3C lasts at most $1.9$ times its mean.
One run reached its step cap.
It used the heuristic threshold at $\delta = 10^{-4}$ and had not stopped after $237\,T^\star(\nu)$ pulls.
Counting that run at the cap gives its configuration a ratio of at least $1.95$ to the limit.

\subsection{Bounds at the algorithms' own error rates}
\label{app:eval-own-error}

At the algorithm's own error rates the expected sample size is bounded by the chain of Proposition~\ref{prop:cl-bai}, stopped before its \textsc{uniform} step (Proposition~\ref{prop:cl-own}),
\[
\mathbb{E}_\nu[\tau] \;\ge\; T^\star(\nu)\, \kl\big(P_\nu(E), P_{\lambda^\star}(E)\big)
\]
Here $E$ is the event of a correct recommendation and $\lambda^\star$ the most confusing alternative at the algorithm's average allocation (Appendix~\ref{app:eval-bai}).
Under $\lambda^\star$ a run seldom stops, so $P_{\lambda^\star}(E)$ is estimated by a change of measure, as the mean of $\mathbf{1}_E\, e^{-L_\tau}$ over runs under $\nu$.
That estimate had to pass a check against an exact control and on the instances before the bound was read.
For Wald's SPRT and the mixture SPRT the bound is read at error rates computed exactly or simulated directly.

\paragraph{Control.}
Four two-arm reductions reproduce the geometry of the instances' most confusing alternatives.
In the first two the arms alternate, so $\lambda^\star$ moves both means to their midpoint, as for round-robin sampling.
The means of the leader and the challenger are $0.75$ and $0.45$ in the first reduction and $0.95$ and $0.80$ in the second.
In the other two the challenger takes the fixed share $0.1$ or $0.05$ of the pulls, so $\lambda^\star$ raises it nearly to the leader's mean.
The means are $0.80$ and $0.40$ in the third reduction and $0.95$ and $0.75$ in the fourth.
At $77$ of the $78$ cross-fitted alternatives of Track-and-Stop and T3C in Appendix~\ref{app:eval-bai}, the challenger is moved at least $95\%$ of the way to the leader.
Each reduction stops by the generalized likelihood ratio rule at the three thresholds of Appendix~\ref{app:eval-bai}.
The number of arms in the union term is $11$ for the first and third reductions and $34$ for the other two, as on the MMLU and CIFAR-10 instances.
The risks are those of Appendix~\ref{app:eval-bai}, except that the fourth reduction runs the threshold of Garivier and Kaufmann and the heuristic threshold at $\delta \in \{10^{-1}, 10^{-3}, 10^{-6}\}$.
This gives $51$ combinations of reduction, threshold and risk.
A dynamic program over the reward sums of the two arms propagates the laws of the continuing paths under $\lambda^\star$, under $\nu$ and under each proposal below.
The probability $P_{\lambda^\star}(E, \tau \le T)$ at the program's horizon $T$ is computed by two routes, by absorption under $\lambda^\star$ and by reweighting the law under $\nu$ with $e^{-L_\tau}$.
Under the rule fixed before the runs, $T$ is grown until this probability moves by less than $10^{-4}$ relative over the last fifth of the steps.
That criterion was out of reach at a usable cost.
Under $\lambda^\star$ at least $84\%$ of the paths are still running at the horizon, so the probability still grows there, by $6 \times 10^{-4}$ to $4 \times 10^{-2}$ relative over the last fifth of the steps.
Each program therefore stops at twice the step at which the continuing mass under $\nu$ falls below $10^{-12}$.
Every estimate below is compared with $P_{\lambda^\star}(E, \tau \le T)$ at that horizon, from runs capped there.

\paragraph{Estimates.}
Under each proposal the leader and the challenger are moved along the line through their means under $\nu$ and under $\lambda^\star$.
The proposals lie a quarter and a half of the way from $\nu$ toward $\lambda^\star$, or on the far side of $\nu$ at one half, one and one and a half times the distance from $\nu$ to $\lambda^\star$.
The proposals on the far side widen the gap between the two arms.
Proposal means are clamped to $[0.01, 0.99]$.
Three proposals of the control were clamped, the two widest of the second reduction and the widest of the third.
No proposal on the instances was clamped.
Under every proposal except the halfway one the recommendation is correct with probability at least $0.99$, and under the halfway one with probability at least $0.93$.
Every estimate was fixed before the runs.
The estimate of Appendix~\ref{app:eval-bai} averages $\mathbf{1}_E\, e^{-L_\tau}$ over $200$ runs under $\nu$.
A second average uses $4{,}000$ runs.
A mixture draws its runs in equal shares from several proposals and weights each run by the likelihood ratio of $\lambda^\star$ to the mixture.
Three mixtures are formed, each with $1{,}000$ runs in all: from $\nu$ and the two nearest widening proposals, from these three with the quarter-way proposal added, and from the three widening proposals.
The second mixture is also formed with $4{,}000$ runs.
The halfway proposal is used alone with $1{,}000$ runs.
The sampling law of each estimate is drawn by resampling from the program's laws of the stopped paths, with $4{,}000$ replicates, or $1{,}000$ at $4{,}000$ runs.
An estimate $\hat P$ meets the tolerance when, for every reduction, threshold and risk, the mean error of $\log \hat P$ lies within $0.2$ nats of zero and $|\log \hat P - \log P_{\lambda^\star}(E, \tau \le T)|$ exceeds $0.5$ nats in at most $10\%$ of the replicates.
The estimate must also agree with a second family on the instances below.

\paragraph{Validation.}
Throughout the control the program meets every tolerance fixed before the runs, on the agreement of its two routes to $P_{\lambda^\star}(E, \tau \le T)$ and on the conservation of mass.
The program's $P_\nu(E)$, mean stopping time and $\mathbb{E}_\nu[L_\tau]$ are reproduced within $3$ standard errors by $4{,}000$ simulated runs for each combination, in $152$ of the $153$ comparisons.
The simulation uses the stopping statistic and the thresholds of Appendix~\ref{app:eval-bai}.
The laws of the stopping times pass a Kolmogorov--Smirnov test against the program throughout, with smallest $p$-value $0.17$ after a Bonferroni correction.
The exception is $\mathbb{E}_\nu[L_\tau]$ in the fourth reduction under the threshold of Garivier and Kaufmann at $\delta = 10^{-6}$, at $3.38$ standard errors.
There $L_\tau$ splits into a part fixed by $\tau$ and a mean-zero martingale part, whose sample means sit at $+1.00$ and $-0.65$ of their standard errors, with correlation $-0.994$.
The standard error of their sum is about a ninth of either part's, so two ordinary draws read as $3.38$ standard errors in the sum.
The divergence decomposition of Theorem~\ref{thm:allocation} holds on the program's laws throughout.
Five further sets of $40{,}000$ runs at that combination give $z$-values from $-0.37$ to $1.67$.
The check, as fixed before the runs, has no correction for multiplicity, so a correct program fails it somewhere with probability about one third.
The verdict below holds without this check.

\paragraph{Readings.}
No estimate meets the tolerance throughout the control.
The second mixture at $4{,}000$ runs fails at $7$ of the $51$ combinations, the fewest, while the estimate of Appendix~\ref{app:eval-bai} fails at $45$.
The decisive case is the fourth reduction under the threshold of Garivier and Kaufmann at $\delta = 10^{-6}$, where the challenger takes $5\%$ of the pulls and $\log P_{\lambda^\star}(E, \tau \le T) = -22.43$.
There every estimate misses by more than half a nat in at least $43\%$ of its replicates, against the $10\%$ allowed, an excess of at least $26$ standard errors of the resampling.
The estimate of Appendix~\ref{app:eval-bai} reads $1.98$ nats low there on average.
By threshold and risk, no estimate passes every reduction under the threshold of Kaufmann and Koolen at any risk, or under that of Garivier and Kaufmann at $\delta = 10^{-6}$.
The weights in the decisive case lie within the range met on the instances: its median effective sample size at $200$ runs, $10.5$, is above that of $12$ of the $78$ halves of Appendix~\ref{app:eval-bai} and below their median of $32.7$.
The misses do not come from stops later than any reached by the runs under $\nu$.
Where the second mixture at $4{,}000$ runs fails, less than $4\%$ of $P_{\lambda^\star}(E, \tau \le T)$ accrues after the continuing mass under $\nu$ falls below $1/200$.
Relative to $|\log P_{\lambda^\star}(E, \tau \le T)|$, the mean error of that mixture is at most $2.6\%$.
The tolerance is set in nats.

\paragraph{Instances.}
Three configurations of Appendix~\ref{app:eval-bai} with uneven weights were rerun, all at $\delta = 10^{-6}$.
They are T3C on the CIFAR-10 instance and Track-and-Stop on the MMLU instance, both under the threshold of Garivier and Kaufmann, and Track-and-Stop on the CIFAR-10 instance under the heuristic threshold.
The alternative $\lambda^\star$ is held at the two cross-fitted alternatives recorded for each configuration.
Every run is fresh, so each alternative is evaluated on runs that played no part in its choice.
Each configuration has $4{,}500$ runs under $\nu$ and $500$ under each proposal other than the halfway one, along the segment to each alternative.
The mean of the estimates from $20$ disjoint blocks of $200$ runs lies $0.01$ to $0.23$ nats below the estimate from the same $4{,}000$ runs.
That is within the tolerance of $0.2$ nats plus two bootstrap standard errors at all six alternatives.
The bootstrap standard errors of these differences are $0.01$ to $0.12$ nats, so a drift below about twice these values would go undetected.
The expected logarithm of a mean of independent copies does not decrease with the number of copies and never exceeds the logarithm of the expectation.
The difference therefore bounds the bias of the $200$-run estimate from below, in expectation.
A comparison between run counts cannot detect a bias shared by both.
The estimate from $4{,}000$ runs under $\nu$ agrees with every mixture within $0.2$ nats plus two combined standard errors at every alternative but one.
At one alternative on the MMLU instance the mixture of the three widening proposals lies $1.79$ nats lower, against a tolerance of $1.07$.
On the instances the second mixture at $4{,}000$ runs is formed from $2{,}000$ runs, $500$ per proposal.
Agreement is weak evidence where the weights are uneven.
At one alternative of the first configuration the second mixture at $1{,}000$ runs has an effective sample size of $2.9$ and lies $1.0$ nats from the $4{,}000$-run estimate, within a tolerance of $1.41$.

One of the $4{,}500$ runs at the third configuration reached the step cap at $951{,}851$ pulls, with the best arm leading.
That run enters every estimate there with weight zero.
Because $e^{-L_t}$ is a nonnegative martingale under $\nu$ and the run's log-likelihood ratio was $62$ nats at the cap against both alternatives, its eventual weight has conditional expectation below $10^{-23}$ of the sum of the weights.
Counting that run at the cap, the mean stopping time of the $4{,}500$ runs is $82{,}424 \pm 596$ (one standard error), against $79{,}774 \pm 1{,}028$ from the $400$ runs of Appendix~\ref{app:eval-bai}, a difference of $2.2$ standard errors.
At the other two configurations the two means agree within $1.5$ standard errors.

\paragraph{Consequence.}
Since no estimate meets the tolerance, the bound at the algorithms' own error rates is not reported for identification.
The estimate of Appendix~\ref{app:eval-bai} is also used to divide $\sum_a \mathbb{E}_\nu[N_a(\tau)]\, d(\mu_a, \mu_a(\lambda^\star)) - \kl(\delta, 1-\delta)$ between the \textsc{uniform} and \textsc{process} terms of Figure~\ref{fig:bai}.
The two reductions with an under-sampled challenger have the geometry of the alternatives of Track-and-Stop and T3C.
In those reductions the estimate reads $\log P_{\lambda^\star}(E, \tau \le T)$ low by $0.07$ to $1.98$ nats over the three thresholds (Table~\ref{tab:own-error-split}).
With $P_\nu(E)$ near $1$, a reading low by $b$ nats moves about $b$ nats from the \textsc{process} term to the \textsc{uniform} term.
The sum of the two terms needs no estimate of $P_{\lambda^\star}(E)$.
The size of the bias on the instances is not determined.
At the three configurations rerun above, the bias is at least the drift found there, in expectation.

\begin{table}[t]
\centering
\caption{\textbf{The sum of the \textsc{uniform} and \textsc{process} terms of Figure~\ref{fig:bai} needs no estimate, while the estimate that splits them reads low in the exact control.} Each row gives ranges in nats over the instances and risks at one stopping threshold. Track-and-Stop runs with all three thresholds; T3C with the threshold of Garivier and Kaufmann only. The columns give the sum of the two terms and the amount by which the estimate of $\log P_{\lambda^\star}(E)$ reads low in the control reductions with an under-sampled challenger, at the same threshold and risks. At the heuristic threshold the sum covers nine configurations, since one run of the omitted configuration did not stop.}
\label{tab:own-error-split}
\begin{tabular}{lcc}
\toprule
threshold & sum of the two terms & low reading in the control \\
\midrule
Garivier and Kaufmann & $14.6$--$20.8$ & $0.07$--$1.98$ \\
Kaufmann and Koolen & $38.6$--$45.0$ & $1.07$--$1.32$ \\
heuristic & $4.0$--$6.5$ & $0.15$--$0.64$ \\
\bottomrule
\end{tabular}

\end{table}

\paragraph{Tests.}
For Wald's SPRT the bound at the test's own error rates is that of Proposition~\ref{prop:cl-wald}, at the realized error rates $\alpha'$ and $\beta'$,
\[
\mathbb{E}_\nu[\tau] \;\ge\; \frac{\kl(1-\beta', \alpha')}{\kl(p_1, p_0)}
\]
The test is computed exactly on both metrics of Appendix~\ref{app:eval-ab}, at $\alpha = \beta \in \{10^{-1}, 0.05, 10^{-2}, 10^{-3}, 10^{-4}\}$.
Under each hypothesis the law of the continuing paths is propagated over the success count of the treatment stream until less than $10^{-15}$ of its mass remains.
The divergence decomposition of Theorem~\ref{thm:allocation}, $\mathbb{E}_\nu[L_\tau] = \mathbb{E}_\nu[\tau]\,\kl(p_1, p_0)$, holds in the program.
The $20{,}000$ runs per level and hypothesis of Appendix~\ref{app:eval-ab} agree with the program's expected sample sizes within $1.6$ standard errors.
At the realized rates the expected sample size exceeds the bound by $0.10\%$ to $0.79\%$, against $2.2\%$ to $13.8\%$ above Wald's bound at the nominal levels, $\kl(1-\alpha, \alpha)/\kl(p_1, p_0)$.
Under the null the expected sample size exceeds the symmetric bound of Proposition~\ref{prop:cl-wald} by at most $0.07\%$.
The defect of Proposition~\ref{prop:cl-wald}, which measures the spread of the overshoot, is $0.009$ to $0.018$ nats.

For the mixture SPRT the bound
\[
\mathbb{E}_\nu[\tau] \;\ge\; \frac{\kl(P_\nu(E_+), \alpha'_+)}{\Kinf(\nu, \mathcal{H}_0)}
\]
follows from Lemma~\ref{lem:relax}(a) at the barycentre null.
Here $E_+$ is the event of rejecting with the treatment ahead and $\alpha'_+$ its probability at the barycentre null (Appendix~\ref{app:eval-ab}).
The probability $\alpha'_{+,\mathrm{cap}}$ of rejecting ahead before a cap is simulated directly at the barycentre null, with $100{,}000$ runs at $\alpha = 10^{-2}$ and $200{,}000$ at $10^{-3}$ on both metrics, and $2{,}000{,}000$ at $10^{-4}$ on the site-visit metric.
At $\alpha \in \{0.1, 0.05\}$ the $10{,}000$ direct runs of Appendix~\ref{app:eval-ab} are used.
Every run is capped at $10$ times the limit plus $100$ pairs, as in Appendix~\ref{app:eval-ab}.
A run that reaches the cap counts as not rejecting.
The bound is then evaluated at the event of rejecting ahead before the cap, with $\alpha'_{+,\mathrm{cap}}$ in place of $\alpha'_+$.
Lemma~\ref{lem:relax}(a) applies to that event, whose probability under the truth is estimated from the $10{,}000$ runs under the truth, all of which stop before the cap.
No direct simulation was run at $\alpha = 10^{-4}$ on the purchase metric, so the ranges below leave that level out.
At the other nine levels the test draws $4.7\%$ to $9.1\%$ more pairs than this bound, against $18\%$ to $62\%$ more than its limit $\log(1/\alpha)/\Kinf(\nu, \mathcal{H}_0)$ of Section~\ref{sec:eval-ab}.
The upper ends of the $95\%$ intervals, from a parametric bootstrap over the mean stopping time and the binomial count of null rejections, are at most $1.13$ times the bound and exceed $1.10$ at four levels.
Figure~\ref{fig:own-error-tests} shows both tests.

Recomputed at the same caps, the change-of-measure estimate of Appendix~\ref{app:eval-ab} lies $2\%$ to $29\%$ below the direct frequencies at eight of the nine levels and $5\%$ above at the ninth, jointly $3.6$ standard errors.
Counted only up to the stopping time of the slowest of the $10{,}000$ runs under the truth, the direct and the change-of-measure values agree, jointly within $0.6$ standard errors and within $1.93$ at every level.
No run under the truth reaches the null's later rejections, which are $2\%$ to $21\%$ of its rejections ahead, so the change of measure gives them no weight.
With the change-of-measure values, the expected sample size would exceed the bound by $2.9\%$ to $6.9\%$.

\begin{figure}[t]
\centering
\includegraphics[width=\linewidth]{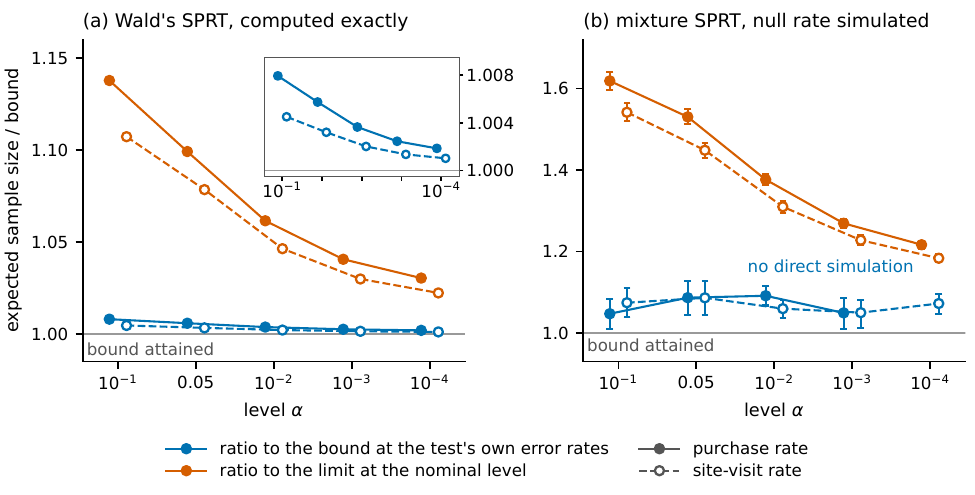}
\caption{\textbf{At the tests' own error rates the expected sample size exceeds the bound by at most $0.8\%$ for Wald's SPRT and $9.1\%$ for the mixture SPRT, against up to $14\%$ and $62\%$ over their limits.} Each marker divides the expected sample size $\mathbb{E}_\nu[\tau]$ on one metric of the Hillstrom experiment by one of two lower bounds: the bound at the test's own error rates (blue) or the limit at the nominal level (vermillion). Filled markers and solid lines show the purchase rate, and open markers and dashed lines the site-visit rate. (a) Wald's SPRT, with the bound $\kl(1-\beta', \alpha')/\kl(p_1, p_0)$ of Proposition~\ref{prop:cl-wald} at the exact realized error rates and the limit $\kl(1-\alpha, \alpha)/\kl(p_1, p_0)$. All values are exact. The inset magnifies the ratios to the bound at the realized error rates over the same five levels. (b) The mixture SPRT, with the bound $\kl(P_\nu(E_+), \alpha'_{+,\mathrm{cap}})/\Kinf(\nu, \mathcal{H}_0)$ at its one-sided rejection rate before the cap, $\alpha'_{+,\mathrm{cap}}$, at the barycentre null, and the limit $\log(1/\alpha)/\Kinf(\nu, \mathcal{H}_0)$. Bars give $95\%$ intervals. At $\alpha = 10^{-4}$ on the purchase metric no direct value of $\alpha'_{+,\mathrm{cap}}$ exists, and the bound is not drawn.}
\label{fig:own-error-tests}
\end{figure}

\subsection{Top-\texorpdfstring{$m$}{m} identification among trained models}
\label{app:eval-topm}

\paragraph{Instances.}
The instances are those of Appendix~\ref{app:eval-bai}.
A recommendation is now a set of $m$ models.
On the MMLU instance $m \in \{2, 3, 5\}$.
On the CIFAR-10 instance $m = 3$.
Its top-$4$ set is not unique, since two models score $93.0$.
At $m = 3$ its third model scores $93.7$, and the next three score $93.0$, $93.0$ and $92.9$.
The limit is $T^\star_m(\nu)\kl(\delta, 1-\delta)$ (Proposition~\ref{prop:cl-topm-exact}).
Its characteristic time is the value of a convex program over the $m(K-m)$ swap pairs.
Each value is certified by a duality bracket of relative width at most $3 \times 10^{-12}$.
On the MMLU instance $T^\star_m(\nu) = 1{,}451.9$, $2{,}070.6$ and $1{,}529.1$ for $m = 2, 3, 5$, and on the CIFAR-10 instance $T^\star_3(\nu) = 23{,}008.0$.
These are $1.65$ to $1.72$ times the explicit sums of Proposition~\ref{prop:cl-topm}.

\paragraph{Algorithms.}
Top-$m$ Track-and-Stop recomputes the optimal top-$m$ weights at the empirical means whenever $t$ grows by $5\%$ and tracks them by D-tracking with forced exploration \cite{garivier2016optimal,kaufmann2021mixture}.
In each round KL-LUCB \cite{kaufmann2013information} and LUCB \cite{kalyanakrishnan2012pac} pull two critical models: the empirical top-$m$ model with the lowest lower confidence bound and the model outside with the highest upper one.
KL-LUCB bounds each mean by the Bernoulli relative entropy.
LUCB uses Hoeffding bounds.
Each uses the exploration rate of its source.
Round-robin sampling cycles through the models.
Every algorithm runs with the generalized likelihood ratio stopping rule.
Its statistic is the minimum over the empirical swap pairs of the pairwise statistic.
The statistic is compared with a threshold.
A wrong recommendation reverses the order of at least one of the $m(K-m)$ swap pairs, so two thresholds take a union over them.
The first is $\log(2t\,m(K-m)/\delta)$, from the pairwise bound in the proof of \cite[Theorem~10]{garivier2016optimal}.
The second is the threshold of Kaufmann and Koolen \cite[(25)]{kaufmann2021mixture} with the same union.
The third, $\log((\log t + 1)/\delta)$, is not proved $\delta$-correct.
LUCB and KL-LUCB also run with their own stopping rules, which stop when the two critical confidence intervals separate.
On the MMLU instance with $m = 3$ the risks are $\delta \in \{10^{-1}, 10^{-2}, 10^{-3}, 10^{-4}, 10^{-6}\}$.
Track-and-Stop runs there under all three thresholds, KL-LUCB under the first and the third, and LUCB and round-robin sampling under the first.
For $m = 2$ and $m = 5$ the risks are $\delta \in \{10^{-1}, 10^{-3}, 10^{-6}\}$, with Track-and-Stop under the first and the third threshold and KL-LUCB under the first.
On the CIFAR-10 instance the risks are $\delta \in \{10^{-1}, 10^{-6}\}$, with Track-and-Stop and KL-LUCB under the first threshold.
The own stopping rules run at $\delta = 0.1$ on the MMLU instance with $m = 3$.
This gives $59$ configurations.
Each runs $400$ times, or $200$ times on the CIFAR-10 instance.
Each run has a step cap, set from the instance's characteristic time and the run's threshold.
A run that reaches the cap is kept.

\paragraph{Split.}
Let $E$ be the event that the recommendation is $\mathcal{S}^\star$.
Let $\lambda^\star$ attain $\inf_{\lambda \in \Alt_m(\nu)} \langle \bar w, k(\lambda)\rangle$ at the algorithm's average allocation $\bar w$.
By Proposition~\ref{prop:cl-topm-exact}(i), $\lambda^\star$ moves one model of $\mathcal{S}^\star$ and one model outside it to their $\bar w$-weighted mean.
As in Appendix~\ref{app:eval-bai},
\[
\frac{\mathbb{E}_\nu[\tau]}{T^\star_m(\nu)} = \kl(\delta, 1-\delta) + \textsc{uniform} + \textsc{process} + \textsc{allocation}
\]
with $\textsc{allocation} = \mathbb{E}_\nu[\tau]\big(T^\star_m(\nu)^{-1} - \langle \bar w, k(\lambda^\star)\rangle\big)$.
The sum $\textsc{uniform} + \textsc{process} = \langle \mathbb{E}_\nu[N(\tau)], k(\lambda^\star)\rangle - \kl(\delta, 1-\delta)$ is computed from the pull counts alone.
It is the information collected at $\lambda^\star$ beyond $\kl(\delta, 1-\delta)$ before the algorithm stops.
Its split into the two terms needs an estimate of $P_{\lambda^\star}(E)$.
No such estimate passed the exact control of Appendix~\ref{app:eval-own-error}, so the two terms are reported together.
The alternative is chosen on half of the runs and evaluated on the other half, and the halves then swap.
Write $T(w)$ for the characteristic time of an allocation $w$, the reciprocal of $\inf_{\lambda \in \Alt_m(\nu)} \langle w, k(\lambda)\rangle$.
The ratio $T(\bar w)/T^\star_m(\nu)$ is reported for every configuration.
The ledger identity $\mathbb{E}_\nu[L_\tau] = \langle \mathbb{E}_\nu[N(\tau)], k(\lambda^\star)\rangle$ is checked on the same runs.
Each expected sample size is reported with a $95\%$ BCa bootstrap interval over runs, and each term with a $95\%$ percentile bootstrap interval.

\paragraph{Readings.}
Of the $22{,}800$ runs, $22{,}786$ stop.
Three of them err, all under the heuristic threshold at $\delta = 0.1$ on the MMLU instance: one of $400$ with $m = 2$ and two of $400$ with $m = 3$.
No run errs under the union thresholds or under the own stopping rules.
The ledger identity holds within $2.1$ standard errors in $56$ of the $57$ configurations without capped runs.
In the last, Track-and-Stop under the first threshold with $m = 5$ at $\delta = 0.1$, the deviation is $3.51$ standard errors.
The largest of $57$ independent deviations reaches that size with probability $0.025$.
The deviation of $3.51$ sits in the Wald martingale $S_a(\tau) - \mu_a N_a(\tau)$ of one model, whose mean over the $55$ configurations on that instance is $0.16$ standard errors.

\begin{figure}[t]
\centering
\includegraphics[width=\linewidth]{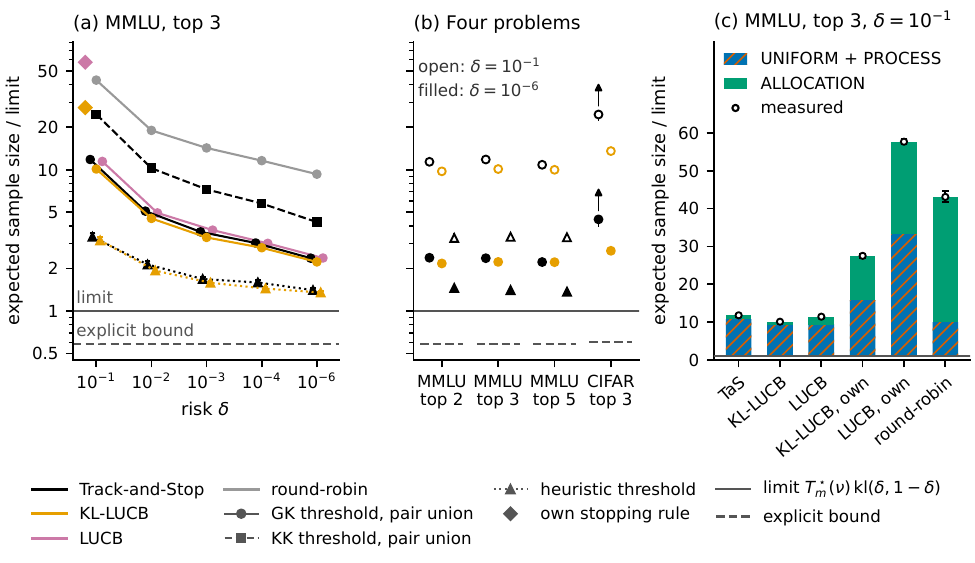}
\caption{\textbf{The exact top-$m$ limit is $1.65$ to $1.72$ times the explicit bound.} Under the union threshold of Garivier and Kaufmann the adaptive algorithms that stop every run cost $2.2$ to $2.7$ times the limit at $\delta = 10^{-6}$. Expected sample sizes relative to the limit $T^\star_m(\nu)\kl(\delta, 1-\delta)$, with $95\%$ intervals, most of them narrower than the markers. Dashed lines mark the explicit bound of Proposition~\ref{prop:cl-topm}. GK and KK are the thresholds of Garivier and Kaufmann and of Kaufmann and Koolen with a union over the swap pairs. (a) The MMLU instance with $m = 3$; diamonds mark the own stopping rules of LUCB and KL-LUCB at $\delta = 0.1$. (b) The four problems at $\delta = 10^{-1}$ (open markers) and $10^{-6}$ (filled). On the CIFAR-10 instance some runs of Track-and-Stop reach the step cap at each risk, so its markers are lower bounds. (c) At $\delta = 0.1$ on the MMLU instance with $m = 3$, the expected sample size split into the limit, the \textsc{uniform}-plus-\textsc{process} term and the \textsc{allocation} term, which add up to the measured ratio. In (c) TaS is Track-and-Stop. Every bar except the two own stopping rules uses the GK threshold.}
\label{fig:topm}
\end{figure}

On the MMLU instance every algorithm sits above the limit at every risk (Figure~\ref{fig:topm}).
At $\delta = 0.1$ Track-and-Stop, KL-LUCB and LUCB draw $9.7$ to $11.8$ times the limit under the first union threshold.
At $\delta = 10^{-6}$ they draw $2.2$ to $2.4$ times the limit.
Their excess is mostly the \textsc{uniform}-plus-\textsc{process} term, $13.3$ to $17.3$ nats at every risk.
Their \textsc{allocation} term is at most $4.9$ nats.
Under the second union threshold the \textsc{uniform}-plus-\textsc{process} term of Track-and-Stop is $38.3$ to $43.5$ nats.
Track-and-Stop then draws $24.7$ times the limit at $\delta = 0.1$ and $4.3$ times at $\delta = 10^{-6}$.
Under the heuristic threshold the term is $1.6$ to $4.8$ nats.
Under it Track-and-Stop and KL-LUCB draw $1.35$ to $1.46$ times the limit at $\delta = 10^{-6}$.
The explicit bound sits $1.65$ to $1.72$ times below the limit.
Every measured expected sample size is at least $2.3$ times the explicit bound.

On the MMLU instance under the first threshold the average allocations of Track-and-Stop and KL-LUCB have characteristic times within $13\%$ of $T^\star_m(\nu)$.
That of LUCB is within $23\%$.
With $m = 5$ at $\delta \le 10^{-3}$ the ratio $T(\bar w)/T^\star_m(\nu)$ of Track-and-Stop is the lower, by $0.04$ and $0.07$.
In the other nine comparisons on the MMLU instance the bootstrap interval of the difference contains $0$ and lies within $\pm 0.08$.
The average allocation of round-robin sampling has the characteristic time of uniform weights, $4.34$ times $T^\star_3(\nu)$.
Its \textsc{allocation} term is $58$ to $99$ nats.
Its expected sample size is $3.65$ to $3.93$ times that of Track-and-Stop, below the ratio $4.34$ of the two characteristic times.
Each algorithm's information is measured at its own $\lambda^\star$.
There the average allocation of Track-and-Stop collects $0.91$ to $0.96$ of the information per pull of the optimal one.
Track-and-Stop also stops after $1.05$ to $1.08$ times the information collected by round-robin sampling.

The own stopping rules of LUCB and KL-LUCB stop later than GLR stopping with the same sampling.
At $\delta = 0.1$ they cost $5.0$ and $2.7$ times as much as GLR stopping.
The \textsc{allocation} term accounts for $48\%$ and $62\%$ of the extra cost.
Their runs last $5.0$ and $2.7$ times as long, while the pulls of the seven weakest models grow only $2.6$ and $1.7$ times.
Those models then hold $11\%$ of the pulls, against $21\%$ and $18\%$ under GLR stopping.
Under these rules the most confusing alternative pairs the third model with one of the three weakest models.

On the CIFAR-10 instance KL-LUCB stops every run, at $13.6$ times the limit at $\delta = 0.1$ and $2.7$ times at $\delta = 10^{-6}$.
Of the $200$ runs of Track-and-Stop at each risk, $7$ reach the cap, so its ratios, $24.6$ and $4.45$, are lower bounds.
Its runs that stop draw $21.6$ and $3.9$ times the limit.
The average allocation of those runs has $1.51$ and $1.43$ times the characteristic time $T^\star_3(\nu)$, against $1.16$ and $1.14$ for KL-LUCB.
The $95\%$ intervals of these ratios, from $200$ runs per configuration, are disjoint at both risks.
At the average allocation of the stopped runs the swap pairs with the least information join the top models to models scoring $86.5$ to $89.7$.
The $11$ models scoring $87.6$ to $91.4$ hold $77\%$ and $83\%$ of their optimal share of the pulls.
The weights tracked by Track-and-Stop also move more on this instance.
After the first $10^4$ pulls, they move by a median $\ell_1$ distance of $0.051$ to $0.055$ per recomputation.
On the MMLU instance under the first threshold the median is $0.023$ to $0.030$.
The largest difference $|t\,w_a - N_a(t)|$ between a model's pulls and its tracked target is a median $5.3\%$ to $5.8\%$ of $t$.
On the MMLU instance it is $0.4\%$ to $1.1\%$.
The $14$ runs at the cap show the same deficit in its extreme form.
In each of them $22$ to $25$ of the $25$ weakest models sit at the forced-exploration floor of $\sqrt{t} - K/2$ pulls.
In $7$ of them a top model has dropped below the two models tied at $93.0$.
The empirical top-$3$ set then holds one model of that pair.
The weights at the empirical means put $96\%$ to $99.9\%$ of their mass on the tied pair.
In $10$ of the $14$ runs the tied pair holds $81\%$ to $98\%$ of all pulls, against $37\%$ under the optimal weights.
Counting the capped runs at their state at the cap, the \textsc{allocation} term of Track-and-Stop is $17.6$ and $22.6$ nats, against $3.3$ and $4.5$ for KL-LUCB.
Its \textsc{uniform}-plus-\textsc{process} term is $23.9$ and $25.1$ nats, against $18.8$ and $18.6$.

\paragraph{Tie tests.}
Two variants of the CIFAR-10 problem with $m = 3$ test the two models tied at $93.0$ and the schedule of the tracked weights.
For each variant the prediction specified before the run was that Track-and-Stop stops every run, with an average allocation of less than $1.25$ times the characteristic time.
In each, Track-and-Stop runs $200$ times at $\delta = 0.1$ and $\delta = 10^{-6}$ under the first union threshold, with the same step caps.
In the first variant the second model at $93.0$ scores $92.8$.
Three models, at $93.0$, $92.9$ and $92.8$, still lie within $0.2$ points of each other.
This counterfactual instance has $T^\star_3(\nu) = 20{,}703.1$, certified by a duality bracket of relative width $2 \times 10^{-12}$, and its ratios are read against it.
On it $2$ and $3$ of the $200$ runs reach the cap, against $7$ and $7$ on the published accuracies.
Over both risks the change from $14$ to $5$ capped runs has a two-sided Fisher exact $p$-value of $0.06$.
At $\delta = 0.1$ the average allocation of the runs that stop has $1.53$ times the characteristic time, with a $95\%$ BCa interval of $1.42$ to $1.64$.
At $\delta = 10^{-6}$ it has $1.37$ times, with $1.27$ to $1.48$.
Counting the capped runs at the cap, Track-and-Stop costs at least $21.5$ and $3.97$ times the limit, with $95\%$ intervals reaching $8\%$ below and $13\%$ above.
KL-LUCB stops every run, at $13.7$ and $2.72$ times the limit, each within $3\%$ at $95\%$.
In the second variant Track-and-Stop recomputes its weights whenever $t$ grows by $1\%$, so a cell makes about $1{,}100$ recomputations against about $240$.
On the published accuracies $4$ and $6$ of its $200$ runs then reach the cap.
Counting them at the cap, it costs at least $24.0$ and $4.52$ times the limit, with $95\%$ intervals reaching $10\%$ below and $13\%$ above.
At $\delta = 0.1$ the average allocation of its runs that stop has $1.63$ times $T^\star_3(\nu)$, with $1.50$ to $1.82$.
At $\delta = 10^{-6}$ it has $1.42$ times, with $1.23$ to $1.54$.
The duality bracket of the tracking solver is checked at one recomputation in $10$.
Under the faster schedule its relative width exceeds the target of $10^{-5}$ at $25$ of $182$ checks at $\delta = 0.1$ and at $43$ of $190$ at $\delta = 10^{-6}$.
The largest relative width is $0.038$, small beside the $63\%$ and $42\%$ by which the characteristic time of the average allocation exceeds $T^\star_3(\nu)$.
No run errs in either variant.
The lag of the pulls behind the tracked weights is as large under the faster schedule.
After the first $10^4$ pulls, the largest difference $|t\,w_a - N_a(t)|$ just before a recomputation is a median $7.0\%$ and $5.7\%$ of $t$, against $5.3\%$ and $5.8\%$ under the $5\%$ schedule.
Of the $15$ runs at the cap in the two variants, $5$ have a top model below the models just under the boundary.
The weights at the empirical means then put more than $97\%$ of their mass on two of those models.
Neither prediction holds.
The deficit therefore persists without the exact tie and with about five times as many recomputations.

\subsection{Gaussian arms from an oat variety trial}
\label{app:eval-field-trial}

\paragraph{Instance.}
For the 2024 oat variety trial at Minot, North Dakota, the grain yields of $22$ varieties are reported, with a trial mean of $121.8$ bushels per acre and a coefficient of variation of $8.9\%$ \cite{bortolon20242024}.
An arm is a variety and a pull is one plot.
The reward of a pull is Gaussian, with the published mean of the variety and the common standard deviation $\sigma = 0.089 \times 121.8 = 10.84$ bushels per acre.
The published least significant differences, $17.6$ at level $0.05$ and $14.7$ at level $0.10$, agree with three replicates and about $80$ residual degrees of freedom.
ND Spilde has the highest mean, $\mu^\star = 157.0$.
SD Titan follows at $152.2$, so the top gap is $4.8$ bushels per acre, or $0.44\,\sigma$.
Deon, Killdeer and AAC Douglas follow at $145.7$, $142.4$ and $136.6$.
Each published mean averages three plots and has standard error $6.3$, so the order of the top varieties is itself uncertain.
The readings are statements about the Gaussian instance at the published means.
A field season is not run plot by plot, so a sequential pull is a stylization.

\paragraph{Characteristic times.}
By the scalar program of Proposition~\ref{prop:scalar}, $T^\star(\nu) = 48.21$ plots.
Round-robin sampling has characteristic time $9.31\,T^\star(\nu)$.
For top-$m$ identification the characteristic time $T^\star_m(\nu)$ of Proposition~\ref{prop:cl-topm-exact} is $34.23$, $105.00$ and $62.01$ plots for $m = 2, 3, 4$.
Each value is certified by two duality brackets of relative width at most $1.2 \times 10^{-12}$.
At $m = 1$ the same solver returns $T^\star(\nu)$.
The explicit sum of Proposition~\ref{prop:cl-topm} is smaller by factors of $1.58$, $1.72$ and $1.56$.
The case $m = 5$ is left out, since its boundary gap is $0.8$ bushels per acre and $T^\star_5(\nu)$ is about $1{,}800$ plots.

\paragraph{Identification.}
The algorithms follow Appendix~\ref{app:eval-bai}, written for Gaussian arms.
Track-and-Stop uses D-tracking toward the weights of the scalar program, recomputed whenever $t$ grows by $5\%$.
It forces every arm with fewer than $\sqrt t - K/2$ pulls \cite{garivier2016optimal}.
T3C draws its leader from the Gaussian posterior and takes as challenger the arm of least transportation cost \cite{shang2020fixed}.
For top-$m$ identification, Track-and-Stop tracks the weights of $T^\star_m(\nu)$.
KL-LUCB pulls in turn the arm of the empirical top $m$ with the smallest lower confidence bound and the arm outside it with the largest upper confidence bound \cite{kaufmann2013information}.
Every algorithm stops by the generalized likelihood ratio rule.
Write $P$ for the number of arm pairs a wrong answer must reverse, $K - 1$ for best-arm and $m(K - m)$ for top-$m$ identification.
Two thresholds are $\delta$-correct.
The first applies the Gaussian calibration of Kaufmann and Koolen to each pair at level $\delta/P$ \cite{kaufmann2021mixture}.
The second is the threshold of Kaufmann and Koolen used in Appendix~\ref{app:eval-bai}, with $P$ in its union term.
The third, $\log((\log t + 1)/\delta)$, is the heuristic of Appendix~\ref{app:eval-bai}.
Best-arm Track-and-Stop runs with all three thresholds.
T3C, KL-LUCB and round-robin sampling run with the first.
Top-$m$ Track-and-Stop runs with the first and the heuristic.
The risks are $\delta \in \{10^{-1}, 10^{-2}, 10^{-3}, 10^{-4}, 10^{-6}\}$, $85$ configurations in all, each run $1{,}000$ times.
No run reached its step cap.

\paragraph{Split.}
Let $E$ be the event that the recommendation is correct.
Let $\lambda^\star$ attain the infimum of $\langle \bar w, k(\lambda)\rangle$ over the alternatives, at the algorithm's average allocation $\bar w$.
It moves one arm of the correct answer and one arm outside it to their $\bar w$-weighted mean.
As in Appendix~\ref{app:eval-bai},
\[
\frac{\mathbb{E}_\nu[\tau]}{T^\star_m(\nu)} = \kl(\delta, 1-\delta) + \textsc{uniform} + \textsc{process} + \textsc{allocation}
\]
with $T^\star_1(\nu) := T^\star(\nu)$ and $\textsc{allocation} = \mathbb{E}_\nu[\tau]/T^\star_m(\nu) - D_\tau$, where $D_\tau = \langle n, k(\lambda^\star)\rangle$ is taken at $\lambda^\star$ (Theorem~\ref{thm:allocation}).
The sum $\textsc{uniform} + \textsc{process}$ is $\mathbb{E}_\nu[L_\tau] - \kl(\delta, 1-\delta)$.
By Theorem~\ref{thm:allocation} it equals $D_\tau - \kl(\delta, 1-\delta)$, the information collected at $\lambda^\star$ beyond $\kl(\delta, 1-\delta)$.
For Gaussian arms $L_\tau$ is a function of the per-run counts and reward sums.
The sum is drawn in Figure~\ref{fig:ident-tightness} as $D_\tau - \kl(\delta, 1-\delta)$ and is given in the identification readings below as $\mathbb{E}_\nu[L_\tau] - \kl(\delta, 1-\delta)$.
The two estimates differ by the sampling error of the ledger identity, which is checked below.
The alternative is chosen on one half of the runs and evaluated on the other, and the halves then swap.
Intervals are $95\%$ percentile bootstrap intervals over runs, with each half resampled on its own and its alternative held fixed.

Splitting the sum needs $P_{\lambda^\star}(E)$.
No estimate of that probability passed the exact control of Appendix~\ref{app:eval-own-error}, so the two terms are reported together.
Two bounds on the split need no estimate.
By Lemma~\ref{lem:relax}(a), \textsc{process} is nonnegative.
Each $\delta$-correct threshold caps at $\delta/P$ the chance that, for a given pair of arms $i, j$, the sum $\sum_{a \in \{i,j\}} N_a(t)\, d(\hat\mu_a(t), \mu_a)$ ever crosses the pair's threshold.
Under $\lambda^\star$ the two moved arms share one mean.
A run that stops in $E$ has their generalized likelihood ratio statistic above the pair's threshold.
That statistic is at most the sum above at the shared mean, so $P_{\lambda^\star}(E) \le \delta/P$ and $\textsc{uniform} \ge \kl(P_\nu(E), \delta/P) - \kl(\delta, 1-\delta)$.
With $P_\nu(E)$ at its one-sided $95\%$ exact binomial lower bound, this floor is $3.0$ to $3.6$ nats for best-arm and $3.6$ to $4.8$ nats for top-$m$ identification.
Between the floor and the sum either term can be the larger.

\paragraph{Identification readings.}
The ledger identity $\mathbb{E}_\nu[L_\tau] = \sum_a \mathbb{E}_\nu[N_a(\tau)]\, d(\mu_a, \mu_a(\lambda^\star))$ holds within $2.5$ standard errors in $84$ of the $85$ configurations.
The exception is best-arm Track-and-Stop with the first threshold at $\delta = 10^{-6}$, at $2.95$ standard errors.
Over $85$ independent configurations a deviation at least that large has probability $0.23$.
No run with a $\delta$-correct threshold recommended a wrong answer.
The heuristic threshold erred in $1$, $1$ and $3$ of $1{,}000$ runs at $\delta = 0.1$, for $m = 1$, $2$ and $3$.

For T3C the expected sample size is $2.2$ to $10.2$ times the limit $T^\star(\nu)\kl(\delta, 1-\delta)$.
The information it collects at $\lambda^\star$ beyond $\kl(\delta, 1-\delta)$ is $15.3$ to $16.0$ nats, $95\%$ to $98\%$ of the excess.
Its \textsc{allocation} term is $0.14$ to $0.65$ nats.
For KL-LUCB the sum $\textsc{uniform} + \textsc{process}$ is $15.8$ to $17.1$ nats.
Its \textsc{allocation} term is $1.3$ to $5.3$ nats, with every interval above zero, and $7\%$ to $24\%$ of the excess.
For round-robin sampling \textsc{allocation} is $81\%$ to $94\%$ of the excess.
For best-arm identification its expected sample size is $9.29$ to $9.30$ times $T^\star(\nu) D_\tau$, which matches its characteristic time, $9.31\,T^\star(\nu)$.

For best-arm Track-and-Stop with a $\delta$-correct threshold the \textsc{allocation} term is $22$ to $49$ nats under the first threshold and $67$ to $88$ nats under the second.
At every risk it exceeds $\textsc{uniform} + \textsc{process}$, which is $16.1$ to $16.5$ and $35.8$ to $41.7$ nats, and the interval of the difference lies above zero.
Both of those terms are nonnegative, so \textsc{allocation} is the largest term with no estimate of $P_{\lambda^\star}(E)$.
The same holds for top-$3$ Track-and-Stop at every risk, with \textsc{allocation} at $21$ to $35$ nats against $20.2$ to $22.3$.
For top-$2$ Track-and-Stop \textsc{allocation} exceeds the sum at $\delta = 10^{-6}$ and falls below it at $\delta \ge 10^{-2}$.
At $\delta = 10^{-3}$ and $10^{-4}$ the interval of their difference contains zero.
For top-$4$ Track-and-Stop \textsc{allocation}, $12.3$ to $18.0$ nats, falls below the sum, $19.3$ to $20.8$ nats, at every risk.

The forced exploration of Track-and-Stop holds every arm at $\sqrt t - K/2$ pulls or more.
With $22$ arms, best-arm Track-and-Stop puts $20\%$ to $35\%$ of its pulls on the $19$ to $20$ varieties at that floor.
Their optimal weights total $12\%$ to $15\%$.
D-tracking pulls whichever arm is furthest below its target in absolute count.
The tracked arms therefore fall short of their targets by one common count.
The two best varieties receive $0.33$ to $0.39$ of the pulls, against optimal weights of $0.42$.
Deon receives $0.014$ to $0.018$ of the pulls, against an optimal weight of $0.042$.
The most confusing pair becomes ND Spilde and Deon.
The characteristic time of the average allocation is $2.2$ to $2.8$ times $T^\star(\nu)$.
The two rules alone, applied at the optimal weights over the measured stopping times, predict this allocation.
The \textsc{allocation} term of the prediction exceeds the measured one by $2\%$ to $7\%$ for best-arm and by $2\%$ to $21\%$ for top-$m$ identification.
The instances of Appendix~\ref{app:eval-bai} have characteristic times $72$ and $116$ times larger.
There the forced floor holds $0.8\%$ to $3.6\%$ of the pulls of Track-and-Stop.
Its \textsc{allocation} term on those instances is $3.1$ to $10.6$ nats.

\paragraph{The bound at the algorithm's own error rates.}
Before its \textsc{uniform} step the chain gives a bound at the algorithm's own error rates:
\[
\mathbb{E}_\nu[\tau] \;\ge\; T^\star_m(\nu)\, D_\tau \;\ge\; T^\star_m(\nu)\, \kl\big(P_\nu(E), P_{\lambda^\star}(E)\big)
\]
Write $T(w)$ for the characteristic time of an allocation $w$, the reciprocal of the infimum of $\langle w, k(\lambda)\rangle$ over the alternatives.
At $\lambda^\star$, $D_\tau = \mathbb{E}_\nu[\tau]/T(\bar w)$, and $T(\bar w) \ge T^\star_m(\nu)$ gives the first inequality.
The second is Lemma~\ref{lem:relax}(a).
The right side needs $P_{\lambda^\star}(E)$ and is not reported.
The ratio of $\mathbb{E}_\nu[\tau]$ to it is still bounded from both sides.
By the first inequality it is at least $T(\bar w)/T^\star_m(\nu)$.
By $P_{\lambda^\star}(E) \le \delta/P$ it is at most $\mathbb{E}_\nu[\tau]/(T^\star_m(\nu)\kl(P_\nu(E), \delta/P))$.
For T3C these bounds are $1.03$ to $1.04$ and $1.8$ to $3.4$.
For KL-LUCB they are $1.06$ to $1.17$ and $1.8$ to $3.4$.
The first inequality is tight for both algorithms: the upper ends of the intervals of $T(\bar w)/T^\star_m(\nu)$ are at most $1.06$ and $1.19$.
For Track-and-Stop with a $\delta$-correct threshold the lower bound is $1.49$ to $2.81$.
Its interval lies above $1.5$ in $24$ of the $25$ configurations.
In those $24$ the expected sample size exceeds the bound at the algorithm's own error rates by more than half of the bound.

\paragraph{The factor of two.}
The instances are one-armed reductions of the trial, as in Appendix~\ref{app:eval-factor-two}.
The arm $a^\star$ is ND Spilde, known at $\mu^\star = 157.0$.
The perturbed arm $b$ is SD Titan, Deon, Killdeer or AAC Douglas.
The alternative $\lambda$ raises arm $b$ to $\mu^\star + \xi$, with margin $\xi \in \{0.1\,\sigma, 0.3\,\sigma\}$.
KL-UCB pulls arm $b$ while $N_b = 0$, while its empirical mean is at least $\mu^\star$, or while $N_b\, d(\hat\mu_b, \mu^\star) \le \log t$ \cite{cappeGarivierMaillardMunosStoltz2013klucb}.
Thompson sampling with a flat prior pulls arm $b$ when a draw from the Gaussian posterior exceeds $\mu^\star$ \cite{thompson1933likelihood}.
By Proposition~\ref{prop:ratio} the per-pull factor is $2$ at every separation.
Each of the $16$ combinations of arm, margin and algorithm runs $200{,}000$ times under $\nu$.
It runs as many times under the tilted law $Q_{1/2}$ of Proposition~\ref{prop:tilt}, and is read at $14$ horizons from $25$ to $3{,}000$.
Under $Q_{1/2}$ arm $b$ draws from $\mathcal{N}(\mu_b + g/2, \sigma^2)$, the normalized geometric mixture of its two laws.
Write $U_T := \sum_a N_a(T)\, c_a(1/2)$.
By the proof of Proposition~\ref{prop:tilt}, $e^{L_T/2} = M_T(1/2)\, e^{-U_T}$, so $C_T(1/2) = -\log \mathbb{E}_{Q_{1/2}}[e^{-U_T}]$, with a bounded integrand.
Since $\min(e^{L_T}, 1) = e^{L_T/2} e^{-|L_T|/2}$, the same step yields
\[
1 - \TV(P_\nu, P_\lambda) \;=\; \mathbb{E}_\lambda\big[\min(e^{L_T}, 1)\big] \;=\; \mathbb{E}_{Q_{1/2}}\!\left[e^{-U_T - |L_T|/2}\right]
\]
Plain Monte Carlo under $\nu$ estimates both quantities a second time.
Its estimate of $C_T(1/2)$ averages the heavy-tailed weight $e^{-L_T/2}$, whose relative variance under $\nu$ is $e^{2C_T(1/2)} - 1$.
Six fixed allocations of $10$, $100$ and $1{,}000$ pulls to arm $b$ serve as positive controls, since there the path factor is exactly $2$.
Intervals are $\pm 1.96$ delta-method standard errors.

\paragraph{Readings on the reductions.}
At a fixed allocation the tilted estimator returns the path factor $2$ exactly.
Its value of $1 - \TV$ lies within $0.8\%$ of the exact value.
Plain Monte Carlo reads path factors of $1.63$ and $1.45$ at $1{,}000$ pulls, so the tilted estimator is the one reported.
Under adaptive play the two estimators of $C_T(1/2)$ agree within $2$ standard errors at $220$ of the $224$ combinations of reduction and horizon.
The two estimators of $1 - \TV$ agree at $217$ of the $224$.

By Proposition~\ref{prop:half} the path factor exceeds $1$ by the Jensen defect $\KL(P_\nu \,\|\, P^{(1/2)})$ over $C_T(1/2)$.
For a Gaussian $L_T$ the defect is $\mathrm{Var}_\nu(L_T)/8$.
At a fixed allocation $\mathrm{Var}_\nu(L_T) = 2D_T$.
Over the $16$ combinations the path factor follows the ratio $\mathrm{Var}_\nu(L_T)/(2D_T)$, with rank correlation $0.93$ at $T = 25$ and $0.86$ at $T = 3{,}000$.
At $T = 25$ the path factor of KL-UCB on SD Titan is $1.64$ and $1.60$ at the two margins, where the ratio is $0.57$ and $0.54$.
At the same horizon the path factor is $1.07$ to $1.45$ on the other $14$ combinations, where the ratio is $0.11$ to $0.47$.
Deon, Killdeer and AAC Douglas lie $1.0\,\sigma$ to $1.9\,\sigma$ below $\mu^\star$.
Under KL-UCB their transported fractions already exceed $1$ by $T = 25$.
SD Titan's is $0.54$ at that horizon.
At $T = 3{,}000$ the path factor is $1.03$ to $1.11$ on $14$ of the $16$ combinations.
On SD Titan at $\xi = 0.3\,\sigma$ it is $1.18$ under Thompson sampling.
Under KL-UCB it is $1.44$, with interval $[1.41, 1.46]$.
For that combination $\mathrm{Var}_\nu(L_T)$ is $52\%$ of $2D_T$.
On the other combinations it is $6\%$ to $29\%$.
Under $P_\nu$, $L_T = N_b(T)\, k_b - (g/\sigma^2)\, S_b$, with $k_b := \KL(\nu_b \,\|\, \lambda_b)$ and $S_b$ the sum of arm $b$'s centered rewards.
Since $\mathbb{E}_\nu[S_b^2] = \sigma^2\, \mathbb{E}_\nu[N_b(T)]$,
\[
\frac{\mathrm{Var}_\nu(L_T)}{2D_T} \;=\; 1 + \frac{k_b\, \mathrm{Var}_\nu(N_b(T))}{2\,\mathbb{E}_\nu[N_b(T)]} - \frac{g}{\sigma^2}\, \frac{\mathrm{Cov}_\nu(N_b(T), S_b)}{\mathbb{E}_\nu[N_b(T)]}
\]
The law of the runs under $P_\nu$ does not depend on the margin.
So the second term scales as $g^2$ and the third as $g$.
Both scalings hold for the measured terms to within $1.2\%$.
On SD Titan under KL-UCB the count has standard deviation $40$ around a mean of $79$ pulls at $T = 3{,}000$.
Between the two margins its term rises from $1.49$ to $2.80$.
The covariance term rises from $2.39$ to $3.28$.

At $T = 3{,}000$ the tilted law pulls the perturbed arm $3.3$ to $7.3$ times as often as $P_\nu$ does on $14$ combinations.
On SD Titan at $\xi = 0.3\,\sigma$ the multiple is $10.5$ under Thompson sampling and $29.2$ under KL-UCB.
Under $Q_{1/2}$ arm $b$ sits $(\Delta_b - \xi)/2$ below $\mu^\star$, which is $0.77$ bushels per acre for SD Titan at that margin.
With pull counts growing as $\log T$ over the per-pull divergence to $\mu^\star$, the multiple is about $(2\Delta_b/(\Delta_b - \xi))^2$.
For SD Titan at $\xi = 0.3\,\sigma$ under KL-UCB this gives $38.5$, a count above the horizon.
There the tilted law pulls SD Titan in $2{,}310$ of the $3{,}000$ rounds.
On the other KL-UCB combinations it gives $4.5$ to $7.9$, which exceeds the measured multiple by $3\%$ to $14\%$.

At $T = 3{,}000$ the affinity bound of Theorem~\ref{thm:limit} lies a factor of $3.2$ to $9.2$ below $1 - \TV$ on $13$ of the $16$ combinations.
On the other three, all under KL-UCB, it lies a factor of $13$, $12$ and $170$ below.
For a Gaussian $L_T$ with the measured mean and variance the factor would be $2.7$ to $9.2$ on $15$ combinations and about $300$ on SD Titan at $\xi = 0.3\,\sigma$ under KL-UCB.
Under KL-UCB on SD Titan at $\xi = 0.1\,\sigma$ and on Deon at $\xi = 0.3\,\sigma$, the measured $1 - \TV$ is $5.2$ and $3.5$ times its Gaussian value.
The Bretagnolle--Huber bound lies below the affinity bound on every combination, by a factor of $1.4$ to $750$.

\paragraph{Regret.}
All $22$ arms are in play.
KL-UCB uses the Gaussian index with exploration $\log t$ and Thompson sampling uses flat priors.
IMED uses its Gaussian index \cite{honda2015nonasymptotic}.
Each algorithm runs $1{,}000$ times to $T = 10{,}000$ on $\nu$ and on each alternative $\lambda^{(b,\xi_\Delta)}$ of Corollary~\ref{cor:cl-gauss}.
The alternative raises one suboptimal arm $b$ to $\mu_b + (1+\xi_\Delta)\Delta_b$, with $\xi_\Delta \in \{1/2, 1\}$, which gives $42$ alternatives.
Counts are read at $T = 100$, $300$, $1{,}000$, $3{,}000$ and $10{,}000$.
For arm $b$ the measured cost is $D_T = \mathbb{E}_\nu[N_b(T)]\, \KL(\nu_b \,\|\, \lambda_b)$.
The error-sum form of Proposition~\ref{prop:cl-finite} is $\log(1/(\varepsilon_T(2 - \varepsilon_T)))$.
Its regret form is $\log(T\underline{\Delta}/(4(R_T(\nu) + R_T(\lambda))))$, with $\underline{\Delta} = \xi_\Delta\Delta_b$ and the measured regrets.
Corollary~\ref{cor:cl-gauss} is evaluated in units of $\sigma$ with $p = 1/2$ and $\Gamma T^p$ replaced by the largest measured regret over the $43$ instances.
A maximum over $43$ instances of the neighborhood can only fall below the maximum over all of it.
So the computed bound can only exceed the bound licensed by the corollary.
In a long-horizon simulation each algorithm runs $10^6$ times on $\nu$ and on $\lambda^{(b,1/2)}$ for SD Titan and Deon, with fresh seeds.
It keeps $N_b(T)$ for those two arms at $T = 1{,}000$, $3{,}000$ and $10{,}000$.
It covers the margin $\xi_\Delta = 1/2$ only.

\paragraph{Regret readings.}
The regret form is negative at every horizon up to $T = 1{,}000$ on the two closest arms, SD Titan and Deon.
It turns positive at $T = 3{,}000$ on Deon, and on SD Titan at $\xi_\Delta = 1$ under Thompson sampling and IMED.
It turns positive at $T = 10{,}000$ on the four remaining combinations of arm, margin and algorithm.
It is positive at $20$ of the $60$ combinations of algorithm, arm, margin and horizon, all at $T \ge 3{,}000$ and each with its interval above zero.
It is positive exactly when $(R_T(\nu) + R_T(\lambda))/T < \xi_\Delta\Delta_b/4$.
The measured sum per round falls from between $1.05$ and $1.34$ bushels per acre at $T = 3{,}000$ to between $0.35$ and $0.47$ at $T = 10{,}000$.
The thresholds $\xi_\Delta\Delta_b/4$ are $0.60$ and $1.20$ on SD Titan and $1.41$ and $2.83$ on Deon.
Where positive, the regret form keeps at most $9\%$ of $D_T$.
On these arms $D_T$ is $11.6$ to $35.4$ nats at $T = 10{,}000$.
The count of Corollary~\ref{cor:cl-gauss} is at most $8.8\%$ of the measured count.
It is zero at $324$ of the $630$ combinations of algorithm, arm, margin and horizon.
Its regret bound is at most $6.6\%$ of the measured regret at $T = 10{,}000$.

The error-sum form needs the event probabilities $P_\nu(N_b(T) > T/2)$ and $P_\lambda(N_b(T) \le T/2)$.
With $1{,}000$ runs per instance, some run crossed $T/2$ in one of the two worlds at only $62$ of the $630$ combinations.
Elsewhere the form is bounded below by $4.43$ nats, its value with each probability at its one-sided $95\%$ exact binomial upper bound.
On the two closest arms $D_T$ grows to $35$ nats.
There the share of $D_T$ kept by the form grows over the short horizons.
On $9$ of the $12$ combinations of algorithm, arm and margin its interval at $T = 300$ lies above the one at $T = 100$.
On SD Titan at $\xi_\Delta = 1/2$ the interval at $T = 1{,}000$ also lies above the one at $T = 300$, under all three algorithms.
There the share rises from $0.04$ to $0.08$ at $T = 100$ to $0.41$ to $0.55$ at $T = 1{,}000$.

The long-horizon simulation resolves the alternatives with $\xi_\Delta = 1/2$ to $T = 10{,}000$.
With $10^6$ runs per instance, between $2$ and $20{,}733$ runs cross $T/2$ in the two worlds together at each combination of algorithm, arm and horizon.
The interval of the share then has half-width at most $0.017$ at $T = 1{,}000$, $0.036$ at $T = 3{,}000$ and $0.096$ at $T = 10{,}000$.
At each horizon its values lie within the intervals from the runs of $1{,}000$ per instance.
Between $T = 1{,}000$ and $T = 3{,}000$ the interval at the later horizon lies above the earlier one on $4$ of the $6$ combinations of algorithm and arm: SD Titan under all three algorithms, and Deon under Thompson sampling.
Between $T = 3{,}000$ and $T = 10{,}000$ this holds only for Thompson sampling on SD Titan, whose share rises from $0.495$ to $0.535$.
At $T = 10{,}000$ the form keeps $0.54$ to $0.71$ of $D_T$.
There $D_T$ is $11.8$ to $19.8$ nats.
The two horizons share their runs, so the change between them can also be read from paired resamples.
On SD Titan from $T = 3{,}000$ to $T = 10{,}000$ the share falls by $0.023$ under KL-UCB, with paired $95\%$ interval $[-0.038, -0.005]$.
Under IMED it rises by $0.003$, with interval $[-0.008, 0.015]$.
On Deon only $2$ to $48$ runs cross $T/2$ at $T = 10{,}000$.
There the share moves by $-0.003$ to $0.015$ between the two horizons.

On SD Titan the error sum falls more slowly at the larger horizons, while $D_T$ keeps its rate of growth.
The error sum falls like $T^{-a}$.
Over $T = 1{,}000$ to $3{,}000$ the local exponent $a$ is $1.6$ to $2.2$.
Over $T = 3{,}000$ to $10{,}000$ it is $1.2$ to $1.5$.
Over both spans $D_T$ grows by $1.6$ to $3.2$ nats per unit of $\log T$, with a change of at most $7\%$ between the spans.
Between $T = 3{,}000$ and $10{,}000$ the slower decay stops the growth of the share under KL-UCB and IMED.
At these error sums the affinity route adds at most $0.002$ nats to the Bretagnolle--Huber form $\log(1/(2\varepsilon_T))$ at $T \ge 3{,}000$.

\begin{figure}[tp]
\centering
\includegraphics[width=\linewidth]{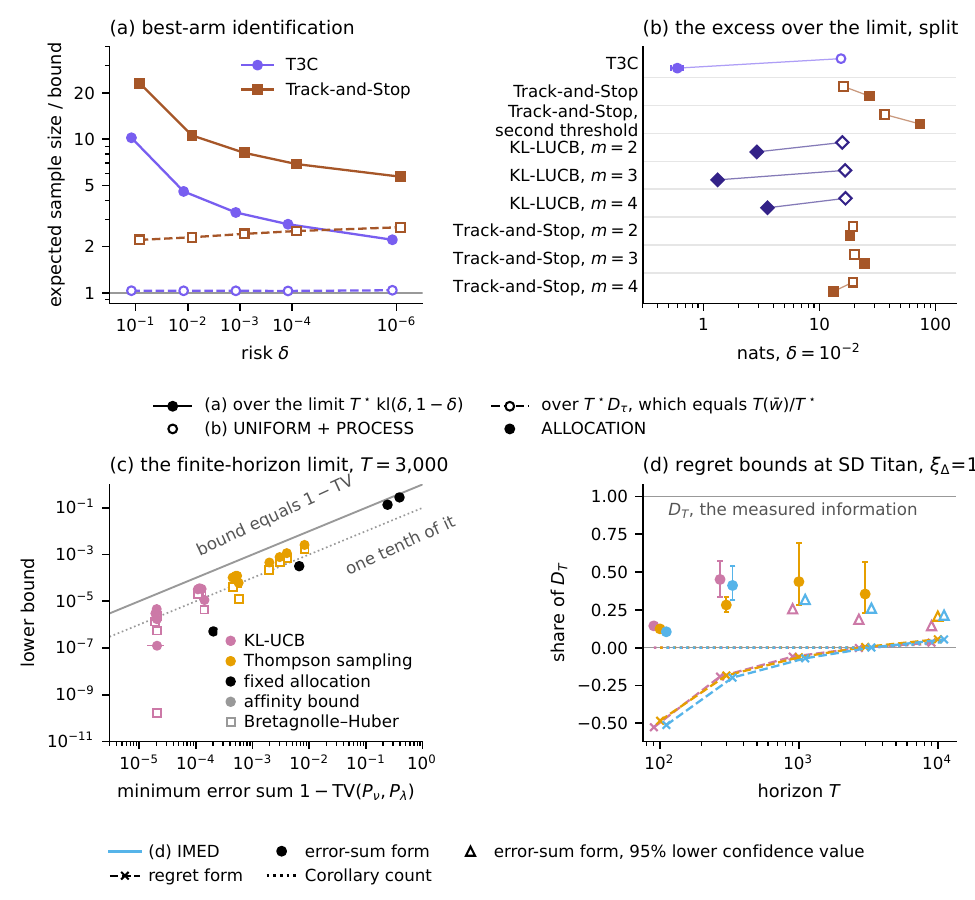}
\caption{\textbf{On an oat variety trial, allocation accounts for most of Track-and-Stop's excess over the identification limit and for at most $4\%$ of T3C's.} (a) Best-arm identification under the first $\delta$-correct threshold. The expected sample size of T3C and of Track-and-Stop is divided by the limit $T^\star(\nu)\kl(\delta, 1-\delta)$ (filled markers, solid) and by $T^\star(\nu) D_\tau$, with $D_\tau$ taken at $\lambda^\star$ (open, dashed). The second ratio equals $T(\bar w)/T^\star(\nu)$. Neither ratio uses an estimate of $P_{\lambda^\star}(E)$. (b) At $\delta = 10^{-2}$, the excess over the limit in nats, split into $\textsc{uniform} + \textsc{process}$ (open) and \textsc{allocation} (filled). Every row uses the first threshold except the one marked as using the second. Round-robin sampling, whose \textsc{allocation} term is $87$ to $170$ nats at this risk, is omitted. (c) The affinity bound of Theorem~\ref{thm:limit} (filled) and the Bretagnolle--Huber bound (open) against the minimum error sum $1 - \TV(P_\nu, P_\lambda)$, on the $16$ one-armed reductions at $T = 3{,}000$ and on four fixed allocations (black). (d) On SD Titan at $\xi_\Delta = 1$, four lower bounds as a share of $D_T$: the error-sum form of Proposition~\ref{prop:cl-finite} (circles), its value with both probabilities at their one-sided $95\%$ upper bounds where no run crossed $T/2$ (open triangles), the regret form (dashed) and the count of Corollary~\ref{cor:cl-gauss} (dotted). Colors in (d) follow (c). Intervals are $95\%$ bootstrap intervals in (a, b), delta-method intervals in (c) and exact binomial intervals in (d).}
\label{fig:field-trial}
\end{figure}

\paragraph{The slack along the chain.}
The identification ratios, the split at $\delta = 10^{-2}$, the testing step and the regret bounds are drawn in Figure~\ref{fig:field-trial}.
For T3C and KL-LUCB most of the excess over the identification limit is the information collected at $\lambda^\star$ beyond $\kl(\delta, 1-\delta)$, $15$ to $17$ nats.
For best-arm and top-$3$ Track-and-Stop most of the excess is the \textsc{allocation} term, set by its forced exploration on $22$ arms.
At the testing step the affinity bound comes within a factor of $3.2$ to $9.2$ of the minimum error sum on $13$ of $16$ reductions.
On the regret chain the regret form and the count of the corollary keep at most $9\%$ of $D_T$.
Where measured at $T \ge 300$ on the two closest arms with $1{,}000$ runs per instance, the error-sum form keeps $23\%$ to $65\%$ of $D_T$.
Between $T = 3{,}000$ and $10{,}000$ its share grows measurably only under Thompson sampling on SD Titan.

\paragraph{Scope.}
The statements are about the Gaussian law at the published means and the common standard deviation of one trial, at one site in one year.
The standard deviation comes from the trial's coefficient of variation, which pools entries the report does not list.
The listed means average $120.8$, against the trial mean of $121.8$.
The one-armed reductions know $a^\star$, so the path factors describe one perturbed arm against a known mean.
At eight interior alternatives with the heuristic threshold at $\delta \in \{0.05, 0.1\}$, the change-of-measure estimate of the error probability agrees with $20{,}000$ direct runs within $0.21$ nats in the log.
There the probabilities are $0.002$ to $0.021$, large enough for direct runs.

\subsection{Sequential A/B tests}
\label{app:eval-ab}

\paragraph{Instance.}
Radcliffe's report of the Hillstrom e-mail experiment gives, for the customers who received an e-mail featuring men's merchandise (21{,}307) and for the control group (21{,}306), a purchase rate of $1.25\%$ against $0.57\%$ and a site-visit rate of $18.28\%$ against $10.62\%$ \cite[Tables~1--2]{radcliffe2008hillstroms}.
Each metric is one instance with control rate $p_0$ and treatment rate $p_1$.

\paragraph{Tests.}
Wald's SPRT observes the treatment stream and tests the rate $p_0$ against the rate $p_1$, with Wald's thresholds $(1-\beta)/\alpha$ and $\beta/(1-\alpha)$ on the likelihood ratio and $\beta = \alpha$.
It runs $20{,}000$ times under each hypothesis at each level.
The mixture SPRT observes (control, treatment) pairs and uses the statistic of Johari, Koomen, Pekelis and Walsh \cite[(12)]{johari2017peeking}.
Its mixing variance is $(p_1 - p_0)^2$, the squared true effect, which tunes the test in its favor.
It rejects when the statistic reaches $1/\alpha$ and runs $10{,}000$ times under the truth; every run rejects.
The levels are $\alpha \in \{10^{-1}, 0.05, 10^{-2}, 10^{-3}, 10^{-4}\}$.

\paragraph{Error probabilities.}
The SPRT's error probabilities are computed by change of measure.
The type-I error is $\alpha' = \mathbb{E}_\nu[\mathbf{1}\{\text{accept } p_1\}\, e^{-L_\tau}]$ over the runs under $p_1$.
The type-II error is $\beta' = \mathbb{E}_\lambda[\mathbf{1}\{\text{accept } p_0\}\, e^{L_\tau}]$ over the runs under $p_0$, with $\lambda$ the instance of rate $p_0$.
Both lie below Wald's bounds $\alpha/(1-\beta)$ and $\beta/(1-\alpha)$ at every level and on both metrics.
The direct error frequencies agree with them within sampling error.
For the mixture test, the null $\mathcal{H}_0$ of equal rates is evaluated at $(q^\star, q^\star)$, where $q^\star = (p_0 + p_1)/2$ is the barycentre of the two rates.
This null is exchangeable and the statistic is symmetric in the two arms, so the null rejects equally often with the treatment ahead and behind.
The runs under the truth almost never reject behind (3 of $100{,}000$).
By the change of measure, the probability of the event $E_+$ of rejecting ahead is $\alpha'_+ = \mathbb{E}_\nu[\mathbf{1}_{E_+} e^{-L_\tau}]$, and the null's rejection probability is $\alpha' = 2\alpha'_+$.
Averaged over the runs under the truth, $\mathbf{1}_{E_+} e^{-L_\tau}$ falls $26\%$ to $50\%$ below $\alpha'_+$, because the importance weights have a heavy tail.
The two probabilities are therefore computed by a program that is exact up to a horizon.
The law of the two success counts under $(q^\star, q^\star)$ is propagated by dynamic programming, with the mass at which the test rejects removed at each step, up to $30$ times the limit for the purchase rate and $60$ times for the site-visit rate.
The two halves are equal, as exchangeability requires.
The probability of rejecting after that horizon, $0.11\alpha$ to $0.16\alpha$, is extrapolated from the first-passage density of the rejection boundary, fitted to the last horizons.
When the extrapolation is fitted at two thirds of the horizon, the program's rejection probability between the two horizons is predicted to within $1\%$.
The same total, within $0.003\alpha$, is given by the martingale mass left at the horizon.
At $\alpha \in \{0.1, 0.05\}$, $10{,}000$ direct runs under $(q^\star, q^\star)$ up to $10$ times the limit agree with the program at that horizon.

\paragraph{Split.}
Under the truth, $\mathbb{E}_\nu[\tau]$ times the divergence per observation equals $\mathbb{E}_\nu[L_\tau]$ by Theorem~\ref{thm:allocation}.
For the SPRT that divergence is $\kl(p_1, p_0)$.
For the mixture test it is the per-pair divergence $\Kinf(\nu, \mathcal{H}_0) = \kl(p_0, q^\star) + \kl(p_1, q^\star)$.
For the SPRT, $\mathbb{E}_\nu[L_\tau]$ is $\kl(1-\alpha, \alpha)$ plus a \textsc{uniform} term and a \textsc{process} term.
The \textsc{uniform} term is $\kl(1-\beta', \alpha') - \kl(1-\alpha, \alpha)$.
It equals the defect of the \textsc{uniform} relaxation of Section~\ref{sec:classical} wherever both realized error rates are at most nominal.
On the purchase metric at $\alpha = 0.1$ and $0.05$ the type-II error is $2.1\%$ and $0.8\%$ above nominal.
The \textsc{uniform} term is positive in every setting.
The defect of data processing to the decision is $\textsc{process} = \mathbb{E}_\nu[L_\tau] - \kl(1-\beta', \alpha')$.
For the mixture test, $\mathbb{E}_\nu[L_\tau]$ is $\log(1/\alpha) + \log 2$ plus a \textsc{uniform} term and a \textsc{process} term.
At $\alpha = 0.1$ the sum misses by at most $0.003$ nats, because a few runs under the truth do not reject ahead.
Its \textsc{uniform} term is $\log(\alpha/\alpha')$, the defect of the \textsc{uniform} relaxation at the barycentre null, and $\textsc{process} = \mathbb{E}_\nu[L_\tau] - \kl(P_\nu(E_+), \alpha'_+)$.
The ledger identity holds within sampling error in all $20$ configurations, with the largest deviation $1.9$ standard errors.

\subsection{Sequential design on test items}
\label{app:eval-design}

\paragraph{Instance.}
An adaptive test places a grade-8 student in an achievement level of the 2019 NAEP science assessment \cite{statistics2019irt}.
The items are the experiments of Appendix~\ref{app:cl-chernoff}.
Abilities on a grid are the states.
The $268$ dichotomous items of the univariate scale enter with their published parameters.
These are treated as the true law.
Their standard errors are not propagated.
For $239$ of them the success probability $\mu_e(\nu)$ of item $e$ at state $\nu$ follows the three-parameter logistic model, and for $29$ the two-parameter model, both with the scaling constant $1.7$.
The $101$ partial-credit items are left out.
The cut scores of the 2009 framework, $141$ for Basic and $170$ for Proficient on the reporting scale, sit at abilities $-0.32$ and $0.46$.
In the two-level design the hypotheses are Proficient or not, with $15$ states at distances $0.25, 0.5, \ldots$ from the cut within $[-2, 2]$.
In the three-level design they are below Basic, Basic, and Proficient or above, with $14$ states.
The Basic level holds the two states $0.25$ and $0.5$ below the Proficient cut.
The indifference half-width $0.25$ is that of Bartroff, Finkelman and Lai \cite{bartroff2008modern}.
An item given again yields a fresh response, independent of the others given the state, as in Chernoff's setting.
Real items that share a stimulus may break that independence.
At every state $T^\star_\Theta(\nu)^{-1}$ and the maximin allocation are solved by linear programming.
At the $28$ states whose allocation uses one item, the value of the program equals that of the best single item.
A search over all $35{,}778$ pairs of items confirms the one mixed state, the Basic state at ability $0.21$ and scale score $161$.
There two items share the allocation with weights $0.35$ and $0.65$, and $T^\star_\Theta(\nu)^{-1} = 0.182$ against $0.155$ for the best single item.

\paragraph{Procedures.}
Chernoff's rule takes the maximum-likelihood state over the grid and draws an item from that state's maximin allocation \cite{chernoff1959sequential,naghshvar2013active}.
Maximum Fisher information at the estimate, the standard adaptive test, gives the most informative item at the maximum-likelihood state.
Maximum Fisher information at the cut gives the most informative item at the cut nearest that state, as in computerized mastery testing \cite{bartroff2008modern}.
Random selection draws an item uniformly from the bank.
A use-once variant of maximum information at the estimate never repeats an item.
A use-once test that exhausts the bank decides the maximum-likelihood level.
Every rule stops by the generalized likelihood ratio over the grid.
The test stops when the best state of the leading level exceeds every state outside that level by a threshold in log-likelihood.
It then decides the leading level.
With the threshold $\log(M/\delta)$, where $M$ is the size of the largest alternative set ($9$ and $12$), the stopping rule is $\delta$-correct for every selection rule, by Corollary~\ref{cor:ville} at $s = 1$ and a union bound over the states outside the true level.
The common threshold $\log(1/\delta)$ is not proved $\delta$-correct.
In the two-level design Wald's SPRT between the two states next to the cut also runs.
The risks are $\delta \in \{10^{-1}, 0.05, 10^{-2}, 10^{-3}, 10^{-4}\}$, with the use-once variant at $\delta \ge 10^{-2}$ only.
Each of the $1{,}679$ configurations runs $2{,}000$ times.

\paragraph{Split.}
Let $E$ be the event that the test decides the level of $\nu$, and let $\lambda^\star$ minimize $\sum_e \bar w_e\, \KL(\nu_e \,\|\, \lambda_e)$ over $\Alt_\Theta(\nu)$ at the average allocation $\bar w$.
By the chain of Proposition~\ref{prop:cl-design},
\[
\mathbb{E}_\nu[\tau]\, T^\star_\Theta(\nu)^{-1} = \kl(\delta, 1-\delta) + \textsc{uniform} + \textsc{process} + \textsc{allocation}
\]
Here $\textsc{allocation} = \mathbb{E}_\nu[\tau]\,T^\star_\Theta(\nu)^{-1} - \sum_e \mathbb{E}_\nu[N_e(\tau)]\,\KL(\nu_e \,\|\, \lambda^\star_e)$, $\textsc{process}$ is that sum minus $\kl(P_\nu(E), P_{\lambda^\star}(E))$, and $\textsc{uniform}$ is the rest.
The alternative $\lambda^\star$ is chosen on half of the runs and evaluated on the other half, and the halves then swap, as in Appendix~\ref{app:eval-bai}.
At every state next to a cut both choices coincide with the minimizer over all runs.
The probability $P_{\lambda^\star}(E) = \mathbb{E}_\nu[\mathbf{1}_E\, e^{-L_\tau}]$ is computed from the log-likelihood of each run at every grid state.
The \textsc{allocation} term and the joint term $\textsc{uniform} + \textsc{process}$ need no estimate of $P_{\lambda^\star}(E)$.
The bound at the rule's own error rates, $T^\star_\Theta(\nu)\,\kl(P_\nu(E), P_{\lambda^\star}(E))$, is evaluated at the realized error probabilities, so it belongs to the rule.

\paragraph{Instrument.}
At the states next to a cut, for Chernoff's rule and maximum information at the estimate under both thresholds at $\delta \in \{0.1, 0.05, 0.01\}$, the probability $P_{\lambda^\star}(E)$ was also counted directly from $100{,}000$ runs under $\lambda^\star$.
These are $72$ configurations.
Agreement within $0.2$ nats or two combined standard errors was required at every configuration, with no drift over nested subsamples of $250$, $1{,}000$ and $2{,}000$ runs.
A drift is a monotone sequence whose total movement exceeds two standard errors of the nested difference.
No configuration drifts.
In the two-level design all $24$ configurations agree.
In the three-level design $1$ of $48$ does not.
The exception is Chernoff's rule with the $\delta$-correct threshold at $\delta = 0.01$ and the state $0.71$.
There the change of measure gives $\log P_{\lambda^\star}(E) = -6.55$ against $-6.79$ from $112$ direct events.
The difference is $0.25$ nats, or $2.2$ standard errors.
The requirement allows no exception among the $72$ configurations.
The check was therefore repeated with fresh seeds and $10$ times as many runs.
In the repeat the change of measure is computed from $20{,}000$ runs under $\nu$.
The direct count is taken from $10^6$ runs under $\lambda^\star$.
The alternative $\lambda^\star$ chosen on the new runs is the original one at all $72$ configurations.
The reading of the repeat was fixed before any of its results were read.
It keeps the requirement and allows at most $4$ of the $48$ three-level configurations to drift.
All $72$ configurations agree.
No difference exceeds $0.08$ nats.
At the exception the difference is $0.04$ nats, or $1.3$ standard errors, from $1{,}276$ direct events.
One three-level configuration drifts, within the allowance.
The three-level design therefore reports the \textsc{uniform} and \textsc{process} terms and the bound at the rules' own error rates, as the two-level design does.
Beyond these $72$ configurations, each alternative $\lambda^\star$ is itself a state of the grid, run with the same rule, threshold and risk.
The main runs therefore also count $P_{\lambda^\star}(E)$ directly, for every rule.
At the $15$ configurations where the effective sample size of the weights is below $150$ of $2{,}000$, $13$ of them for the use-once variant, the differences have standard deviation $1.42$ and no consistent sign.
There the change of measure varies more than its standard error states.
Against the repeat, the main runs' estimates at the $72$ configurations also vary more than their standard errors state.
Their standardized differences from the repeat have standard deviation $1.17$.
At these configurations the bound at Chernoff's own error rates was also computed from the repeat alone, with the direct count in place of the change of measure.
For the use-once variant at $\delta = 0.01$ the \textsc{process} term is therefore also reported from the direct count.

\paragraph{Exact control.}
In the two-level design maximum information at the cut gives the same item at every step, the most informative item at the one cut.
That test is a walk on the numbers of responses and successes.
Its law is computed exactly.
Over the $225$ configurations of the rule, the standardized differences between the simulated and the exact expected lengths have mean $0.01$ and standard deviation $1.09$.
The largest, $-4.2$ at the far state $-0.79$, appears again in Wald's identity on the same runs.
At the two states next to the cut the change-of-measure estimates of $\log P_{\lambda^\star}(E)$ lie within $0.08$ nats of the exact values.
At $1{,}000$ runs that estimate has a mean error below $0.005$ nats at every risk, down to $\delta = 10^{-4}$, where no direct count is possible.
The exact law also gives the rule's largest error probability over the states under each stopping rule.
It is $0.20\,\delta$ for the $\delta$-correct threshold, $0.93\,\delta$ for Wald's SPRT and $1.88\,\delta$ for the common threshold.

\begin{figure}[t]
\centering
\includegraphics[width=\linewidth]{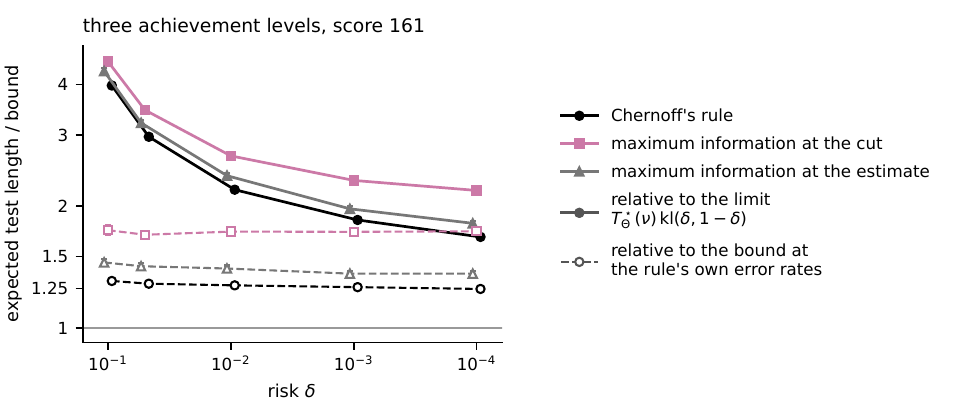}
\caption{\textbf{At the Basic state where the maximin allocation mixes two items, Chernoff's rule sits $25\%$ to $31\%$ above the bound at its own error rates, while maximum information at the cut sits $70\%$ to $74\%$ above its bound.} Three achievement levels, the Basic state just below the Proficient cut (scale score $161$). The expected test length is drawn relative to the limit $T^\star_\Theta(\nu)\,\kl(\delta, 1-\delta)$ (solid) and relative to the bound at the rule's own error rates (dashed), for three item-selection rules at the $\delta$-correct threshold. The three rules need $1.7$ to $4.6$ times the limit. Chernoff's rule and maximum information at the cut keep their colors of Figure~\ref{fig:ab}(c). Error bars are $95\%$ bootstrap intervals over runs; where none shows, the interval is narrower than the marker.}
\label{fig:design-three-levels}
\end{figure}

\paragraph{Readings.}
At $\delta = 10^{-4}$ Chernoff's rule with the $\delta$-correct threshold needs $1.29$ to $1.68$ times the limit $T^\star_\Theta(\nu)\,\kl(\delta, 1-\delta)$ at the six states next to a cut.
At $\delta = 0.1$ it needs $2.59$ to $3.98$ times the limit.
The ratio falls at every state as $\delta$ shrinks (Figures~\ref{fig:ab}(c) and~\ref{fig:design-three-levels}).
No configuration of the $\delta$-correct threshold falls below the limit; over all $667$ of them the smallest ratio is $1.23$.
Six configurations of the common threshold fall below the limit, all for maximum information at the cut in the two-level design.
The lowest of them is $0.93$ [$0.91$, $0.96$].
The exact law places five of the six below the limit and the sixth at $1.008$.
That threshold is not $\delta$-correct on this bank, so Proposition~\ref{prop:cl-design} does not bind it.

In the two-level design \textsc{uniform} is the largest term at every risk, $64\%$ to $81\%$ of Chernoff's excess over the limit.
The $\delta$-correct threshold holds the error probability at $\lambda^\star$ to $0.11\,\delta$ to $0.19\,\delta$.
The bound at the rule's own error rates is tight there.
Chernoff's rule sits $8\%$ to $15\%$ above that bound at every risk.
The upper ends of its $95\%$ intervals are at most $1.18$.
Maximum information at the cut sits $1.5\%$ to $6.2\%$ above its bound.
By the exact law its excess is $2.3\%$ to $5.5\%$.
In the three-level design \textsc{uniform} is also the largest term at every state and risk, $50\%$ to $80\%$ of Chernoff's excess.
There the $\delta$-correct threshold holds the error probability at $\lambda^\star$ to $0.04\,\delta$ to $0.19\,\delta$.
Chernoff's rule sits $9\%$ to $31\%$ above the bound at its own error rates.
That bound is tight at $15$ of the $20$ configurations.
At the Basic state $0.21$ the ratio is $1.25$ to $1.31$ at the five risks.
The upper ends of its $95\%$ intervals there are at most $1.33$.
Maximum information at the cut sits $8\%$ to $15\%$ above its bound at the outer states $-0.57$ and $0.71$, where that bound is tight.
At the two Basic states its excess is $33\%$ to $74\%$.
The ratio to the bound is the allocation factor $\mathbb{E}_\nu[\tau]\,T^\star_\Theta(\nu)^{-1}/\sum_e \mathbb{E}_\nu[N_e(\tau)]\,\KL(\nu_e \,\|\, \lambda^\star_e)$ times the ratio to the bound over the rule's own allocation.
The second factor is at least $1$ by data processing.
At the Basic states the cut rule's allocation factor, which needs no change of measure, is $1.29$ to $1.60$.
The rule's draws follow the cut nearest the maximum-likelihood state.
The item of the Proficient cut separates a Basic state little from the binding alternative, the below-Basic state $-0.57$.
At $0.21$ its divergence from that state is $0.08$ nats per response, against $T^\star_\Theta(\nu)^{-1} = 0.18$.
There the item takes $60\%$ to $71\%$ of the draws.
At $-0.04$ its divergence is $0.02$ nats, against $0.12$.
There the item takes $20\%$ to $24\%$ of the draws, those made while the maximum-likelihood state is $0.21$ or above.
At $0.21$ maximum information at the estimate sits $36\%$ to $45\%$ above its bound.
Under the common threshold the ratios grow at $\delta = 0.1$, to $1.51$ for Chernoff's rule and $2.05$ for the cut rule.
The ratio to the limit is the product of three factors: the threshold over $\kl(\delta, 1-\delta)$, the expected evidence against $\lambda^\star$ at stopping over the threshold, and the allocation factor.
At $\delta = 0.1$ the first is $2.72$ in the three-level design.
At the Basic state $0.21$ the other two are $1.24$ and $1.18$.
The ratio there reaches $3.98$ [$3.91$, $4.04$].
That is above the range specified before the run, $1.2$ to $3.5$.

The \textsc{allocation} term measures how far a rule's average allocation falls short of the maximin allocation of the true state.
Chernoff's rule and maximum information at the estimate fall short while the maximum-likelihood state is wrong.
The cut rule follows only the cut nearest the maximum-likelihood state, so in the two-level design it gives the same item throughout.
That item is the maximin item at both states next to the cut.
The cut rule's \textsc{allocation} term is therefore zero there, while Chernoff's term is $0.21$ to $0.52$ nats over the three stopping rules, from its early wrong states.
At the four lowest two-level states the same item reaches only $0.28$ to $0.45$ of $T^\star_\Theta(\nu)^{-1}$.
There the cut rule's \textsc{allocation} term is $0.55$ to $0.72$ of $\mathbb{E}_\nu[\tau]\,T^\star_\Theta(\nu)^{-1}$, its design-time share to within $10^{-6}$.
It is larger there than the term of either other information rule.
The order specified before the run put Chernoff's term at or below the cut rule's, the cut rule's at or below the estimate rule's, and the estimate rule's below random selection's.
It holds in $11$ of the $70$ configurations at the states next to a cut: all $10$ at the below-Basic state $-0.57$ and $1$ of $10$ at the Basic state $0.21$.
At a second configuration at $0.21$ the order fails by a margin not resolved by its interval.
The cut rule has the smallest term in every configuration at the two two-level states and at the three-level state $0.71$.
The estimate rule has it in $9$ of $10$ at the Basic state $-0.04$, where the maximin allocation is that rule's own single item.
Chernoff's rule has it at the states $-0.57$ and $0.21$.
At $-0.04$ a search over item pairs confirms that single item.
There the estimate rule's term at $\delta = 10^{-4}$ is $0.99$ [$0.96$, $1.02$] nats.
Chernoff's term is $1.24$ [$1.19$, $1.29$] nats.
At the Basic state $0.21$ the cut rule's item reaches $0.44$ of $T^\star_\Theta(\nu)^{-1}$.
Its test runs $5.6$ to $25.7$ items longer than Chernoff's under the $\delta$-correct threshold.
Chernoff's rule spends $66\%$ to $75\%$ of its draws there on the two items of the mixed allocation.
Random selection reaches $0.16$ to $0.27$ of $T^\star_\Theta(\nu)^{-1}$ over the grid.
Its allocation factor is $4.1$ to $5.7$ at the states next to a cut.

Chernoff's \textsc{allocation} term stays below $1$ nat at $\delta \le 10^{-3}$ at four of the six states next to a cut.
At the two Basic states it reaches $1.11$ to $1.75$ nats.
There its share of $\mathbb{E}_\nu[\tau]\,T^\star_\Theta(\nu)^{-1}$ falls as $\delta$ shrinks, while $\mathbb{E}_\nu[\tau]$ grows.
At $-0.04$ the share falls from $0.14$ to $0.095$.
At $0.21$ it falls from $0.15$ to $0.11$.
First-order asymptotic optimality drives that share to zero and leaves the term in nats free to grow.

Near the cuts only $6$ to $27$ items reach half the best item's information.
The whole bank holds $7.7$ to $10.9$ nats against the binding alternative.
At $\delta = 0.01$ the $\delta$-correct threshold asks for $6.8$ to $7.1$ nats, so $238$ to $681$ of the $2{,}000$ use-once tests exhaust the bank.
They decide correctly $90\%$ to $97\%$ of the time.
At the three-level Basic state $-0.04$ these forced decisions raise the error rate to $0.032$, against $\delta = 0.01$.
At the below-Basic state $-0.57$ they raise it to $0.035$.
The tests that stop at the threshold err at rates $0.0005$ and $0$ at these two states.
The use-once variant's \textsc{allocation} term exceeds that of maximum information at the estimate by $1.4$ to $16.4$ nats at every state next to a cut, stopping rule and risk.
At the two-level states next to the cut the \textsc{process} term stays below $1.4$ nats for every rule but the use-once variant, under all three stopping rules.
By the change of measure the use-once variant exceeds $2$ nats there only at the state $0.71$ and $\delta = 0.01$.
Its \textsc{process} term is $2.09$ [$1.80$, $2.34$].
There the $314$ tests that exhaust the bank stop with a mean $3.5$ nats of evidence against $\lambda^\star$, below the threshold of $6.8$.
They supply $84\%$ of $P_{\lambda^\star}(E)$.
With the direct count in place of the change of measure the term is $1.94$ [$1.51$, $2.32$] there.
At the state $0.21$ the change of measure gives $1.47$ [$1.16$, $1.74$].
The direct count gives $2.04$ [$1.64$, $2.41$] there.
At the three-level states next to a cut the \textsc{process} term is at most $1.32$ nats for every rule but the use-once variant.
With the $\delta$-correct threshold at $\delta = 0.01$ that variant's term is $2.2$ to $3.3$ nats at the four states next to a cut.
Its interval lies above $2$ nats at three of them.
At the four states the tests that exhaust the bank stop with a mean $3.5$ to $4.6$ nats of evidence against $\lambda^\star$, below the threshold of $7.1$.
Their forced decisions supply $88\%$ to $98\%$ of $P_{\lambda^\star}(E)$.
A direct count of $P_{\lambda^\star}(E)$ in the use-once runs at $\lambda^\star$ also puts the term above $2$ nats at those three states.

Where the limit is below three items, at states far from the cuts and large risks, the excess over it is item granularity.
The limit asks for $0.80$ to $2.99$ items there.
At $T^\star_\Theta(\nu)^{-1}$ per item the $\delta$-correct threshold needs $1.5$ to $2.7$ times as many.
Chernoff's shortest test there lasts $2$ to $4$ items.
Chernoff's rule needs $2.2$ to $6.0$ times the limit at these configurations.

The ledger identity $\mathbb{E}_\nu[L_\tau] = \sum_e \mathbb{E}_\nu[N_e(\tau)]\,\KL(\nu_e \,\|\, \lambda_e)$ is checked at every alternative in each half and in the full sample.
Of the $12{,}972$ full-sample tests $1.3\%$ exceed $2.5$ standard errors, against $1.2\%$ for a normal law.
No configuration of the $\delta$-correct threshold has an unforced error rate whose $95\%$ Clopper--Pearson interval lies above $\delta$.
The common threshold has an error rate whose interval lies above $\delta$ at $46$ configurations.
At none of them does the interval lie above that threshold's union bound $|\Alt_\Theta(\nu)|\,\delta$.
Wald's SPRT has an error rate whose interval lies above $\delta$ at two configurations, both of the use-once variant at $\delta = 0.01$.
Of their $41$ and $30$ errors, $19$ and $17$ are made by tests that exhaust the bank.
The unforced error rates there, $0.011$ and $0.0065$, have intervals that contain $\delta$.
In the risk form of Proposition~\ref{prop:cl-chernoff} with unit loss at $\kappa = \delta$, the risk of Chernoff's rule at $\kappa = 10^{-4}$ is $1.29$ to $1.68$ times $\kappa\log(1/\kappa)\,T^\star_\Theta(\nu)$.
That bound is asymptotic and is read here at a finite cost.

Off the grid the abilities lie outside Chernoff's finite-state model.
No bound is claimed there.
The ability law of grade-8 students is the normal law that matches the 2019 percentages below Basic and at or above Proficient.
Averaged over it, Chernoff's rule needs $18.3$ items in the two-level design and $45.2$ in the three-level design.
It places $97.5\%$ and $95.1\%$ of students correctly.
In the two-level design the cut rule needs $0.70$ [$0.55$, $0.85$] items fewer.
It places $0.15$ [$0.03$, $0.27$] percentage points fewer correctly.
In the three-level design it needs $0.90$ [$0.56$, $1.25$] items fewer.
There it places $0.02$ [$-0.15$, $0.18$] percentage points fewer correctly.
That difference is not resolved.

\subsection{Regret on a maintenance Markov decision process}
\label{app:eval-mdp-regret}

\paragraph{Instance.}
The Markov decision process is the machine-maintenance example of Hillier and Lieberman \cite{hillier2001introduction}.
Its four states are good as new, minor deterioration, major deterioration and inoperable.
The operator may do nothing in every state but inoperable, overhaul in major deterioration, and replace the machine in every state but good as new, so there are seven state-action pairs.
Costs are known to the learner.
The learner estimates the transition laws, including their supports.
Exact rational arithmetic reproduces the four published average costs of the deterministic policies considered there ($25/13$, $5/3$, $19/11$ and $3$) and the stationary law $(2/21,\allowbreak 5/7,\allowbreak 2/21,\allowbreak 2/21)$ of the optimal policy.
The optimal gain is $g^\star = -5/3$ and the bias is $h^\star = (0,\allowbreak -4/3,\allowbreak -11/3,\allowbreak -13/3)$.
Three state-action pairs are suboptimal.
Replacing in minor deterioration ($\Delta_{x,a} = 3$) and in major deterioration ($\Delta_{x,a} = 2/3$) already leads to the state of largest bias, so no transition law makes them optimal and $\Kinf(x,a) = +\infty$.
Doing nothing in major deterioration has $\Delta_{x,a} = 5/3$ and $\Kinf(x,a) = \tfrac12 \log(143/49) = 0.5355$, the infimum of the relative entropy from its transition law to one that makes the pair optimal; three independent computations confirm it.
Hence $c^\star_\Phi(\nu) = \Delta_{x,a} / \Kinf(x,a) = 3.112$ for the class $\Phi$ of all transition laws (Proposition~\ref{prop:cl-mdp-decoupled}).

\paragraph{Algorithms.}
IMED-RL \cite{pesquerel2022imedrl} uses Dirichlet pseudo-counts $1/|\mathcal{X}|$ and exact policy iteration.
PSRL \cite{osband2013more} uses uniform Dirichlet priors and the doubling episodes of UCRL2.
UCRL2 \cite{jaksch2010} uses confidence parameter $0.05$.
Each algorithm runs $1{,}000$ times for $T = 10^5$ steps from the state good as new.
Counts are read at $16$ log-spaced checkpoints from $t = 100$ to $10^5$.

\paragraph{Alternatives.}
Write $(x,a)$ for doing nothing in major deterioration, so that $\Kinf(x,a) = 0.5355$.
For each margin $\xi_\Delta \in \{0.02,\allowbreak 0.05,\allowbreak 0.1\}$ the alternative $\lambda^{(\xi_\Delta)}$ agrees with $\nu$ except at $(x,a)$.
There its transition law minimizes $\KL(\nu_{x,a} \,\|\, \lambda^{(\xi_\Delta)}_{x,a})$ subject to
\[
r(x,a) + \lambda^{(\xi_\Delta)}(\cdot \mid x,a) \cdot h^\star = g^\star + h^\star(x) + \xi_\Delta\, \Delta_{x,a}
\]
with $g^\star$ and $h^\star$ of $\nu$ and $r(x,a) = -3$ the reward of the pair.
The minimizer moves mass from major deterioration and inoperable to good as new.
Then $\KL(\nu_{x,a} \,\|\, \lambda^{(\xi_\Delta)}_{x,a}) - \Kinf(x,a) = \log\frac{7}{7 - 5\xi_\Delta}$, which is $0.0144$, $0.0364$ and $0.0741$.
Every deterministic policy is irreducible under $\lambda^{(\xi_\Delta)}$, and $\nu \ll \lambda^{(\xi_\Delta)}$ at every pair.
The unique optimal policy of $\lambda^{(\xi_\Delta)}$ does nothing in every state but inoperable.
Its gain exceeds $g^\star$ by $0.0041$, $0.0102$ and $0.0201$.
Overhauling in major deterioration then has a Bellman gap of $0.04$ to $0.21$.
Its $\Kinf$, computed under $\lambda^{(\xi_\Delta)}$, is $0.03$ to $0.17$.
Each algorithm runs $1{,}000$ further times under $\nu$ and under each $\lambda^{(\xi_\Delta)}$, with the parameters and checkpoints above.

\paragraph{Read-outs.}
The regret in count form is $\sum_{x,a} \mathbb{E}_\nu[N_{x,a}(t)]\,\Delta_{x,a}$.
It equals the pathwise regret plus the bias of the initial state minus that of the state after round $t$, a term between $0$ and $13/3$, plus a martingale.
The martingale has mean zero within sampling error at every checkpoint.
The information transported through the pair with finite $\Kinf(x,a)$ is $I_{x,a}(t) = \mathbb{E}_\nu[N_{x,a}(t)]\,\Kinf(x,a)$.
The identity of Theorem~\ref{thm:allocation} with state-action pairs as arms, the relative-entropy endpoint of Theorem~\ref{thm:mdp}, was checked at $T = 2{,}000$ with $400$ runs per algorithm against two alternatives.
The first is the most confusing single-pair alternative.
The second differs from $\nu$ at every pair.
The identity holds within sampling error for IMED-RL and PSRL against both alternatives.
UCRL2's first draw of $400$ runs deviated by $2.4$ and $2.8$ standard errors against the two alternatives.
Two fresh batches of $3{,}200$ and $9{,}600$ runs give pooled deviations within $1.7$ standard errors, so the deviation does not persist.
A negative control acts on the next transition before it is drawn in major deterioration, so its action law given $\mathcal{F}_{t-1}$ depends on the instance, and it fails the identity against the first alternative by $209$ standard errors.
Against the second the deviation is $0.13$ standard errors, since there every transition out of major deterioration has a log-likelihood ratio equal to its pair's relative entropy, whatever the next state, so acting on the next transition changes nothing.

The count-form regret ratio splits exactly.
The statistic is $Z_t := N_{x,a}(t)/N_x(t)$, the fraction of visits to major deterioration at which nothing is done, with $Z_t := 0$ when $N_x(t) = 0$.
The normalized count $N_{x,a}(t)/t$ would lose the $\log t$ scale, since its mean under $\lambda^{(\xi_\Delta)}$ tends to the stationary probability of major deterioration, about $0.12$.
Write $n_{x,a} := \mathbb{E}_\nu[N_{x,a}(t)]$, $z_\nu := \mathbb{E}_\nu[Z_t]$, $z_\lambda := \mathbb{E}_{\lambda^{(\xi_\Delta)}}[Z_t]$ and $\underline{\kl} := (1-z_\nu)\log\frac{1}{1-z_\lambda} - \log 2$.
Then
\begin{align*}
n_{x,a}\,\Kinf(x,a) - \log t
 &= \big(n_{x,a} \KL(\nu_{x,a} \,\|\, \lambda^{(\xi_\Delta)}_{x,a}) - \kl(z_\nu, z_\lambda)\big) + \big(\kl(z_\nu, z_\lambda) - \underline{\kl}\big) \\
 &\quad + \big(\underline{\kl} - \log t\big) - n_{x,a}\big(\KL(\nu_{x,a} \,\|\, \lambda^{(\xi_\Delta)}_{x,a}) - \Kinf(x,a)\big)
\end{align*}
The four terms are the \textsc{process}, \textsc{binary}, joint \textsc{uniform}-and-\textsc{asymptotic} and \textsc{boundary} terms.
The two replacing pairs, whose $\Kinf$ is infinite, add the \textsc{allocation} term: their expected visit counts weighted by their Bellman gaps.
The count-form regret ratio minus one equals the left side divided by $\log t$ plus the \textsc{allocation} term divided by $c^\star_\Phi(\nu)\log t$.
The split is exact wherever $z_\lambda < 1$.
Standard errors come from the delta method over runs.
At $t = 10^5$, estimates from $10$ blocks of $100$ runs agree with them within a factor of $1.6$.
Over all checkpoints they agree within a factor of $3.1$.

\paragraph{Readings.}
At $T = 10^5$ the regret is $29.0$ for IMED-RL, $14.9$ for PSRL and $2{,}981$ for UCRL2, against $c^\star_\Phi(\nu) \log T = 35.8$.
IMED-RL's transported fraction rises over the whole range.
Both pairs with infinite $\Kinf$ receive about one visit under IMED-RL and fewer under PSRL.

With the closest alternative at $T = 10^5$, UCRL2's regret ratio of $84.3$ comes almost all from the \textsc{process} term, which adds $86.4$ times the limit.
Under $\nu$ it transports about $1{,}000$ nats through doing nothing in major deterioration, of which the statistic retains $1.9$.
At the farther alternatives IMED-RL's mean overhaul fraction falls to $7\%$ and $2\%$, its joint term rises to $-0.83$ and $-0.70$, and its \textsc{process} term falls to $0.52$ and $0.44$.
PSRL's overhaul fraction falls to $22\%$ and $17\%$, and its terms move less.
IMED-RL's joint term rises with $t$ at every alternative.
At the checkpoints before $t = 25{,}119$, and before $t = 39{,}811$ at $\xi_\Delta = 0.1$, every run of UCRL2 under the alternative does nothing at every visit to major deterioration, so $z_\lambda = 1$ and its split is undefined there.
The further runs under $\nu$ reproduce the ratios of the first set within $1.9$ standard errors at every checkpoint.
The identity $\mathbb{E}_\nu[L_t] = n_{x,a}\,\KL(\nu_{x,a} \,\|\, \lambda^{(\xi_\Delta)}_{x,a})$ is checked as well, in one check that covers the three alternatives, because the log-likelihood ratio of a transition out of $(x,a)$ minus $\KL(\nu_{x,a} \,\|\, \lambda^{(\xi_\Delta)}_{x,a})$ does not depend on $\xi_\Delta$.
For PSRL and UCRL2 the deviation is within $1.7$ standard errors at every checkpoint, and for IMED-RL within $0.9$ at $t = 10^5$.
From $t = 100$ to $10{,}000$ IMED-RL drew $1.6$ to $3.0$ standard errors, all negative and correlated across checkpoints because $L_t$ accumulates.
With $3{,}200$ fresh runs its deviation at $t = 100$ to $398$ is at most $0.8$ standard errors.

\subsection{Regret on an engine-replacement model fitted to fleet data}
\label{app:eval-bus-engine}

\paragraph{Instance.}
The Markov decision process is Rust's model of engine replacement for a fleet of city buses \cite{rust1987optimal}.
Its mileage law and costs are Rust's estimates for the $4{,}292$ bus-months of his fourth group of buses, as restated in a replication by Ferrall \cite{ferrall2023was}.
In Rust's coding the mileage since the last replacement falls in bins of $5{,}000$ miles.
In a month it rises by $0$, $1$ or $2$ of these bins with probabilities $0.3919$, $0.5953$ and $0.0128$.
Maintenance costs $0.002293\,x$ a month in bin $x$, and a replacement costs $10.08$, in the units of Rust's estimated utility.
Here the mileage is grouped in $18$ bins of $25{,}000$ miles, with the position inside a bin taken as uniform.
In a month, keeping the engine moves it up one bin with probability $0.1242$, the mean monthly increment of $0.6209$ fine bins divided by $5$.
Otherwise it stays in its bin.
Replacing the engine leads to bin $0$.
Keeping in bin $x$ costs the maintenance cost at the bin's mean fine index $5x + 2$.
Costs are divided by the replacement cost, so rewards lie in $[-1, 0]$.
They are known to the learner.
The learner estimates the transition laws, including their supports.
Rust's extreme-value shocks are dropped.
The criterion is the average reward.

Replacement is forced in bin $17$, at $425{,}000$ miles and above, as in the inoperable state of the machine of Hillier and Lieberman \cite{hillier2001introduction}.
With this change alone the model is not ergodic, since a policy that replaces below bin $17$ never reaches the bins above its threshold.
Replacement is therefore unavailable in bin $0$, where keeping moves the engine to a uniformly drawn bin with probability $0.01$ a month, as their good-as-new machine reaches every condition in one period.
Under these changes each of the $65{,}536$ deterministic policies makes the bins one irreducible chain.
None of them alters the law of a suboptimal pair.

The unique optimal policy keeps the engine in bins $0$ to $14$ and replaces it from bin $15$, at $375{,}000$ miles.
Rational arithmetic, policy iteration, relative value iteration and the enumeration of all deterministic policies agree on it.
Its gain is $g^\star = -0.01700$.
The second-best policy is $4.8 \times 10^{-7}$ a month worse.
The optimal policy is not a model of the replacement decisions in Rust's data; the instance takes his estimated costs and mileage law as given.
Replacing in bins $1$ to $14$ already leads to bin $0$, the bin of largest bias, so no transition law makes these pairs optimal and their $\Kinf(x,a)$ is infinite.
Among the suboptimal pairs, only keeping in bin $15$ and keeping in bin $16$ have finite $\Kinf(x,a)$.
Keeping in bin $15$ has $\Delta_{x,a} = 5.270 \times 10^{-4}$ and $\Kinf(x,a) = 5.363 \times 10^{-4}$.
Keeping in bin $16$ has $\Delta_{x,a} = 1.665 \times 10^{-3}$ and $\Kinf(x,a) = 1.695 \times 10^{-3}$.
Both next states of keeping in these bins have the same bias, $h^\star(15) = h^\star(16) = h^\star(17) = -0.9830$, so $\Kinf(x,a) = -\log(1 - \Delta_{x,a}/|h^\star(15)|)$.
The dual, two closed forms and a direct minimization give the same value.
On this ergodic instance the program of Proposition~\ref{prop:cl-mdp} reduces to one constraint per pair \cite[Theorem~2]{ok2018exploration}.
For the class $\Phi$ of all transition laws, $c^\star_\Phi(\nu) = \sum_{x,a} \Delta_{x,a}/\Kinf(x,a) = 1.965$.

Each of the two pairs contributes close to $0.983$ to the constant.
The constant thus counts the bins between the optimal threshold and the forced bin.
It changes with the grouping.
On Rust's own bins, replacing just below the threshold also has finite $\Kinf(x,a)$, since a replacement there lands in one of three fine bins of different bias.
The grouping removes those pairs.
The gap at the threshold, $5.27 \times 10^{-4}$, is $46\%$ of the increase in the cost of keeping from one bin to the next.

\paragraph{Horizon.}
A policy that never learns has linear regret.
The policy that replaces from bin $16$ has regret $3.4$ at $t = 10^5$ and $33.7$ at $t = 10^6$, against $c^\star_\Phi(\nu)\log t = 22.6$ and $27.1$.
It crosses the limit at $t = 7.9 \times 10^5$.
Before that round a regret ratio below $1$ does not show learning, so the readings are taken at $T = 10^6$.
There every threshold policy except the optimal one lies above the limit.
The limit asks for $\log T / \Kinf(x,a) = 25{,}760$ samples of keeping in bin $15$ by $T = 10^6$, three times the $8{,}490$ visits made to that bin by the optimal policy.
The readings describe this fitted, grouped and ergodic model at $T = 10^6$ and are not extrapolated.

\paragraph{Algorithms and alternatives.}
IMED-RL, PSRL and UCRL2 use the parameters of Appendix~\ref{app:eval-mdp-regret}.
PSRL plans by policy iteration, which returns the policy of relative value iteration on $1{,}000$ sampled bus kernels.
Each algorithm runs $500$ times for $T = 10^6$ steps from bin $0$.
Counts are read at $21$ log-spaced checkpoints from $t = 100$.
For each pair $(x,a)$ with finite $\Kinf(x,a)$ and each margin $\xi_\Delta \in \{0.02, 1\}$, the alternative $\lambda^{(x,a,\xi_\Delta)}$ is built as in Appendix~\ref{app:eval-mdp-regret}.
It agrees with $\nu$ except at $(x,a)$, where its law minimizes the relative entropy from $\nu_{x,a}$ subject to a margin of $\xi_\Delta\,\Delta_{x,a}$.
The minimizer moves mass to bin $0$: $5.5 \times 10^{-4}$ at bin $15$ and $1.7 \times 10^{-3}$ at bin $16$ when $\xi_\Delta = 0.02$.
Every deterministic policy is irreducible under each alternative.
The unique optimal policy of each alternative keeps the engine at $(x,a)$.
Under the alternatives with $\xi_\Delta = 0.02$, IMED-RL and PSRL run $500$ times and UCRL2 $200$ times.
Under those with $\xi_\Delta = 1$, IMED-RL and PSRL run $300$ times.
IMED-RL and PSRL also run $300$ times at the wear-shock rates $0.005$ and $0.02$.
These rates move the gap at the threshold to $7.42 \times 10^{-4}$ and $1.11 \times 10^{-4}$, while $c^\star_\Phi(\nu)$ stays between $1.964$ and $1.966$.

\paragraph{Read-outs.}
The regret in count form is $\sum_{x,a} \mathbb{E}_\nu[N_{x,a}(t)]\,\Delta_{x,a}$.
It equals the pathwise regret plus the bias of the initial state minus that of the state after round $t$, a term between $0$ and $0.983$, plus a martingale.
For six of the seven combinations of algorithm and wear-shock rate under $\nu$, the martingale has mean zero within $2.8$ standard errors at every checkpoint.
For PSRL at the rate $0.01$ it reaches $3.6$ standard errors at $t = 2.5 \times 10^5$.
A Brownian path read at the same $21$ checkpoints reaches this level with probability $0.005$, and one of seven such paths with probability $0.04$.
The information transported through each pair with finite $\Kinf(x,a)$ is $I_{x,a}(t) = \mathbb{E}_\nu[N_{x,a}(t)]\,\Kinf(x,a)$.
Its transported fraction is $I_{x,a}(t)/\log t$.
The divergence decomposition, with state-action pairs as arms, was checked at $T = 2{,}000$ with $400$ runs per algorithm against two alternatives at which the log-likelihood ratio of a transition varies with the next state.
The alternatives $\lambda^{(x,a,\xi_\Delta)}$ give no such check, since the two next states of keeping in bins $15$ and $16$ have the same bias and each transition there has a log-likelihood ratio equal to its pair's relative entropy.
The deviations lie within $1.41$ standard errors for the three algorithms and both alternatives, while a control that keeps in bin $14$ exactly when the next state would be bin $14$ fails the identity by $79$ standard errors.

The count-form regret ratio splits along the chain of Proposition~\ref{prop:cl-mdp}, one pair at a time.
Write $n_{x,a} := \mathbb{E}_\nu[N_{x,a}(t)]$.
Then
\[
\frac{\sum_{x,a} n_{x,a}\Delta_{x,a}}{c^\star_\Phi(\nu)\log t} - 1
 = \sum_{(x,a)} \frac{\Delta_{x,a}}{\Kinf(x,a)\,c^\star_\Phi(\nu)} \cdot \frac{n_{x,a}\Kinf(x,a) - \log t}{\log t}
 + \frac{\textsc{allocation}}{c^\star_\Phi(\nu)\log t}
\]
where the sum on the right runs over the two pairs with finite $\Kinf(x,a)$.
Their weights $\Delta_{x,a}/(\Kinf(x,a)\,c^\star_\Phi(\nu))$, which sum to one, are $0.5001$ and $0.4999$.
The \textsc{allocation} term is $\sum_{x,a} n_{x,a}\Delta_{x,a}$ over the pairs with infinite $\Kinf(x,a)$.
Each numerator $n_{x,a}\Kinf(x,a) - \log t$ splits into the \textsc{process}, \textsc{binary}, joint \textsc{uniform}-and-\textsc{asymptotic} and \textsc{boundary} terms of Appendix~\ref{app:eval-mdp-regret}.
The split of a pair uses its own alternative and the statistic $Z_t := N_{x,a}(t)/N_x(t)$ of its own bin, with $Z_t := 0$ when $N_x(t) = 0$.
Under $\nu$, $59\%$ of IMED-RL's runs have not visited bin $16$ by $t = 1{,}000$, but every run has visited both bins by $t = 1{,}585$.
The split holds wherever $z_\lambda < 1$ at both pairs.
Standard errors come from the delta method over runs.
At $T = 10^6$ block estimates from $100$ runs agree with them within a factor of $1.8$ at $\xi_\Delta = 0.02$ and $2.2$ at $\xi_\Delta = 1$, where the largest discrepancy is the \textsc{boundary} term.
From $t = 10^4$ on, the checkpoints at which the two differ by more than a factor of two are as frequent as chance gives for the five and three blocks available.

\paragraph{Readings.}
At $T = 10^6$ IMED-RL's regret is $7.62$, with a $95\%$ interval from $7.51$ to $7.73$.
This is $0.281$ of $c^\star_\Phi(\nu)\log T = 27.15$.
Every run replaces from bin $15$ at $T$.
The regret lies $467$ standard errors below the $33.7$ of the policy that replaces from bin $16$.
With the closest alternatives the \textsc{process} term adds $0.056$ of the limit, the \textsc{binary} term $0.021$ and the \textsc{allocation} term $0.220$.
The \textsc{boundary} term subtracts $0.001$.
The joint \textsc{uniform}-and-\textsc{asymptotic} term subtracts $1.015$, more than the regret ratio falls short of one.
Each of these has a delta-method standard error of at most $0.0021$.
Against those alternatives the two pairs transport $1.28$ and $0.44$ nats, of which the statistic retains $0.14$ and $0.03$.
At both pairs the binary bound $\underline{\kl}$ at IMED-RL's own behavior is negative, against an asymptotic requirement of $13.8$ nats.
Of IMED-RL's regret of $7.62$, $5.97$ is the \textsc{allocation} term.
Of that, $4.66$ is the one replacement IMED-RL tries in each of bins $1$ to $12$, a cost that does not grow with $t$.
Its transported fractions are $0.091$ at bin $15$ and $0.031$ at bin $16$, up from $0.020$ and $0.007$ at $t = 10^5$.
Over that decade its regret grew by $1.35$, which is $0.30$ of the growth of the limit.
The joint term rises with the margin, by $0.035 \pm 0.002$ from $\xi_\Delta = 0.02$ to $\xi_\Delta = 1$.
From $t = 10^5$ to $10^6$ it rises by $0.011 \pm 0.001$ at $\xi_\Delta = 0.02$.
At $\xi_\Delta = 1$ it rises by $0.041 \pm 0.002$ over the same decade.
At $\xi_\Delta = 1$ the \textsc{process} term adds $0.088$ and the \textsc{binary} term $0.013$.
The \textsc{boundary} term subtracts $0.061$ and the joint term $0.980$.
IMED-RL visits bin $16$ a total of $546$ times by $T = 10^6$, 15 times as often as the optimal policy.
Of these visits, about $290$ follow a keep in bin $15$.

PSRL and UCRL2 never replace below bin $17$.
At every checkpoint their regrets lie within $1.3$ and $1.8$ standard errors of the regret of keeping until forced, which is $132.2$ at $T = 10^6$.
Both incur $4.87$ times the limit.
Under both closest alternatives PSRL and UCRL2 also keep at every visit to bins $15$ and $16$, so $z_\lambda = 1$ and their split is undefined from $t = 251$ on.
For PSRL the two pair terms $n_{x,a}\Kinf(x,a) - \log t$ are $18.4$ and $88.5$ nats.
The \textsc{allocation} term is zero, so the regret ratio minus one, $3.87$, is their weighted sum divided by $\log t$.
For UCRL2 the two pair terms are $18.4$ and $88.6$ nats.
UCRL2's regret grows by $1.323 \times 10^{-4}$ a step for $t \ge 10^4$, against $1.322 \times 10^{-4}$ for keeping until forced.
It crosses the limit near $t = 1.8 \times 10^5$.
Its confidence radius lets its optimistic model move more mass to bin $0$ than the shift that makes keeping optimal.
At the counts produced by keeping until forced, its own planner first replaces before bin $17$ after about $10^9$ rounds.

PSRL never tries replacing below bin $17$, so the laws of those pairs stay at the uniform Dirichlet prior over the $18$ bins.
Under the laws of keeping of $\nu$, a sampled replacement improves on keeping only if almost all of its mass falls on bin $0$.
This has probability $7.6 \times 10^{-28}$ for each planning step.
PSRL plans with sampled laws of keeping, whose Dirichlet prior adds a pseudo-count of one to every bin.
In a re-simulation of $4{,}000$ runs to $t = 12{,}000$ the probability that a planning step replaces is zero before $t = 1{,}000$.
After that its largest value is $5.1 \times 10^{-7}$.
These probabilities have a heavy upper tail.
A second stream of $500$ runs, drawn on the seeds of the batch in which an earlier version of this simulation found its largest value, holds one step at $t = 3{,}555$ that replaces with probability $0.023$.
There the sampled law of keeping in bin $0$ puts $13\%$ of its mass above bin $1$, against $0.9\%$ under $\nu$; with that law set to its value under $\nu$ the probability falls to $3.2 \times 10^{-8}$.
A sampled replacement at bin $15$ beats keeping in $40\%$ of draws only after $700$ observed replacements there.

At the wear-shock rates $0.005$ and $0.02$, IMED-RL's regret ratios are $0.319$ and $0.242$.
Every run replaces from bin $15$ at $T$.
The fraction at bin $15$ falls as the gap at the threshold closes, from $0.100$ at the rate $0.005$ to $0.041$ at the rate $0.02$.
At the rate $0.02$ the fraction at bin $16$, $0.045$, exceeds that at bin $15$ by $0.0035 \pm 0.0004$, in both halves of the runs.
The ratio of the two fractions is the ratio of the two keep counts times the ratio of the two values of $\Kinf(x,a)$.
The count at bin $15$ stays near 10 times that at bin $16$ at the three rates.
The ratio of the values of $\Kinf(x,a)$ rises from $2.5$ and $3.2$ to $11.3$ as the gap at bin $15$ closes.
At the rate $0.02$ the policy that replaces from bin $16$ has regret $7.28$ at $T = 10^6$ and lies below the limit until $t = 4.1 \times 10^6$, so the regret ratio does not show learning there.
IMED-RL's regret of $6.57$ lies $19$ standard errors below that policy's.
PSRL keeps until forced at both rates as well, with regret ratios $5.74$ and $3.12$.
Under $\nu$ the first and the second $250$ runs agree within $1.4$ standard errors on every reading at $T$, for all three algorithms.
Over all $21$ checkpoints the largest difference is $3.2$ standard errors, in PSRL's transported fraction at bin $15$ at $t = 2.5 \times 10^5$.

\paragraph{A lighter prior.}
PSRL was run again with a Dirichlet weight of $1/18$ on every next state, the pseudo-count of IMED-RL, and with nothing else changed.
It ran $500$ times for $T = 10^6$ steps under $\nu$ and under each closest alternative with $\xi_\Delta = 0.02$.
Policy iteration again returns the policy of relative value iteration on $1{,}000$ bus kernels sampled under the new prior.
No planner reached its iteration cap.
The divergence decomposition was not checked again, since the simulator is the one checked above.
Uncertainties below are one standard error over runs.
Under $\nu$, $374$ of the $500$ runs replace the engine below bin $17$ at least once by $T$, against none of the $2{,}700$ runs with weight $1$.
Under the three environments together the count is $1{,}169$ of $1{,}500$.
Of the $263$ runs under $\nu$ that replace in bins $14$ to $16$, $225$ first do so by $t = 1{,}000$, when the sampled laws of keeping rest on few observations.
In $88$ of the $374$ runs the replacements lie only in bins $1$ to $12$.
At $T$ the \textsc{allocation} term is $0.237 \pm 0.019$.
Of the runs that do not replace in bins $14$ to $16$ by $t = 10^4$, $22$ of $686$ over the three environments first replace there later, a share of $3.2\%$ with a $95\%$ Clopper--Pearson interval of $2.0\%$ to $4.8\%$.
This comparison was not specified before the run.
Its predicted share is $3.3\%$, from the $7.4$ planning steps that follow $t = 10^4$ on the line of keeping until forced, at a probability of $4.5 \times 10^{-3}$ a step under the laws of keeping of $\nu$.

At $T = 10^6$ the regret under $\nu$ is $126.3 \pm 1.1$ in count form, $4.65 \pm 0.04$ times $c^\star_\Phi(\nu)\log T$.
Without the \textsc{allocation} term it is $4.645 \pm 0.039$ times the limit, $4.6\%$ below the $4.871$ of keeping until forced.
The reading specified before the run put this ratio at $4.82$ or above.
In that reading a run replaces only in an unbroken chain of replacing steps, which ends at the first step that keeps.
The whole shortfall comes from $34$ runs.
Each of them replaces at bins $15$ and $16$ at least $1{,}000$ times by $T$ and ends with a policy that replaces before bin $17$.
Without the \textsc{allocation} term their regret averages $1.53 \pm 0.15$ times the limit, with a range of $0.10$ to $4.33$.
The other $466$ runs stay on the line of keeping until forced, within $0.7$ standard errors of its count form of $132.2$.
A sampled law of replacing keeps its observations through the planning steps that keep, and each later step samples it again.
A run that has replaced once at a bin can therefore replace there again at any later step.
At the laws of keeping of $\nu$ a law of replacing at bin $15$ beats keeping in $3.3\%$ of draws after one observed replacement, in $20\%$ after eight and in $78\%$ after $64$.
Under $\nu$, $50$ runs replace at least $16$ times at one of bins $14$ to $16$, and in $41$ of them an interval of at least $1{,}000$ steps without such a replacement lies between two intervals with one.
By $T = 10^6$, $7\%$ of the runs have left keeping until forced.

\section{Proofs}
\label{app:proofs}

\subsection{Proofs for Section~\ref{sec:allocation}}
\label{app:proofs-the-allocation-ledger}

\begin{proof}[Proof of Theorem~\ref{thm:allocation}]
The policy factors cancel in the likelihood ratio, so the increment of $L$ at time $t$ is $\ell_{A_t}(X_t)$, where $A_t$ is drawn given $\mathcal{F}_{t-1}$ with probabilities that do not depend on the instance.
Its $P_\nu$-conditional mean given $\mathcal{F}_{t-1}$ and $A_t$ is $\KL(\nu_{A_t} \,\|\, \lambda_{A_t})$, so $W_t := L_t - \sum_a N_a(t)\,\KL(\nu_a \,\|\, \lambda_a)$ is a $P_\nu$-martingale with $W_0 = 0$.
Using $\log y \le y - 1$,
\[
\mathbb{E}_{\nu_a}\big[(-\ell_a)^+\big]
 \;\le\; \mathbb{E}_{\nu_a}\!\left[\left(\tfrac{d\lambda_a}{d\nu_a} - 1\right)^{\!+}\right]
 \;\le\; \mathbb{E}_{\nu_a}\!\left[\tfrac{d\lambda_a}{d\nu_a}\right] \;\le\; 1
\]
from which $\mathbb{E}_{\nu_a}|\ell_a| = \KL(\nu_a \,\|\, \lambda_a) + 2\,\mathbb{E}_{\nu_a}[(-\ell_a)^+] \le \KL(\nu_a \,\|\, \lambda_a) + 2$.
The conditional increments of $W$ are therefore bounded in $L^1$ by $2\max_a \KL(\nu_a \,\|\, \lambda_a) + 2$.
Optional stopping at a time with $\mathbb{E}_\nu[\tau] < \infty$ gives $\mathbb{E}_\nu[W_\tau] = 0$.
The left side of \eqref{eq:allocation} is $\mathbb{E}_\nu[L_\tau]$, which the same bound renders absolutely convergent.
\end{proof}

\subsection{Proofs for Section~\ref{sec:sufficiency}}
\label{app:proofs-sufficiency-closes-the-data-processing-s}

\begin{proof}[Proof of Theorem~\ref{thm:sufficiency}]
With $\psi$ the log-partition function and $\theta_a, \eta_a$ the natural parameters of $\nu_a, \lambda_a$, the policy factors cancel and $L_\tau = \sum_a \big[ (\theta_a - \eta_a) S_a(\tau) - N_a(\tau)\,(\psi(\theta_a) - \psi(\eta_a)) \big]$,
a function of $Y^\star$ alone; the factorization criterion gives sufficiency.
Equation~\eqref{eq:chain} is the chain rule for relative entropy.
Because $\tau$ is finite almost surely under both laws and the arms' laws under $\nu$ and $\lambda$ have the same support, $P_\nu$ and $P_\lambda$ are mutually absolutely continuous on $\mathcal{F}_\tau$, and a statement that holds $P_\nu$-almost surely holds $P_\lambda$-almost surely.
For the equivalence, suppose first that $L_\tau$ is a function of $Y$.
Then $dP_\nu/dP_\lambda = e^{L_\tau}$ is a function of $Y$ on $\mathcal{F}_\tau$, so for every $E \in \mathcal{F}_\tau$,
\[
P_\nu(E \mid Y)
 \;=\; \frac{\mathbb{E}_\lambda\!\left[\mathbf{1}_E e^{L_\tau} \mid Y\right]}
            {\mathbb{E}_\lambda\!\left[e^{L_\tau} \mid Y\right]}
 \;=\; P_\lambda(E \mid Y)
\]
and the conditional divergences vanish.
Conversely, if the conditional laws agree $P_\nu$-almost surely then disintegrating both measures against the law of $Y$ gives $dP_\nu/dP_\lambda = (dP_\nu^{Y}/dP_\lambda^{Y})(Y)$, which is a function of $Y$.
Since the second term of \eqref{eq:chain} is a $P_\nu$-average of non-negative quantities, it vanishes exactly when those conditional laws agree.
\end{proof}

\subsection{Proofs for Section~\ref{sec:coincidence}}
\label{app:proofs-the-coincidence-ledger}

\begin{proof}[Proof of Theorem~\ref{thm:ledger}]
Conditionally on $\mathcal{F}_{t-1}$ and the chosen arm $a$, the increment of $sL$ is $s\,\ell_a(X_t)$ and the increment of $\sum_a N_a c_a(s)$ is $c_a(s)$.
Since $\mathbb{E}_{\lambda_a}[e^{s \ell_a}] = \int \nu_a^{\,s}\lambda_a^{\,1-s} = e^{-c_a(s)}$, this gives
\[
\mathbb{E}_\lambda[ M_t(s) \mid \mathcal{F}_{t-1}] = M_{t-1}(s) \cdot e^{-c_a(s)} e^{c_a(s)} = M_{t-1}(s)
\]
The action's conditional law given $\mathcal{F}_{t-1}$ does not depend on the instance, so averaging over it leaves $M_{t-1}(s)$, and $M_0(s) = 1$.
\end{proof}

\begin{proof}[Proof of Corollary~\ref{cor:ledger-stopped}]
$M(s)$ is a martingale, so the stopped process $M_{\tau \wedge t}(s)$ is a martingale with the same initial value, which gives the first claim and, for bounded $\tau$, the second.
For general $\tau$, $M_{\tau \wedge t}(s) \to M_\tau(s)$ almost surely on $\{\tau < \infty\}$ and $M(s) \ge 0$, so Fatou's lemma gives $\mathbb{E}_\lambda[M_\tau(s)] \le \liminf_t \mathbb{E}_\lambda[M_{\tau \wedge t}(s)] = 1$.
\end{proof}

\begin{proof}[Proof of Corollary~\ref{cor:endpoint}]
The integral defining $c_a(s)$ equals $1$ at $s = 0$ and at $s = 1$, which gives the endpoint values.
On the interval where it is finite, the integrand's $s$-derivative $\nu_a^s \lambda_a^{1-s} \log(\nu_a/\lambda_a)$ is dominated, locally uniformly in $s$, by an integrable envelope built from the integrand at two interior points.
So differentiation under the integral is legitimate and $c_a'(s) = -\int \nu_a^s \lambda_a^{1-s} \log(\nu_a/\lambda_a) \big/ \int \nu_a^s \lambda_a^{1-s}$,
which at $s=0$ is $-\mathbb{E}_{\lambda_a}[\log(\nu_a/\lambda_a)] = \KL(\lambda_a \,\|\, \nu_a)$, finite because the integral defining $c_a(s)$ is finite at some $s < 0$, and at $s=1$ is $-\mathbb{E}_{\nu_a}[\log(\nu_a/\lambda_a)] = -\KL(\nu_a \,\|\, \lambda_a)$.
The same domination licenses interchanging the derivative with $\mathbb{E}_\lambda$ in \eqref{eq:ledger}; using $\mathbb{E}_\nu[\,\cdot\,] =
\mathbb{E}_\lambda[e^{L_T}\,\cdot\,]$ then returns \eqref{eq:allocation} at $T$.
At a stopping time $\tau \le T$ the same computation applies to $\mathbb{E}_\lambda[M_\tau(s)] = 1$ (Corollary~\ref{cor:ledger-stopped}).
\end{proof}

\begin{proof}[Proof of Corollary~\ref{cor:ville}]
Ville's inequality \cite{ramdas2023statistical} applied to the nonnegative mean-one martingale $M_t(s)$ of Theorem~\ref{thm:ledger}, at threshold $1/\alpha$; the inequality bounds the supremum over the whole trajectory, so no horizon is needed.
\end{proof}

\subsection{Proofs for Section~\ref{sec:tightening}}
\label{app:proofs-per-pull-divergences-at-the-bhattacharyy}

\begin{proof}[Proof of Proposition~\ref{prop:ratio}]
With $F(y) := \kl(\mu_b,y)$ we have $F(\mu_b) = F'(\mu_b) = 0$, $F''(\mu_b) = 1/v$ and $F'''(\mu_b) = 2(1-\mu_b)^{-2} - 2\mu_b^{-2}$, so $\kl(\mu_b, \mu_b+g) = g^2/2v + \tfrac{g^3}{3}\big((1-\mu_b)^{-2} - \mu_b^{-2}\big) + O(g^4)$.
With $J(y) := \sqrt{\mu_b y} + \sqrt{(1-\mu_b)(1-y)}$ we have $J(\mu_b) = 1$, $J'(\mu_b) = 0$, $J''(\mu_b) = -1/4v$ and $J'''(\mu_b) = \tfrac38\big(\mu_b^{-2} - (1-\mu_b)^{-2}\big)$, so $c_b(1/2) = -\log J(\mu_b+g) = g^2/8v + \tfrac{g^3}{16}\big((1-\mu_b)^{-2} - \mu_b^{-2}\big) + O(g^4)$.
Dividing the two expansions gives \eqref{eq:ratio}.
In the Gaussian case $\int \nu_b^s \lambda_b^{1-s}$ is a Gaussian integral evaluating to $\exp\{-s(1-s)g^2/2\sigma^2\}$, and $\KL(\nu_b \,\|\, \lambda_b) = g^2/2\sigma^2$.
\end{proof}

\subsection{Proofs for Section~\ref{sec:limit}}
\label{app:proofs-the-finite-horizon-limit}

\begin{proof}[Proof of Theorem~\ref{thm:limit}]
The infimum of $P_\nu(E^c) + P_\lambda(E)$ over $E \in \mathcal{F}_T$ is $1 - \TV(P_\nu,
P_\lambda)$.
Writing $f_\nu, f_\lambda$ for densities against a common dominating measure, so that $\rho_T = \int \sqrt{f_\nu f_\lambda}$,
\[
\TV^2 = \tfrac14 \Big( \int |\sqrt{f_\nu} - \sqrt{f_\lambda}|\,|\sqrt{f_\nu} + \sqrt{f_\lambda}| \Big)^{\!2}
 \;\le\; \tfrac14 \int (\sqrt{f_\nu} - \sqrt{f_\lambda})^2 \int (\sqrt{f_\nu} + \sqrt{f_\lambda})^2
 \;=\; 1 - \rho_T^2
\]
by the Cauchy--Schwarz inequality, so $1 - \TV \ge 1 - \sqrt{1 - \rho_T^2}$; substituting $\rho_T = e^{-C_T(1/2)}$ gives \eqref{eq:limit}.
For $\varepsilon \le 1$, rearranging $1 - \sqrt{1-\rho_T^2} \le \varepsilon$ gives $\rho_T^2 \le \varepsilon(2-\varepsilon)$, which is the stated requirement.
For the two-point pair, $\rho_T = 2\sqrt{q(1-q)}$ and $\TV = |2q-1| = \sqrt{1-\rho_T^2}$.
\end{proof}

\begin{proof}[Proof of Proposition~\ref{prop:s-half}]
Let $f_\nu, f_\lambda$ be densities against a common dominating measure.
Pointwise $f_\nu^s f_\lambda^{1-s} \le (f_\nu \wedge f_\lambda)^{\underline{s}} (f_\nu \vee f_\lambda)^{1-\underline{s}}$, so H\"older gives $e^{-C_T(s)} \le (\int f_\nu \wedge f_\lambda)^{\underline{s}} (\int f_\nu \vee f_\lambda)^{1-\underline{s}} \le 2^{1-\underline{s}} (\int f_\nu \wedge f_\lambda)^{\underline{s}}$, using $\int f_\nu \vee f_\lambda = 2 - \int f_\nu \wedge f_\lambda \le 2$; rearranging and using $\inf_E (P_\nu(E^c) + P_\lambda(E)) = \int f_\nu \wedge f_\lambda$ gives \eqref{eq:family}.
The map $s \mapsto \int f_\nu^s f_\lambda^{1-s}$ is log-convex by H\"older, so $C_T$ is concave, and $C_T(0) = C_T(1) = 0$.
Concavity with $C_T(0) = 0$ makes $s \mapsto C_T(s)/s$ non-increasing on $(0,1]$, and concavity with $C_T(1) = 0$ makes $u \mapsto C_T(1-u)/u$ non-increasing on $(0,1]$;
hence $C_T(s)/\underline{s}$ is minimized over $(0,1)$ at $s = 1/2$, where it equals $2 C_T(1/2)$.
The prefactor $2^{1-1/\underline{s}}$ is increasing in $\underline{s}$ and so is also maximized at $s = 1/2$.
\end{proof}

\begin{proof}[Proof of Proposition~\ref{prop:half}]
The measure $P^{(1/2)}$ is a probability measure because $e^{C_T(1/2)}$ is by definition the reciprocal of $\int \sqrt{dP_\nu \, dP_\lambda}$. Therefore
\[
\KL\!\left(P_\nu \,\big\|\, P^{(1/2)}\right)
 = \mathbb{E}_\nu\!\left[\log \frac{dP_\nu}{\sqrt{dP_\nu\, dP_\lambda}}\right] - C_T(1/2)
 = \tfrac12\, \mathbb{E}_\nu\!\left[\log \frac{dP_\nu}{dP_\lambda}\right] - C_T(1/2)
 = \tfrac12 D_T - C_T(1/2)
\]
which is \eqref{eq:half}; non-negativity of relative entropy gives the inequality.
The per-arm statement is the same computation on a single observation.
\end{proof}

\begin{proof}[Proof of Proposition~\ref{prop:tilt}]
Write $U := \sum_a N_a(T)\, c_a(1/2)$.
By Theorem~\ref{thm:ledger},
$\rho_T = \mathbb{E}_\lambda[e^{L_T/2}] = \mathbb{E}_\lambda[M_T(1/2)\, e^{-U}]
= \mathbb{E}_{Q_{1/2}}[e^{-U}]$, so $C_T(1/2) = -\log \mathbb{E}_{Q_{1/2}}[e^{-U}]$.
Let $\widetilde{Q}$ have $d\widetilde{Q}/dQ_{1/2} = e^{-U}/\mathbb{E}_{Q_{1/2}}[e^{-U}]$; then
\[
\KL\!\left(Q_{1/2} \,\big\|\, \widetilde{Q}\right)
 = \mathbb{E}_{Q_{1/2}}\!\left[U + \log \mathbb{E}_{Q_{1/2}}[e^{-U}]\right]
 = \mathbb{E}_{Q_{1/2}}[U] - C_T(1/2)
\]
which is the left side of \eqref{eq:tilt}.
Finally $d\widetilde{Q}/dP_\lambda = M_T(1/2)\, e^{-U}/\rho_T = e^{L_T/2}/\rho_T$, so $\widetilde{Q} = P^{(1/2)}$.
\end{proof}

\subsection{Proofs for Section~\ref{sec:certificate}}
\label{app:proofs-a-per-arm-efficiency-certificate}

\begin{proof}[Proof of Proposition~\ref{prop:cert}]
Substitute $\mathbb{E}_\nu[N_b(T)] = I_b(T)/d_b$ into $R_T = \sum_b \Delta_b
\mathbb{E}_\nu[N_b(T)]$ and divide.
Equal gaps make every $\Delta_b$ and every $d_b$ equal, so the weights coincide; a mean equals each of its arguments exactly when they are all equal.
\end{proof}

\subsection{Proofs for Section~\ref{sec:alternative}}
\label{app:proofs-the-alternative-set-and-the-allocation-s}

\begin{proof}[Proof of Proposition~\ref{prop:mbary}]
For an exponential family, $d(\mu, \mu') = \psi^\star(\mu) - \psi^\star(\mu') - (\psi^\star)'(\mu')(\mu - \mu')$.
Write $\bar\mu := \sum_i w_i \mu_i$.
Since the means of the family form an interval and $w$ is a probability vector, $\bar\mu$ is again a mean of the family.
Summing against $w$ and subtracting $d(\bar\mu, \mu')$ cancels every term that involves $\mu'$, because $\sum_i w_i (\mu_i - \mu') = \bar\mu - \mu'$.
Hence, for every $\mu'$,
\[
\sum_i w_i \, d(\mu_i, \mu') \;=\; \sum_i w_i \psi^\star(\mu_i) - \psi^\star(\bar\mu) \;+\; d(\bar\mu, \mu')
\]
The last term is a relative entropy, so it is non-negative.
It vanishes at $\mu' = \bar\mu$.
The minimum over $\mu'$ is therefore attained at $\bar\mu$, where it equals the right side of \eqref{eq:mbary}.
\end{proof}

\begin{proof}[Proof of Proposition~\ref{prop:scalar}]
On the half-space in which arm $b$ is at least as good as $a^\star$, the arms other than $a^\star$ and $b$ keep their laws at no cost.
Each $d(\mu, \cdot)$ decreases below $\mu$ and increases above it, and $\mu_b < \mu_{a^\star}$.
So a pair of means with arm $b$ strictly ahead can be moved to a common mean at no greater cost.
By Proposition~\ref{prop:mbary} the best common mean is the $w$-weighted mean of $\mu_{a^\star}$ and $\mu_b$.
For Gaussian arms the infimum is then $w_{a^\star} w_b d_b / (w_{a^\star} + w_b)$, read as $0$ when $w_{a^\star} = w_b = 0$.
Hence
\[
T^\star(\nu)^{-1} = \sup_w \min_{b \ne a^\star} \frac{w_{a^\star} w_b d_b}{w_{a^\star} + w_b}
\]
Each term is continuous on the simplex, since $uv/(u+v) \le u \wedge v$, so the supremum is attained.
Every term is positive at the allocation with every $w_a = 1/K$, so the supremum is positive.
Every term vanishes at $w_{a^\star} = 0$, so a maximizer has $w_{a^\star} > 0$.
There each term is strictly increasing in its $w_b$ and involves no other $w_{b'}$.
So at a maximizer all $K-1$ terms are equal: were they not all equal, moving a little mass from an arm whose term exceeds the minimum to the arms attaining it, in equal parts, would raise the minimum.
Write $V$ for the common value and $\gamma := V / w_{a^\star}$.
Then $V(w_{a^\star} + w_b) = w_{a^\star} w_b d_b$ gives $w_b/w_{a^\star} = \gamma/(d_b - \gamma)$, so $\gamma < \min_b d_b$.
With $F(\gamma) := 1/\gamma + \sum_{b \ne a^\star} 1/(d_b - \gamma)$, the condition $\sum_a w_a = 1$ gives $1/V = F(\gamma)$.
Conversely, every $\gamma \in (0, \min_b d_b)$ gives the allocation $w(\gamma)$ with $w_{a^\star} = 1/\big(\gamma F(\gamma)\big)$ and $w_b = 1/\big(F(\gamma)(d_b - \gamma)\big)$.
It sums to one.
Every term at $w(\gamma)$ equals $1/F(\gamma)$, so $T^\star(\nu) \le F(\gamma)$.
A maximizer attains this bound at its own $\gamma$, so $T^\star(\nu) = \min_\gamma F(\gamma)$, which is \eqref{eq:scalar}.
Each summand of $F$ is strictly convex on $(0, \min_b d_b)$, and $F$ diverges at both ends, so its minimizer $\gamma^\star$ is unique.
The maximizer is therefore $w(\gamma^\star)$.
Differentiating gives $\gamma^{-2} = \sum_b (d_b - \gamma)^{-2}$ at $\gamma^\star$, and substituting the weights turns this into $w_{a^\star}^2 = \sum_b w_b^2$.
\end{proof}

\begin{proof}[Proof of Proposition~\ref{prop:scalar-general}]
Normalizing the weights $(1, r)$ by their sum puts Proposition~\ref{prop:mbary} in force.
So the minimization defining $G_b$ is attained at $\bar\mu_b(r) := (\mu_{a^\star} + r \mu_b)/(1+r)$, and $G_b(r) = \psi^\star(\mu_{a^\star}) + r \psi^\star(\mu_b) - (1+r) \psi^\star\big(\bar\mu_b(r)\big)$.
Since $(1+r)\, \bar\mu_b'(r) = \mu_b - \bar\mu_b(r)$, differentiating gives $G_b'(r) = d\big(\mu_b, \bar\mu_b(r)\big) > 0$.
As $r \to \infty$, $\bar\mu_b(r) \to \mu_b$ and $G_b(r) \to d(\mu_{a^\star}, \mu_b)$.
So $G_b$ is continuous and strictly increasing with the stated supremum.
Its inverse $G_b^{-1}$ is continuous on $[0, d(\mu_{a^\star}, \mu_b))$.
The objective is homogeneous of degree one in the weights, so its value at $(w_{a^\star}, w_b)$ is $w_{a^\star} G_b(w_b/w_{a^\star})$ when $w_{a^\star} > 0$.
That value is at most $w_{a^\star}\, d(\mu_{a^\star}, \mu_b)$, so it extends continuously by $0$ to $w_{a^\star} = 0$.
The proof of Proposition~\ref{prop:scalar} then applies with $G_b$ in place of $r \mapsto r d_b/(1+r)$.
The minimum of the $K-1$ terms attains its maximum, at an allocation with $w_{a^\star} > 0$, and the terms are equal there.
Writing $\gamma$ for the common value divided by $w_{a^\star}$ gives $w_b/w_{a^\star} = G_b^{-1}(\gamma)$.
Imposing $\sum_a w_a = 1$ gives $T^\star(\nu) = \big(1 + \sum_{b \ne a^\star} G_b^{-1}(\gamma)\big)/\gamma$.
Conversely, every $\gamma \in (0, \min_b d(\mu_{a^\star}, \mu_b))$ gives the allocation with $w_b/w_{a^\star} = G_b^{-1}(\gamma)$ and $\sum_a w_a = 1$.
Its $K-1$ terms all equal $\gamma/\big(1 + \sum_{b \ne a^\star} G_b^{-1}(\gamma)\big)$, so $T^\star(\nu) \le \big(1 + \sum_{b \ne a^\star} G_b^{-1}(\gamma)\big)/\gamma$.
Hence $T^\star(\nu)$ is the minimum over $\gamma$, which is \eqref{eq:scalar-general}, and the minimizers are the values of $\gamma$ at the optimal allocations.
The minimizer is unique \cite[Theorem~5]{garivier2016optimal}.
For Gaussian arms $G_b(r) = r d_b/(1+r)$, so $G_b^{-1}(\gamma) = \gamma/(d_b - \gamma)$ and \eqref{eq:scalar-general} reduces to \eqref{eq:scalar}.
\end{proof}

\begin{proof}[Proof of Proposition~\ref{prop:altsets}]
An infimum over a union is the minimum of the infima over its parts.
An intersection is contained in each part, and an infimum is non-increasing in its domain.
\end{proof}

\subsection{Proofs for Appendix~\ref{app:classical}}
\label{app:proofs-classical}

\begin{proof}[Proof of Lemma~\ref{lem:relax}]
(a) By Theorem~\ref{thm:allocation} the left side is $D_\tau$.
The chain rule for relative entropy, which is \eqref{eq:chain} and needs no assumption on the arms' families or on $\tau$, splits $D_\tau$ with $Y = \mathbf{1}_E$ into $\KL(P_\nu^{Y} \,\|\, P_\lambda^{Y})$ and the conditional term.
A $\{0,1\}$-valued statistic has Bernoulli laws, so the first term is $\kl(P_\nu(E), P_\lambda(E))$.

(b) Given the history up to $\tau$, draw $B \in \{0,1\}$ with $P(B = 1 \mid \mathcal{F}_\tau) = Z$, by the same kernel under both instances.
The kernel cancels in the density ratio, so the joint laws of the history and $B$ have relative entropy $D_\tau$.
The chain rule on the history enlarged by $B$, with $Y = B$, then gives $D_\tau \ge \KL(P_\nu^B \,\|\, P_\lambda^B) = \kl(\mathbb{E}_\nu[Z], \mathbb{E}_\lambda[Z])$.
The discarded term is the conditional divergence of the history given $B$.

(c) Write $\kl(u,v) = (1-u)\log\frac{1}{1-v} + u \log\frac1v - H(u)$ with $H(u) := -u\log u - (1-u)\log(1-u) \le \log 2$; the middle term is non-negative.
The second inequality is the first applied to $(1-u, 1-v)$.
For the third, $\partial_v \big[\kl(u,v) - 2(u-v)^2\big] = (v-u)\big(1/(v(1-v)) - 4\big)$ has the sign of $v - u$, and the bracket vanishes at $v = u$.
The monotonicity follows from $\partial_u \kl(u,v) = \log\frac{u(1-v)}{v(1-u)}$ and $\partial_v \kl(u,v) = \frac{v-u}{v(1-v)}$.

(d) As a function of its first argument, $\kl(\cdot, v)$ vanishes with its derivative at $v$ and has second derivative $y \mapsto 1/(y(1-y))$.
Taylor's formula with integral remainder therefore gives the identity, at $u = 0$ by continuity.
On $[u, v]$, $y(1-y) \le y \le v$, so the integrand is at least $(y-u)/v$, whose integral is $(v-u)^2/(2v)$.
The bound discards $\int_u^v (y-u)\big(1/(y(1-y)) - 1/v\big)\,dy \ge 0$.
On $\{u \le v\}$ the partial derivative above in $u$ is non-positive and the one in $v$ is non-negative.
\end{proof}

\begin{proof}[Proof of Proposition~\ref{prop:cl-twopoint}]
The first bound is Theorem~\ref{thm:limit}.
Its proof processes the history to $\mathbf{1}_E$ in total variation (\textsc{process}) and then applies the Cauchy--Schwarz step (\textsc{test}).
The defect of that step is the term in parentheses of Lemma~\ref{lem:affinity}.
When $P_\lambda(E) > P_\nu(E)$ the first step also discards $2(P_\lambda(E) - P_\nu(E))$.
The error sum then exceeds $1$.
The second follows from $\sqrt{1-u} \le 1 - u/2$ on $[0,1]$ (\textsc{test}).
It discards $1 - \sqrt{1-\rho_T^2} - \rho_T^2/2$.
For the third (\textsc{test}), Proposition~\ref{prop:half} gives
\[
\rho_T^2 = e^{-2C_T(1/2)} = e^{-D_T}\, e^{2\KL(P_\nu \| P^{(1/2)})} \ge e^{-D_T}
\]
The discarded factor is $e^{2\KL(P_\nu \| P^{(1/2)})}$.
For Pinsker's bound let $E_{\TV} := \{f_\nu > f_\lambda\}$, with the densities $f_\nu$ and $f_\lambda$ of the proof of Theorem~\ref{thm:limit}, so that $\TV(P_\nu, P_\lambda) = P_\nu(E_{\TV}) - P_\lambda(E_{\TV})$.
Lemma~\ref{lem:relax}(a) (\textsc{process}) and Lemma~\ref{lem:relax}(c) (\textsc{binary}) give $D_T \ge \kl(P_\nu(E_{\TV}), P_\lambda(E_{\TV})) \ge 2\,\TV(P_\nu, P_\lambda)^2$.
\end{proof}

\begin{proof}[Proof of Lemma~\ref{lem:affinity}]
Since $\nu_a \ll \lambda_a$ for every arm and the action probabilities cancel, $P_\nu \ll P_\lambda$ on $\mathcal{F}_T$ with density $e^{L_T}$, and $L_T$ is finite $P_\nu$-almost surely.
Hence $\rho_T = \int \sqrt{dP_\nu\, dP_\lambda} = \mathbb{E}_\nu[e^{-L_T/2}]$ and $\TV(P_\nu, P_\lambda) = \int (dP_\nu - dP_\lambda)^+ = \mathbb{E}_\nu[(1 - e^{-L_T})^+]$.
The infimum of the error sum is $1 - \TV(P_\nu, P_\lambda)$, as in the proof of Theorem~\ref{thm:limit}, which gives the identity.
For the sign, let $\bar P := \tfrac12 (P_\nu + P_\lambda)$, let $Z$ be the density of $P_\nu$ with respect to $P_\nu + P_\lambda$, and let $W := 2\sqrt{Z(1-Z)}$.
Then $dP_\nu/d\bar P = 2Z$ and $dP_\lambda/d\bar P = 2(1-Z)$, so $\rho_T = \mathbb{E}_{\bar P}[W]$ and $\TV(P_\nu, P_\lambda) = \mathbb{E}_{\bar P}|2Z - 1| = \mathbb{E}_{\bar P}[\sqrt{1 - W^2}]$, since $(2Z - 1)^2 = 1 - W^2$.
The map $y \mapsto \sqrt{1 - y^2}$ is strictly concave on $[0,1]$, so Jensen's inequality gives $\mathbb{E}_{\bar P}[\sqrt{1 - W^2}] \le \sqrt{1 - \rho_T^2}$ (\textsc{test}).
The term in parentheses is the defect of this step.
It vanishes exactly when $W$ is constant $\bar P$-almost surely, that is, when $Z \in \{z, 1 - z\}$ almost surely for some $z \in [0, 1/2]$.
Since $P_\nu \ll P_\lambda$, the null sets of $\bar P$ are those of $P_\lambda$, and $Z < 1$ almost surely.
Since $P_\nu(Z > 0) = 1$, $z = 0$ is excluded.
Then $Z \in (0,1)$ and $L_T = \log(Z/(1-Z))$ almost surely, so the condition reads $|L_T| = \log((1-z)/z)$ almost surely under $P_\lambda$.
Conversely, if $|L_T|$ is almost surely constant under $P_\lambda$, the constant is finite because $P_\nu \ll P_\lambda$, so $Z = e^{L_T}/(1 + e^{L_T})$ lies in a set $\{z, 1 - z\}$ almost surely.
\end{proof}

\begin{proof}[Proof of Proposition~\ref{prop:cl-split}]
For $\varepsilon \in (0,1]$ the map $u \mapsto \kl(1-u, \varepsilon-u)$ is convex on $[0, \varepsilon]$, because $\kl$ is jointly convex and both arguments are affine in $u$.
It is finite on $[0, \varepsilon)$.
For $\varepsilon < 1$ it tends to $\infty$ as $u \to \varepsilon$.
The minimum defining $\underline{D}(\varepsilon)$ is therefore attained, at some $u^\star \in [0, \varepsilon)$ when $\varepsilon < 1$.

(i) Let $\varepsilon \le \varepsilon' \le 1$.
The point $u^\star$ of $\varepsilon$ is admissible for $\varepsilon'$, and $\varepsilon - u^\star \le \varepsilon' - u^\star \le 1 - u^\star$.
On $\{u \ge v\}$, $\kl(u,v)$ is non-increasing in $v$ (Lemma~\ref{lem:relax}(c)), so $\underline{D}(\varepsilon') \le \kl(1-u^\star, \varepsilon'-u^\star) \le \underline{D}(\varepsilon)$.
At $\varepsilon = 1$ every $u$ gives $\kl(1-u, 1-u) = 0$.
The choice $u = 0$ gives $\underline{D}(\varepsilon) \le \kl(1, \varepsilon) = \log(1/\varepsilon)$.
For the lower bounds let $\varepsilon < 1$, and let $B_\nu$ and $B_\lambda$ be the Bernoulli laws with means $1-u^\star$ and $\varepsilon-u^\star$.
Then $\TV(B_\nu, B_\lambda) = 1-\varepsilon$ and $\KL(B_\nu \,\|\, B_\lambda) = \underline{D}(\varepsilon)$.
Their Bhattacharyya coefficient is $\rho_\star := \sqrt{(1-u^\star)(\varepsilon-u^\star)} + \sqrt{u^\star(1-\varepsilon+u^\star)}$.
Proposition~\ref{prop:half}, applied to this pair with one observation, gives $\underline{D}(\varepsilon) = -2\log\rho_\star + 2\KL(B_\nu \,\|\, B^{(1/2)})$, with $B^{(1/2)}$ the normalized geometric mixture of the two laws, and the last term is non-negative (\textsc{test}).
On the two-point space the loss of the Cauchy--Schwarz step in the proof of Theorem~\ref{thm:limit} is given by the Lagrange identity,
\[
1 - \rho_\star^2 - (1-\varepsilon)^2 \;=\; \Big(\sqrt{u^\star(1-u^\star)} - \sqrt{(\varepsilon-u^\star)(1-\varepsilon+u^\star)}\Big)^2
\]
So $-2\log\rho_\star = \log\frac{1}{\varepsilon(2-\varepsilon)} + \log\frac{\varepsilon(2-\varepsilon)}{\rho_\star^2}$, and the last term is non-negative (\textsc{test}).
The two \textsc{test} defects do not vanish together.
Since $(\varepsilon-u^\star)(1-\varepsilon+u^\star) - u^\star(1-u^\star) = (1-\varepsilon)(\varepsilon - 2u^\star)$, the Lagrange identity shows that the second vanishes only at $u^\star = \varepsilon/2$.
There $B_\nu$ and $B_\lambda$ charge both points and differ, so $B^{(1/2)} \ne B_\nu$ and $\KL(B_\nu \,\|\, B^{(1/2)}) > 0$.
This gives the strict inequality.
Finally $\log\frac{1}{\varepsilon(2-\varepsilon)} = \log\frac{1}{2\varepsilon} + \log\frac{2}{2-\varepsilon}$, and the last term is non-negative (\textsc{test}).

(ii) Let $\TV(P,Q) = 1-\varepsilon$, let $p$ and $q$ be densities of $P$ and $Q$ against a common dominating measure, and let $A := \{p > q\}$, so that $P(A) - Q(A) = 1-\varepsilon$.
Data processing to $\mathbf{1}_A$ gives $\KL(P \,\|\, Q) \ge \kl(P(A), Q(A))$ (\textsc{process}).
With $u := P(A^c) \in [0, \varepsilon]$ the right side is $\kl(1-u, \varepsilon-u) \ge \underline{D}(\varepsilon)$ (\textsc{binary}).
The pair $(B_\nu, B_\lambda)$ of (i) attains the infimum.
At $\varepsilon = 1$ the pair $P = Q$ attains it.

(iii) By Theorem~\ref{thm:allocation}, $D_\tau$ is finite, so $P_\nu \ll P_\lambda$ on $\mathcal{F}_\tau$.
Were $\varepsilon = 0$, the event $E$ would have $P_\nu(E) = 1$ and $P_\lambda(E) = 0$, which $P_\nu \ll P_\lambda$ excludes.
Lemma~\ref{lem:relax}(a) splits $D_\tau$ into $\kl(P_\nu(E), P_\lambda(E))$ and the conditional divergence, the \textsc{process} defect.
With $u := P_\nu(E^c) \in [0, \varepsilon]$, $\kl(P_\nu(E), P_\lambda(E)) = \kl(1-u, \varepsilon-u) \ge \underline{D}(\varepsilon)$ (\textsc{binary}), with the difference as its defect.
The last claim follows from (i).
\end{proof}

\begin{proof}[Proof of Lemma~\ref{lem:chernoff-point}]
The standing assumption $\nu_a \ll \lambda_a$ makes $c_a(s)$ finite for $s \in [0,1]$, so $Q_s$ is defined.
Write $U := \sum_a N_a(T)\, c_a(s)$.
By Theorem~\ref{thm:ledger}, $dQ_s/dP_\lambda = e^{sL_T + U}$ on $\mathcal{F}_T$, so $dP_\lambda/dQ_s = e^{-sL_T - U}$ and $dP_\nu/dQ_s = e^{(1-s)L_T - U}$.
Hence $(dP_\nu \wedge dP_\lambda)/dQ_s = e^{-U - ((sL_T) \vee (-(1-s)L_T))}$, and integrating gives the identity, since $1 - \TV(P_\nu, P_\lambda) = \int dP_\nu \wedge dP_\lambda$.
For the upper bound, $x \wedge y \le x^{s}y^{1-s}$ gives $1 - \TV(P_\nu, P_\lambda) \le e^{-C_T(s)}$ at every $s \in [0,1]$.
Under $Q_s$ each pull of arm $a$ yields an observation with law $\nu_a^{(s)}$ given the past, because $M_T(s)$ is the product over rounds of $(d\nu_{A_t}^{(s)}/d\lambda_{A_t})(X_t)$.
So the increments of $L_T$ have conditional means $\mathbb{E}_{\nu_a^{(s)}}[\ell_a] = -c_a'(s)$ and conditional variances $\operatorname{Var}_{\nu_a^{(s)}}(\ell_a)$ on a pull of arm $a$.
With deterministic counts $U$ is deterministic, and $e^{-C_T(s)} = \mathbb{E}_\lambda[e^{sL_T}] = \mathbb{E}_{Q_s}[e^{-U}] = e^{-U}$, so $U = \sum_a n_a c_a(s) = C_T(s)$.
The conditional means then sum to $-\sum_a n_a c_a'(s) = -C_T'(s)$, also deterministic.
So $L_T + C_T'(s)$ is a sum of martingale differences under $Q_s$.
Their orthogonality gives $\operatorname{Var}_{Q_s}(L_T) = \sum_a n_a \operatorname{Var}_{\nu_a^{(s)}}(\ell_a)$, whatever the order of the pulls.
Mutual absolute continuity gives $C_T(0) = C_T(1) = 0$; if $C_T \equiv 0$ then $P_\nu = P_\lambda$ and the bounds are trivial.
Otherwise the concave function $C_T$ attains its maximum at some $s^\star \in (0,1)$, where $\mathbb{E}_{Q_{s^\star}}[L_T] = -C_T'(s^\star) = 0$.
By Jensen's inequality (\textsc{test}),
\[
1 - \TV(P_\nu, P_\lambda) \;\ge\; e^{-C_T(s^\star)}\, e^{-\mathbb{E}_{Q_{s^\star}}[(s^\star L_T) \vee (-(1-s^\star) L_T)]}
\]
Then $(sx) \vee (-(1-s)x) \le (s \vee (1-s))|x|$ (\textsc{test}) and $\mathbb{E}_{Q_{s^\star}}|L_T| \le (\operatorname{Var}_{Q_{s^\star}} L_T)^{1/2}$, by the Cauchy--Schwarz inequality and $\mathbb{E}_{Q_{s^\star}}[L_T] = 0$ (\textsc{test}), give the lower bound.
These moments are finite, because the moment generating function of $\ell_a$ under $\nu_a^{(s^\star)}$ is $e^{c_a(s^\star) - c_a(s^\star + r)}$, finite for $r \in [-s^\star, 1 - s^\star]$.
For the comparison after the lemma, concavity and $C_T(0), C_T(1) \ge 0$ give $C_T(1/2) \ge \tfrac12 C_T(s^\star)$, by writing $1/2$ as a convex combination of $s^\star$ and the farther endpoint.
\end{proof}

\begin{proof}[Proof of Proposition~\ref{prop:cl-lai-robbins}]
If $\Kinf(\nu_b, \mu^\star) = \infty$ there is nothing to prove.
Otherwise fix $\xi > 0$ and choose $\lambda_b \in \mathcal{M}_b$ with mean above $\mu^\star$ and $\KL(\nu_b \,\|\, \lambda_b) \le \Kinf(\nu_b, \mu^\star) + \xi$.
Let $\lambda$ equal $\nu$ except at arm $b$, whose law is $\lambda_b$.
Each arm's law under $\lambda$ lies in that arm's class, so $\lambda$ is an instance of $\mathcal{M}_1 \times \dots \times \mathcal{M}_K$ and consistency applies at $\lambda$.
Under $\lambda$, arm $b$ is the unique optimal arm, and every other arm has gap at least $g - \Delta_b > 0$, where $g := \mu_b(\lambda) - \mu_b$ is the separation of arm $b$.
Let $E := \{N_b(T) > T/2\}$.
\begin{enumerate}
\item Identity (Theorem~\ref{thm:allocation}): $\mathbb{E}_\nu[N_b(T)]\,\KL(\nu_b \,\|\, \lambda_b) = D_T$.
\item \textsc{process} (Lemma~\ref{lem:relax}(a)): $D_T \ge \kl(P_\nu(E), P_\lambda(E))$.
\item \textsc{binary} (Lemma~\ref{lem:relax}(c)): $\kl(P_\nu(E), P_\lambda(E)) \ge (1 - P_\nu(E)) \log(1/P_\lambda(E^c)) - \log 2$.
\item \textsc{markov}: $P_\nu(E) \le 2\,\mathbb{E}_\nu[N_b(T)]/T \le 2R_T(\nu)/(T\Delta_b)$, and $P_\lambda(E^c) = P_\lambda\big(\sum_{a \ne b} N_a(T) \ge T/2\big)$, which is at most $2R_T(\lambda)/(T(g - \Delta_b))$.
\item \textsc{uniform}: consistency at $\nu$ and at $\lambda$ gives $R_T(\nu) = o(T^p)$ and $R_T(\lambda) = o(T^p)$ for every $p > 0$.
\item \textsc{asymptotic}: by steps 4 and 5, $P_\nu(E) \to 0$ and $\log(1/P_\lambda(E^c)) \ge (1-p)\log T + \log((g - \Delta_b)/2)$ for all large $T$.
Dividing steps 1 to 3 by $\log T$ gives $\liminf_T \mathbb{E}_\nu[N_b(T)]/\log T \ge (1-p)/(\Kinf(\nu_b, \mu^\star) + \xi)$ for every $p > 0$.
\item \textsc{boundary}: let $\xi \to 0$.
\end{enumerate}
The regret bound follows from the identity $R_T = \sum_b \Delta_b\, \mathbb{E}_\nu[N_b(T)]$.
\end{proof}

\begin{proof}[Proof of Proposition~\ref{prop:cl-finite}]
Let $E := \{N_b(T) > T/2\}$, so that $\varepsilon_T - 1 = P_\nu(E) - P_\lambda(E)$.
The identity of Theorem~\ref{thm:allocation} gives $\mathbb{E}_\nu[N_b(T)]\,\KL(\nu_b \,\|\, \lambda_b) = D_T$.
Proposition~\ref{prop:half} gives $D_T \ge 2C_T(1/2)$ (\textsc{test}).
The Cauchy--Schwarz step in the proof of Theorem~\ref{thm:limit} gives $\TV(P_\nu, P_\lambda)^2 \le 1 - e^{-2C_T(1/2)}$ (\textsc{test}).
Its defect, $1 - e^{-2C_T(1/2)} - \TV(P_\nu, P_\lambda)^2$, is the term in parentheses of Lemma~\ref{lem:affinity} times $\sqrt{1 - e^{-2C_T(1/2)}} + \TV(P_\nu, P_\lambda)$.
Data processing to $\mathbf{1}_E$ in total variation gives $\TV(P_\nu, P_\lambda) \ge |P_\nu(E) - P_\lambda(E)| = |\varepsilon_T - 1|$ (\textsc{process}).
Together these give $e^{-2C_T(1/2)} \le 1 - (\varepsilon_T - 1)^2 = \varepsilon_T(2 - \varepsilon_T)$, which is the first inequality.
For the second, $\varepsilon_T(2-\varepsilon_T) \le 2\varepsilon_T$ (\textsc{test}) discards $\log(2/(2-\varepsilon_T))$ from the bound.
Markov's inequality gives $R_T(\nu) \ge \tfrac T2 \Delta_b P_\nu(E)$ (\textsc{markov}).
Under $\lambda$ every arm other than $b$ has gap at least $g - \Delta_b$, so Markov's inequality also gives $R_T(\lambda) \ge \tfrac T2 (g - \Delta_b) P_\lambda(E^c)$ (\textsc{markov}).
Bounding both gaps below by $\underline{\Delta}$ gives $\varepsilon_T \le 2(R_T(\nu) + R_T(\lambda))/(T\underline{\Delta})$ (\textsc{markov}).
\end{proof}

\begin{proof}[Proof of Corollary~\ref{cor:cl-gauss}]
For each suboptimal $b$ let $\lambda$ equal $\nu$ except that arm $b$ has mean $\mu_b + (1+\xi_\Delta)\Delta_b$ (\textsc{boundary}).
Then $\lambda \in \mathcal{U}(\nu)$, $\underline{\Delta} = \xi_\Delta\Delta_b$ and $\KL(\nu_b \,\|\, \lambda_b) = (1+\xi_\Delta)^2\Delta_b^2/2$.
The factor $(1+\xi_\Delta)^2$ is the ratio of $\KL(\nu_b \,\|\, \lambda_b)$ to its boundary value $\Delta_b^2/2$.
The margin $\xi_\Delta$ is not sent to $0$, since $\log(\xi_\Delta\Delta_b/8\Gamma) \to -\infty$ as $\xi_\Delta \to 0$.
Since $\nu \in \mathcal{U}(\nu)$ as well, $R_T(\nu) + R_T(\lambda) \le 2\Gamma T^p$ (\textsc{uniform}).
Proposition~\ref{prop:cl-finite} then gives $\mathbb{E}_\nu[N_b(T)] \ge 2\big((1-p)\log T + \log(\xi_\Delta\Delta_b/8\Gamma)\big)/\big((1+\xi_\Delta)^2\Delta_b^2\big)$.
Counts are non-negative, which gives the positive part, and $R_T(\nu) = \sum_b \Delta_b\, \mathbb{E}_\nu[N_b(T)]$.
\end{proof}

\begin{proof}[Proof of Proposition~\ref{prop:cl-gms-initial}]
For an arm $a$ with $\Delta_a = 0$ the hypothesis at $\nu$ gives $\mathbb{E}_\nu[N_a(T)] \ge T/K$, which is at least both right sides.
Let $a$ be suboptimal, fix $\lambda_a \in \mathcal{M}$ with mean above $\mu^\star$ and $\KL(\nu_a \,\|\, \lambda_a) < \infty$, which exists because $\Kinf(\nu_a, \mu^\star) < \infty$, and let $\lambda$ equal $\nu$ except at arm $a$, whose law is $\lambda_a$.
Under $\lambda$, arm $a$ is the unique optimal arm.
Let $Z := N_a(T)/T$, with means $z_\nu$ under $\nu$ and $z_\lambda$ under $\lambda$.
If $z_\nu \ge 1/K$ both bounds hold, since their right sides are at most $T/K$.
Let $z_\nu < 1/K$.
\begin{enumerate}
\item Identity (Theorem~\ref{thm:allocation}): $T z_\nu\, \KL(\nu_a \,\|\, \lambda_a) = D_T$.
\item \textsc{process} (Lemma~\ref{lem:relax}(b)): $D_T = \kl(z_\nu, z_\lambda)$ plus the conditional divergence of the history given the coin of the proof of Lemma~\ref{lem:relax}(b).
\item \textsc{uniform}: the hypothesis at $\lambda$ gives $z_\lambda \ge 1/K > z_\nu$, so $\kl(z_\nu, z_\lambda) = \kl(z_\nu, 1/K)$ plus the defect $\kl(z_\nu, z_\lambda) - \kl(z_\nu, 1/K)$, which is non-negative by Lemma~\ref{lem:relax}(d).
\item \textsc{binary} (Lemma~\ref{lem:relax}(d)): $\kl(z_\nu, 1/K) = \tfrac{K}{2}(1/K - z_\nu)^2$ plus the defect $\int_{z_\nu}^{1/K} (y - z_\nu)\big(1/(y(1-y)) - K\big)\,dy \ge 0$.
\end{enumerate}
Steps 1 to 4 write $T z_\nu\,\KL(\nu_a \,\|\, \lambda_a)$ as $\tfrac{K}{2}(1/K - z_\nu)^2$ plus three non-negative defects.
On $[0, 1/K)$ the resulting inequality holds exactly when $z_\nu \ge \tfrac1K\big(1 + T\KL(\nu_a \,\|\, \lambda_a) - \sqrt{T^2\KL(\nu_a \,\|\, \lambda_a)^2 + 2T\KL(\nu_a \,\|\, \lambda_a)}\big)$, a decreasing function of $\KL(\nu_a \,\|\, \lambda_a)$.
\begin{enumerate}
\setcounter{enumi}{4}
\item \textsc{boundary}: letting $\lambda_a$ range over the laws in $\mathcal{M}$ with mean above $\mu^\star$ and finite relative entropy from $\nu_a$ gives the first bound.
\end{enumerate}
Bounding $T z_\nu$ above by $T/K$ on the left of the inequality discards the non-negative term $T(1/K - z_\nu)\KL(\nu_a \,\|\, \lambda_a)$ and gives $(1/K - z_\nu)^2 \le 2T\KL(\nu_a \,\|\, \lambda_a)/K^2$.
The same supremum then gives the second bound.
If $T \le 1/(8\Kinf(\nu_a,\mu^\star))$ then $\sqrt{2T\Kinf(\nu_a,\mu^\star)} \le 1/2$.
\end{proof}

\begin{proof}[Proof of Proposition~\ref{prop:cl-gms-finite}]
Write $u_1$, $u_2$ and $u_3$ for the three terms subtracted from $1$ in the statement.
If $\Kinf(\nu_b, \mu^\star) = \infty$ or the right side is negative there is nothing to prove.
Let $\Kinf(\nu_b, \mu^\star) < \infty$ and the right side be non-negative, so that $u_1 + u_2 + u_3 < 1$.
Let $\xi := (\log T)^{-4}$.
Since $\Kinf(\nu_b, \mu^\star + \xi) \le \Kinf(\nu_b, \mu^\star) + \xi\,\mathrm{Lip}_{\mathcal{M}}(\nu_b, \mu^\star) < \infty$, there is $\lambda_b \in \mathcal{M}$ with mean above $\mu^\star + \xi$ and $\KL(\nu_b \,\|\, \lambda_b) < \infty$.
Fix one, and let $\lambda$ equal $\nu$ except at arm $b$, whose law is $\lambda_b$.
Under $\lambda$ every optimal arm of $\nu$ has gap at least $\xi$.
Let $Z := T^{-1}\sum_{a : \Delta_a = 0} N_a(T)$, with means $z_\nu$ and $z_\lambda$, and let $H(u) := -u\log u - (1-u)\log(1-u)$.
\begin{enumerate}
\item Identity (Theorem~\ref{thm:allocation}): $\mathbb{E}_\nu[N_b(T)]\,\KL(\nu_b \,\|\, \lambda_b) = D_T$.
\item \textsc{process} (Lemma~\ref{lem:relax}(b)): $D_T = \kl(z_\nu, z_\lambda)$ plus the conditional divergence of the history given the coin of the proof of Lemma~\ref{lem:relax}(b).
\item \textsc{binary} (second bound of Lemma~\ref{lem:relax}(c)): $\kl(z_\nu, z_\lambda) = z_\nu\log(1/z_\lambda) - \log 2$ plus the defect $(1 - z_\nu)\log\frac{1}{1 - z_\lambda} + \log 2 - H(z_\nu) \ge 0$.
\item \textsc{uniform}: the count guarantee at $\nu$ gives $z_\nu \ge 1 - u_2$.
At $\lambda$ it gives $z_\lambda \le \Upsilon\log T/(\xi^2 T)$ times the number of optimal arms of $\nu$, which is at most $K$, so $\log(1/z_\lambda) \ge \log T - \log\big(K\Upsilon(\log T)^9\big) = (1 - u_3)\log T$.
Hence $z_\nu\log(1/z_\lambda)$ equals $(1 - u_2)(1 - u_3)\log T$ plus a non-negative defect.
Steps 1 to 4 then give $\mathbb{E}_\nu[N_b(T)]\,\KL(\nu_b \,\|\, \lambda_b) \ge (1 - u_2)(1 - u_3)\log T - \log 2$.
\item \textsc{boundary}: since $u_2u_3 \ge 0$, the right side of step 4 is at least $(1 - u_1 - u_2 - u_3)\log T - \log 2 \ge 0$.
Letting $\lambda_b$ range over these laws therefore puts $\Kinf(\nu_b, \mu^\star + \xi)$ in place of $\KL(\nu_b \,\|\, \lambda_b)$.
The margin $\xi$ is kept.
Since $\xi < \xi_{\mathcal{M}}(\mu^\star)$, its defect $\Kinf(\nu_b, \mu^\star + \xi) - \Kinf(\nu_b, \mu^\star)$ is at most $\xi\,\mathrm{Lip}_{\mathcal{M}}(\nu_b, \mu^\star) = u_1\Kinf(\nu_b, \mu^\star)$.
\end{enumerate}
Hence $\mathbb{E}_\nu[N_b(T)] \ge \big((1 - u_2)(1 - u_3)\log T - \log 2\big)/\big((1 + u_1)\Kinf(\nu_b,\mu^\star)\big)$.
The statement follows from $1/(1 + u_1) \ge 1 - u_1$, $(1 - u_1)(1 - u_2)(1 - u_3) \ge 1 - u_1 - u_2 - u_3$ and $(1 - u_1)\log 2 \le \log 2$.
They discard the non-negative terms $u_1^2/(1 + u_1)$, $u_1u_2 + u_2u_3 + u_3u_1 - u_1u_2u_3$ and $u_1\log 2$, the second because $u_1, u_2, u_3 \in [0,1]$.
\end{proof}

\begin{proof}[Proof of Proposition~\ref{prop:cl-etc}]
Let $a^\star$ and $b$ be the optimal and the suboptimal arm of $\nu$.
Each part uses one alternative $\lambda$ in the class of its hypothesis, under which $b$ is the unique optimal arm.
In (i), $\lambda$ gives arm $b$ the mean $\mu^\star + \Delta$ and keeps arm $a^\star$, so $\KL(\nu_b \,\|\, \lambda_b) = 2\Delta^2$ and $\KL(\nu_{a^\star} \,\|\, \lambda_{a^\star}) = 0$.
In (ii), $\lambda$ exchanges the two laws, so both per-pull divergences are $\Delta^2/2$.
In (iii), for a margin $\xi > 0$, $\lambda$ gives arm $a^\star$ the mean $(\mu^\star + \mu_b - \xi)/2$ and arm $b$ the mean $(\mu^\star + \mu_b + \xi)/2$, so both per-pull divergences are $(\Delta + \xi)^2/8$.
Under $\lambda$ the gap of arm $a^\star$ is $\Delta$ in (i) and (ii) and $\xi$ in (iii).
Let $\tau' := T$ in (i) and $\tau' := \tau \wedge T$ in (ii) and (iii), and let $Z := N_b(T)/T$, with means $z_\nu$ and $z_\lambda$.
In (ii) and (iii), $Z$ is $\mathcal{F}_{\tau'}$-measurable, because after round $\tau \wedge T$ the algorithm pulls $\hat a$, which is $\mathcal{F}_\tau$-measurable, at every round up to $T$.
Let $H(u) := -u\log u - (1-u)\log(1-u)$.
\begin{enumerate}
\item Identity (Theorem~\ref{thm:allocation} at $\tau'$): in (i), $D_T = 2\Delta^2\,\mathbb{E}_\nu[N_b(T)]$.
In (ii) and (iii) the two per-pull divergences are equal and $N_{a^\star}(\tau') + N_b(\tau') = \tau'$, so $D_{\tau'}$ is $\mathbb{E}_\nu[\tau \wedge T]$ times their common value.
\item \textsc{process} (Lemma~\ref{lem:relax}(b) at $\tau'$): $D_{\tau'} = \kl(z_\nu, z_\lambda)$ plus the conditional divergence of the history given the coin of the proof of Lemma~\ref{lem:relax}(b).
\item \textsc{binary} (first bound of Lemma~\ref{lem:relax}(c)): $\kl(z_\nu, z_\lambda) = (1 - z_\nu)\log\frac{1}{1 - z_\lambda} - \log 2$ plus the defect $z_\nu\log(1/z_\lambda) + \log 2 - H(z_\nu) \ge 0$.
\item \textsc{uniform}: consistency at $\nu$ and at $\lambda$ makes both $z_\nu = R_T(\nu)/(\Delta T)$ and $1 - z_\lambda = R_T(\lambda)/(\Delta_{a^\star}(\lambda)\,T)$ of order $o(T^{p-1})$ for every $p > 0$.
\item \textsc{asymptotic}: by step 4, $z_\nu \to 0$ and $\log(1/(1 - z_\lambda)) \ge (1 - p)\log T$ for all large $T$, so steps 1 to 3 give $\liminf_{T \to \infty} D_{\tau'}/\log T \ge 1 - p$ for every $p > 0$, hence at least $1$.
\end{enumerate}
In (i), $R_T = \Delta\,\mathbb{E}_\nu[N_b(T)] = D_T/(2\Delta)$, which gives $1/(2\Delta)$.
In (ii) and (iii), $N_b(T) = N_b(\tau \wedge T) + (T - \tau \wedge T)\mathbf{1}\{\hat a = b\}$, and the alternation gives $N_b(\tau \wedge T) \ge (\tau \wedge T - 1)/2$, so $R_T \ge \tfrac{\Delta}{2}(\mathbb{E}_\nu[\tau \wedge T] - 1)$.
This discards the non-negative commit term $\Delta\,\mathbb{E}_\nu[(T - \tau \wedge T)\mathbf{1}\{\hat a = b\}]$ and a rounding term between $0$ and $\Delta$.
Steps 1 and 5 then give $\liminf_{T \to \infty} R_T/\log T \ge (\Delta/2)/(\Delta^2/2) = 1/\Delta$ in (ii) and at least $4\Delta/(\Delta + \xi)^2$ in (iii).
In (iii), letting $\xi \to 0$ gives $4/\Delta$ (\textsc{boundary}).
\end{proof}

\begin{proof}[Proof of Proposition~\ref{prop:cl-etc-fixed}]
Let $b$ be the suboptimal arm of $\nu$; under $\lambda$ it is optimal, and arm $a^\star$ has gap $\Delta$.
(i) With $2n \le T$, each arm is pulled $n$ times in the first $2n$ rounds, and the arm $\hat a$ at the other $T - 2n$.
Hence $R_T(\nu) = \Delta\big(n + (T - 2n)P_\nu(\hat a = b)\big)$ and $R_T(\lambda) = \Delta\big(n + (T - 2n)P_\lambda(\hat a = a^\star)\big)$, whose sum is the identity.
Let $E := \{\hat a = a^\star\}$, an event of $\mathcal{F}_{2n}$.
\textsc{process}: on $\mathcal{F}_{2n}$, $P_\nu(E^c) + P_\lambda(E)$ equals $1 - \TV(P_\nu, P_\lambda)$ plus the defect $\TV(P_\nu, P_\lambda) - \big(P_\nu(E) - P_\lambda(E)\big) \ge 0$.
The log-likelihood ratio on $\mathcal{F}_{2n}$ is $L_{2n} = \Delta\big(S_{a^\star}(2n) - S_b(2n)\big)$, Gaussian with variance $2n\Delta^2$ and mean $n\Delta^2$ under $\nu$ and $-n\Delta^2$ under $\lambda$.
The event $\{L_{2n} > 0\}$ attains the total variation, which is therefore $1 - \mathrm{erfc}(\Delta\sqrt{n}/2)$, and $C_{2n}(1/2) = -\log\mathbb{E}_\nu[e^{-L_{2n}/2}] = n\Delta^2/4$.
On $\{L_{2n} > 0\}$ the arm with the larger sample mean is $a^\star$, so for that arm the \textsc{process} defect vanishes, and by symmetry its two error probabilities are equal, which gives its regret.

(ii) By (i), $R_T(\nu) \ge \Delta n$ when $2n \le T$.
Otherwise the algorithm alternates up to $T$ and $R_T(\nu) \ge \Delta(T - 1)/2$.
Consistency at $\nu$ therefore gives $4n \le T$ for all large $T$.
For such $T$, $T - 2n \ge T/2$, and (i) gives $R_T(\nu) + R_T(\lambda) \ge \tfrac{\Delta T}{2}\,\mathrm{erfc}(\Delta\sqrt{n}/2)$.
The step discards $2\Delta n$, $\Delta(T - 2n)$ times the \textsc{process} defect, and $\Delta(T/2 - 2n)\,\mathrm{erfc}(\Delta\sqrt{n}/2)$.
\textsc{uniform}: consistency on $\mathcal{M}^2_\Delta$ at $\nu$ and at $\lambda$ makes the left side $o(T^p)$ for every $p > 0$, so $\mathrm{erfc}(\Delta\sqrt{n}/2) = o(T^{p-1})$.
\textsc{asymptotic}: then $\Delta\sqrt{n}/2 \to \infty$, and $-\log\mathrm{erfc}(z) = z^2 + \log z + O(1)$ as $z \to \infty$.
With $n = o(T^p)$ this gives $n\Delta^2/4 \ge (1 - 2p)\log T$ for all large $T$.
Hence $\liminf_{T \to \infty} R_T/\log T \ge \liminf_{T \to \infty} \Delta n/\log T \ge 4(1 - 2p)/\Delta$ for every $p > 0$.
With Theorem~\ref{thm:limit} and the second bound of Proposition~\ref{prop:cl-twopoint} in place of the exact value (\textsc{test}), the error sum is at least $\tfrac12 e^{-2C_{2n}(1/2)}$.
The same steps then give $n\Delta^2/2 \ge (1 - 2p)\log T$ and the constant $2/\Delta$.
With the Bretagnolle--Huber bound $\tfrac12 e^{-D_{2n}}$ they give $n\Delta^2 \ge (1 - 2p)\log T$ and $1/\Delta$.
\end{proof}

\begin{proof}[Proof of Proposition~\ref{prop:cl-wald}]
With one stream, Theorem~\ref{thm:allocation} reads $\mathbb{E}_\nu[\tau]\,\KL(\nu \,\|\, \lambda) = D_\tau$.
Lemma~\ref{lem:relax}(a) with the decision event (\textsc{process}) gives $D_\tau = \kl(P_\nu(E), P_\lambda(E))$ plus the conditional term, and $P_\nu(E) = 1 - \beta'$, $P_\lambda(E) = \alpha'$.
For $E' \in \{E, E^c\}$, change of measure on $\mathcal{F}_\tau$ gives $P_\lambda(E') = \mathbb{E}_\nu[\mathbf{1}_{E'} e^{-L_\tau}]$.
Hence $\KL(P_\nu(\cdot \mid E') \,\|\, P_\lambda(\cdot \mid E')) = \mathbb{E}_\nu[L_\tau \mid E'] - \log(P_\nu(E')/P_\lambda(E')) = \mathbb{E}_\nu[L_\tau \mid E'] + \log\mathbb{E}_\nu[e^{-L_\tau} \mid E']$.
Each such term is a Jensen defect and is non-negative.
\end{proof}

\begin{proof}[Proof of Proposition~\ref{prop:cl-ww}]
Under each law the steps $L_t - L_{t-1}$ are independent and identically distributed.
Since $\nu \ne \lambda$ and the two laws are mutually absolutely continuous, the steps are not almost surely zero.
So under each law there is $h > 0$ such that a step exceeds $h$ with positive probability, or falls below $-h$ with positive probability.
Let $J$ be an integer with $Jh$ larger than the length of the interval.
From any point of the interval, $J$ such steps in a row take $L_t$ out of it.
Each block of $J$ consecutive rounds is such a run with the same positive probability, independently of the earlier blocks.
So $P(\tau^\star > kJ)$ decays geometrically in $k$ under each law.
Hence $\mathbb{E}_\nu[\tau^\star]$ and $\mathbb{E}_\lambda[\tau^\star]$ are finite.
In particular $\tau^\star$ is finite almost surely under both laws, so the test is a sequential test.
On $E^\star$, $L_{\tau^\star} > 0$, so change of measure on $\mathcal{F}_{\tau^\star}$ gives $\alpha = \mathbb{E}_\nu[\mathbf{1}_{E^\star} e^{-L_{\tau^\star}}] \le P_\nu(E^\star) = 1 - \beta$.
Fix a sequential test with $\alpha' \le \alpha$ and $\beta' \le \beta$, and let $E$ be its event of deciding for $\nu$.
If $\mathbb{E}_\nu[\tau] = \infty$ there is nothing to prove.
\begin{enumerate}
\item Identity (Theorem~\ref{thm:allocation} with one stream, and Lemma~\ref{lem:relax}(a) at $E^\star$): $\mathbb{E}_\nu[\tau^\star]\,\KL(\nu \,\|\, \lambda)$ is $\kl(1-\beta, \alpha)$ plus the numerator of the bound.
\item Identity (Theorem~\ref{thm:allocation} with one stream): $\mathbb{E}_\nu[\tau]\,\KL(\nu \,\|\, \lambda) = D_\tau$.
\item \textsc{process} (Lemma~\ref{lem:relax}(a)): $D_\tau = \kl(1-\beta', \alpha') + \mathbb{E}_\nu[\KL(P_\nu(\cdot \mid \mathbf{1}_E) \,\|\, P_\lambda(\cdot \mid \mathbf{1}_E))]$, and the second term is the defect.
\item \textsc{uniform}: $\alpha' \le \alpha \le 1 - \beta \le 1 - \beta'$, so the monotonicity in Lemma~\ref{lem:relax}(c) gives $\kl(1-\beta', \alpha') \ge \kl(1-\beta, \alpha)$, with the difference as the defect.
\end{enumerate}
By steps 1 to 4, $(\mathbb{E}_\nu[\tau^\star] - \mathbb{E}_\nu[\tau])\,\KL(\nu \,\|\, \lambda)$ equals the numerator of the bound minus the defects of steps 3 and 4.
These defects are non-negative, and $\KL(\nu \,\|\, \lambda) > 0$ since $\nu \ne \lambda$, which gives the bound.
By the proof of Proposition~\ref{prop:cl-wald}, the numerator is the sum over $E' \in \{E^\star, (E^\star)^c\}$ of $P_\nu(E')\big(\mathbb{E}_\nu[L_{\tau^\star} \mid E'] + \log \mathbb{E}_\nu[e^{-L_{\tau^\star}} \mid E']\big)$.
By Jensen's inequality for the strictly convex exponential, the term of $E'$ is zero exactly when $L_{\tau^\star}$ is $P_\nu$-almost surely constant on $E'$.
If $L_{\tau^\star}$ is $P_\nu$-almost surely constant on both events, the numerator is zero, and step 1 gives $\mathbb{E}_\nu[\tau^\star]\,\KL(\nu \,\|\, \lambda) = \kl(1-\beta, \alpha)$.
Under $\lambda$ the same steps hold with $\nu$ and $\lambda$ exchanged and with $E^\star$ and $E$ replaced by their complements, the events of deciding for $\lambda$, since $1 - \alpha' \ge 1 - \alpha \ge \beta \ge \beta'$.
On $\mathcal{F}_{\tau^\star}$ the laws $P_\nu$ and $P_\lambda$ are mutually absolutely continuous, so the condition for a zero defect is the same under both.
\end{proof}

\begin{proof}[Proof of Proposition~\ref{prop:cl-multitest}]
Fix $j \ne i$ with $\alpha_i + \alpha_j \le 1$ and $\lambda \in \mathcal{H}_j$ with $\KL(\nu \,\|\, \lambda) < \infty$, and let $E$ be the event that the test decides $\mathcal{H}_i$.
\begin{enumerate}
\item Identity (Theorem~\ref{thm:allocation} with one stream): $\mathbb{E}_\nu[\tau]\,\KL(\nu \,\|\, \lambda) = D_\tau$.
\item \textsc{process} (Lemma~\ref{lem:relax}(a)): $D_\tau = \kl(P_\nu(E), P_\lambda(E)) + \mathbb{E}_\nu[\KL(P_\nu(\cdot \mid \mathbf{1}_E) \,\|\, P_\lambda(\cdot \mid \mathbf{1}_E))]$, and the second term is the defect.
\item \textsc{uniform}: under $\nu$ every decision other than $\mathcal{H}_i$ is an error, so $P_\nu(E) \ge 1 - \alpha_i$, and under $\lambda$ the decision $\mathcal{H}_i$ is an error, so $P_\lambda(E) \le \alpha_j \le 1 - \alpha_i$.
The monotonicity in Lemma~\ref{lem:relax}(c) gives $\kl(P_\nu(E), P_\lambda(E)) \ge \kl(1 - \alpha_i, \alpha_j)$, with the difference as the defect.
\item \textsc{boundary}: steps 1 to 3 hold for every $\lambda \in \mathcal{H}_j$ with $\KL(\nu \,\|\, \lambda) < \infty$, and the infimum of $\KL(\nu \,\|\, \lambda)$ over these $\lambda$ is $\Kinf(\nu, \mathcal{H}_j)$, which gives (i).
At a fixed $\lambda$ the defect is $\mathbb{E}_\nu[\tau]\,(\KL(\nu \,\|\, \lambda) - \Kinf(\nu, \mathcal{H}_j))$.
\item \textsc{binary} (second bound of Lemma~\ref{lem:relax}(c)): $\kl(1 - \alpha_i, \alpha_j) \ge (1 - \alpha_i)\log(1/\alpha_j) - \log 2$.
\item \textsc{asymptotic}: dividing by $\log(1/\alpha_j)$ and letting $\alpha_i, \alpha_j \to 0$ gives (ii).
\end{enumerate}
\end{proof}

\begin{proof}[Proof of Proposition~\ref{prop:cl-power-one}]
Fix $\lambda \in \mathcal{H}_0$ with $\KL(\nu \,\|\, \lambda) < \infty$, and let $E := \{\tau < \infty\}$, an event in $\mathcal{F}_\tau$.
Since $\mathbb{E}_\nu[\tau] < \infty$, $P_\nu(E) = 1$.
\begin{enumerate}
\item Identity (Theorem~\ref{thm:allocation} with one stream): $\mathbb{E}_\nu[\tau]\,\KL(\nu \,\|\, \lambda) = D_\tau$.
\item \textsc{process} (Lemma~\ref{lem:relax}(a)): $D_\tau = \kl(P_\nu(E), P_\lambda(E)) + \mathbb{E}_\nu[\KL(P_\nu(\cdot \mid \mathbf{1}_E) \,\|\, P_\lambda(\cdot \mid \mathbf{1}_E))]$.
Since $P_\nu(E) = 1$, the first term is $\kl(1, P_\lambda(E)) = \log(1/P_\lambda(E))$ with no \textsc{binary} step, and the second is $\KL(P_\nu \,\|\, P_\lambda(\cdot \mid E))$.
This is the identity of (i), with its second term as the defect.
\item \textsc{uniform}: the level gives $P_\lambda(E) \le \alpha$, so $\log(1/P_\lambda(E)) = \log(1/\alpha) + \log(\alpha/P_\lambda(E))$, and the second term is the defect.
\item \textsc{boundary}: steps 1 to 3 hold for every $\lambda \in \mathcal{H}_0$ with $\KL(\nu \,\|\, \lambda) < \infty$, and the infimum of $\KL(\nu \,\|\, \lambda)$ over these $\lambda$ is $\Kinf(\nu, \mathcal{H}_0)$.
At a fixed $\lambda$ the defect is $\mathbb{E}_\nu[\tau]\,(\KL(\nu \,\|\, \lambda) - \Kinf(\nu, \mathcal{H}_0))$.
If no such $\lambda$ exists, the bound holds because $\mathbb{E}_\nu[\tau] > 0$: for $\lambda \in \mathcal{H}_0$, $P_\nu(\tau = 0) = P_\lambda(\tau = 0) \le \alpha < 1$.
\end{enumerate}
For (ii), $Y$ is $\mathcal{F}_\tau$-measurable and takes the values $\pm 1$ almost surely under $P_\nu$ and under $P_\lambda(\cdot \mid E)$.
The chain rule for relative entropy, applied to these two laws with the statistic $Y$, splits $\KL(P_\nu \,\|\, P_\lambda(\cdot \mid E))$ into the relative entropy between the laws of $Y$, which is $\kl(P_\nu(E_+), P_\lambda(E_+ \mid E))$, and a conditional term.
The events $\{Y = 1\} = E_+$ and $\{Y = -1\} = E_-$ lie in $E$, so conditioning $P_\lambda(\cdot \mid E)$ on $Y$ gives $P_\lambda(\cdot \mid Y)$, and the conditional term is $\mathbb{E}_\nu[\KL(P_\nu(\cdot \mid Y) \,\|\, P_\lambda(\cdot \mid Y))]$.
\end{proof}

\begin{proof}[Proof of Proposition~\ref{prop:cl-chernoff}]
Fix $\kappa$ small enough that every risk is finite, and let $\tau$ and the decision be those of the procedure for $\kappa$.
Then $\mathbb{E}_\nu[\tau] < \infty$.
Fix $\lambda \in \Alt_\Theta(\nu)$ and let $E$ be the event of deciding the hypothesis that contains $\nu$.
Theorem~\ref{thm:allocation}, with experiments as arms, gives $\sum_e \mathbb{E}_\nu[N_e(\tau)]\, \KL(\nu_e \,\|\, \lambda_e) = D_\tau$.
Lemma~\ref{lem:relax}(a) (\textsc{process}) and the second bound of Lemma~\ref{lem:relax}(c) (\textsc{binary}) give $D_\tau \ge (1 - P_\nu(E^c)) \log(1/P_\lambda(E)) - \log 2$.
Markov's inequality on the terminal loss gives $P_\lambda(E) \le \mathcal{R}_\kappa(\lambda)/r_{\min}$ and $P_\nu(E^c) \le \mathcal{R}_\kappa(\nu)/r_{\min}$ (\textsc{markov}), after the sampling cost is added to the terminal loss.
Both bounds are $O(\kappa\log(1/\kappa))$ (\textsc{uniform}).
With $\bar w := \mathbb{E}_\nu[N(\tau)]/\mathbb{E}_\nu[\tau]$, the left side equals $\mathbb{E}_\nu[\tau] \sum_e \bar w_e\, \KL(\nu_e \,\|\, \lambda_e)$, and $\min_{\lambda} \sum_e \bar w_e\, \KL(\nu_e \,\|\, \lambda_e) \le T^\star_\Theta(\nu)^{-1}$ (\textsc{allocation}).
Taking the minimum over $\lambda$ gives $\mathbb{E}_\nu[\tau] \ge (1 - o(1))\, T^\star_\Theta(\nu) \log(1/\kappa)$ as $\kappa \to 0$ (\textsc{asymptotic}).
Finally $\mathcal{R}_\kappa(\nu) \ge \kappa\,\mathbb{E}_\nu[\tau]$, which discards the expected terminal loss.
\end{proof}

\begin{proof}[Proof of Corollary~\ref{cor:cl-active}]
Every risk is non-negative, so $\pi_\lambda \mathcal{R}_\kappa(\lambda) \le \sum_{\nu} \pi_\nu \mathcal{R}_\kappa(\nu)$ for every $\lambda \in \Theta$, which drops the other states' risks.
Since $\pi_\lambda > 0$, $\mathcal{R}_\kappa(\lambda) = O(\kappa \log(1/\kappa))$ for every $\lambda \in \Theta$, which is the hypothesis of Proposition~\ref{prop:cl-chernoff}.
That proposition holds with finitely many hypotheses.
Its chain gives $\mathcal{R}_\kappa(\nu) \ge (1 - o(1))\, \kappa \log(1/\kappa)\, T^\star_\Theta(\nu)$ at each $\nu \in \Theta$.
Weighting by $\pi_\nu$ and summing over the finite set $\Theta$ gives the claim.
\end{proof}

\begin{proof}[Proof of Proposition~\ref{prop:cl-design}]
If $\Alt_\Theta(\nu)$ is empty, then $T^\star_\Theta(\nu) = 0$ and there is nothing to prove.
Otherwise let $\nu \in \Theta_i$, fix $\lambda \in \Alt_\Theta(\nu)$, and let $E$ be the event that the procedure stops and decides $\Theta_i$.
\begin{enumerate}
\item Identity (Theorem~\ref{thm:allocation}, with experiments as arms): $\sum_e \mathbb{E}_\nu[N_e(\tau)]\, \KL(\nu_e \,\|\, \lambda_e) = D_\tau$.
\item \textsc{process} (Lemma~\ref{lem:relax}(a)): $D_\tau = \kl(P_\nu(E), P_\lambda(E)) + \mathbb{E}_\nu[\KL(P_\nu(\cdot \mid \mathbf{1}_E) \,\|\, P_\lambda(\cdot \mid \mathbf{1}_E))]$, and the second term is the defect.
\item \textsc{uniform}: $\delta$-correctness at $\nu$ gives $P_\nu(E) \ge 1 - \delta$, and at $\lambda$, whose hypothesis is not $\Theta_i$, it gives $P_\lambda(E) \le \delta$.
So $P_\nu(E) \ge 1 - \delta \ge 1/2 \ge \delta \ge P_\lambda(E)$, and the monotonicity in Lemma~\ref{lem:relax}(c) gives $\kl(P_\nu(E), P_\lambda(E)) \ge \kl(1 - \delta, \delta) = \kl(\delta, 1-\delta)$, with the difference as the defect.
\item \textsc{boundary}: if $\mathbb{E}_\nu[\tau] = 0$, steps 1 to 3 give $\kl(\delta, 1-\delta) \le 0$, and the bound holds.
Otherwise write $\bar w := \mathbb{E}_\nu[N(\tau)]/\mathbb{E}_\nu[\tau]$, so that the left side of step 1 is $\mathbb{E}_\nu[\tau] \sum_e \bar w_e\, \KL(\nu_e \,\|\, \lambda_e)$.
Steps 1 to 3 hold at every $\lambda \in \Alt_\Theta(\nu)$, so
\[
\mathbb{E}_\nu[\tau] \inf_{\lambda \in \Alt_\Theta(\nu)} \sum_e \bar w_e\, \KL(\nu_e \,\|\, \lambda_e) \;\ge\; \kl(\delta, 1-\delta)
\]
At a fixed $\lambda$ the defect is $\mathbb{E}_\nu[\tau]$ times the difference between its sum and the infimum.
\item \textsc{allocation}: the infimum is at most $T^\star_\Theta(\nu)^{-1}$, with $\mathbb{E}_\nu[\tau]$ times the difference as the defect.
So $\mathbb{E}_\nu[\tau]\, T^\star_\Theta(\nu)^{-1} \ge \kl(\delta, 1-\delta)$, and $T^\star_\Theta(\nu) < \infty$ gives the bound.
\end{enumerate}
\end{proof}

\begin{proof}[Proof of Proposition~\ref{prop:cl-bai}]
Let $\lambda \in \Alt(\nu)$ have a unique optimal arm, and let $E := \{\hat a = a^\star(\nu)\}$.
The identity of Theorem~\ref{thm:allocation} gives $\sum_a \mathbb{E}_\nu[N_a(\tau)]\, d(\mu_a, \mu_a(\lambda)) = D_\tau$, and Lemma~\ref{lem:relax}(a) gives $D_\tau \ge \kl(P_\nu(E), P_\lambda(E))$ (\textsc{process}).
By $\delta$-correctness at $\nu$ and at $\lambda$, $P_\nu(E) \ge 1 - \delta \ge 1/2 \ge \delta \ge P_\lambda(E)$, so Lemma~\ref{lem:relax}(c) gives $\kl(P_\nu(E), P_\lambda(E)) \ge \kl(1-\delta, \delta) = \kl(\delta, 1-\delta)$ (\textsc{uniform}).
With $\bar w := \mathbb{E}_\nu[N(\tau)]/\mathbb{E}_\nu[\tau]$ this reads $\mathbb{E}_\nu[\tau] \sum_a \bar w_a\, d(\mu_a, \mu_a(\lambda)) \ge \kl(\delta, 1-\delta)$.
The instances with a unique optimal arm are dense in $\Alt(\nu)$ and the sum is continuous in $\lambda$, so the bound holds with the infimum over $\Alt(\nu)$ (\textsc{boundary}).
That infimum is at most $T^\star(\nu)^{-1}$ (\textsc{allocation}).
\end{proof}

\begin{proof}[Proof of Proposition~\ref{prop:cl-topm}]
Fix an arm $a$ and $\xi > 0$.
Let $\lambda^{(a)}$ equal $\nu$ except at arm $a$, whose mean becomes $\mu_{[m+1]} - \xi$ if $a \in \mathcal{S}^\star$ and $\mu_{[m]} + \xi$ otherwise; no top-$m$ set of $\lambda^{(a)}$ is $\mathcal{S}^\star$.
If $\lambda^{(a)}$ has no unique top-$m$ set, several arms of $\nu$ tie at the new boundary: at $\mu_{[m+1]}$ if $a \in \mathcal{S}^\star$ and at $\mu_{[m]}$ otherwise.
Moving one of them by $\xi' > 0$, up in the first case and down in the second, gives an instance $\lambda^{(a,\xi')}$ with a unique top-$m$ set other than $\mathcal{S}^\star$.
If $\lambda^{(a)}$ has a unique top-$m$ set, let $\lambda^{(a,\xi')} := \lambda^{(a)}$.
Theorem~\ref{thm:allocation} gives $D_\tau$ as $\mathbb{E}_\nu[N_a(\tau)]\, d(\mu_a, \mu_a(\lambda^{(a)}))$ plus the moved arm's term, which is at most $\mathbb{E}_\nu[\tau]$ times that arm's per-pull divergence.
Lemma~\ref{lem:relax}(a) with $E := \{\text{the recommendation is } \mathcal{S}^\star\}$ (\textsc{process}), $\delta$-correctness at $\nu$ and at $\lambda^{(a,\xi')}$ (\textsc{uniform}) and Lemma~\ref{lem:relax}(c) give $D_\tau \ge \kl(\delta, 1-\delta)$.
Since $\mathbb{E}_\nu[\tau] < \infty$, the moved arm's term vanishes as $\xi' \to 0$.
Letting $\xi' \to 0$ and then $\xi \to 0$ (\textsc{boundary}), and summing $\mathbb{E}_\nu[N_a(\tau)]$ over $a$, gives the bound, since $\sum_a \mathbb{E}_\nu[N_a(\tau)] = \mathbb{E}_\nu[\tau]$.
\end{proof}

\begin{proof}[Proof of Proposition~\ref{prop:cl-topm-exact}]
(i) The set $\Alt_m(\nu)$ is the union over $a \in \mathcal{S}^\star$ and $b \notin \mathcal{S}^\star$ of the half-spaces $\{\lambda : \mu_b(\lambda) \ge \mu_a(\lambda)\}$, so Proposition~\ref{prop:altsets} reduces the infimum to a minimum over the pairs $(a, b)$.
On the half-space of $(a,b)$ the other arms can keep their laws, so only the terms of arms $a$ and $b$ remain.
Let $\mu_a(\lambda) \le \mu_b(\lambda)$ and $x := \mu_a \wedge (\mu_a(\lambda) \vee \mu_b)$.
Since $\mu_b < \mu_a$, the point $x$ lies in the closed interval between $\mu_a$ and $\mu_a(\lambda)$ and in the one between $\mu_b$ and $\mu_b(\lambda)$.
Since $\partial_{\mu'} d(\mu, \mu') = (\psi^\star)''(\mu')(\mu' - \mu)$, the function $d(\mu, \cdot)$ is non-increasing below $\mu$ and non-decreasing above it.
So $d(\mu_a, x) \le d(\mu_a, \mu_a(\lambda))$ and $d(\mu_b, x) \le d(\mu_b, \mu_b(\lambda))$, and moving both arms to $x$ stays in the half-space.
The infimum over the half-space is therefore the minimum over $\mu'$ of $w_a\, d(\mu_a, \mu') + w_b\, d(\mu_b, \mu')$.
When $w_a + w_b > 0$, Proposition~\ref{prop:mbary} with the weights normalized by their sum places the minimum at the weighted mean.

(ii) The proof of Proposition~\ref{prop:cl-bai} applies with $E := \{\text{the recommendation is } \mathcal{S}^\star\}$, $\Alt_m(\nu)$ in place of $\Alt(\nu)$ and $T^\star_m(\nu)$ in place of $T^\star(\nu)$.
For $\lambda \in \Alt_m(\nu)$ with a unique top-$m$ set, $\delta$-correctness at $\lambda$ gives $P_\lambda(E) \le \delta$, because $\mathcal{S}^\star$ is not that set.
The steps are the identity, \textsc{process}, \textsc{uniform}, \textsc{boundary} and \textsc{allocation}.
For \textsc{boundary}, the instances in $\Alt_m(\nu)$ with a unique top-$m$ set are dense in it.
Raising $\mu_b(\lambda)$ slightly and separating the remaining ties gives such an instance, with an arm outside $\mathcal{S}^\star$ above an arm of $\mathcal{S}^\star$.

(iii) Moving arm $a \in \mathcal{S}^\star$ to $\mu_{[m+1]}$, or arm $a \notin \mathcal{S}^\star$ to $\mu_{[m]}$, gives an instance in $\Alt_m(\nu)$.
Write $d_a$ for the per-pull divergence of that move.
Then $\inf_{\lambda \in \Alt_m(\nu)} \langle w, k(\lambda)\rangle \le \min_a w_a d_a$ for every allocation $w$, and $\sup_w \min_a w_a d_a = (\sum_a 1/d_a)^{-1}$, attained at $w_a \propto 1/d_a$.
\end{proof}

\begin{proof}[Proof of Proposition~\ref{prop:cl-own}]
(i) Theorem~\ref{thm:allocation} at $\lambda^\star$ gives $\mathbb{E}_\nu[\tau]\, \langle \bar w, k(\lambda^\star)\rangle = D_\tau$, the divergence of $P_\nu$ from $P_{\lambda^\star}$ on $\mathcal{F}_\tau$.
Lemma~\ref{lem:relax}(a) splits $D_\tau$ into $\kl(P_\nu(E), P_{\lambda^\star}(E))$ and the conditional divergence (\textsc{process}), which is the first display.
By the choice of $\lambda^\star$, $\langle \bar w, k(\lambda^\star)\rangle = \inf_{\lambda \in \Alt(\nu)} \langle \bar w, k(\lambda)\rangle \le \sup_w \inf_{\lambda \in \Alt(\nu)} \langle w, k(\lambda)\rangle = T^\star(\nu)^{-1}$ (\textsc{allocation}).
Adding $\mathbb{E}_\nu[\tau]\big(T^\star(\nu)^{-1} - \langle \bar w, k(\lambda^\star)\rangle\big)$ to both sides of the first display gives the second.
Its first two terms are a relative entropy and an average of relative entropies.
The third is non-negative by the inequality just shown.
(ii) The same steps apply with $\Alt_m(\nu)$ and $T^\star_m(\nu)$.
(iii) The laws of an exponential family are mutually absolutely continuous, so on $\{\tau = t\}$ the density of $P_{\lambda^\star}$ with respect to $P_\nu$ on $\mathcal{F}_t$ is $e^{-L_t}$.
Summing over $t$ on $E \subseteq \{\tau < \infty\}$ gives the formula.
\end{proof}

\begin{proof}[Proof of Proposition~\ref{prop:cl-weights}]
$\Alt(\nu)$ is the union over $b \ne a^\star$ of the sets on which arm $b$ is at least as good as $a^\star$, so Proposition~\ref{prop:altsets} reduces the infimum to a minimum over $b$.
On each such set the other arms keep their laws.
By the monotonicity argument in the proof of Proposition~\ref{prop:scalar}, arms $a^\star$ and $b$ move to a common mean.
Proposition~\ref{prop:mbary} places it at their weighted mean, which gives $w_{a^\star} G_b(w_b/w_{a^\star})$ by homogeneity.
The objective of (i) is an infimum of linear functions of $w$, so it is upper semicontinuous on the simplex and attains its maximum.
That maximum is positive, since the objective is positive at the allocation with every $w_a = 1/K$.
The objective vanishes at $w_{a^\star} = 0$, so every optimal allocation has $w_{a^\star} > 0$ and (i) applies there.
Items (ii) and (iii) are then the proofs of Propositions~\ref{prop:scalar} and~\ref{prop:scalar-general}.
\end{proof}

\begin{proof}[Proof of Lemma~\ref{lem:jensen-pair}]
Write $\psi_k := \psi^{(k)}(\theta_a)$, $u := \eta_a - \theta_a$ and $\bar\mu := s\mu_a + (1-s)\mu_a(\lambda)$, and let $\tilde\theta$ solve $\psi'(\tilde\theta) = \bar\mu$.
For exponential-family laws $c_a(s) = s\psi(\theta_a) + (1-s)\psi(\eta_a) - \psi(s\theta_a + (1-s)\eta_a)$.
(i) On the interior of the natural parameter space $\psi$ is analytic and $\psi'' > 0$, so $\tilde\theta$ is analytic in $u$.
Taylor expansion of $c_a(s) = s\psi(\theta_a) + (1-s)\psi(\theta_a + u) - \psi(\theta_a + (1-s)u)$ gives $c_a(s) = \tfrac12 s(1-s)\psi_2 u^2 + \tfrac16 s(1-s)(2-s)\psi_3 u^3 + O(u^4)$.
By Proposition~\ref{prop:mbary}, $\mathcal{J}_a(s) = s\, d(\mu_a, \bar\mu) + (1-s)\, d(\mu_a(\lambda), \bar\mu)$, and $d(\psi'(t), \psi'(\tilde\theta)) = \psi(\tilde\theta) - \psi(t) - (\tilde\theta - t)\psi'(t)$.
Matching powers of $u$ in $\psi'(\tilde\theta) = \bar\mu$ gives $\tilde\theta - \theta_a = (1-s)u + \tfrac12 s(1-s)(\psi_3/\psi_2)u^2 + O(u^3)$.
Substituting gives $\mathcal{J}_a(s) = \tfrac12 s(1-s)\psi_2 u^2 + \tfrac16 s(1-s)(1+s)\psi_3 u^3 + O(u^4)$.
Subtracting the two expansions gives (i).
(ii) With $\psi(t) = \sigma^2 t^2/2$ and $\psi^\star(\mu) = \mu^2/2\sigma^2$, both $c_a(s)$ and $\mathcal{J}_a(s)$ equal $s(1-s)\sigma^2 u^2/2$, which is $s(1-s)g^2/2\sigma^2$ since $g = \sigma^2 u$.
(iii) For exponential laws $\psi(t) = -\log(-t)$ on $t < 0$, the means are $-1/\theta_a$ and $-1/\eta_a$, and $\psi^\star(\mu) = -1 - \log\mu$.
Then $c_a(s) = \log(-s\theta_a - (1-s)\eta_a) - s\log(-\theta_a) - (1-s)\log(-\eta_a)$ and $\mathcal{J}_a(1-s) = \log\big((1-s)/(-\theta_a) + s/(-\eta_a)\big) + (1-s)\log(-\theta_a) + s\log(-\eta_a)$.
The two agree because $(1-s)/(-\theta_a) + s/(-\eta_a) = (-s\theta_a - (1-s)\eta_a)/(\theta_a\eta_a)$.
(iv) For Bernoulli laws $\psi^\star(\mu) = \mu\log\mu + (1-\mu)\log(1-\mu)$ is the negative entropy.
So $\mathcal{J}_a(s)$ is the mutual information between a Bernoulli label of mean $s$ and an observation drawn from $\nu_a$ when the label is $1$ and from $\lambda_a$ when it is $0$.
It is at most the entropy of the label, which is at most $\log 2$.
With $\mu_a = q$ and $\mu_a(\lambda) = 1-q$, $e^{-c_a(s)} = q^s(1-q)^{1-s} + (1-q)^s q^{1-s}$.
Each term is at most $q^{\underline{s}}$.
The term whose exponent on $q$ is $\underline{s}$ is at least $q^{\underline{s}}(1-q)$.
These two estimates give the two bounds.
\end{proof}

\begin{proof}[Proof of Proposition~\ref{prop:cl-tolerance}]
Fix an arm $a$ and let $E := \{\hat a \ne a\}$.
Under $\lambda^{(a)}$ every arm other than $a$ has gap at least $\Delta_{\mathrm{tol}}$, so $(\Delta_{\mathrm{tol}}, \delta)$-correctness at $\lambda^{(a)}$ gives $P_{\lambda^{(a)}}(E) \le \delta$.
Only arm $a$ differs, so Theorem~\ref{thm:allocation} gives $\mathbb{E}_\nu[N_a(\tau)]\, k_a(\lambda^{(a)}) = D_\tau$.
Lemma~\ref{lem:relax}(a) bounds $D_\tau$ below by the binary divergence $\kl(P_\nu(E), P_{\lambda^{(a)}}(E))$ (\textsc{process}).

(i) Correctness at $\nu$ gives $P_\nu(E) \ge 1 - \delta$, since $a \notin \mathcal{S}_{\mathrm{tol}}$.
As $1 - \delta \ge 1/2 \ge \delta$, Lemma~\ref{lem:relax}(c) bounds the binary divergence below by $\kl(1-\delta, \delta) = \kl(\delta, 1-\delta)$ (\textsc{uniform}).

(ii) If $P_\nu(E) \ge \delta$, Lemma~\ref{lem:relax}(c) bounds the binary divergence below by $\kl(P_\nu(E), \delta)$ (\textsc{uniform}).
Its second bound gives $\kl(P_\nu(E), \delta) \ge P_\nu(E)\log(1/\delta) - \log 2$ (\textsc{binary}).
Here $P_\nu(E) = 1 - P_\nu(\hat a = a)$.
If $P_\nu(E) < \delta$, the right side of (ii) is below $\delta\log(1/\delta) - \log 2$, which is negative.

(iii) The expected sample size is the sum of the expected pull counts.
By (i) the arms outside $\mathcal{S}_{\mathrm{tol}}$ contribute at least the first term.
By (ii), and since pull counts are non-negative, the arms of $\mathcal{S}_{\mathrm{tol}}$ contribute at least the positive part of
\[
\log\frac1\delta \sum_{a \in \mathcal{S}_{\mathrm{tol}}} \frac{1 - P_\nu(\hat a = a)}{k_a(\lambda^{(a)})} - \log 2 \sum_{a \in \mathcal{S}_{\mathrm{tol}}} \frac{1}{k_a(\lambda^{(a)})}
\]
The probabilities $P_\nu(\hat a = a)$ sum to at most one.
So the sum of $P_\nu(\hat a = a)/k_a(\lambda^{(a)})$ over $\mathcal{S}_{\mathrm{tol}}$ is at most the largest $1/k_a(\lambda^{(a)})$ there, which drops a non-negative term.
\end{proof}

\begin{proof}[Proof of Proposition~\ref{prop:cl-fb-twopoint}]
(i) Let $E := \{\hat{\mathcal{S}}_T = \mathcal{S}^\star(\nu)\}$.
Let $Z$ be its conditional probability given $\mathcal{F}_T$, which is the same under both instances.
Since $\mathcal{S}^\star(\lambda) \ne \mathcal{S}^\star(\nu)$, the error probability under $\lambda$ is $P_\lambda(E) + P_\lambda(\hat{\mathcal{S}}_T \notin \{\mathcal{S}^\star(\nu), \mathcal{S}^\star(\lambda)\})$.
As $\max(u, v) = \tfrac12(u + v) + \tfrac12|u - v|$, the maximum of the two error probabilities is $\tfrac12\big(P_\nu(E^c) + P_\lambda(E)\big)$ plus two non-negative terms, which are dropped.
Then $P_\nu(E^c) + P_\lambda(E) = 1 - \big(\mathbb{E}_\nu[Z] - \mathbb{E}_\lambda[Z]\big)$, which is at least $1 - \TV(P_\nu, P_\lambda)$ on $\mathcal{F}_T$ (\textsc{process}).
The proofs of Theorem~\ref{thm:limit} and Proposition~\ref{prop:cl-twopoint} bound $1 - \TV(P_\nu, P_\lambda)$ below by $1 - \sqrt{1 - \rho_T^2}$, by $\tfrac12\rho_T^2$ and by $\tfrac12 e^{-D_T}$ (\textsc{test}).
Here $\rho_T^2 = e^{-2C_T(1/2)}$.

(ii) Proposition~\ref{prop:tilt} gives $C_T(1/2) = \sum_a \mathbb{E}_{Q_{1/2}}[N_a(T)]\, c_a(1/2) - \KL(Q_{1/2} \,\|\, P^{(1/2)})$.
Since $\sum_a N_a(T) = T$, the sum equals $T\max_a c_a(1/2) - \sum_a \mathbb{E}_{Q_{1/2}}[N_a(T)]\big(\max_{a'} c_{a'}(1/2) - c_a(1/2)\big)$.
\end{proof}

\begin{proof}[Proof of Proposition~\ref{prop:cl-fb-gauss}]
For Gaussian laws with a common variance, moving a mean by twice a distance multiplies its divergence by four.
(i) Arm $b$ of $\lambda^{(b)}$ has mean $\mu^\star + \Delta_b$, so $b$ is its unique optimal arm.
Its per-pull divergence is $k_b(\lambda^{(b)}) = 4d_b$.
Since $\sum_{b \ne a^\star} \mathbb{E}_\nu[N_b(T)] \le T$, some suboptimal arm $b$ has $\mathbb{E}_\nu[N_b(T)] \le T d_b^{-1}/\sum_{b' \ne a^\star} d_{b'}^{-1}$ (\textsc{allocation}).
Only arm $b$ differs, so Theorem~\ref{thm:allocation} gives $D_T = 4d_b\, \mathbb{E}_\nu[N_b(T)] \le 4T/\sum_{b'} d_{b'}^{-1}$.
Proposition~\ref{prop:cl-fb-twopoint}(i) with $m = 1$ gives the bound.
The \textsc{boundary} defect is the difference $3d_b\, \mathbb{E}_\nu[N_b(T)]$ between $D_T$ and its value at the boundary $\mu_b(\lambda) = \mu^\star$ of $\Alt(\nu)$, beyond which $\lambda^{(b)}$ keeps the margin $\Delta_b$.

(ii) In $\lambda^{(a,b)}$ arm $a$ falls below $\mu_{[m+1]}$.
Arm $b$ rises above $\mu_{[m]}$.
Every other arm of $\mathcal{S}^\star$ keeps a mean of at least $\mu_{[m]}$.
Every other arm outside it keeps a mean of at most $\mu_{[m+1]}$.
So the top-$m$ set of $\lambda^{(a,b)}$ is unique and equals $\mathcal{S}^{(a,b)}$.
Write $U := \sum_{c \in [K]} 1/d^{(m)}_c$, $U^+ := \sum_{c \in \mathcal{S}^\star} 1/d^{(m)}_c$ and $U^- := U - U^+$.
Since $\sum_c \mathbb{E}_\nu[N_c(T)] = T$, some arm $c$ has $\mathbb{E}_\nu[N_c(T)] \le T (d^{(m)}_c)^{-1}/U$ (\textsc{allocation}).
If $c \in \mathcal{S}^\star$, let $a := c$.
The arms outside $\mathcal{S}^\star$ receive at most $T$ pulls, so some $b \notin \mathcal{S}^\star$ has $\mathbb{E}_\nu[N_b(T)] \le T (d^{(m)}_b)^{-1}/U^-$ (\textsc{allocation}).
If $c \notin \mathcal{S}^\star$, let $b := c$ and choose $a \in \mathcal{S}^\star$ with $\mathbb{E}_\nu[N_a(T)] \le T (d^{(m)}_a)^{-1}/U^+$ in the same way.
Arms $a$ and $b$ move by twice their distances to the boundary means $\mu_{[m+1]}$ and $\mu_{[m]}$.
So $k_a(\lambda^{(a,b)}) = 4d^{(m)}_a$ and $k_b(\lambda^{(a,b)}) = 4d^{(m)}_b$.
Theorem~\ref{thm:allocation} then gives
\[
D_T \;=\; 4d^{(m)}_a\, \mathbb{E}_\nu[N_a(T)] + 4d^{(m)}_b\, \mathbb{E}_\nu[N_b(T)] \;\le\; \frac{4T}{U} + \frac{4T}{\min(U^+, U^-)}
\]
Proposition~\ref{prop:cl-fb-twopoint}(i) gives the bound.
The \textsc{boundary} defect is the difference $3d^{(m)}_a\, \mathbb{E}_\nu[N_a(T)] + 3d^{(m)}_b\, \mathbb{E}_\nu[N_b(T)]$ between $D_T$ and its value with arms $a$ and $b$ at the boundary means.
\end{proof}

\begin{proof}[Proof of Proposition~\ref{prop:cl-fb-rate}]
(i) Fix $\lambda \in \Alt(\nu)$ with a unique optimal arm, and let $E := \{\hat a_T = a^\star\}$.
In a one-parameter exponential family $\KL(\lambda_a \,\|\, \nu_a)$ is finite at every arm.
Theorem~\ref{thm:allocation} with the roles of $\nu$ and $\lambda$ exchanged gives, on $\mathcal{F}_T$,
\[
\sum_a \mathbb{E}_\lambda[N_a(T)]\, \KL(\lambda_a \,\|\, \nu_a) \;=\; \KL(P_\lambda \,\|\, P_\nu)
\]
Lemma~\ref{lem:relax}(a) with the roles exchanged (\textsc{process}) and the first bound of Lemma~\ref{lem:relax}(c) (\textsc{binary}) give
\[
\KL(P_\lambda \,\|\, P_\nu) \;\ge\; \kl\big(P_\lambda(E), P_\nu(E)\big) \;\ge\; \big(1 - P_\lambda(E)\big)\log\frac{1}{P_\nu(\hat a_T \ne a^\star)} - \log 2
\]
Since $a^\star(\lambda) \ne a^\star$, $P_\lambda(E) \le P_\lambda(\hat a_T \ne a^\star(\lambda))$, which tends to $0$ (\textsc{uniform}).
Dividing by $T(1 - P_\lambda(E))$ and letting $T \to \infty$ (\textsc{asymptotic}) bounds the left side of (i) by $\limsup_{T \to \infty} \sum_a \big(\mathbb{E}_\lambda[N_a(T)]/T\big)\KL(\lambda_a \,\|\, \nu_a)$.
This is at most $\max_a \KL(\lambda_a \,\|\, \nu_a)$ (\textsc{allocation}).
With deterministic counts it equals $\sum_a w_a \KL(\lambda_a \,\|\, \nu_a)$.
The instances with a unique optimal arm are dense in $\Alt(\nu)$.
On $\Alt(\nu)$, $\KL(\lambda_a \,\|\, \nu_a) = d(\mu_a(\lambda), \mu_a)$ is continuous in $\mu_a(\lambda)$, so both bounds hold with the infimum over $\Alt(\nu)$ (\textsc{boundary}).

(ii) Write $\theta'$ for the natural parameter of the mean $\mu'$ and $\bar\theta := s\theta_{a^\star} + (1-s)\theta_b$.
For a mean $\mu$ with natural parameter $\theta$, $d(\mu', \mu) = (\theta' - \theta)\mu' - \psi(\theta') + \psi(\theta)$.
So the left side of (ii) minus $d(\mu', \psi'(\bar\theta))$ is $s\psi(\theta_{a^\star}) + (1-s)\psi(\theta_b) - \psi(\bar\theta)$.
That is $c_{\nu_{a^\star}, \nu_b}(s)$, since $\int \nu_{a^\star}^{\,s} \nu_b^{\,1-s} = e^{\psi(\bar\theta) - s\psi(\theta_{a^\star}) - (1-s)\psi(\theta_b)}$.

(iii) $\Alt(\nu)$ is the union over $b \ne a^\star$ of the sets on which arm $b$ is at least as good as $a^\star$.
An infimum over a union is the minimum of the infima over its parts, as in the proof of Proposition~\ref{prop:altsets}, so each infimum is a minimum over $b$.
On the set of $b$ the other arms keep their laws at no cost.
Each $d(\cdot, \mu)$ decreases below $\mu$ and increases above it, and $\mu_b < \mu_{a^\star}$.
So means of arms $a^\star$ and $b$ with arm $b$ ahead can be moved to a common point between $\mu_b$ and $\mu_{a^\star}$ at no greater cost.
For the first infimum, (ii) with $s := w_{a^\star}/(w_{a^\star} + w_b)$ gives
\[
w_{a^\star}\, d(\mu', \mu_{a^\star}) + w_b\, d(\mu', \mu_b) \;=\; (w_{a^\star} + w_b)\Big(c_{\nu_{a^\star}, \nu_b}(s) + d\big(\mu', \psi'(\bar\theta)\big)\Big)
\]
Its minimum over $\mu'$ is attained at $\mu' = \psi'(\bar\theta)$, which lies between $\mu_b$ and $\mu_{a^\star}$.
For the second infimum, (ii) gives, for every $\mu'$ and every $s$,
\[
\max\big(d(\mu', \mu_{a^\star}),\, d(\mu', \mu_b)\big) \;\ge\; s\, d(\mu', \mu_{a^\star}) + (1-s)\, d(\mu', \mu_b) \;\ge\; c_{\nu_{a^\star}, \nu_b}(s)
\]
As $s$ runs from $0$ to $1$, the point $\psi'(\bar\theta)$ moves continuously from $\mu_b$ to $\mu_{a^\star}$.
So for some $s$ the two divergences $d(\psi'(\bar\theta), \mu_{a^\star})$ and $d(\psi'(\bar\theta), \mu_b)$ are equal.
There (ii) makes both equal to $c_{\nu_{a^\star}, \nu_b}(s)$.
So the minimum over $\mu'$ of the maximum is at most $\max_s c_{\nu_{a^\star}, \nu_b}(s)$.
By the displayed bound it is also at least that value.
\end{proof}

\begin{proof}[Proof of Proposition~\ref{prop:cl-mdp-decomposition}]
Write $\ell_{x,a} := \log(d\nu_{x,a}/d\lambda_{x,a})$, a function of the reward and the next state.
The initial-state laws are equal.
The action probabilities given $\mathcal{F}_{t-1}$ do not depend on the instance.
Both therefore cancel in the likelihood ratio.
So the increment of $L$ at round $t$ is $\ell_{x,a}$ at the reward and the next state of that round, where $(x,a)$ is its state-action pair.
Given $\mathcal{F}_{t-1}$ and the action of round $t$, that pair is fixed.
The reward and the next state then have the law $\nu_{x,a}$ under $P_\nu$.
The conditional mean of the increment is therefore $\KL(\nu_{x,a} \,\|\, \lambda_{x,a})$.
Hence $W_t := L_t - \sum_{x,a} N_{x,a}(t)\, \KL(\nu_{x,a} \,\|\, \lambda_{x,a})$ is a $P_\nu$-martingale with $W_0 = 0$.
As in the proof of Theorem~\ref{thm:allocation}, $\mathbb{E}_{\nu_{x,a}}|\ell_{x,a}| \le \KL(\nu_{x,a} \,\|\, \lambda_{x,a}) + 2$ at every pair.
The conditional increments of $W$ are therefore bounded in $L^1$ by twice the largest of these finitely many relative entropies, plus $2$.
Optional stopping at a time with $\mathbb{E}_\nu[\tau] < \infty$ gives $\mathbb{E}_\nu[W_\tau] = 0$.
The left side of the identity is $\mathbb{E}_\nu[L_\tau]$, which the same bound renders absolutely convergent.
\end{proof}

\begin{proof}[Proof of Proposition~\ref{prop:cl-mdp}]
Fix $\lambda \in \Alt_\Phi(\nu)$ with $\KL(\nu_{x,a} \,\|\, \lambda_{x,a}) < \infty$ at every pair; step 5 returns to the other alternatives.
It is ergodic, because $\nu \ll \lambda$ only adds transitions to those of $\nu$.
If every state $x$ had an action in $\mathcal{A}^\star_x \cap \mathcal{A}^\star_x(\lambda)$, the stationary policy choosing such actions would lie in $\Pi^\star \cap \Pi^\star(\lambda)$.
So some state $x$ has every action of $\mathcal{A}^\star_x$ suboptimal under $\lambda$.
By the count form at $\lambda$ and uniform goodness, $\mathbb{E}_\lambda\big[\sum_{a \in \mathcal{A}^\star_x} N_{x,a}(T)\big] = o(T^p)$ for every $p > 0$ (\textsc{uniform}).
Let $E := \{N_x(T) \ge \chi T,\ \sum_{a \notin \mathcal{A}^\star_x} N_{x,a}(T) \le \sqrt T\}$.
\begin{enumerate}
\item Identity: the two instances start from the same initial state, so Proposition~\ref{prop:cl-mdp-decomposition} at the fixed time $T$ gives
\[
\mathbb{E}_\nu[L_T] = \sum_{x,a} \mathbb{E}_\nu[N_{x,a}(T)] \KL(\nu_{x,a} \,\|\, \lambda_{x,a})
\]
Pairs with $a \in \mathcal{A}^\star_x$ contribute nothing.
\item \textsc{process}: $\mathbb{E}_\nu[L_T] \ge \kl(P_\nu(E), P_\lambda(E))$, as in Lemma~\ref{lem:relax}(a).
\item \textsc{markov}: by Markov's inequality and the visit bound, $1 - P_\nu(E) \le O(e^{-\chi T/2}) + \mathbb{E}_\nu\big[\sum_{a \notin \mathcal{A}^\star_x} N_{x,a}(T)\big]/\sqrt T$.
The expectation is $o(T^p)$ by the count form at $\nu$ and uniform goodness (\textsc{uniform}), so $P_\nu(E) \to 1$.
On $E$ the optimal actions at $x$ are taken at least $\chi T - \sqrt T$ times, so $P_\lambda(E) \le \mathbb{E}_\lambda\big[\sum_{a \in \mathcal{A}^\star_x} N_{x,a}(T)\big]/(\chi T - \sqrt T)$.
\item \textsc{binary} and \textsc{asymptotic}: Lemma~\ref{lem:relax}(c) gives $\kl(P_\nu(E), P_\lambda(E)) \ge P_\nu(E)\log(1/P_\lambda(E)) - \log 2$, so for every $p > 0$
\[
\liminf_{T \to \infty} \frac{1}{\log T} \sum_{x,a} \mathbb{E}_\nu[N_{x,a}(T)] \KL(\nu_{x,a} \,\|\, \lambda_{x,a}) \;\ge\; 1 - p
\]
\item \textsc{asymptotic} and \textsc{allocation}: the constraint holds for every alternative in $\Alt_\Phi(\nu)$ with finite divergences at every pair.
Take $T_k \to \infty$ realizing $\liminf_T R_T/\log T$, which may be assumed finite.
Pass to a subsequence along which $\zeta_k := \mathbb{E}_\nu[N(T_k)]/\log T_k$ converges in $[0,\infty]^{\mathcal{X} \times \mathcal{A}}$, to $\zeta$.
By the count form at $\nu$ its infinite coordinates lie at optimal pairs, where the Bellman gap and the relative entropy $\KL(\nu_{x,a} \,\|\, \lambda_{x,a})$ both vanish, so setting them to zero changes neither sum.
The other coordinates converge to finite values, so for every alternative with finite divergences, letting $k \to \infty$ in its constraint gives $\sum_{x,a} \zeta(x,a) \KL(\nu_{x,a} \,\|\, \lambda_{x,a}) \ge 1$.
An alternative in $\Alt_\Phi(\nu)$ with an infinite divergence has it at a suboptimal pair, since divergences vanish at optimal pairs.
Its constraint holds for every point that is positive at that pair.
So adding the same positive constant to every coordinate of $\zeta$ gives a feasible point of the program defining $c^\star_\Phi(\nu)$.
Letting the constant tend to $0$ gives $c^\star_\Phi(\nu) \le \sum_{x,a} \zeta(x,a)\, \Delta_{x,a}$.
By the count form at $\nu$, $R_{T_k}/\log T_k \to \sum_{x,a} \zeta(x,a)\, \Delta_{x,a}$, which is therefore at least $c^\star_\Phi(\nu)$.
\end{enumerate}
\end{proof}

\begin{proof}[Proof of Proposition~\ref{prop:cl-mdp-decoupled}]
Write $\Sigma$ for the right side and $r_\lambda(x,a)$ for the mean reward of an instance $\lambda$ at $(x,a)$, and let
\[
J_\lambda(x,a) \;:=\; r_\lambda(x,a) + \sum_y \lambda(y \mid x,a)\, h^\star(y)
\]
so that $J_\nu(x,a) = g^\star + h^\star(x) - \Delta_{x,a}$.
In an ergodic instance an optimal stationary policy takes only optimal actions \cite[Section~3]{ok2018exploration}.

For the inequality, fix a suboptimal pair $(x,a)$ and an instance $\lambda \in \Phi$ that differs from $\nu$ only at $(x,a)$, with finite relative entropy there and with $J_\lambda(x,a) > g^\star + h^\star(x)$.
Then $\nu \ll \lambda$.
The instance $\lambda$ agrees with $\nu$ at the optimal pairs.
It is ergodic, because $\nu \ll \lambda$ only adds transitions to those of $\nu$.
Let $f \in \Pi^\star$.
It takes only optimal actions, so its chain and its rewards are the same under $\lambda$ as under $\nu$.
Its gain under $\lambda$ is therefore $g^\star$.
Let $f'$ equal $f$ except that $f'(x) := a$, and let $\pi'$ be the stationary law of $f'$ under $\lambda$, which charges every state because $\lambda$ is ergodic.
Then $J_\lambda(x', f'(x')) \ge g^\star + h^\star(x')$ at every state $x'$, with equality at $x' \ne x$ and strict inequality at $x$.
Averaging against $\pi'$ gives the gain of $f'$ under $\lambda$ as $\sum_{x'} \pi'(x')\big(J_\lambda(x', f'(x')) - h^\star(x')\big) > g^\star$, so $f \notin \Pi^\star(\lambda)$.
Since $f$ was arbitrary, $\lambda \in \Alt_\Phi(\nu)$.
Every $\zeta$ feasible for the program defining $c^\star_\Phi(\nu)$ therefore has $\zeta(x,a)\, \KL(\nu_{x,a} \,\|\, \lambda_{x,a}) \ge 1$, the only nonzero term of its constraint at $\lambda$.
The infimum of this relative entropy over such $\lambda$ is $\Kinf(x,a)$, so $\zeta(x,a) \ge 1/\Kinf(x,a)$.
Hence $\sum_{x,a} \zeta(x,a)\, \Delta_{x,a} \ge \Sigma$ for every feasible $\zeta$, which is the inequality.

For the equality let $\Phi$ be a product class and suppose $\Sigma < \infty$.
Then $\Kinf(x,a) > 0$ at every suboptimal pair.
Let $\zeta^\star(x,a) := 1/\Kinf(x,a)$ at the suboptimal pairs and $\zeta^\star(x,a) := 0$ at the optimal ones, and fix $\xi > 0$ and $\lambda \in \Alt_\Phi(\nu)$.
Some pair $(x,a)$ has $J_\lambda(x,a) > g^\star + h^\star(x)$, by display~(18) of the proof of Theorem~2 of Ok, Prouti\`ere and Tranos \cite[Appendix~D]{ok2018exploration}.
Indeed, otherwise $r_\lambda(x',a') \le g^\star + h^\star(x') - \sum_y \lambda(y \mid x',a')\, h^\star(y)$ at every pair $(x',a')$, with equality at the pairs of any $f \in \Pi^\star$, at which $\lambda$ agrees with $\nu$.
Summing along any history under $\lambda$ bounds the expected total reward of every algorithm over $T$ rounds by $Tg^\star$ plus the span of $h^\star$.
The policy $f$ attains the gain $g^\star$, so $f \in \Pi^\star(\lambda)$, which contradicts $\lambda \in \Alt_\Phi(\nu)$.
The pair $(x,a)$ is suboptimal, since at an optimal pair $\lambda$ agrees with $\nu$ and $J_\lambda(x,a) = g^\star + h^\star(x)$.
The instance equal to $\nu$ except that its law at $(x,a)$ is $\lambda_{x,a}$ lies in $\Phi$, because $\Phi$ is a product class, so it enters the infimum defining $\Kinf(x,a)$.
If $\KL(\nu_{x,a} \,\|\, \lambda_{x,a}) < \infty$, it is at least $\Kinf(x,a)$.
Then $\sum_{x',a'} (\zeta^\star(x',a') + \xi)\, \KL(\nu_{x',a'} \,\|\, \lambda_{x',a'}) \ge \zeta^\star(x,a)\, \Kinf(x,a) = 1$.
If it is infinite, the same sum is infinite, because $\xi > 0$.
So $\zeta^\star$ plus $\xi$ at every pair is a feasible point.
Hence $c^\star_\Phi(\nu) \le \Sigma + \xi \sum_{x,a} \Delta_{x,a}$.
Letting $\xi \to 0$ gives $c^\star_\Phi(\nu) \le \Sigma$.
\end{proof}

\begin{proof}[Proof of Corollary~\ref{cor:cl-structured}]
View each instance as a Markov decision process with one state, whose actions are the arms.
At the pair formed by that state and arm $a$, the next state is the one state and the reward has the law $\nu_a$, so the relative entropy at the pair is $\KL(\nu_a \,\|\, \lambda_a)$.
Every stationary deterministic policy plays one arm, under which the one state forms an irreducible chain.
So every instance is ergodic.
The optimal gain is $\mu^\star$, the bias is constant, the Bellman gap of arm $a$ is $\Delta_a$, and the optimal stationary policies are the optimal arms.
The alternative set of Proposition~\ref{prop:cl-mdp} is therefore $\Alt_\Phi(\nu)$.
The constant of Proposition~\ref{prop:cl-mdp} is then $c^\star_\Phi(\nu)$.
The regret in Proposition~\ref{prop:cl-mdp} is that of Section~\ref{sec:notation}, so an algorithm consistent on $\Phi$ is uniformly good.
The one state is visited in every round.
The count form therefore holds with no $O(1)$ term.
The visit bound holds for every $\chi < 1$.
Proposition~\ref{prop:cl-mdp} then gives the bound.
Its proof applies \textsc{process}, \textsc{binary}, \textsc{markov}, \textsc{uniform}, \textsc{asymptotic} and \textsc{allocation}.
\end{proof}

\begin{proof}[Proof of Corollary~\ref{cor:cl-linear}]
Corollary~\ref{cor:cl-structured} gives the bound with $c^\star_\Phi(\nu)$ as defined there.
It remains to evaluate its constraint.
Write $\omega$ for the parameter of $\nu$ and $\omega + u$ for that of $\lambda \in \Phi$.
Then $\KL(\nu_a \,\|\, \lambda_a) = \langle \varphi_a, u \rangle^2/2$ at every arm, so the sum in the constraint at $\lambda$ is $\tfrac12 u^\top V(\zeta) u$, and $\nu \ll \lambda$ always.
When $\langle \varphi_{a^\star}, u \rangle = 0$, the mean of arm $b$ under $\lambda$ exceeds that of $a^\star$ by $\langle \varphi_b, u \rangle - \Delta_b$.
So $\Alt_\Phi(\nu)$ is the union over $b \ne a^\star$ of the sets of $u$ with $\langle \varphi_{a^\star}, u \rangle = 0$ and $\langle \varphi_b, u \rangle > \Delta_b$.
By Proposition~\ref{prop:altsets} and the continuity of the quadratic form, the constraint of $c^\star_\Phi(\nu)$ reads $q_b(\zeta) \ge 1$ for every $b \ne a^\star$, where
\[
q_b(\zeta) := \inf\big\{ \tfrac12 u^\top V(\zeta) u : \langle \varphi_{a^\star}, u \rangle = 0,\ \langle \varphi_b, u \rangle \ge \Delta_b \big\}
\]
with $\inf \emptyset := \infty$.
Two facts compare it with the stated condition.
First, for $M$ positive definite and $\Delta > 0$, the Cauchy--Schwarz inequality gives $\inf\{\tfrac12 u^\top M u : \langle \varphi, u \rangle \ge \Delta\} = \Delta^2/(2\varphi^\top M^{-1}\varphi)$, attained at a multiple of $M^{-1}\varphi$.
Second, the infimum $\tilde q_b(c)$ of $\tfrac12 u^\top \big(V(\zeta) + c\, \varphi_{a^\star}\varphi_{a^\star}^\top\big) u$ over $\langle \varphi_b, u \rangle \ge \Delta_b$ increases to $q_b(\zeta)$ as $c \to \infty$.
Indeed $\tilde q_b(c)$ is non-decreasing in $c$ and at most $q_b(\zeta)$.
Scaling a point down to the hyperplane $\langle \varphi_b, u \rangle = \Delta_b$ only lowers the form.
Let $q_\infty$ be the supremum of $\tilde q_b(c)$ over $c$, and suppose $q_\infty < q_b(\zeta)$.
Points $u_c$ on that hyperplane with form at most $q_\infty + 1/c$ keep $V(\zeta)^{1/2}u_c$ bounded and send $\langle \varphi_{a^\star}, u_c \rangle$ to $0$.
The image of $u \mapsto \big(V(\zeta)^{1/2}u, \langle \varphi_{a^\star}, u \rangle, \langle \varphi_b, u \rangle\big)$ is a closed subspace, so an accumulation point of these images is the image of a point $u$ with $\langle \varphi_{a^\star}, u \rangle = 0$, $\langle \varphi_b, u \rangle = \Delta_b$ and $\tfrac12 u^\top V(\zeta) u \le q_\infty$, a contradiction.
Now let $\zeta$ meet the stated condition.
The first fact with $M = V(\zeta)$ gives $\tilde q_b(0) \ge 1$.
Since $q_b(\zeta) \ge \tilde q_b(0)$, the point $\zeta$ is feasible for $c^\star_\Phi(\nu)$.
Conversely let $\zeta$ be feasible for $c^\star_\Phi(\nu)$, fix $\xi > 0$, and let $\zeta'$ add $\xi$ at every arm.
Then $V(\zeta')$ is positive definite, because the vectors $\varphi_a$ span $\mathbb{R}^d$.
Added weight can only raise each infimum, so $q_b(\zeta') \ge q_b(\zeta) \ge 1$.
By the second fact, applied to $\zeta'$, there is $c$ with $\tilde q_b(c) \ge 1/(1+\xi)$ for every $b \ne a^\star$.
Let $\zeta''$ add $c$ to $\zeta'$ at $a^\star$.
The vector $(1+\xi)\zeta''$ has the positive definite matrix $(1+\xi)\big(V(\zeta') + c\, \varphi_{a^\star}\varphi_{a^\star}^\top\big)$, so by the first fact it meets the stated condition.
Its value is $(1+\xi)\big(\sum_{b \ne a^\star} \zeta_b\Delta_b + \xi \sum_{b} \Delta_b\big)$.
Letting $\xi \to 0$ shows that the two infima are equal.
\end{proof}

\begin{proof}[Proof of Corollary~\ref{cor:cl-lipschitz}]
Corollary~\ref{cor:cl-structured} gives the bound with $c^\star_\Phi(\nu)$ as defined there.
It remains to evaluate its constraint.
For $b \ne a^\star$ let $\mathcal{B}_b$ hold the $\lambda \in \Alt_\Phi(\nu)$ with $\mu_b(\lambda) > \mu^\star$.
An alternative keeps the mean $\mu^\star$ at $a^\star$, which is not optimal under it, so $\Alt_\Phi(\nu) = \bigcup_{b \ne a^\star} \mathcal{B}_b$.
Proposition~\ref{prop:altsets} then reduces the constraint to one infimum for each $b$.
Write $m_b(a) := \mu_a \vee (\mu^\star - \mathrm{dist}(a,b))$.
Every $\lambda \in \mathcal{B}_b$ has $\mu_a(\lambda) \ge \mu_b(\lambda) - \mathrm{dist}(a,b) > \mu^\star - \mathrm{dist}(a,b)$ at every arm $a$.
The map $v \mapsto \kl(\mu_a, v)$ is non-decreasing on $[\mu_a, 1]$, since its derivative is $(v - \mu_a)/(v(1-v))$.
So $\kl(\mu_a, \mu_a(\lambda)) \ge \kl(\mu_a, m_b(a))$ wherever $m_b(a) > \mu_a$.
Wherever $m_b(a) = \mu_a$ the right side is $0$.
Hence $\sum_a \zeta_a \kl(\mu_a, \mu_a(\lambda)) \ge \sum_a \zeta_a \kl(\mu_a, m_b(a))$ on $\mathcal{B}_b$.
Conversely, for $0 < \xi < (1 - \mu^\star) \wedge \mathrm{dist}(a^\star, b)$ let $\lambda^{(b,\xi)}$ have Bernoulli laws with means $\mu_a \vee (\mu^\star + \xi - \mathrm{dist}(a,b))$.
These means lie in $(0,1)$ and satisfy the Lipschitz condition, as the maximum of two functions that do, the second by the triangle inequality.
The mean at $a^\star$ is $\mu^\star$, because $\xi < \mathrm{dist}(a^\star, b)$.
The mean at $b$ is $\mu^\star + \xi$.
So $\lambda^{(b,\xi)} \in \mathcal{B}_b$, with $\nu \ll \lambda^{(b,\xi)}$ because every mean lies in $(0,1)$.
As $\xi \to 0$ its sum tends to $\sum_a \zeta_a \kl(\mu_a, m_b(a))$, by the continuity of $\kl$ on $(0,1)^2$.
The infimum over $\mathcal{B}_b$ is therefore that sum.
So the constraint is the stated one.
\end{proof}

\section*{Acknowledgments}

We acknowledge the use of large language models in preparing this work.

\bibliography{sequential_limits_ledger}

\end{document}